%% file: thesis_main.tex
\def\firstsub{first}%
\def\arxivsub{arxiv}%
\ifx\submissiontype\undefined%
\def\submissiontype{final}%
\fi%
\ifx\submissiontype\firstsub%
\PassOptionsToPackage{twoside}{geometry}
\fi%
\documentclass[titlepage,shield]{hwthesis}

\usepackage[table]{xcolor}
\usepackage{amsthm,graphicx,a4wide,exscale,relsize,setspace,lineno}
\usepackage[square,numbers,sort&compress]{natbib}
\usepackage{times}
\usepackage{epsfig}
\usepackage{subcaption}
\usepackage{authblk}
\usepackage{array}
\usepackage{eso-pic}
\usepackage{xspace}
\usepackage{geometry}
\usepackage[nottoc]{tocbibind}
\usepackage[title,titletoc]{appendix}
\usepackage{mathtools} %
\usepackage{amsmath} %
\usepackage{tensor}   %
\usepackage{mathrsfs} %
\usepackage{bbm}      %
\usepackage{booktabs} %
\usepackage{pdfpages} %
\usepackage{enumitem} %

\usepackage{url}

\usepackage{breakurl}
\usepackage[breaklinks]{hyperref}

\usepackage{float}

\usepackage{multirow}
\usepackage{arydshln}

\usepackage{siunitx}
\usepackage{tabularx}

\usepackage{comment}

\usepackage{color}
\usepackage{arydshln}
\usepackage{relsize}
\usepackage{makecell}
\usepackage{tabularx}
\usepackage{array}
\usepackage{tabularx}

\usepackage{notoccite}

\usepackage{pifont}%
\usepackage{listings}
\usepackage{color}
\usepackage{mdframed} %
\usepackage{arydshln}
\usepackage{mathtools}

\usepackage{epsfig}
\usepackage{amsmath}
\usepackage{amssymb}

\usepackage{subcaption}
\usepackage{eurosym}

\usepackage{makecell}
\usepackage{algorithm}
\usepackage[noend]{algpseudocode}

\usepackage{arydshln}
\usepackage{pifont}%
\newcommand{\cmark}{\ding{51}}%
\newcommand{\xmark}{\ding{55}}%
\newcommand{\rulesep}{\unskip\ \vrule\ }

\usepackage{amssymb}
\usepackage{scalerel}
\usepackage{stackengine}
\usepackage{pgf}
\newcounter{iloop}
\newcommand\openbigstar[1][0.7]{%
  \scalerel*{%
    \stackinset{c}{-.125pt}{c}{}{\scalebox{#1}{\color{white}{$\bigstar$}}}{%
      $\bigstar$}%
  }{\bigstar}
}
\newcommand{\Stars}[1]{\ensuremath{%
\pgfmathtruncatemacro{\imax}{ifthenelse(int(#1)==#1,#1-1,#1)}%
\pgfmathsetmacro{\xrest}{0.9*(1-#1+\imax)}%
\setcounter{iloop}{0}%
\loop\stepcounter{iloop}\ifnum\value{iloop}<\the\numexpr\imax+1
\bigstar\repeat
\openbigstar[\xrest]%
\setcounter{iloop}{0}%
\loop\stepcounter{iloop}\ifnum\value{iloop}<\the\numexpr5-\imax\relax
\openbigstar[.9]\repeat}}

\newcommand{\hyperfootnote}[1][]{\def\ArgI\hyperfootnoteRelay}
\newcommand\hyperfootnoteRelay[2][]{\href{#1#2}{\ArgI}\footnote{\href{#1#2}{#2}}}

\newcommand\todo[1]{\textcolor{black}{#1}}
\newcommand\newtext[1]{\textcolor{black}{#1}}
\newcommand{\etal}{{\textit{et al}}.\@ }

\setlist{nosep}        %

\definecolor{Gray}{gray}{0.9}
\definecolor{maroon}{cmyk}{0,0.87,0.68,0.32}		       

\ifx\submissiontype\arxivsub%
\else%
\fi%

\definecolor{MyDarkBlue}{rgb}{0.15,0.25,0.45}

\usepackage{hyperref}
\ifx\submissiontype\arxivsub%
\hypersetup{
hypertexnames=false,
colorlinks=true,
citecolor=MyDarkBlue,
linkcolor=MyDarkBlue,
urlcolor=MyDarkBlue,
pdfauthor={Marcel Sheeny de Moraes},                     %
pdftitle={All-Weather Object Recognition Using Radar and Infrared Sensing},            %
pdfsubject={hep-th math-ph},              %
breaklinks=true
}
\else%
\hypersetup{
hypertexnames=false,
colorlinks=true,
citecolor=black,
linkcolor=black,
urlcolor=black,
pdfauthor={Marcel Sheeny de Moraes},                     %
pdftitle={All-Weather Object Recognition Using Radar and Infrared Sensing},            %
pdfsubject={hep-th math-ph},              %
breaklinks=true
}
\fi%

\ifx\submissiontype\arxivsub%
\relax
\else%
\fi%

\usepackage{macros} %

\global\hbadness=100000 %

\let\oldapp\appendix
\renewcommand{\appendix}{\oldapp\renewcommand*{\chaptername}{\appendixname}}

\begin{document}

\title{All-Weather Object Recognition Using Radar and Infrared Sensing}

\date{\today} %

\author{Marcel Sheeny} %

\maketitle
\pagenumbering{gobble}
\ifx\submissiontype\firstsub%
\mbox{}
\fi%

\begin{abstract}
\input{Chapters/Abstract.tex}
\end{abstract}
\ifx\submissiontype\firstsub%
\mbox{}
\fi%

\clearpage
\vspace*{12cm}
\hfil
\hspace{6.5cm}
\textit{Dedicated to my mom Nadra}
\clearpage
\ifx\submissiontype\firstsub%
\mbox{}
\clearpage
\fi%

\begin{acknowledgements}
\input{Chapters/Acknowledgements.tex}

\end{acknowledgements}
\ifx\submissiontype\firstsub%
\mbox{}
\fi%

\ifx\submissiontype\arxivsub%
\relax
\else%
\includepdf[pages={1,2,3}]{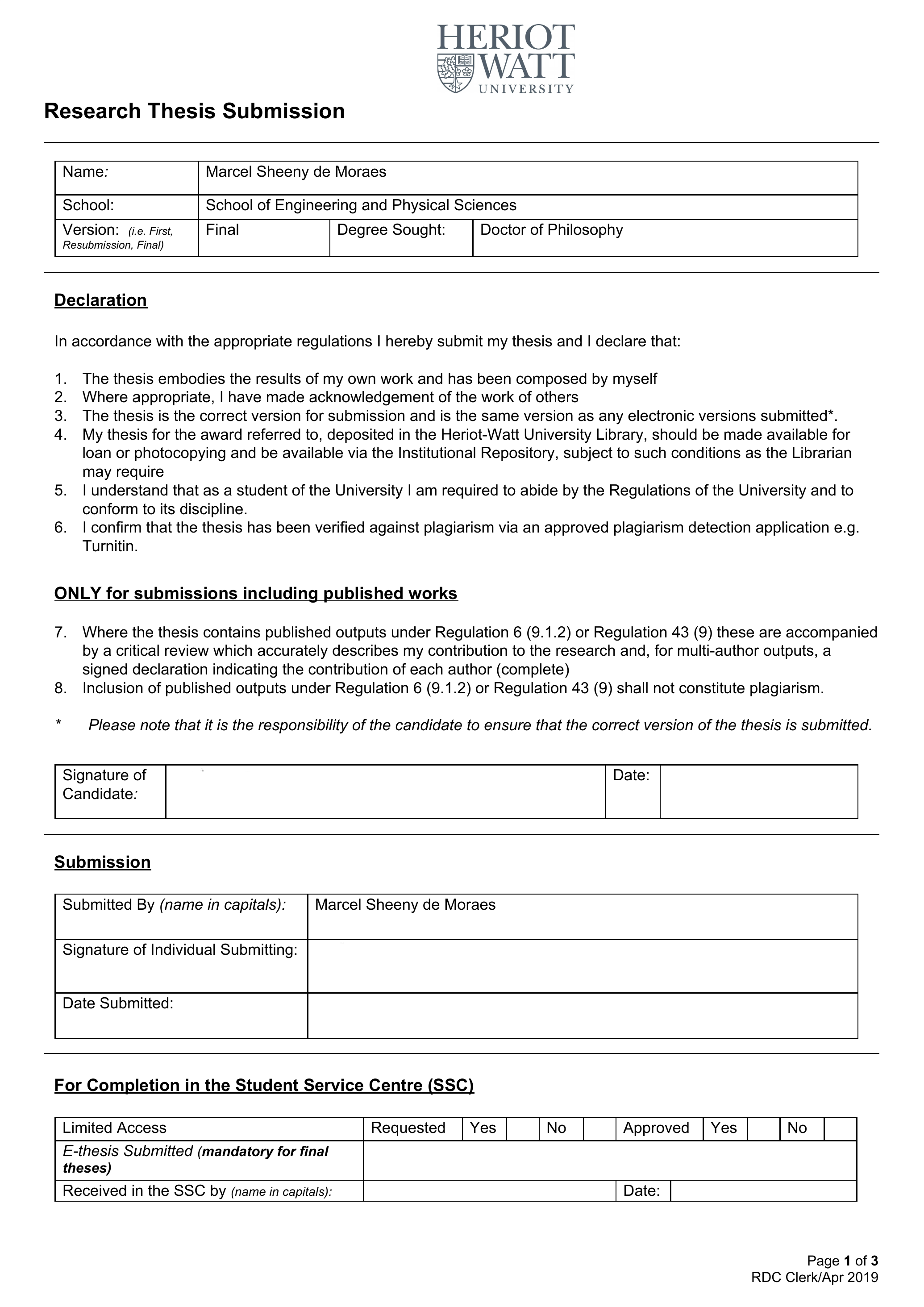}
\ifx\submissiontype\firstsub%
\mbox{}
\fi%
\fi%

\clearpage
\setcounter{page}{1}
\pagenumbering{roman}

\tableofcontents
\listoftables %
\listoffigures %

\clearpage
\pagenumbering{arabic}
\input{Chapters/Chapter1_Introduction.tex}

\input{Chapters/Chapter2_background.tex}

\input{Chapters/Chapter3_infrared.tex}

\input{Chapters/Chapter5_300ghz_DL.tex}

\input{Chapters/Chapter6_300ghz_DA_mdpi_version.tex}
\input{Chapters/Chapter7_navtech_data_bmvc.tex}

\input{Chapters/Chapter8_navtech_results_bmvc.tex}

\input{Chapters/Chapter9_conclusions.tex}

\appendix 
\input{Chapters/Appendix_150ghz.tex}

\lhead{}

\bibliographystyle{ieeetr}
\bibliography{bibliography}  %

\clearpage

\end{document}

%% file: Chapters/Abstract.tex
Autonomous cars are an emergent technology which has the capacity to change human lives. The current sensor systems which are most capable of perception are based on optical sensors. For example, deep neural networks show outstanding results in recognising objects when used to process data from cameras and Light Detection And Ranging (LiDAR) sensors. However these sensors perform poorly under adverse weather conditions such as rain, fog, and snow due to the sensor wavelengths. This thesis explores new sensing developments based on long wave polarised infrared (IR) imagery and imaging radar to recognise objects. First, we developed a methodology based on Stokes parameters using polarised infrared data to recognise vehicles using deep neural networks. Second, we explored the potential of using only the power spectrum captured by low-THz radar sensors to perform object recognition in a controlled scenario. This latter work is based on a data-driven approach together with the development of a data augmentation method based on attenuation, range and speckle noise. Last, we created a new large-scale dataset in the "wild" with many different weather scenarios (sunny, overcast, night, fog, rain and snow) showing radar robustness to detect vehicles in adverse weather. High resolution radar and polarised IR imagery, combined with a deep learning approach, are shown as a potential alternative to current automotive sensing systems based on visible spectrum optical technology as they are more robust in severe weather and adverse light conditions.

%% file: Chapters/Acknowledgements.tex
Firstly, I would like to thank my main supervisor Professor Andrew Wallace. During my PhD journey he helped me to develop critical thinking and inspired new ideas. Because of his support, I have gained a great understanding of research and the subject area. He helped to develop my writing skills and always helped to edit and improve my manuscripts. He was always happy to meet up and discuss ideas, address issues, and straighten my research path - all of which greatly contributed in developing the content of this thesis.

I also would like to thank my second supervisor Dr. Sen Wang. He always supported and motivated me throughout this journey. His very high standard of research helped me to work very hard and helped the development of my PhD.

I would like to thank Mehryar Emambakhsh and Alireza Ahrabian, the post-docs from the computer vision lab, who always helped me, gave me excellent advice and made the lab a very pleasant place to work in.

I also would like to thank Emanuele De Pellegrin and Georgios Kalokyris, who assisted by labelling and developing the website for the large scale dataset developed for this thesis.

I also would like to thank Marina Gashinova, Liam Daniel and Dominic Phippen from the University of Birmingham. They helped to collect data using their new sensor which was a huge contribution to this thesis.

I also would like to thank Bernie Mulgrew, Shahzad Gishkori and David Wright from the University of Edinburgh. They  assisted me with their expertise in signal processing and provided the code for phase correction.

A thanks to Nvidia, which donated a GPU and thus enabled me to train deep neural networks that were developed during this thesis.

This work was supported by Jaguar Land Rover and the UK Engineering and Physical Research Council, grant reference EP/N012402/1 (TASCC: Pervasive low-TeraHz and Video Sensing for Car Autonomy and Driver Assistance (PATH CAD)).

I would like to thank my family and especially my mom Nadra Sheeny de Moraes, who always dedicated her life to help and support me in all of my decisions. 

Lastly, I would like to thank the love of my life, my partner, my fiancée Gabriele Janušonyte who always supported me and also helped to proofread this thesis. Aš tave myliu! \ensuremath\heartsuit

This thesis is a work of many sleepless nights and my passion about the topic of computer vision. I always dreamed about contributing to this scientific community and I hope this work can contribute with further development in the field.

%% file: Chapters/Chapter1_Introduction.tex
\chapter{Introduction} %
\label{chapter:introduction} %

\lhead{Chapter 1. \emph{Introduction}} %

In recent years, many car manufacturers were promising to release fully autonomous cars by 2021 \cite{Ford, Volvo, Nissan, BMW}. Companies like Tesla, Uber, Ford, Volvo, Nissan and BMW are investing millions of dollars to develop new sensors and artificial intelligence technologies to release a fully autonomous car. But are we ready to release a fully autonomous vehicle? Most self driving cars use cameras and LiDAR systems (Light Detection and Ranging, also stylised as LIDAR) and these sensors are  vulnerable to adverse weather conditions \cite{korean}. During the Future Automobile Technology Competition in South Korea in 2014 \cite{korean} twelve autonomous cars teams needed to complete a series of tasks in an urban scenario. The first day had good weather and it was sunny and the four teams completed the tasks proposed. However on the second day it was raining, and the road was wet and slippery, and two of the successful teams from the first day crashed. This competition showed that we are still not ready to release fully autonomous cars on public roads in bad weather. Adverse weather is a challenging condition that needs to be addressed when developing autonomous vehicles.

 This project aims to develop an autonomous car perception system combining low-THz radar, video and LiDAR for adverse weather scenarios. This thesis was developed as part of the Pervasive low-TeraHz and Video Sensing for Car Autonomy and Driver Assistance project (PATH CAD), which is a collaboration between Heriot-Watt University, the Universities of Birmingham and Edinburgh and Jaguar Land Rover.  The University of Birmingham was responsible for developing novel low-THz radars (0.15 THz and 0.3 THz), The University of Edinburgh was responsible for developing signal processing algorithms to improve the radar image resolution and Heriot-Watt University developed sensor fusion and scene interpretation algorithms. This thesis, in particular, developed novel object recognition systems based on polarised infrared and imaging radar. 

Infrared is an alternative sensor for day and night perception \cite{dickson2015long, dickson2013improving, abbott2019multi}. Objects emit heat radiation based on their temperature. This radiation is invisible to the human eye and this radiation is at infrared wavelengths. Infrared images can sense the temperature of objects independent of an external light source. Polarised infrared is used to retrieve extra information such as the material refractive index, surface orientation and angle of observation, supplementing standard infrared sensors. This thesis explored machine learning on the basis of the Stokes parameters computed from polarised IR imagery. 

Radar (Radio Detection and Ranging, also stylised as RADAR) sensors are known for penetrating fog, rain and snow \cite{skolnik2001introduction}. Developing reliable methods that rely on radar will lead to a safer perception system that will help to achieve full autonomy in all-weather scenarios. Radar produces lower resolution images when compared to video and LiDAR. Designing an object recognition system for bad weather is a challenging problem. Low-THz radar is being designed to improve the resolution, with the intention to be able to improve scene interpretation. These low-THz sensors can offer bandwidth up to 20 GHz, which means a 0.75 cm range resolution. Scanning 79GHz imaging radars can provide a better azimuth resolution over \textit{Multiple Input Multiple Output (MIMO)} radars. In this thesis we used 79 GHz, 150 GHz and 300 GHz scanning radars to recognise objects without Doppler or temporal information, and we also showed its robustness in adverse weather conditions.

\section{Problem definition}

\begin{figure}[H]
\begin{center}
\begin{subfigure}{0.4\textwidth}
  \centering
  \includegraphics[width=\linewidth]{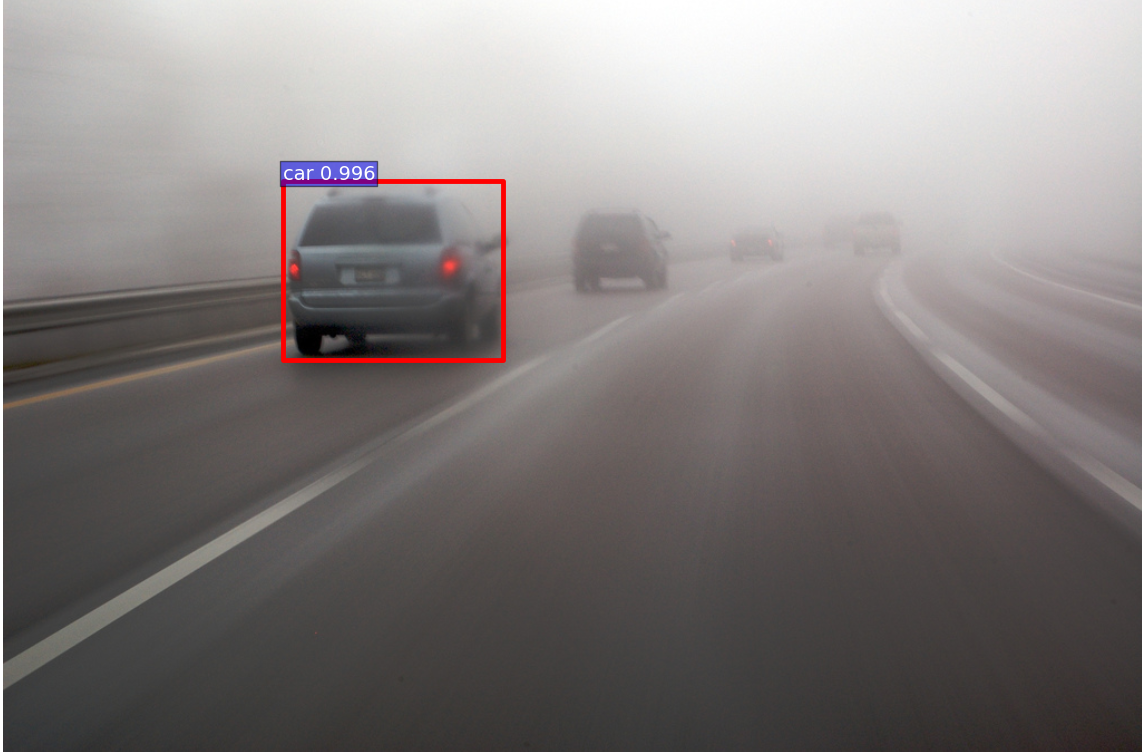}
  \caption{}
\end{subfigure}
\begin{subfigure}{0.4\textwidth}
  \centering
  \includegraphics[width=\linewidth]{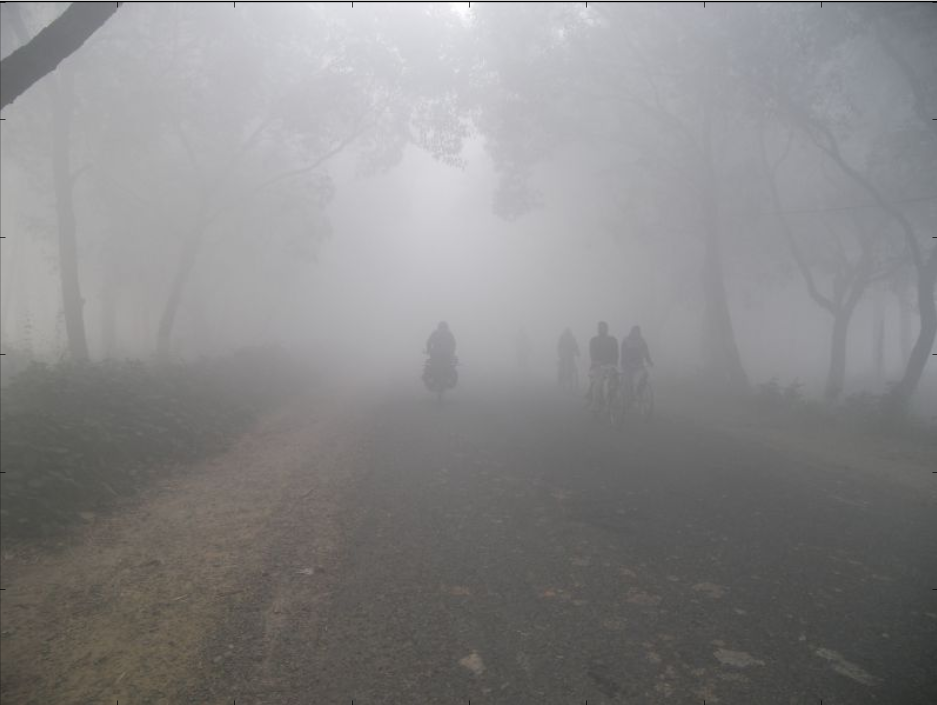}
  \caption{}
\end{subfigure}

\begin{subfigure}{0.4\textwidth}
  \centering
  \includegraphics[width=\linewidth]{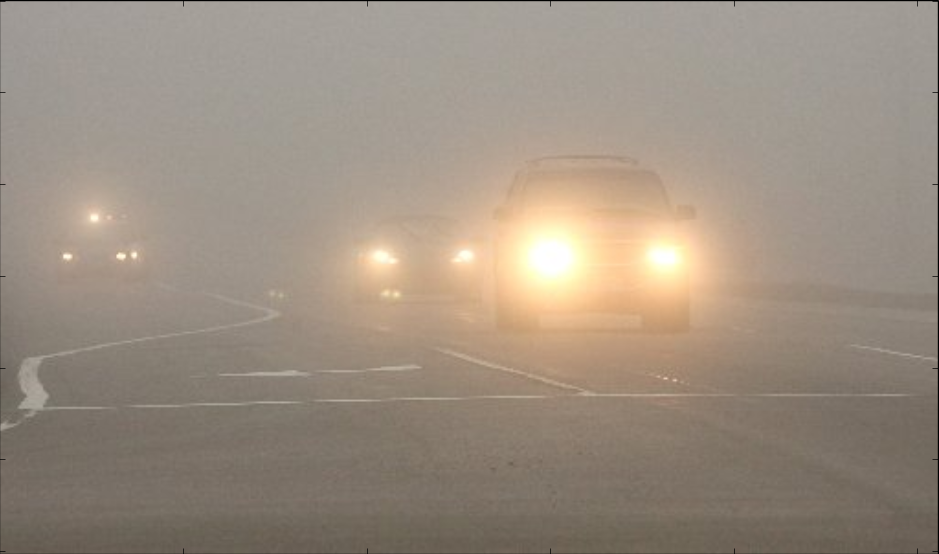}
  \caption{}
\end{subfigure}
\begin{subfigure}{0.4\textwidth}
  \centering
  \includegraphics[width=\linewidth]{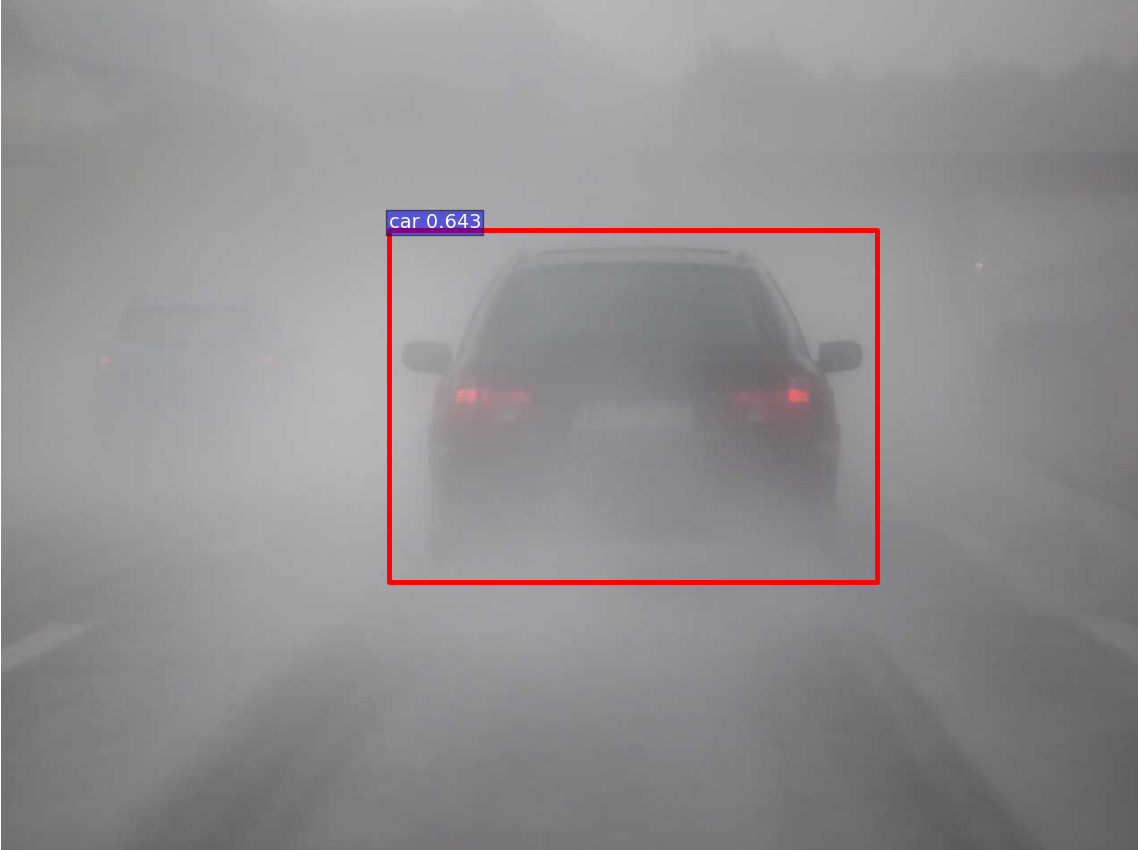}
  \caption{}
\end{subfigure}

\begin{subfigure}{0.4\textwidth}
  \centering
  \includegraphics[width=\linewidth]{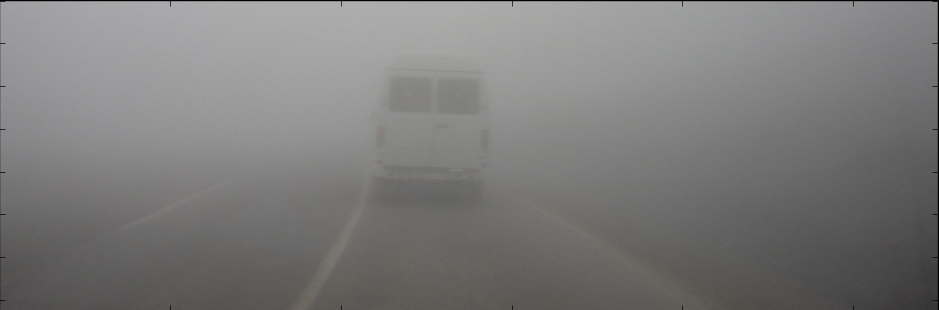}
  \caption{}
  \label{fig:sfig1}
\end{subfigure}
\begin{subfigure}{0.4\textwidth}
  \centering
  \includegraphics[width=\linewidth]{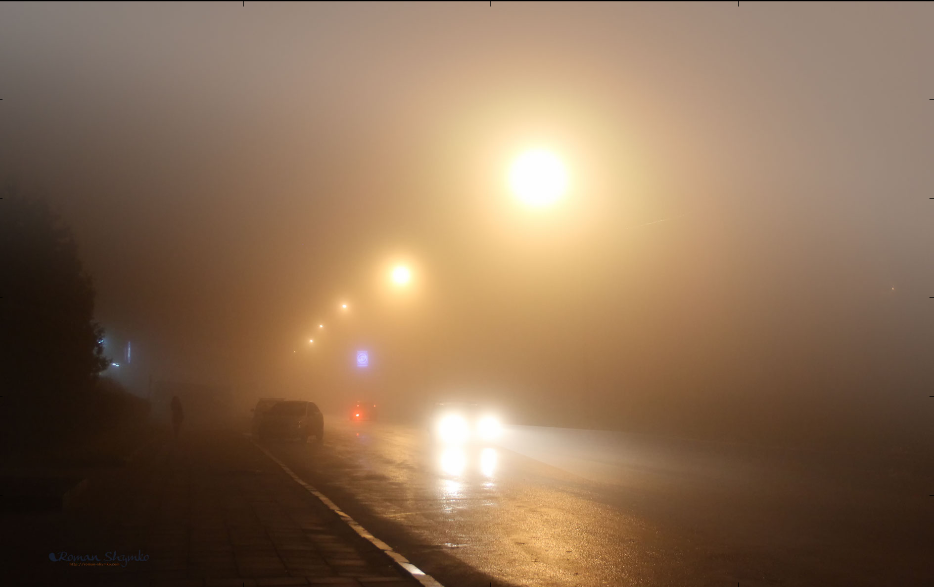}
  \caption{}
  \label{fig:sfig2}
\end{subfigure}

\begin{subfigure}{0.4\textwidth}
  \centering
  \includegraphics[width=\linewidth]{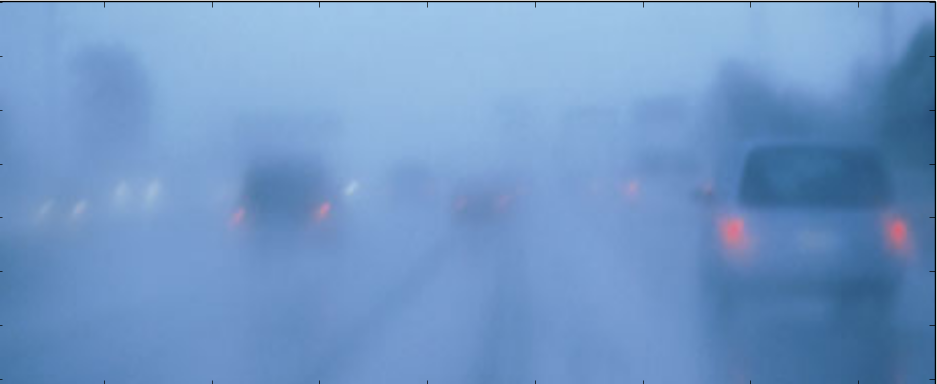}
  \caption{}
\end{subfigure}
\begin{subfigure}{0.4\textwidth}
  \centering
  \includegraphics[width=\linewidth]{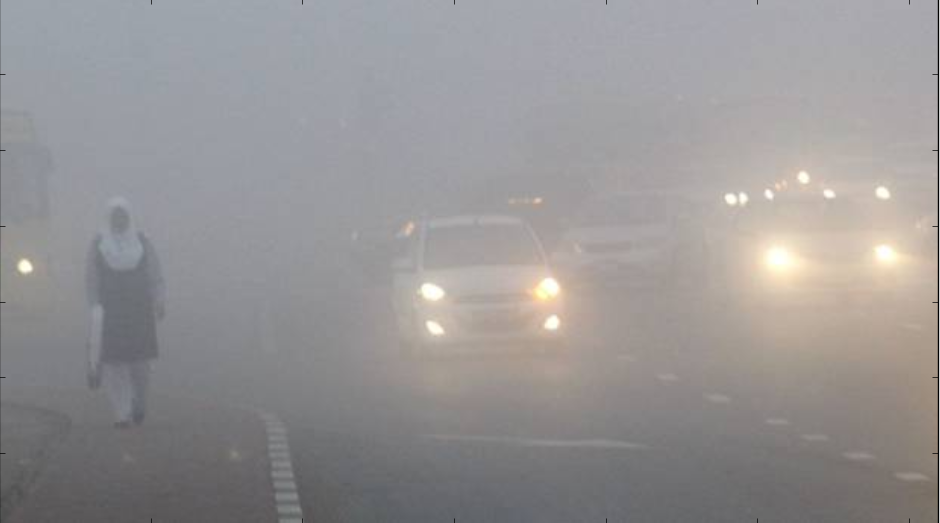}
  \caption{}
\end{subfigure}
\caption{Faster R-CNN algorithm applied on several images in bad weather. This network was trained on MS-COCO \cite{mscoco}. This figure shows that camera sensor fails to recognise objects under adverse weather conditions.}
\label{fig:faster_rcnn}
\end{center}
\end{figure}

The main sensors used in most autonomous cars systems are video and LiDAR. Can we use video and LiDAR in bad weather? Figure \ref{fig:faster_rcnn} shows the state-of-the-art object detection algorithm (Faster R-CNN \cite{ren2015faster}) trained on MS-COCO \cite{mscoco} applied to several foggy road environments. As we can see, in just two images was the network capable of detecting vehicles - it failed to recognize vehicles or pedestrians in the other images. A perception system based on video would fail to detect important objects in bad weather due to signal attenuation and alternative sensing, such as radar, is a key aspect to tackle the adverse weather problem. The problem addressed in this thesis is object recognition for all weather scenarios (sunny, night, outcast, fog, rain and snow) based on infrared and radar sensing.

\clearpage

Even though radar sensors are capable of penetrating rain, fog and snow, they provide relatively poor spatial resolution (especially in cross-range). High resolution radar systems such as Synthetic Aperture Radar (SAR) and Inverse Synthetic Aperture Radar (ISAR) need lateral movements for large aperture simulation. The implementation of SAR techniques on forward-looking radar is still a topic of research. Even with the development of forward looking SAR in automotive radar, in real-time automotive applications, we usually have to recognise vehicles the first moment they are sensed.

\begin{figure}[ht]
     \centering
     \begin{subfigure}[b]{1.0\textwidth}
         \centering
         \includegraphics[width=\textwidth]{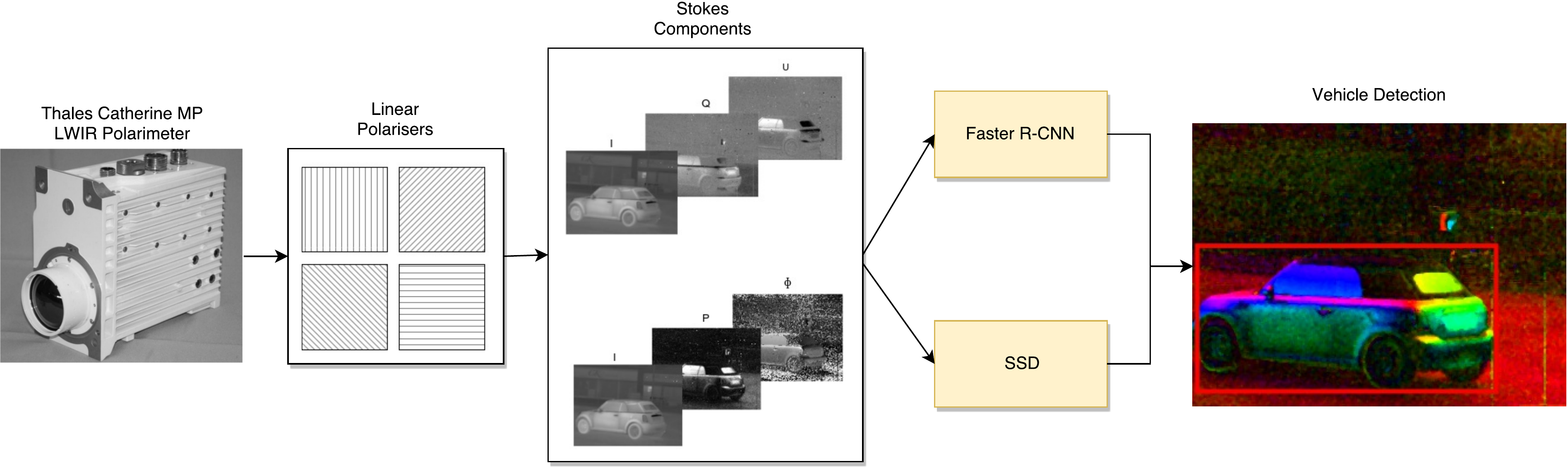}
         \caption{POL-LWIR Vehicle Detection: Convolutional Neural Networks Meet Polarised Infrared.}
     \end{subfigure}

     \begin{subfigure}[b]{0.45\textwidth}
         \centering
         \includegraphics[width=\textwidth]{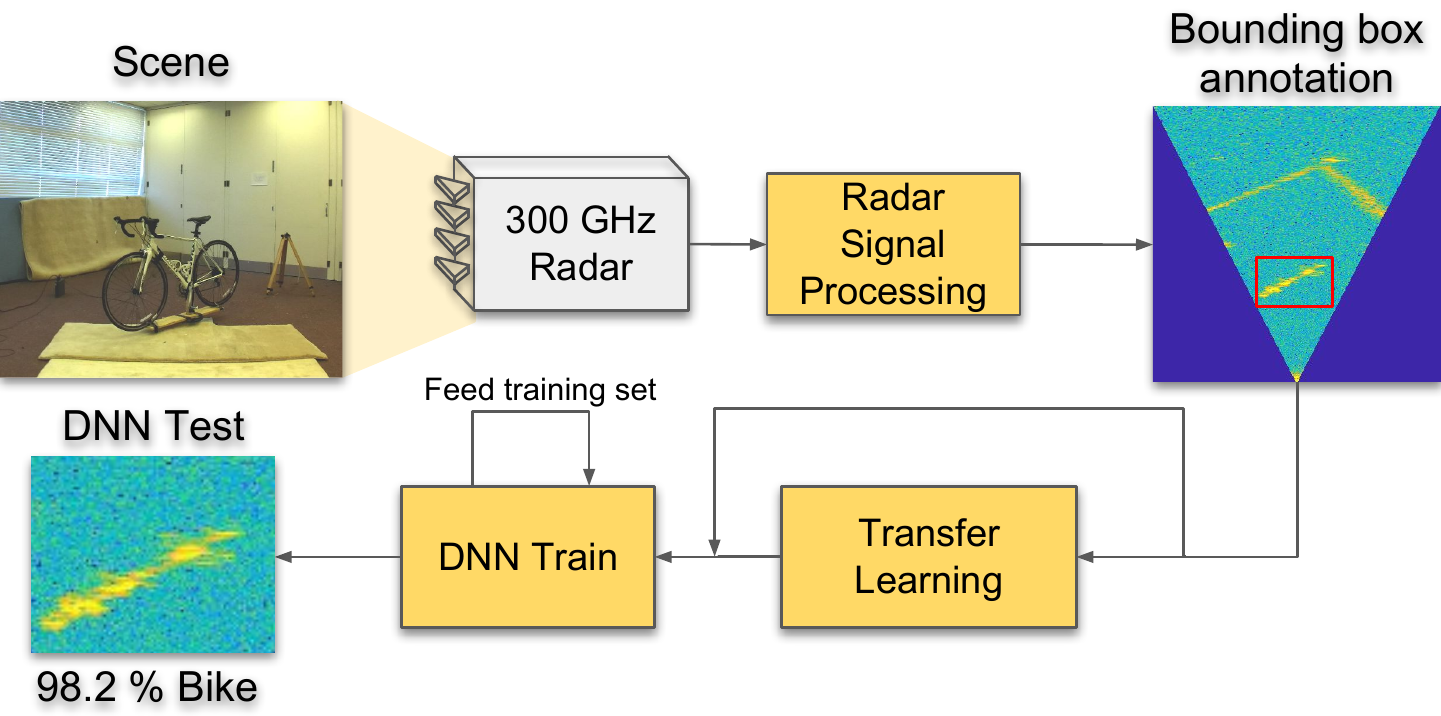}
         \caption{300 GHz FMCW Radar Object Recognition based on Deep Neural Networks and Transfer Learning}
     \end{subfigure}
    \hfill
     \begin{subfigure}[b]{0.45\textwidth}
         \centering
         \includegraphics[width=\textwidth]{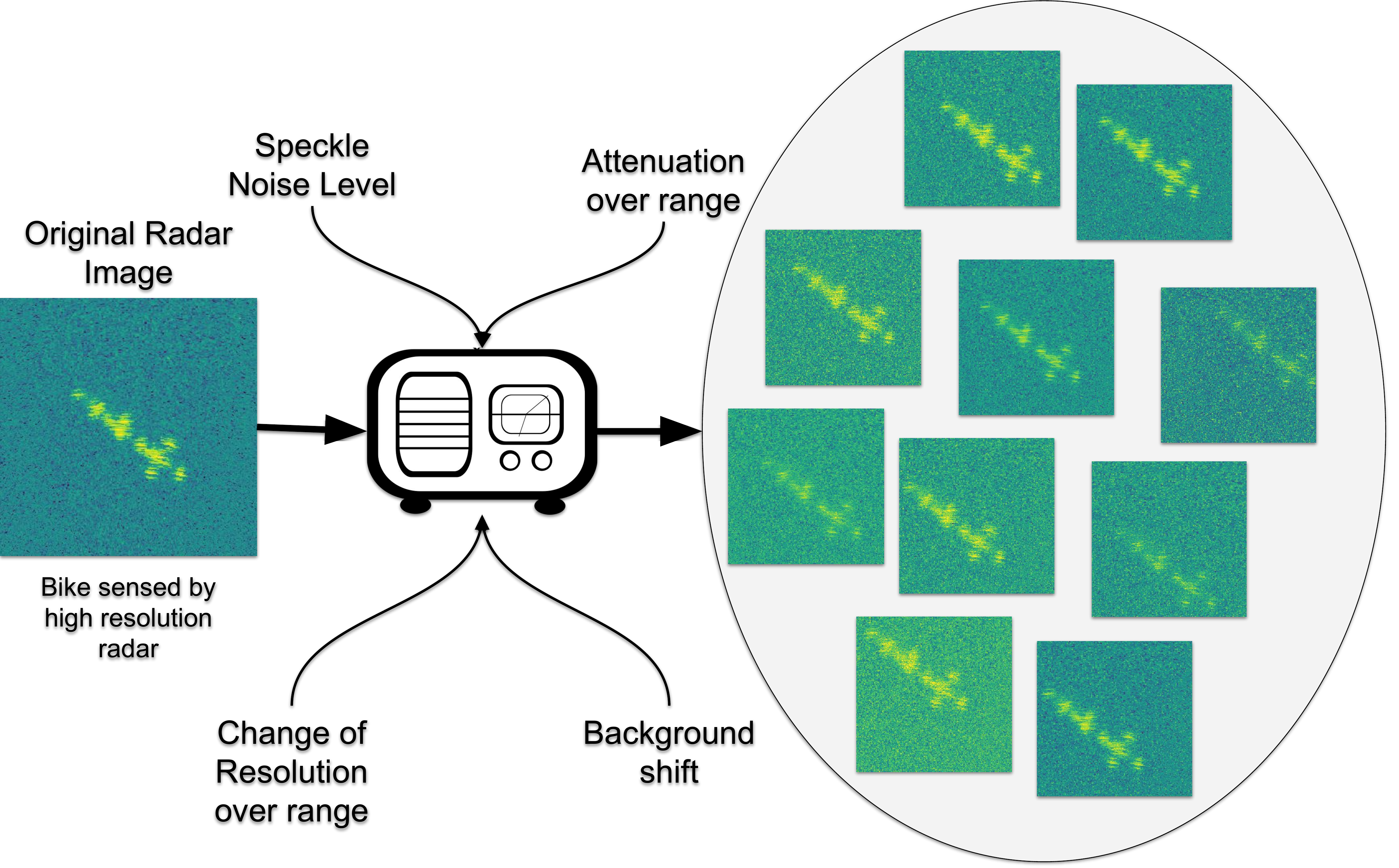}
         \caption{RADIO: RAdar Data augmentIOn.}
     \end{subfigure}
     
     \begin{subfigure}[b]{0.8\textwidth}
         \centering
         \includegraphics[width=\textwidth]{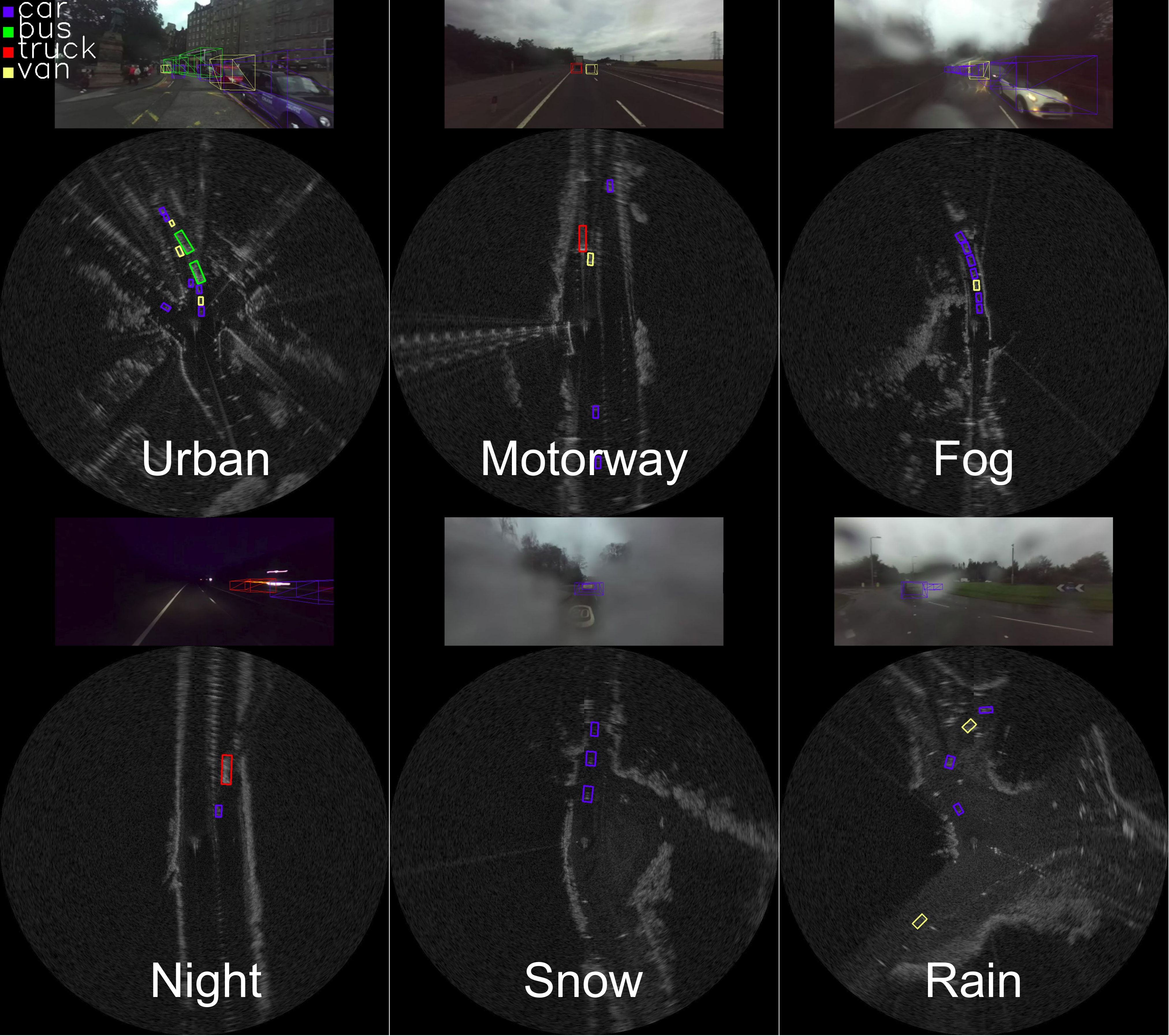}
         \caption{RADIATE: RAdar Dataset In Adverse weaThEr.}
     \end{subfigure}
        \caption{Pictorial representation of the methods developed in this thesis}
        \label{fig:thesis}
\end{figure}

The main objective of this thesis is to develop a perception system that detects and recognizes objects for automotive applications using infrared images, robust during the day and night, and radar images, robust in all-weather scenarios. Figure \ref{fig:thesis} shows illustrative images of the methods developed for this thesis.

\section{Nomenclature}

For this thesis, we detect and classify different objects. However, in the computer vision and radar communities the term \textit{detection} means different things. In the radar community, \textit{detection} means only detecting regions without classifying them. Constant False Alarm Rate (CFAR) \cite{skolnik2001introduction} is an example of a detection algorithm. In the computer vision community detection means localising one or more regions (usually rectangular boxes) and classifying each region. Faster R-CNN \cite{ren2015faster} is an example of a detection algorithm for computer vision. For this thesis we used:

\begin{itemize}
    \item \textbf{Classification}: We use the term \textit{classification} when the whole image is classified, without any localisation where the object is. AlexNet \cite{krizhevsky2012imagenet} is an example of image classification method. The main metric used for classification is accuracy.
    \item \textbf{Detection}: We use \textit{detection} by localising potential regions and classifying them (Faster R-CNN \cite{ren2015faster}, RetinaNet \cite{lin2017focal}, YOLO \cite{redmonyolo2016}). The main metric used for classification is Average Precision (AP) \cite{everingham2010pascal}.
    \item \textbf{Recognition}: The term \textit{recognition} is used in a more generic context, when either classification occurs in a detection context or not.

\end{itemize}

\section{Contributions}

The thesis developed new techniques of object recognition for both polarised infrared and imaging radar sensors. Most similar research either concentrates on the sensing \cite{jasteh2015low, jastehthesis, patole2017automotive} or machine learning \cite{schumann2018semantic, Dong_2020_CVPR_Workshops, Major_2019_ICCV, Sless_2019_ICCV} aspects with very little overlap. This thesis tries to fill the gap between sensing, machine learning and robotics. The main contributions are listed below:

\subsubsection*{Polarised Infrared Vehicle Detection using CNNs}

A novel methodology was developed for polarised infrared vehicle detection based on Stokes parameters and deep neural networks. We employed two different image decompositions: the first based on the polarisation ellipse, and the second on the Stokes parameters themselves. We used these two approaches as input to object detection based on neural networks. We showed that polarised infrared using the Stokes parameters methodology developed can achieve a better vehicle recognition over standard infrared sensors (Chapter \ref{chapter:pol_lwir}).

\subsubsection*{300 GHz Radar Object Recognition Based on Deep Neural Networks and Transfer Learning}

Most object recognition methods using radar rely on Doppler and temporal features.
However, as both the ego-vehicle and the target vehicles may be stationary, this is not always possible. An object recognition system using a 300 GHz radar using only the power spectrum based on deep neural networks was developed. Since we have a small dataset, we also worked by transferring knowledge from a bigger \textit{Synthetic Aperture Radar} (SAR) dataset to improve its results (Chapter \ref{ch:300dl}).

\subsubsection*{300 GHz Radar Object Recognition Based on Effective Data Augmentation}
Another strategy to avoid overfitting and deal with small datasets is to use data augmentation. A novel data augmentation technique for low-THz radar with small datasets was developed. The data augmentation developed leverages physical properties of radar signals, such as attenuation, azimuthal beam divergence and speckle noise, for data generation and augmentation. We showed that the method developed achieves better results when compared to conventional deep learning and transfer learning  (Chapter \ref{ch:radio}).

\subsubsection*{Public Multi-Modal Dataset in Adverse Weather}

We contributed substantially to the establishment of a novel public multi-modal dataset including radar, LiDAR and stereo camera data captured in fair and adverse weather (night, fog, rain and snow). We called this dataset RADIATE (RAdar Dataset In Adverse weaThEr), this is the first public dataset that includes relatively high resolution imaging radar data in adverse weather. Several sequences have also been labelled with the constituent actors (cars, trucks etc.) to aid with object recognition experiments. We hope this dataset will help the community in research into object detection, tracking and domain adaptation using heterogeneous sensors (Chapter \ref{ch:navtech}). A website\footnote{\url{http://pro.hw.ac.uk/radiate/}} for the RADIATE dataset was developed. A Software Development Kit (SDK) to use RADIATE was developed\footnote{\url{https://github.com/marcelsheeny/radiate_sdk}}. Also a documentation on how to use RADIATE SDK\footnote{\url{https://marcelsheeny.github.io/radiate_sdk/radiate.html}}.

\subsubsection*{79 GHz Radar Vehicle Detection in Adverse Weather}

For the lower frequency, a vehicle detection system only using the radar data, again without use of Doppler and temporal information, was developed. A comparison between radar robustness depending on the weather conditions was done. We showed that the radar image interpretation was relatively less influenced by adverse weather, comparing quantitatively in good \textit{vs.} bad weather for vehicle detection. And also qualitatively when compared empirically to the video and LiDAR data (Chapter \ref{ch:navtech2}). The code developed for vehicle detection in this chapter is publicly available\footnote{\url{https://github.com/marcelsheeny/radiate_sdk/tree/master/vehicle_detection}}.

\clearpage

\section{Thesis Structure}

\begin{itemize}
    \item {\bf Chapter \ref{chapter:literature_review}:} This is a foundational literature review on sensing for autonomous cars, recognition in video, LiDAR, infrared and radar images. More detailed review of specific topics related to the several contributions can be found in the later chapters.
    \item {\bf Chapter \ref{chapter:pol_lwir}:} A methodology to recognise vehicles using polarised infrared based on Stokes parameters and deep neural networks is described.
        \item {\bf Chapter \ref{ch:300dl}:} A 300 GHz radar was used to collect image data for several objects in a laboratory setting. A data-driven approach using deep neural networks and transfer learning was developed to perform object recognition.
    \item {\bf Chapter 5:} A novel data augmentation technique based on attenuation, range and speckle is described.
    \item {\bf Chapter \ref{ch:navtech}:} A large-scale multi-modal (camera, radar and LiDAR) dataset "in the wild" is presented.
    \item {\bf Chapter \ref{ch:navtech2}:} Using only the radar data from this dataset, a recognition network for operation in adverse weather is described.
    \item {\bf Chapter \ref{ch:conclusions}:} Conclusions and potential future work are presented.
    \item {\bf Appendix \ref{chapter:150ghz}:} We have included a preliminary study of an object recognition system applied to prototypical 150 GHz radar based on hand-crafted features.
    
\end{itemize}

\clearpage

\section{List of Publications}

\begin{itemize}
    \item Marcel Sheeny, Andrew Wallace, Mehryar Emambakhsh, Sen Wang and Barry Connor. POL-LWIR Vehicle Detection: Convolutional Neural Networks Meet Polarised Infrared Sensors. In Proceedings of the IEEE Conference on Computer Vision and Pattern Recognition Workshops 2018 (pp. 1247-1253), Salt Lake City, USA.

    \item Marcel Sheeny, Andrew Wallace and Sen Wang. 300 GHz Radar Object Recognition based on Deep Neural Networks and Transfer Learning. IET Radar, Sonar and Navigation, 2020, DOI: 10.1049/iet-rsn.2019.0601.
    
    \item Marcel Sheeny, Andrew Wallace and Sen Wang. RADIO: Parameterized Generative Radar Data Augmentation for Small Datasets. Applied Sciences 2020, 10, 3861. 
    
    \item Marcel Sheeny, Emanuele De Pellegrin, Saptarshi Mukherjee, Alireza Ahrabian, Andrew Wallace and Sen Wang. RADIATE: A Radar Dataset for Automotive Perception.  (Under Review).

\end{itemize}

%% file: Chapters/Chapter2_background.tex
\chapter{Background and Literature Review} %
\label{chapter:literature_review} %

\lhead{Chapter 2. \emph{Literature Review}} %

    This thesis develops an object recognition perception system for autonomous cars in all weather scenarios. This is a crucial task in developing this intelligent system. Knowing the object that we are detecting, we can predict where it is going in order to create a safe perception system. This chapter will give a brief explanation on the theoretical background used for this thesis (Section \ref{sec:background}) and also review the literature of object recognition using different sensors (Section \ref{sec:litrev}).

    \section{Background}\label{sec:background}
    
    In this part of this chapter, we will give a brief explanation about the main technologies and mathematical tools used in this thesis. Section \ref{sec:electro_lr} gives a brief introduction on electromagnetic waves used for sensing and the different wavelengths. Section \ref{sec:dnn_lr} will show the mathematical background of deep neural networks, which is the main tool used for this thesis.  Section \ref{sec:ac} will review the history of autonomous cars and the state of current development.
    
    \subsection{Electromagnetic Sensing}\label{sec:electro_lr}
    
        \begin{figure}[t]
            \centering
            \includegraphics[width=1.0\textwidth]{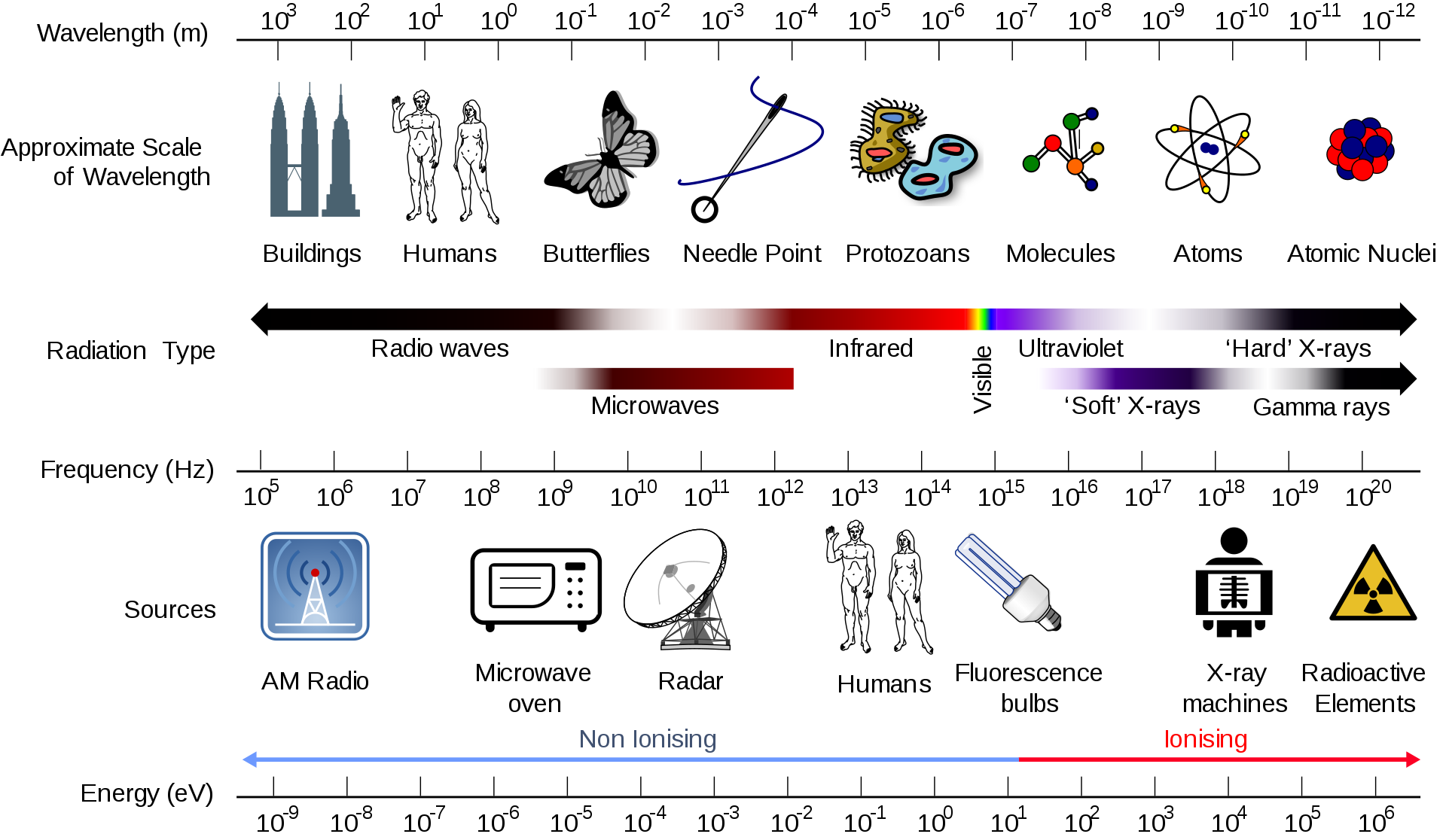}
            \caption{ Electromagnetic Spectrum \cite{electro_fig}.}
            \label{fig:electro_fig}
        \end{figure}
    
    Sensing using electromagnetic (EM) waves is the main action of physical phenomena used for sensing in self-driving cars. LiDARs, stereo cameras, infrared cameras and radars are examples of sensors based on EM waves. Infrared and stereo cameras are passive sensors - it means that they receive EM waves naturally emitted by the environment. LiDARs and radars are active sensors - it means that they transmit EM into the environment and receive it back to sense their surroundings.
    
    Electromagnetic waves are a natural phenomena which occurs by the excitement of electrically charged particles. This was modelled mathematically by the Scottish physicist James Maxwell \cite{Jackson:100964}. According to contemporary physics, electromagnetic waves are particles without mass - also known as photons.
    
    Electromagnetic waves have a very wide spectrum (Figure \ref{fig:electro_fig}). EM travel at the speed of light in vacuum, so a range of uses are developed especially for long range applications. The frequency is related to the wavelength - Equation \ref{eq:freq_lr} shows its relation.
    
    \begin{equation}\label{eq:freq_lr}
        f = \frac{c}{\lambda}
    \end{equation}
    
    \noindent In this Equation, $c$ is the speed of light constant (299,792,458 m/s) and $\lambda$ is the wavelength in meters.
    
    We can define the wave mathematically as a function of time and frequency,
    
    \begin{equation}\label{eq:signal}
        y(t) = A \sin (2 \pi f t + \theta)
    \end{equation}
    
    \noindent where $A$ is the amplitude, $f$ is the frequency, $t$ is the time and $\theta$ is the phase.
    
    The frequencies captured by the human eye are between $4 \times 10^{14}$ Hz to $8 \times 10^{14}$ Hz or 380 to 740 nanometers. The human retina is sensitive to those frequencies, which is on the visible spectrum. The visible spectrum is the wavelength used for camera and LiDAR sensors. 
    
    Camera sensors have very similar responses to the human vision, being widely used as a perception system.   LiDAR, on the other hand, is an active sensor, providing a high resolution detection map.
    
   However, there is a wide spectrum of frequencies that are not captured by human eyes. For this thesis, we use radio and infrared sensing to recognise objects.
    
    Infrared frequencies are in the range of $1.9 \times 10^{13}$ Hz to $1.2\times 10^{14}$ Hz. These frequencies are quite useful for military applications, especially night vision. Any object whose temperature is higher than 0 K, emits radiation. This emission can be used to measure temperature.
    
    Radio frequencies are in the range of 30 Hz to 300 GHz. Radio waves have a wide range of applications, being mostly used for communication purposes. Radar uses radio waves to measure range.

    \subsubsection{Camera Sensing}
    
    Camera sensor is probably one of the most used sensors for robotic perception. Since it uses the visible spectrum for sensing, it generates images similar to the human vision. Camera sensors are relatively cheap ( $\approx$ £200), compared to LiDAR ($\approx$ £15,000), infrared ($\approx$ £10,000). By being a cheap sensor, it allows to be widely researched. Nowadays, everyone has a camera in their phone, so it is a popular sensor which can easily collect huge amounts of data.
    
    \begin{figure}[h]
    \centering
    \begin{minipage}{.5\textwidth}
      \centering
               \includegraphics[width=\textwidth]{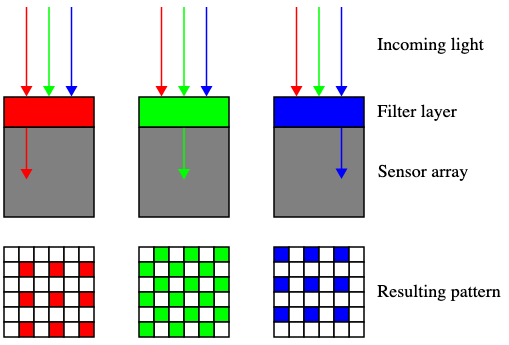}
                \caption{ Bayer Filter \cite{wiki_image_sensor}.}
                \label{fig:bayer_filter}
    \end{minipage}%
    \begin{minipage}{.5\textwidth}
      \centering
      \includegraphics[width=\textwidth]{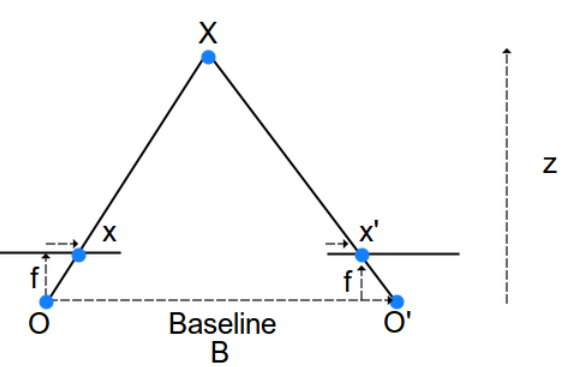}
                \caption{Disparity from 2 cameras. Where O and O' are the two cameras.}
                \label{fig:disparity}
    \end{minipage}
    \end{figure}
    
    \clearpage
    
    Cameras use a 2D array of sensors to capture light emitted by the environment. The sensor does it by converting light waves into electric signals, and this electric signal is used to generate the image.
    
    The main technology used by camera sensors are CCD (charge-coupled device) and CMOS (complementary  metal–oxide–semiconductor), with CMOS being the main technology used in most of current cameras \cite{daniel2014high}.
    
    In order to generate each coloured pixel of the camera image, the camera has a pattern of 2D array of CMOS sensors. In order to capture colour information, camera sensors use the Bayer filter (Figure \ref{fig:bayer_filter}). This is a low-cost way to capture RGB information using 1 red, 2 green and 1 blue sensor per pixel.

    Even being a passive sensor, camera can also capture depth information by using a stereo camera system. In order to capture depth information, we need to find the correspondent pixel in the other camera. Depending on the disparity measure, we can measure the distance from the sensor. This is illustrated in Figure \ref{fig:disparity}.

\begin{figure}[h]
            \centering
            \includegraphics[width=0.9\textwidth]{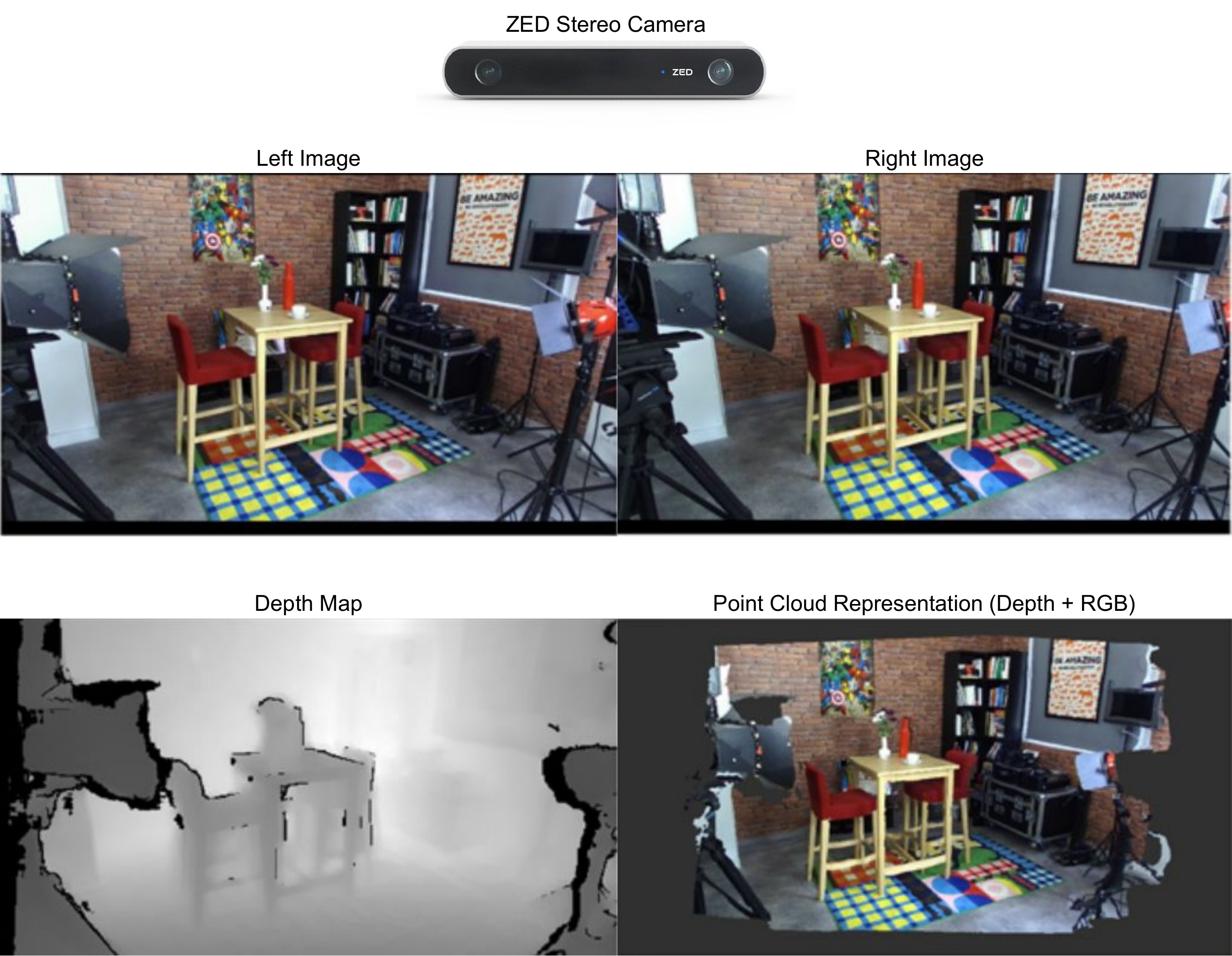}
            \caption{ Example of images captured by a ZED Camera.}
            \label{fig:zed_camera}
        \end{figure}

    In this thesis, we used a ZED stereo camera system. An example of information captured by the ZED camera can be seen in Figure \ref{fig:zed_camera}
    
    \subsubsection{LiDAR Sensing}
    
    LiDAR (light detection and ranging) is an active sensor which measures range using visible light wavelengths. LiDAR has been widely used in self-driving cars because it creates high resolution 3D maps, being the main sensor for localisation, mapping and collision avoidance.
    
    \begin{figure}[h]
        \centering
        \includegraphics[width=1.0\textwidth]{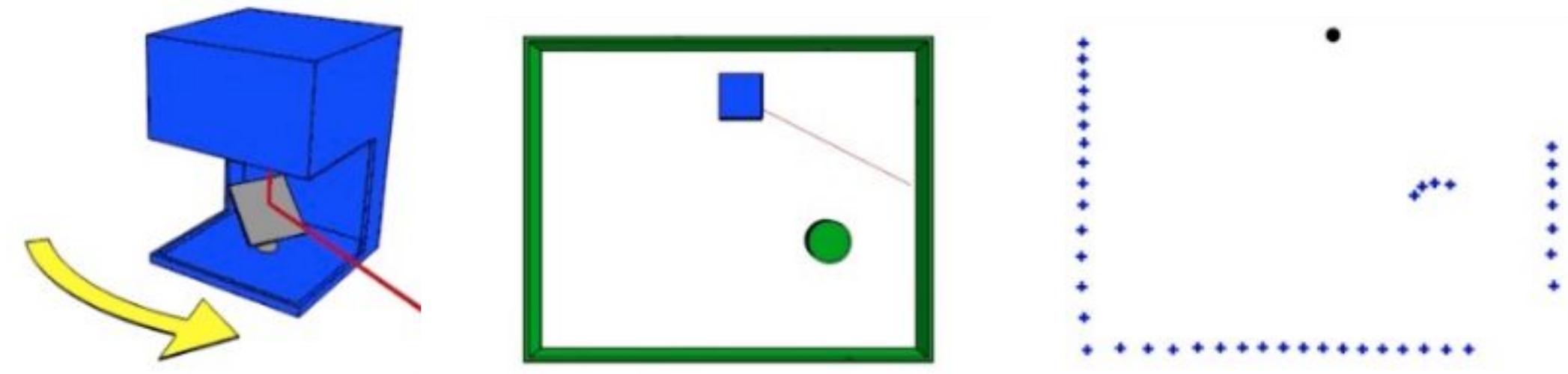}
        \caption{Example of LiDAR scanner \cite{wiki_lidar}.}
        \label{fig:lidar_scan}
    \end{figure}
    
    For robotic/self-driving car applications, the common wavelength used is 905 nm. This frequency is used because requires less energy and it is a eye-safe laser. The LiDAR technology is based on laser pulse and the time of flight is measured to retrieve the distance.
    
    A common approach to create a high resolution 3D point cloud with LiDAR is by using a mechanical scanner. A laser emitter is constantly transmitting the signal, while a mirror rotates to capture the scene. Figure \ref{fig:lidar_scan} illustrates how the scanner works where the right image is the lidar sensor with a mechanical scanner, the image in the center shows the scenario being sensed, and the image in the right shows the point cloud being formed by the mechanical lidar scanner.
    
    \begin{figure}[b]
            \centering
            \includegraphics[width=\textwidth]{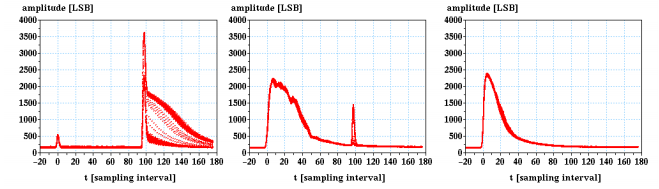}
            \caption{Echoes from the full waveform with different visibilities. No fog scenario is showed on the left, 40 meters visibility in the middle and 10 meters visibility in the right \cite{pfennigbauer2014online}.}
            \label{fig:riegl}
    \end{figure}

\begin{figure}[t]
            \centering
            \includegraphics[width=0.7\textwidth]{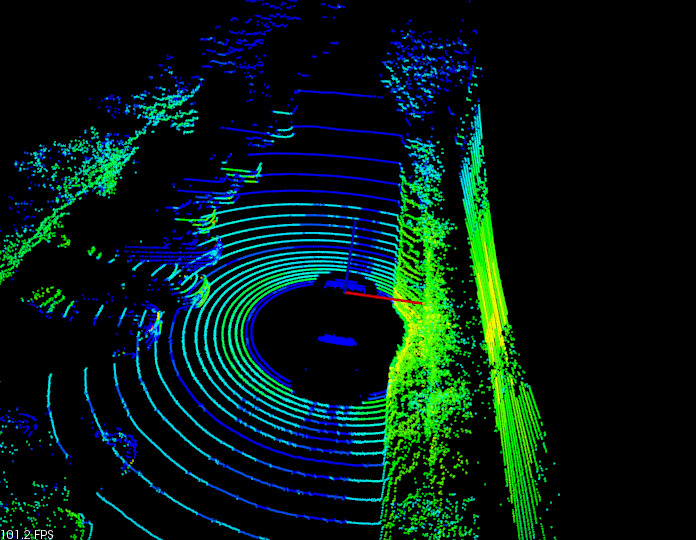}
            \caption{Example of point cloud given by LiDAR of a road scenario.}
            \label{fig:lidar_car}
        \end{figure}

     LiDAR produces a very small beamwidth due to its small wavelength, being capable of achieving very high range (2 cm) and azimuth ($0.1^o$) resolution. An example of LiDAR point cloud can be visualised in Figure \ref{fig:lidar_car}. This Figure shows a example of a road scene sensed by lidar, which can be used as a perception sensor.

     However, LiDAR cannot penetrate weather artifacts, such as rain, fog and snow; since it uses the visible spectrum, any weather artifact will be reflected by the laser beam. This can be a problem when trying to achieve a high level of autonomy in all weather scenarios.
     
     There is current research on how to remove the weather artifacts from LiDAR \cite{satat2018towards}, however we need to apply heavy signal processing and work with full waveform, which are normally not available in current commercial LiDARs.
     
     The paper by Pfennigbauer, \textit{et al} \cite{pfennigbauer2014online} (Figure \ref{fig:riegl}) shows the LiDAR capabilities in foggy scenarios. They conducted an experiment on a chamber with fog with different visibilities with an object at 100 meters. The experiment showed that the full waveform of the Riegl LiDAR has a strong unwanted return signal which represents the fog. However, it also returned the signal from objects as well with a measured 40 meters visibility (Figure \ref{fig:riegl}). As shown in the graph, LiDAR has potential of sensing in moderate foggy weather. By doing the correct full waveform processing, we can remove the fog effects from the image. With highly foggy scenarios, objects at long range cannot be detected.

    \subsubsection{Infrared Sensing}
    
    Infrared cameras (also called thermal cameras) capture wavelength in the infrared radiation spectrum. Common wavelengths used in infrared cameras are about 8000 nm to 12000 nm.
    
\begin{figure}[ht]
            \centering
            \includegraphics[width=0.8\textwidth]{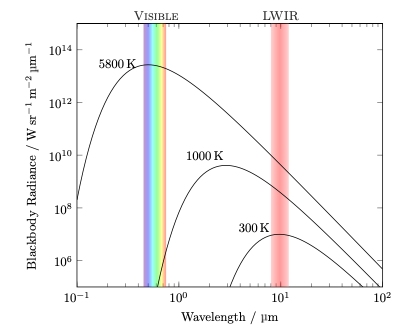}
            \caption{Amount radiance emitted vs. wavelength.}
            \label{fig:blackbody}
        \end{figure}

    Any object with a temperature of 0 K or above emits radiation \cite{hecht2002optics}. Figure \ref{fig:blackbody} shows the amount of radiance emitted in different wavelengths. As we observed, a blackbody at 300 K ($27^o $ C) has its peak of radiance emission in the long-wave infrared (LWIR) spectrum. The temperature of the object can be measured by the amount of infrared radiation that is captured by the sensor. 
    
        \begin{figure}[h]
            \centering
            \includegraphics[width=0.5\textwidth]{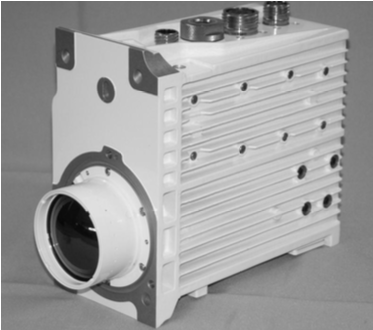}
            \caption{Catherine MP LWIR sensor.}
            \label{fig:catherine}
        \end{figure}
        
            \begin{figure}[h]
            \centering
            \includegraphics[width=\textwidth]{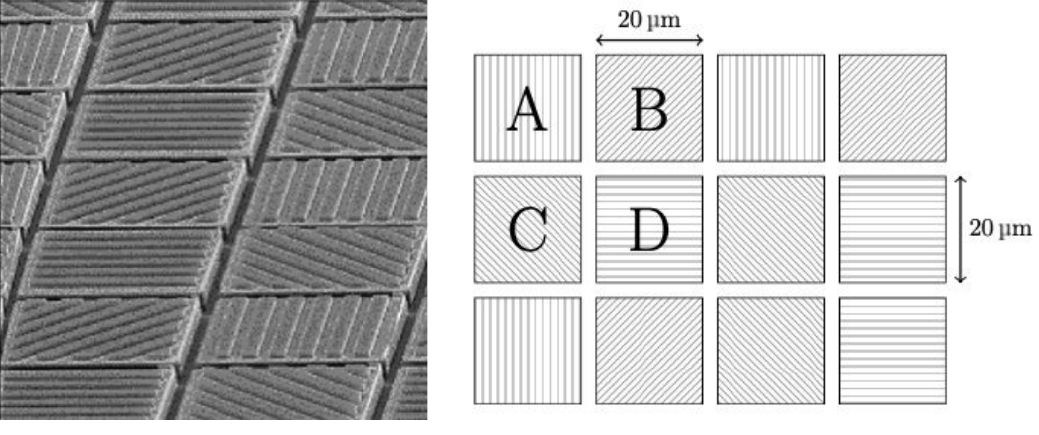}
            \caption{Image of the actual infrared physical sensor with 4 linear polarisers (left) and the diagram representation (right).}
            \label{fig:pola_caterine}
        \end{figure}
    
    The main application of infrared sensing is in environments that lack illumination, and especially for military applications. More recently, infrared has begun to be considered as an alternative sensor to self-driving cars \cite{dickson2013improving}. Since objects reflect different colours depending on illumination, infrared is an alternative sensor when tackling lack of illumination.
    
    More recently, polarisation is being used in infrared images \cite{dickson2015long}. The polarisation can capture better signatures which leads to a better recognition rate \cite{dickson2015long}. In this thesis, we used the Catherine MP LWIR sensor (Figure \ref{fig:catherine}) for vehicle detection using deep neural networks for vehicle detection (Chapter \ref{chapter:pol_lwir}). This sensor uses 4 linear polarisers (0$^o$,45$^o$,90$^o$,135$^o$). Figure \ref{fig:pola_caterine} shows the image of the 4 linear polarisers and the diagram representation.

    \subsubsection{Radar Sensing}
            
\begin{figure}[ht]
            \centering
            \includegraphics[width=0.7\textwidth]{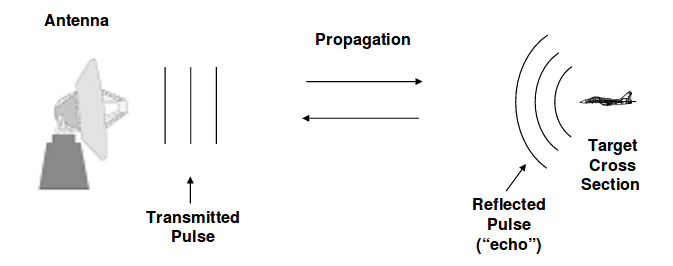}
            \caption{ Basic scheme from a radar system  \cite{skolnik2001introduction}.}
            \label{fig:radar_scheme}
        \end{figure}

        A basic radar system sends an electromagnetic pulse and detects the range of an object by measuring its time-of-flight ($r = \frac{c\tau}{2}$, where $r$ is the range in meters, $c$ is the speed of the light constant and $\tau$ is the round trip time). Figure \ref{fig:radar_scheme} shows the basic scheme of a radar system.
        
        The return power ($P_r$) from the radar is modeled physically by Equation \ref{eq:radar_equation} \cite{skolnik2001introduction}, where $P_t$ is the power transmitted, $G$ is the antenna gain,  $\sigma$ is the radar cross section, $\lambda$ is the radar wavelength, $L$ is the loss factor and $R$ is the distance between the sensor and the target. 
        
        \begin{equation}\label{eq:radar_equation}
            P_r = \frac{P_t G^2 \sigma \lambda^2}{(4 \pi)^3  R^4 L}
        \end{equation}

        The radar cross section means how detectable an object is. It is influenced by many factors \cite{skolnik2001introduction}: target material, target size, target shape, incident angle and transmitted polarization. 
        
        The return signal is also affected by attenuation. The signal loses its power over distance due to atmospheric influences.

        \subsubsection{FMCW Radar}
        
\begin{figure}[b]
            \centering
            \includegraphics[width=0.8\textwidth]{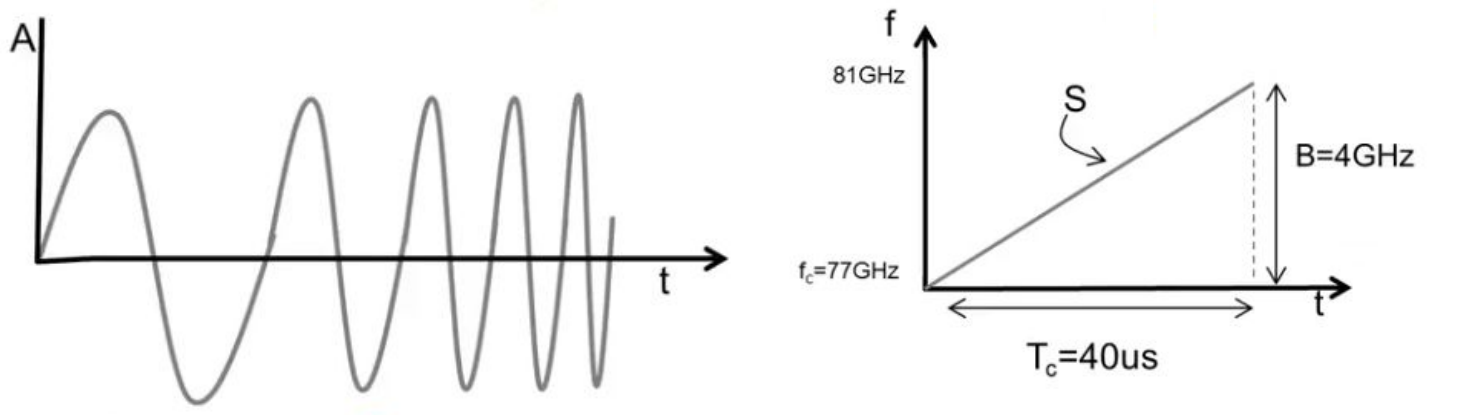}
            \caption{Example of chirp generated by a FMCW automotive radar.}
            \label{fig:chirp}
        \end{figure}

        Automotive radar technology uses a frequency modulated continuous wave (FMCW) radar. The main frequencies used on automotive radar are 24, 77 and 79 GHz with 4 GHz bandwidth \cite{tiradar}. In a FMCW system, the transmitter generates a waveform with a starting frequency $f_0$, bandwidth $B$  and duration $T_c$. An example of a chirp generated by the transmitter can be seen in Figure \ref{fig:chirp}
        
\begin{figure}[ht]
            \centering
            \includegraphics[width=0.8\textwidth]{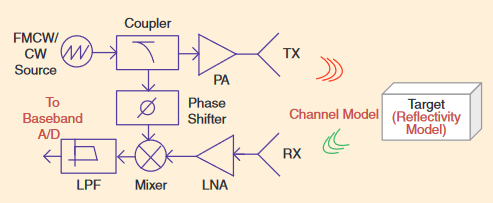}
            \caption{Standard FMCW Diagram \cite{patole2017automotive}}
            \label{fig:fmcw_diagram_lr}
        \end{figure}

        A diagram with the components of a FMCW radar can be seen in Figure \ref{fig:fmcw_diagram_lr}.

\begin{figure}[ht]
    \centering
    \includegraphics[width=0.6\textwidth]{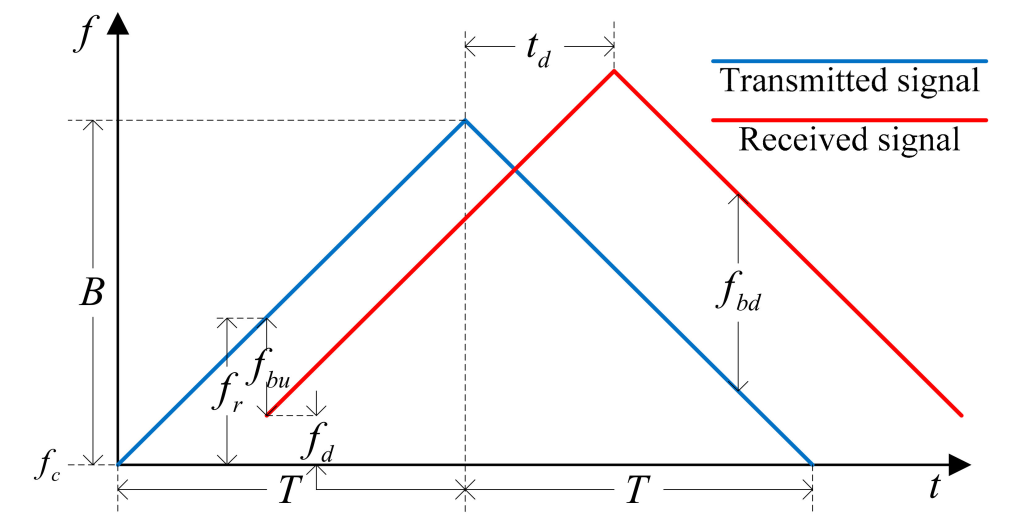}
    \caption{FMCW triangular chirp transmitted and received diagram.}
    \label{fig:fmcw_chirp}
\end{figure}

\begin{figure}[ht]
    \centering
    \includegraphics[width=1.0\textwidth]{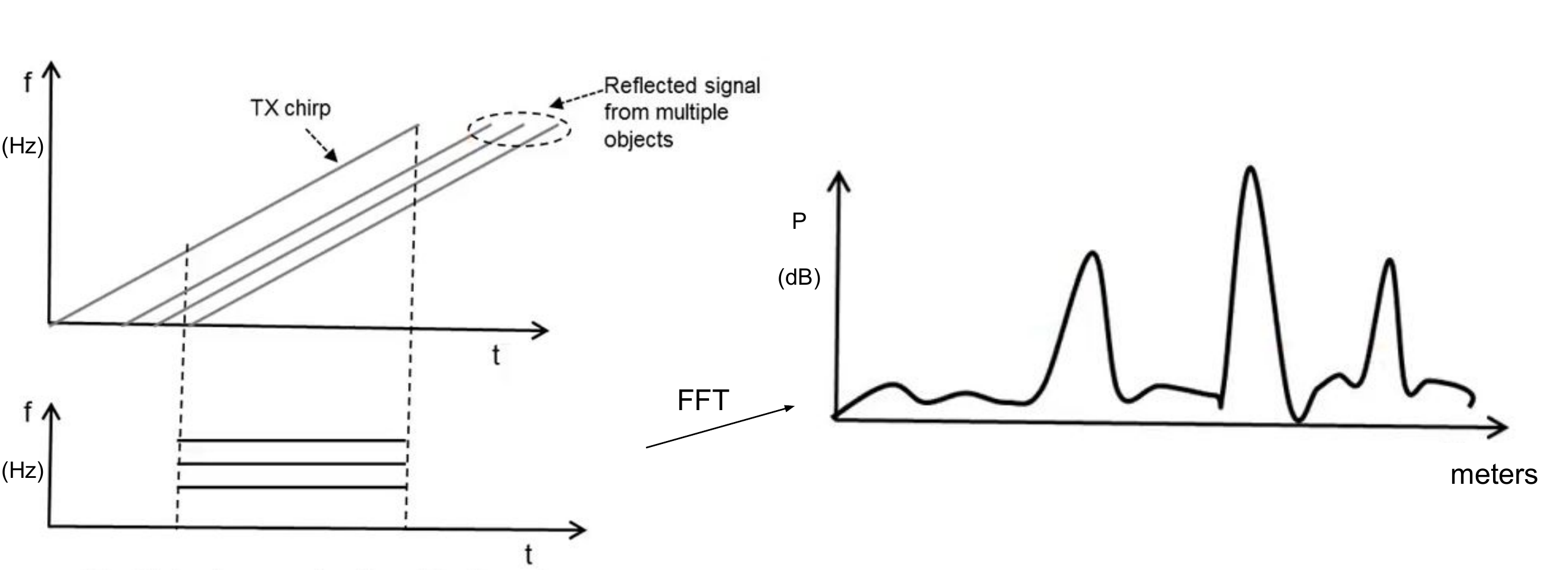}
    \caption{Illustration showing how to convert the signal in time domain to range information \cite{tiradar}. This shows a graph on top right a saw-tooth pattern being and 3 returns after some time. The graph on bottom right shows the output from the mixer. Image on the right shows the range profile after applying FFT in the mixed response. }
    \label{fig:range_from_freq}
\end{figure}    

 From the diagram from Figure \ref{fig:fmcw_diagram_lr}, we can describe the main components as follows:
        
\begin{itemize}
    \item \textbf{FMCW Source Generator}
    
    This block generates the waveform that will be transmitted. The modulation generated captures the range information, and the size of the bandwidth represents how much information it can carry, which means more range resolution. To generate the waveform, Equation \ref{eq:signal} can be extended to add the modulation part:
    
    \begin{equation}
        y(t) = Asin(2\pi f_0 t + \pi k t^2 + \theta)
    \end{equation}
    
    \noindent where $k=\frac{f_1 - f_0}{\tau_p}$, $f_0$ and $f_1$ are the initial and final frequencies respectively, and $\tau_p$ is time difference from one frequency to another. The most used types of modulations are Sawtooth wave and Triangular wave.

    \item \textbf{Mixer}
    
    The mixer is used to mix the input signal with the received signal. By mixing the signals, we can retrieve the range by the beat frequency. Given two signals $x_1$ and $x_2$, with frequencies $w_1$ and $w_2$ and phase $\theta_1$ and $\theta_2$, we can define the transmitted and received signal as:
    
    \begin{equation}
    \begin{split}
        x_1 &= sin(w_1 t + \theta_1) \\
        x_2 &= sin(w_2 t + \theta_2)
    \end{split}
    \end{equation}

    \noindent The mixer captures the instantaneous frequency difference and phase difference of 2 input signals. The result of the mixer is given by:
    
    \begin{equation}
        x_{mixer} = sin[(w_2 - w_1)t + (\theta_2 - \theta_1)]
    \end{equation}

\end{itemize}

Applying \textit{Fast Fourier Transform} (FFT) to the mixed signal, we get the beat frequency up and down as $f_{bu}$ and $f_{bd}$ and the Doppler frequency $f_d$ (Figure \ref{fig:fmcw_chirp}). Figure \ref{fig:range_from_freq} illustrates an example of converting a signal in time domain to range domain.

        The range resolution of a radar depends only on its bandwidth. Equation \ref{eq:range_res} shows the relation of the range resolution to the bandwidth.
        
        \begin{equation}\label{eq:range_res}
            d_{res} = \frac{c}{2B}    
        \end{equation}

        \noindent where $c$ is the speed of light and $B$ is the bandwidth in Hz.

        Most automotive radars are based on multiple input multiple output (MIMO). MIMO radar is a type of radar with multiple transmitters and receivers. By using multiple antennas, the direction of arrival (DoA) can be computed. The angular resolution in MIMO radars is given in Equation \ref{eq:doa}:
        
        \begin{equation}\label{eq:doa}
            \theta_{res} = \frac{\lambda}{N d cos(\theta)},
        \end{equation}
        
        \noindent where $N$ is the number of antennas, $d$ is the distance between the antennas, $\lambda$ is the wavelength and $\theta$ is the angle of where the object is.
        Assuming we have $d = \frac{\lambda}{2}$ and $\theta = 0$, we now get:
        
        \begin{equation}\label{eq:doa2}
            \theta_{res} = \frac{2}{N}.
        \end{equation}
        
        As an example, consider a MIMO radar with 3 transmitters and 4 receivers, using Equation \ref{eq:doa2} we can measure the angular resolution
        
        \begin{equation}
            \theta_{res} = \frac{2}{3+4} = 0.28 rad = 16^o .  
        \end{equation}

\begin{figure}[ht]
            \centering
            \includegraphics[width=0.8\textwidth]{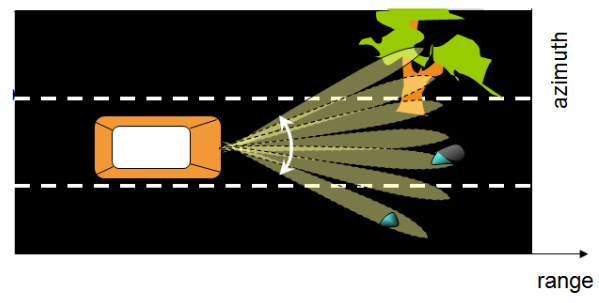}
            \caption{Scanning radar \cite{jastehthesis}.}
            \label{fig:scan_radar}
        \end{figure}

        An angular resolution of $16^o$ degrees is not suitable for imaging applications. A way to achieve high resolution imaging radar, is by using a mechanical scan. The angular resolution of a scanning radar is based on its beamwidth. The azimuth resolution of a scanning radar is given by the beamwidth of the signal at -3 dB. Figure \ref{fig:scan_radar} shows an illustrative example of a scanning radar. 
        
        \begin{figure}[ht]
            \centering
            \includegraphics[width=1\textwidth]{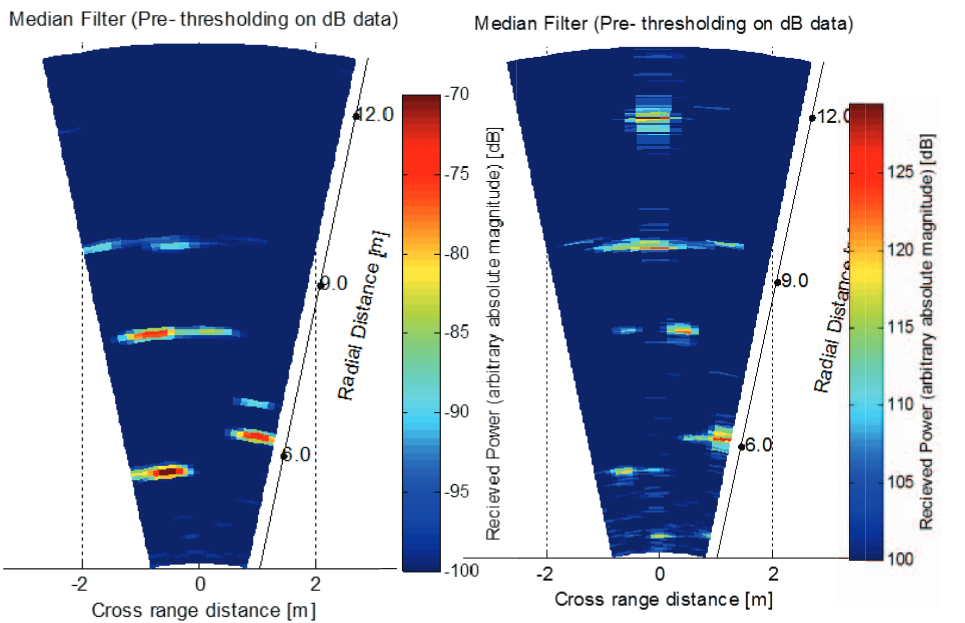}
            \caption{The radar imaging difference between 30 GHz (a) and 150 GHz (b) \cite{jasteh2015low}}
            \label{fig:jasteh150low}
        \end{figure}
        
        \begin{figure}[ht]
            \centering
            \includegraphics[width=0.7\textwidth]{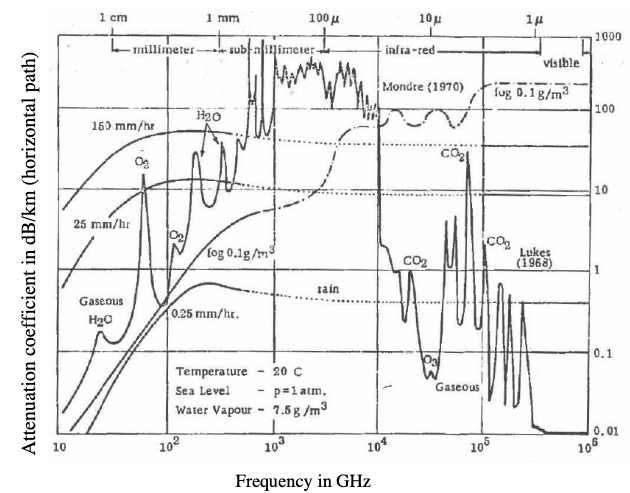}
            \caption{ Atmospheric attenuation vs. frequency under different atmospheric conditions \cite{stove2015potential}.}
            \label{fig:low-thz}
        \end{figure}
        
     Chapters \ref{ch:300dl}, \ref{ch:radio} and Appendix \ref{chapter:150ghz} use low-THz sensing technology (150 GHz and 300 GHz). Automotive radar systems use mmwave wavelengths (24 GHz and 79 GHz).
     
     There are previous works on low-THz radars to improve the radar resolution. Jasteh, \etal \cite{jasteh2015low} compares 30 GHz and 150 GHz radar imaging for automotive scenarios. The paper shows that 150 GHz can improve the range and azimuth resolution, creating a better representation of the scenario comparing to the 30 GHz (Figure \ref{fig:jasteh150low}). The Equation \ref{eq:azimuth_res} \cite{skolnik2001introduction} shows the relation between beamwidth (azimuth resolution) and frequency, where $\theta$ is the beamwidth in radians, $\lambda$ is the wavelength, $\omega_0$ is the antenna aperture diameter.
     
     \begin{equation}
         \theta = \frac{\lambda}{\pi \omega_0}
     \end{equation}\label{eq:azimuth_res}
     
    \noindent From this equation, we can see that the higher the frequency, the lower is the beamwidth. The development of higher frequencies improves the resolution, with the drawback of higher attenuation due to some atmospheric conditions. Improving radar resolution can lead to a better recognition, without using Doppler information.

     Low-THz sensing suffers from high absorption of EM energy, which can transform part of the EM energy into thermal energy \cite{skolnik2001introduction}. But with recent technology advancements, a new system was developed that overcomes EM absorption and low power issues.
        
        Low-THz sensing can provide smaller beamwidth than mmwave radar due to smaller antennas. Low-THz sensing can also provide up to 20 GHz bandwidth. 20 GHz bandwidth represents 0.75 cm, which is a very high resolution for radar applications.
        
        The main limitation of using low-THz sensing is the atmospheric attenuation. Figure \ref{fig:low-thz} shows the atmospheric attenuation for different frequencies. This Figure shows the different phenomena of different frequencies in different types of conditions. As we can see from this graph, rain does not attenuate the radar signal over different GHz and THz frequencies, being THz radar sensing a good alternative frequency which worked under adverse weather.

        In Appendix \ref{chapter:150ghz} and Chapters \ref{ch:300dl} and \ref{ch:radio} we use 150 GHz and 300 GHz radars which were developed to achieve a high resolution. In Chapters \ref{ch:navtech} and \ref{ch:navtech2} we use a commercial 79 GHz imaging radar to detect vehicles in the "wild".
        
         \subsubsection{Synthetic Aperture Radar}
        
\begin{figure}[ht]
            \centering
            \includegraphics[width=1.0\textwidth]{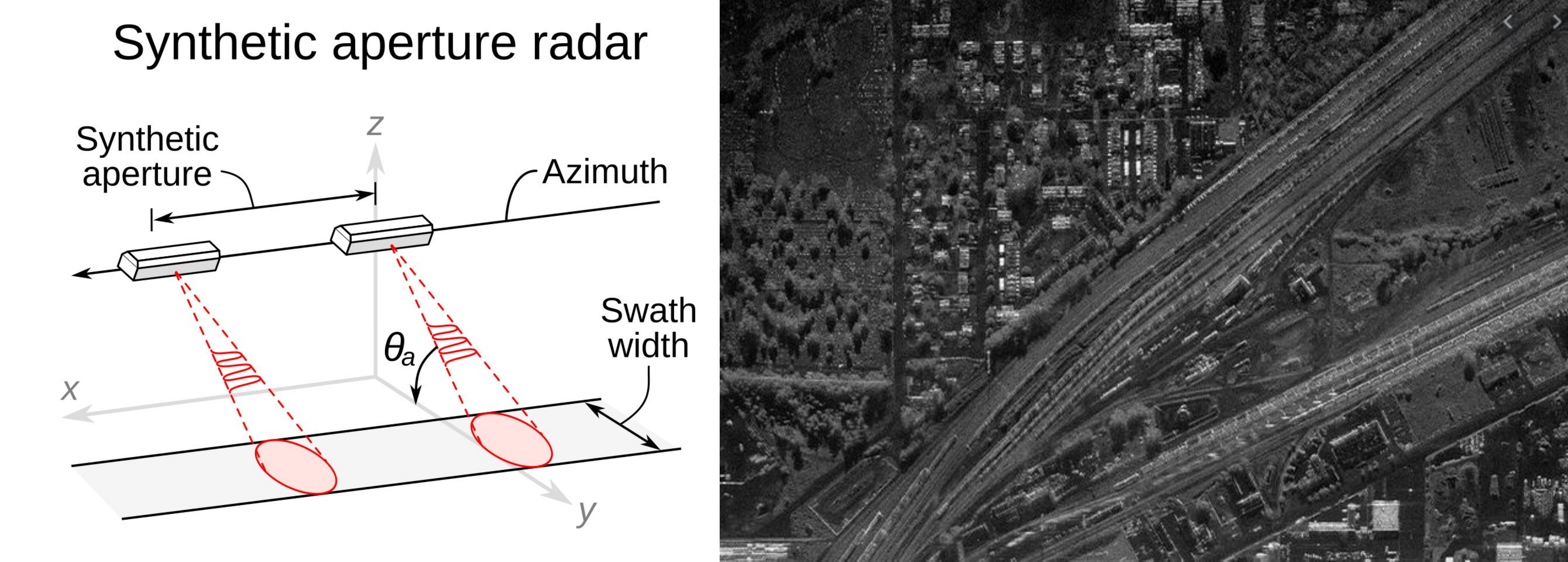}
            \caption{ Scheme how a SAR image is captured, and an example of SAR image \cite{wiki_sar}.}
            \label{fig:sar_image}
        \end{figure}

        The most common imaging radar technique is based on Synthetic Aperture Radar (SAR). SAR images are reconstructed after a series of consecutive signals are captured from a moving airplane \cite{sar_book}. The side move of the antenna creates a synthetic aperture which can be used to improve the resolution.  Figure \ref{fig:sar_image} shows a scheme how a SAR captures the image by moving the sensor, and an example of a SAR image.
        
        To reconstruct a high resolution image, a common method used is the backprojection algorithm \cite{yegulalp1999fast}.
        
        Another similar technique to SAR images, is Inverse SAR (ISAR). Instead of moving the sensor, we capture a moving target that is sensed in different aspects. The movement of the target is known or estimated, so a reconstruction algorithm is employed to achieve a high resolution imaging.
        
        Since large SAR datasets are already available, in this thesis we use SAR images in a transfer learning context to improve object recognition on our small low-THz radar dataset.
        
\subsection{Deep Neural Networks}\label{sec:dnn_lr}
	
\begin{figure}[ht]
    \centering
    \includegraphics[width=0.6\textwidth]{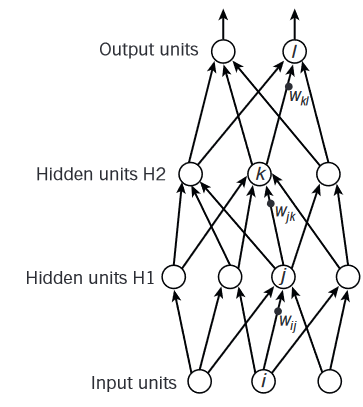}
    \caption{Graphical representation of a simple neural network \cite{lecun2015deep}.}
    \label{fig:lit_neural_net}
\end{figure}

Neural networks (NNs) are going to be used throughout this thesis. In this section we will introduce the theoretical aspects of neural networks. The graphical representation of a neural network can be visualised in Figure \ref{fig:lit_neural_net}. It is composed of an input layer, a hidden layer and an output layer \cite{rumelhart1988learning}. The neural network graph is represented by a set of the nodes connections, where the connections are the weights to be learned.

The idea of developing an artificial brain has been studied since the 40's \cite{mcculloch1943logical} by McCulloch and Pitts. The paper tries to explain how the brain might work and the neural network was modelled by using electrical circuits. The biological brain and neuron connections have always been a motivation for the development of neural networks.

Widrow and Hoff \cite{widrow1960adaptive} developed ADALINE to recognise binary pattern from phone line connections. It was one of the first papers to use a neural network in a real world problem. They used ADALINE to predict the next bit in a phone line connection.

The main challenge has always been to develop an effective learning algorithm. The breakthrough for training neural networks was introduced by the development of the backpropagation method \cite{rumelhart1986learning}. The backpropagation, like the name suggests, propagates the error over each layer of the neural network backwards.

We can mathematically formalize a neural network as set of layers,

\begin{equation}\label{eq:yfun}
y_l = f(x^l; W^l) 
\end{equation}

\noindent where $y_l$ is the output for each layer, $f$ is the activation function, $W^l$ is a set of weights at layer $l$ and $x^l$ is the input at layer $l$. The neural network learns the weights $W$ which should be generalized to any input. Architectures may have several types of layers; convolutional, rectified linear units (ReLU), max pooling, dropout and softmax \cite{Goodfellow:2016:DL:3086952}. When using sequential data Recurrent Neural Networks (RNN) and Long-Short Term Memory (LSTM) are alternative architectures which are not covered in this thesis.

\subsubsection{Convolution Layer}

\begin{figure}[ht]
    \centering
    \includegraphics[width=0.8\textwidth]{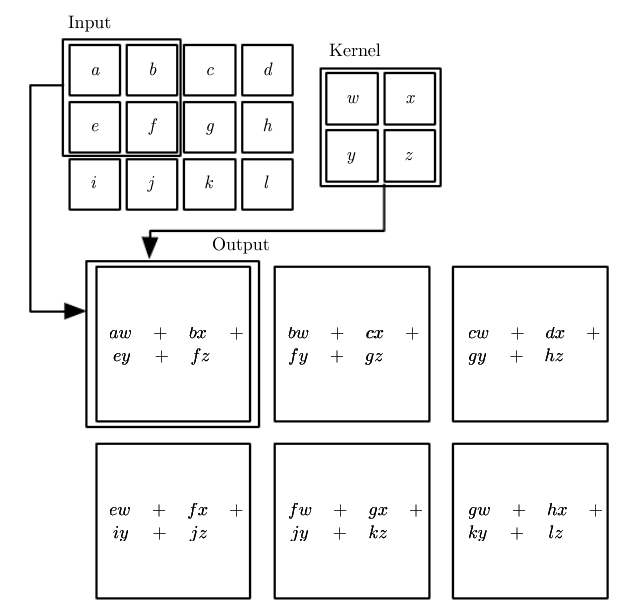}
    \caption{Convolutional Layer. This figure shows an image as input, an example of convolutional kernel with corresponding weights. In the bottom, it shows the equation how the output is computed \cite{Goodfellow:2016:DL:3086952}.}
    \label{fig:conv_layer}
\end{figure}

The most common layer in deep neural networks applied to computer vision is the convolutional layer. It learns convolution masks which are used to extract features based on spatial information. Equation \ref{eq:conv} shows the convolution layer computation for each mask learned,

\begin{equation}\label{eq:conv}
h(i,j,k) = \sum_{u=0}^{M-1} \sum_{v=0}^{N-1} \sum_{w=0}^{D-1} W^l(u,v) X(i - u, j - v, k - w) + b^l,
\end{equation}

\noindent where $M$ is the mask width, $N$ is the mask height, $D$ is the mask depth, $W^l$ is the convolution mask learned, $b^l$ is the bias and $X$ is our image. Figure \ref{fig:conv_layer} shows a graphical representation of a convolutional layer.

\subsubsection{Rectified Linear Unit}

\begin{figure}[ht]
    \centering
    \includegraphics[width=0.5\textwidth]{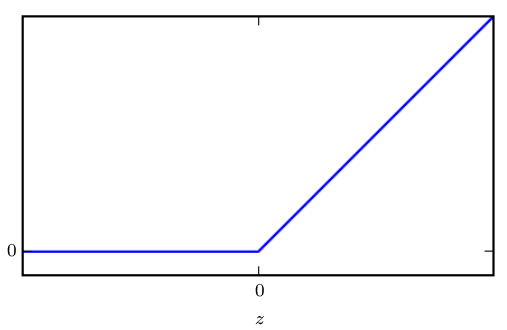}
    \caption{Plot of a ReLU function \cite{Goodfellow:2016:DL:3086952}.}
    \label{fig:relu}
\end{figure}

The activation function $f$ is usually a non-linear function that maps the output of current layer. A simple method that is computationally cheap and approximates more complicated non-linear functions, such as, $tanh$ and $sigmoid$, is the Rectified Linear Unit (ReLU).
Equation \ref{eq:fgfun} shows the ReLU function where $X$ is the output from the current layer. Figure \ref{fig:relu} shows the plot of a ReLU activation function.

\begin{gather}\label{eq:fgfun}
ReLU(X) = max(0, X)
\end{gather}

\subsubsection{Max Pooling}

\begin{figure}[ht]
    \centering
    \includegraphics[width=0.7\textwidth]{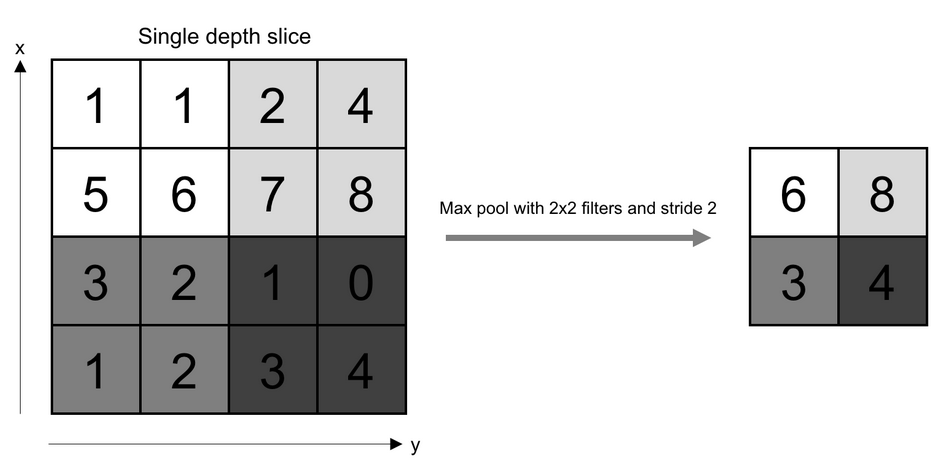}
    \caption{Max Pooling. This figure shows an example image and a max-pooling of 2$\times$2 is applied getting the maximum value of each patch  \cite{shanmugamani2018deep}.}
    \label{fig:max_pool}
\end{figure}

To reduce the image dimensions, max pooling can be used. It simply looks at a region and gets the maximum value, since this value has the greatest activation from previous layers. Figure \ref{fig:max_pool} shows a visual example of max pooling on a $8 \times 8$ input.

\subsubsection{Dropout}

\begin{figure}[ht]
    \centering
    \includegraphics[width=0.9\textwidth]{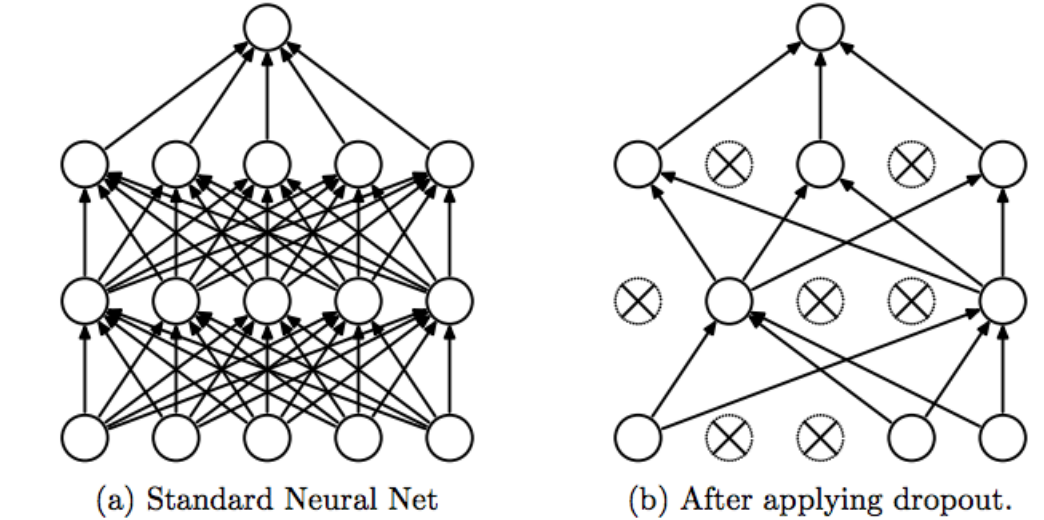}
    \caption{Dropout. Dropout set random weights to zero, forcing the network to find other path during training \cite{gal2016dropout}. This technique is only applied during training.}
    \label{fig:dropout}
\end{figure}

The dropout technique was introduced by the papers \cite{krizhevsky2012imagenet, srivastava2014dropout}. This technique sets random weights to 0 during training, forcing other paths to train the neural network. This technique avoids overfitting. Figure \ref{fig:dropout} shows an example of how dropout randomly removes some connections.

\subsubsection{Softmax}

The softmax layer converts the output from a previous layer to pseudo-probabilities, normalising the output vector to 1. Thus, it will give a likelihood for each class. Equation \ref{eq:softmax} shows the softmax layer,

\begin{equation}\label{eq:softmax}
    Softmax(x, i) = \frac{e^{x_i}}{\sum_j e^{x_j}}
\end{equation}

\noindent where $x_i$ is the output for the current class, $x_j$ is the output for all classes. 

After we feed forward the neural network, we need to compute an error metric, which is usually called cost or loss function.
A common cost function used for classification is the categorical cross-entropy (Equation \ref{eq:loss_c}).

\begin{equation}\label{eq:loss_c}
    C(\hat y,y) = - \sum_i y_i log(\hat y_i)
\end{equation}

In Equation \ref{eq:loss_c} $\hat y$ is the predicted vector from softmax output and $y$ is the ground truth.

Those are the main layers used in deep neural networks. Technically a neural network to become "deep", it needs more than one hidden layer, however usually the "deep" term is used when hand-crafted features are not extracted - the raw image is used as input inside a deep network. Deep neural networks is a very active area of research, and new layers are constantly being researched for many types of applications.

To make a neural network learn, we need to apply an optimisation method. A common method is {\it Gradient Descent} (GD). GD updates the weights of the network depending on the gradient of the function that represents the current layer, as in Equation \ref{eq:sgd_background}.

\begin{equation}\label{eq:sgd_background}
W_{t+1} = W_{t} - \alpha \nabla f(x;W) + \eta \Delta W 
\end{equation}

In Equation \ref{eq:sgd_background}, $\eta$ is the momentum, $\alpha$ is the learning rate, $t$ is the current time step, $W$ defines the weights of the network and $\nabla f(x;W)$ is the derivative of the function that represents the network. To compute the derivative for all layers, we need to apply the chain rule, so we can compute the gradient through the whole network. Since we work with many layers and huge datasets, GD can be very hard to compute, since it needs to give the whole input to the algorithm to compute one step.  In a deep neural network context, \textit{Stochastic Gradient Descent} (SGD) is more popular, since it uses batches of data, which is easier to compute and an approximation of Gradient Descent \cite{Goodfellow:2016:DL:3086952}. 
	
Equation \ref{eq:sgd_background} is used for the whole neural network. In order to apply the derivative for each layer, the backpropagation algorithm was developed \cite{rumelhart1988learning}, which is the most popular method to train neural networks. Backpropagation propagates the error by using the chain-rule over each layer updating its weights.
Mathematically, it works as follows: we have a pair $(x,y)$, where $x$ is the input and $y$ is the desired result. We can use a loss function $C$ and a network $f$ which represents a series of composite functions at each layer (Equation \ref{eq:chain_rule1}).

\begin{equation}\label{eq:chain_rule1}
    C(y, f^L(W^L f^{L-1}(W^{L-1} \cdots f^1(W^1 x))))
\end{equation}

After we compute the loss, we can apply the chain-rule of derivatives to compute the error for each layer of the neural network. The total derivative for the the loss function $C$ is given in the Equation \ref{eq:chain_rule2},

\begin{equation}\label{eq:chain_rule2}
    \frac{d C}{d a^L}\cdot \frac{d a^L}{d z^L}\cdot \frac{d z^L}{d a^{L-1}} \cdot \frac{d a^{L-1}}{d z^{L-1}}\cdot \frac{d z^{L-1}}{d a^{L-2}} \cdots \frac{d a^1}{d z^1} \cdot \frac{\partial z^1}{\partial x}.
\end{equation}

In the Equation \ref{eq:chain_rule2}, we denote at a layer $l$ as $z^l=f^l(x)$ and $a^l = activation(z^l)$. Equation \ref{eq:chain_rule2} shows the total derivative using chain-rule. We can simplify this equation by using an auxiliary term $\delta^l$ which stores the derivative error for each layer which can be propagated over the previous layers.

The recursive implementation of the backpropagation is given in the Equation \ref{eq:delta}.

\begin{equation}\label{eq:delta}
    \delta^{l-1} = (f^{l-1})' \cdot (W^l)^T \cdot \delta^l
\end{equation}

A very big discussion in optimising neural networks is how is it able to generalise. Many researchers see deep neural networks as a black box which is hard to explain why it works. The main creators of deep learning methods: Yann Lecun, Geoffrey Hinton and Yoshua Bengio claim that usually there are several local minima. The weights trained which finds one of those local minima, is able to generalizes well. Also large quantities of data help the deep neural network to learn a function which is more general - it means that it learns a more meaningful manifold which generalises to the test set, without overfitting to the training set.

    \subsection{An Overview of Autonomous Vehicles}\label{sec:ac}

    Nowadays we hear in the news that autonomous vehicles will be released in the next 5 years \cite{Ford, Volvo, Nissan, BMW}. Tesla is currently leading it, which a level 3 of autonomy is already implemented in their products. Autonomous car systems are becoming more reliable however to achieve full autonomy we need to take many factors into consideration.

    A report published by SAE International \cite{sae_levels} defines the levels of autonomy that an autonomous car can achieve, divided into 5 levels.
    
    \begin{itemize}
        \item \textbf{Level 0} : The car system automatically warns the driver if the car is close to collision, but has no control of the vehicle. Currently already implemented by many car manufacturers.
        
        \item \textbf{Level 1} : Driver uses the wheel to control, however the car system has some automation, such as Adaptive Cruise Control (ACC). For ACC the driver controls the wheel, but the speed is controlled by the car. Parking assistant is also part of the first level of autonomy. Lane Keeping Assistance (LKA) also fits into this level, but the driver should maintain attention to the road in case the driver needs to fully control the car.
        
        \item \textbf{Level 2} : This level is close to what the current Tesla cars offer. The system takes control of speed, braking and steering, but the driver should be always prepared to take control of the car at any time.
        
        \item \textbf{Level 3} : In this level the driver can text or watch a movie while the system controls the car. However the system can call driver's attention or ask for the driver to use the car under some circumstances like bad weather or poor road scenarios.
        
        \item \textbf{Level 4} : This level is like the level 3, however if the driver does not retake control of the car when necessary, the system will find a safe way to park the car until the driver takes control of the situation.
        
        \item \textbf{Level 5} : In this level, the system has full autonomy and the steering wheel is optional. Human intervention is not required.
        
    \end{itemize}
    
\begin{figure}[ht!]
            \centering
            \includegraphics[width=1\textwidth]{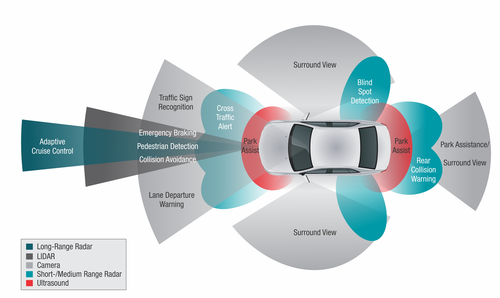}
            \caption{Set of sensors used by an autonomous car \cite{car_sensors}.}
            \label{fig:car_sensor}
        \end{figure}

    Sensor design and algorithm development are crucial for the development of reliable perception system for autonomous cars. The current research on autonomous vehicles uses a set of sensors, such as LiDAR, camera, radar, infrared and ultra-sound. In the Figure \ref{fig:car_sensor} there is an example of a set of sensors used and its coverage.

    The development of autonomous cars has a long history. Since the 80's there has been intense research in the field, involving researchers from many fields, such as computer vision, sensors, control and robotics.
    
\begin{figure}[ht!]
            \centering
            \includegraphics[width=0.7\textwidth]{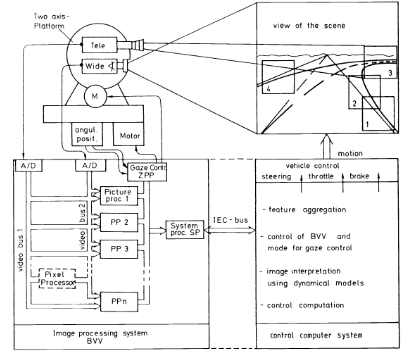}
            \caption{Diagram of the methodology of the first autonomous cars \cite{dickmanns1987curvature}.}
            \label{fig:dickmanns1987curvature}
    \end{figure}

    The first autonomous vehicle was developed in 1987 by Dickmanns \cite{dickmanns1987curvature} where it used the lane recognition from empty autobahns in Germany to drive autonomously. It basically just detected the lane curvature in order compute the control of the steering wheel and the pedals of the car. Figure \ref{fig:dickmanns1987curvature} shows the scheme of how it was developed. This paper lead to the biggest autonomous vehicles research project, the Eureka Prometheus Project which received \euro749M to develop state-of-the-art autonomous cars during the time.
    
\begin{figure}[ht!]
            \centering
            \includegraphics[width=0.7\textwidth]{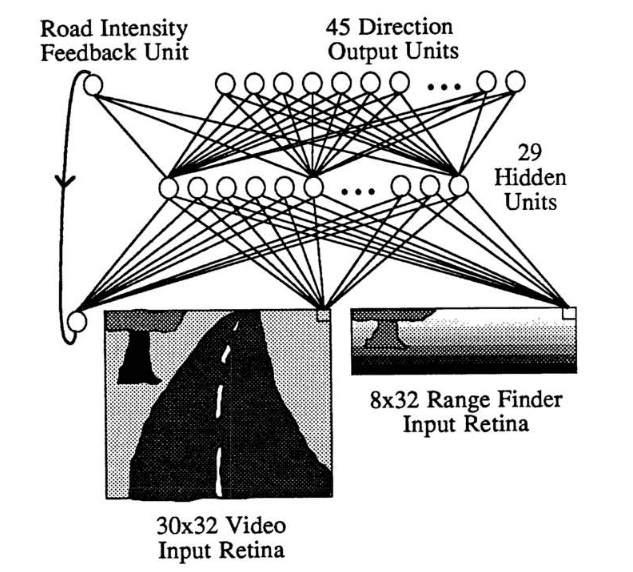}
            \caption{Network architecture developed in the ALVINN project \cite{pomerleau1989alvinn}.}
            \label{fig:pomerleau1989alvinn}
    \end{figure}

    Another famous project in autonomous cars was the ALVINN Project in 1989 \cite{pomerleau1989alvinn} where it recorded a human driving a car and used it as an input to an artificial neural network. The network used a $30 \times 32$ raw input from a camera and a $8 \times 32$ raw range finder. The output from the neural network is the control from the steering wheel. After the neural network was trained, the car successfully drove along the road. In Figure \ref{fig:pomerleau1989alvinn} we can visualize the architecture of the neural network.
    
    A talk about autonomous cars given by Jitendra Malik at the Berkeley EECS Annual Research Symposium \cite{malik_talk} discusses that those previous research papers \cite{pomerleau1989alvinn, dickmanns1987curvature} solved 95\% of the problems which are keeping in lane and avoiding obstacles. Problems like the weather influence, different illuminations, bad road conditions, lack of lane information are the next steps to be solved to achieve full autonomy. 
    
    Following the last two decades the research advancement on autonomous cars has been fast. The DARPA Grand Challenge is a competition that has the intention of sponsoring high technological advance in military use. The DARPA Grand Challenges from 2004, 2005 and 2007 were focused on developing autonomous cars for long distance on urban and off-road scenarios. Many universities took part in the competition. This challenge helped the advancement of state-of-the-art tasks such as localization, mapping and obstacle detection \cite{thrun2006stanley}.
    
    Google also decided to join the autonomous cars research and development in 2009 focusing on the development of intelligent vehicles in urban scenarios \cite{litman2014autonomous}. The Google car completed over 1,498,000 miles completely autonomously over regions of California, Texas and Washington. It registered 14 accidents where 13 of them were caused by other drivers. In 2016 the Google car project become a separate company called Waymo.
    
    To evaluate the performance of current methods in many tasks of autonomous cars, the KITTI dataset was developed \cite{geiger2012we}. The KITTI dataset is a collection of data from stereo camera and LiDAR from a mid-sized city in Germany. It defined tasks such as object detection, stereo reconstruction, optical flow, SLAM, multi-target tracking, road/lane detection and semantic segmentation. KITTI does not provide the test set publicly, therefore it is a fair system that allows to compare different methods.
    
    Most of the new algorithm development of autonomous cars uses KITTI dataset as benchmark. State-of-the-art methods in almost all tasks are based on deep neural networks \cite{xiang2016subcategory, zbontar2016stereo, ilg2016flownet}.
    
    More recently Nvidia started a new approach of solving the perception system of autonomous cars called {\it end-to-end learning} \cite{bojarski2016end}. The {\it end-to-end learning} algorithm receives video as the input and the steering wheels and pedals as the output. It means that we do not need to identify pedestrians, cars, or map the free space - the convolutional neural network is capable of learning and reacting based on raw data. It tries to create a method that imitates the human perception. However {\it end-to-end learning} can be quite dangerous, since it is hard to explain why the car took  a certain decision.
    
    The development of new methods for autonomous cars are still really active in the research community, however datasets on more challenging scenarios, such as bad weather, should be addressed in order to create better benchmarks on more heterogeneous scenarios.

\section{Literature Review}\label{sec:litrev}

    This part of the chapter will cover the main literature on object recognition. Recognising the object in an automotive application is crucial to achieve full autonomy. We review the literature on object recognition using video, LiDAR, infrared and radar sensors (Section \ref{sec:obj_rec}). Finally, Section \ref{sec:sensor_comparison} shows the comparison between the sensors presented in this chapter.
    
    \subsection{Object Recognition}\label{sec:obj_rec}
    
    Object recognition is probably one of the fundamental problems in computer vision. How to answer the question: how to represent an object? There are many challenges on how to represent an object depending on the sensor. We can automatize many tasks done by a human to a computer using object recognition, such as industrial inspection, medical diagnosis, face detection, human-computer interaction and robot navigation. 
    
    Object recognition is one of the main tasks of autonomous cars. When we create a virtual map of the scenario, we need to identify some key actors, such as pedestrians and vehicles. We need to give special attention to those objects, and also predict their movement in order to create a safe autonomous car perception system.
    
    As we can see from the previous section, to solve levels 3, 4 and 5 of autonomy we need to take the weather scenario into consideration. Video and LiDAR are the primary sensors in the current development of autonomous cars. Video can provide high colour resolution and LiDAR provides high accuracy 3D point cloud estimation. Video can also provide 3D information if two cameras are used. Both sensors can be used for pedestrian recognition \cite{kidono2011pedestrian, ren2015faster}, vehicle detection \cite{premebida2007lidar} and 3D mapping \cite{krishnan2014intelligent, cadena2016past}. Since video can capture colour information, it can also be used also for traffic sign recognition \cite{wu2013traffic} and lane detection \cite{ghafoorian2018gan}. For better representation of the scene, often a fused representation of both sensors are used \cite{premebida2007lidar}.
    
    However video and LiDAR both have bad performance in severe weather since these sensors use optical electromagnetic spectrum which does not penetrate fog, rain and snow. On the other hand, radar sensors use radio waves which are capable to penetrate these types of weather. The drawback of radar is that it gives poor resolution compared to LiDAR and video. To take advantage of both types of sensors, radar sensors have been used for sensor fusion approaches, like seen at \cite{radecki2016all, chavez2016multiple, cho2014multi}.

        \subsubsection{Video}\label{sec:vs}
        
        Video is probably the most researched sensor for object recognition, since it is cheap and it enables acquiring huge amount of data. Video is the closest sensor to the human eyes, so methods using biological inspirations are more common. Object recognition in humans is a complex topic with many unknowns \cite{dicarlo2012does}. Neuroscientists argue that information such as shape texture and colour are processed in hierarchical neural layers in order to identify an object. It raises a question: how to develop a mathematical model that will create invariant features to different scenarios to represent an object in video images? There are many difficulties in creating a perfect method which will cover all possibilities, but slowly the research community is improving the methods and uncovering new results.
        
        Many aspects should be taken into consideration when designing an object recognition method. The challenges of object recognition are listed below \cite{cheng2016survey}:
        
        \begin{itemize}
            \item Viewpoint variation: The object will have different shapes depending on the viewpoint.
            \item Illumination: The object changes its colour properties depending on the scene illumination.
            \item Occlusion: The object changes its shape properties when parts of it are occluded.
            \item Scale: Depending how close your object is to the sensor, the size relative to the camera can have huge variations.
            \item Deformation: Animals, for example, can deform according to the movement of their body, creating variant shapes from the same object.
            \item Background clutter: The background can have a lot of unwanted colour information which can make it hard to segment different objects.
        \end{itemize}
        
        As we can see there are many aspects to take into consideration and it is a really challenging problem in computer vision.
        
\begin{figure}[ht]
            \centering
            \includegraphics[width=1\textwidth]{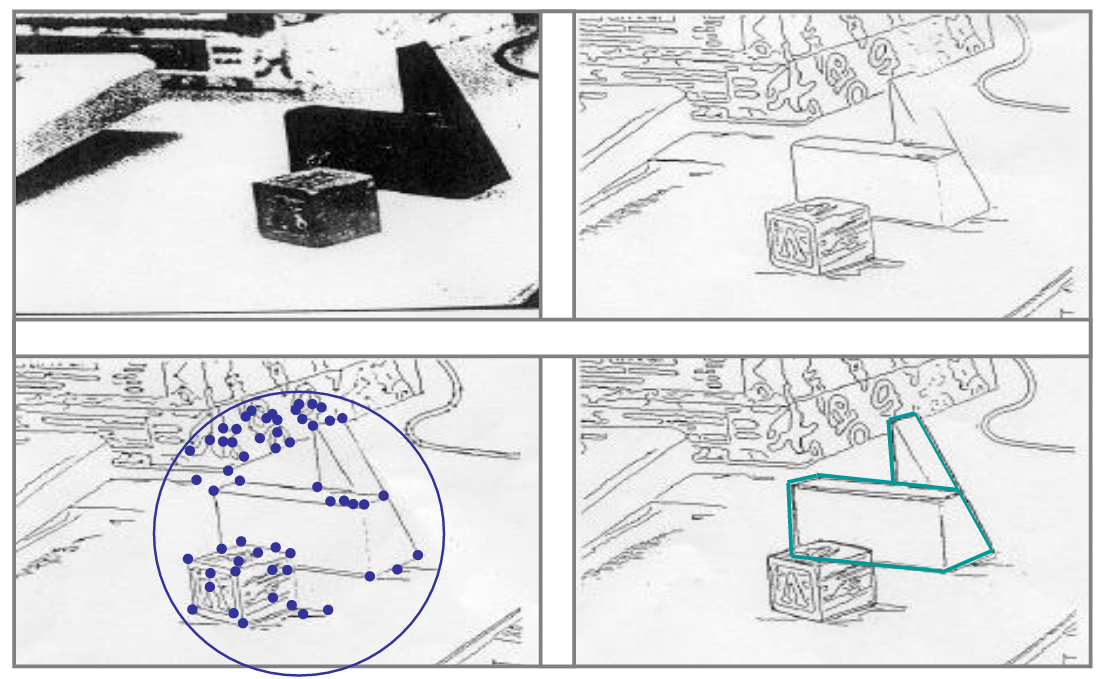}
            \caption{Aligning edges and corner to a fitting model to recognize an object \cite{huttenlocher1987object}.}
            \label{fig:huttenlocher1987object}
        \end{figure}

        The first works on object recognition dealt with known object geometry using edges and corners (\cite{roberts1963machine, lowe1987three, faugeras1986representation, huttenlocher1987object}). These works primarily investigated fitting a shape model to the desired object to recognize it (example in Figure \ref{fig:huttenlocher1987object}).
        
\begin{figure}[ht] 
            \centering
            \includegraphics[width=1\textwidth]{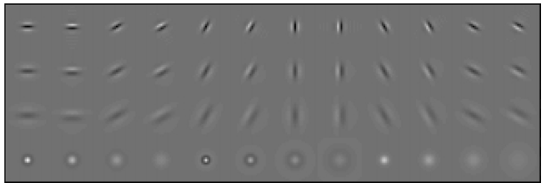}
            \caption{Leung and Malik created a set of 48 filters to classify different types of terrain \cite{leung2001representing}.}
            \label{fig:filters}
        \end{figure}

        In the 90's there has been a lot of research done on how to extract representations from objects by using linear filter for texture analysis \cite{perona1990scale, leung2001representing, schmid1997local, viola2001robust} (Figure \ref{fig:filters}).
        
\begin{figure}[ht]
            \centering
            \includegraphics[width=0.5\textwidth]{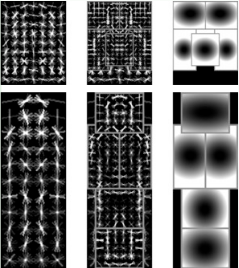}
            \caption{Histogram of Oriented Gradients (HOG) used in Deformable Part Models \cite{felzenszwalb2010object}.}
            \label{fig:dpm}
        \end{figure}

        In the 2000's features based on histograms were successful methods \cite{belongie2002shape, lowe1999object, dalal2005histograms, felzenszwalb2010object}. Scale-Invariant feature transform (SIFT) \cite{lowe1999object} was probably the most famous one and many research papers were using SIFT for feature extraction in various applications. Techniques like Histogram of oriented Gradients (HoG) \cite{dalal2005histograms} and Deformable Part Models (DPM) \cite{felzenszwalb2010object} (Figure \ref{fig:dpm}) are still often used in today's research papers.
        
        Together with the development of feature extraction algorithms, machine learning algorithms were also being developed. The feature extraction methods need a good classification algorithm to decide which class the feature vector represents. Discriminative classifiers need to maximize the margin between classes being general enough to avoid overfitting. Methods like AdaBoost \cite{freund1999short}, Support Vector Machines (SVM) \cite{cortes1995support} and Random Forest \cite{breiman2001random} were between the most popular ones. SVM, in particular, is still widely used when feature extraction is needed. SVM defines support vectors that maximize the margins between the classes, creating a good generalization from the classification. SVM also forms the basis for kernel methods which expands data to higher dimensions in order to be linearly separable.
        
        With the introduction of larger datasets other methods were being developed. The PASCAL Visual Objects Classes challenge \cite{everingham2010pascal} created in 2007 motivated the development of new methodologies. One popular approach developed during this time was the bag-of-visual-words \cite{csurka2004visual}, using which objects were represented by a histogram of visual features.
        
\begin{figure}[ht]
            \centering
            \includegraphics[width=1\textwidth]{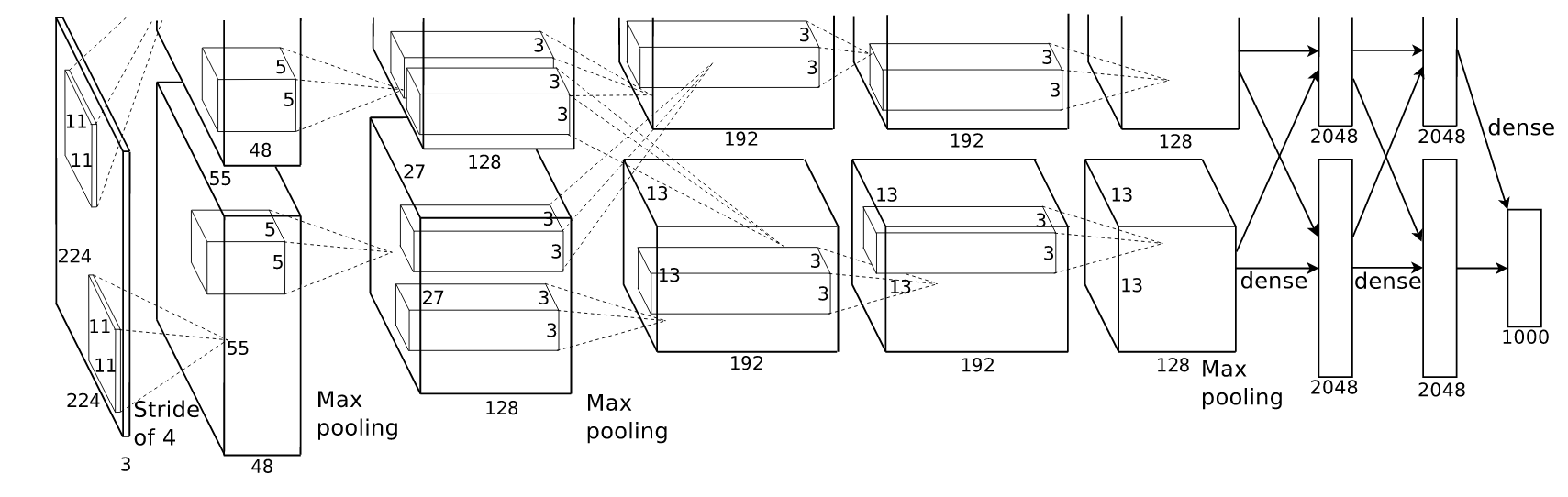}
            \caption{AlexNet Network structure that won the ImageNet 2012 Challenge  \cite{krizhevsky2012imagenet}.}
            \label{fig:alexnet}
        \end{figure}

        The ImageNet challenge was introduced in 2009 \cite{deng2009imagenet} following  the same structure of Pascal VOC challenge \cite{everingham2010pascal}, but adding much more data. The current ImageNet dataset contains 3.2 million images with more than 5000 classes. In 2012 the introduction of using deep convolutional neural networks (DCNN) \cite{krizhevsky2012imagenet} for large scale image recognition changed the field of computer vision (Figure \ref{fig:alexnet}). AlexNet achieved 84.70\% top-5 accuracy on ImageNet.
        
\begin{figure}
            \centering
            \includegraphics[width=\textwidth]{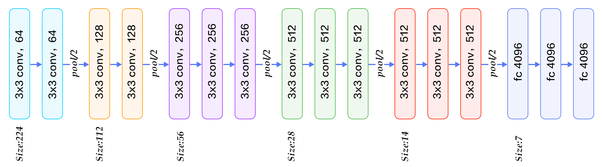}
            \caption{VGG-16 Network \cite{simonyan2014very}.}
            \label{fig:vgg16}
        \end{figure}

        After AlexNet, several other network architectures were achieving even better results. VGGNet is an extension of AlexNet with more layers and fixed kernel convolutions of $3 \times 3$ \cite{simonyan2014very} (Figure \ref{fig:vgg16}). The authors of VGGNet realised that they could combine 3 convolutions which has the same effect as a $7 \times 7$ convolution, but with fewer number of parameters and a lower computational cost. This network achieved 92.30\% top-5 accuracy on ImageNet.
        
\begin{figure}
            \centering
            \includegraphics[width=0.6\textwidth]{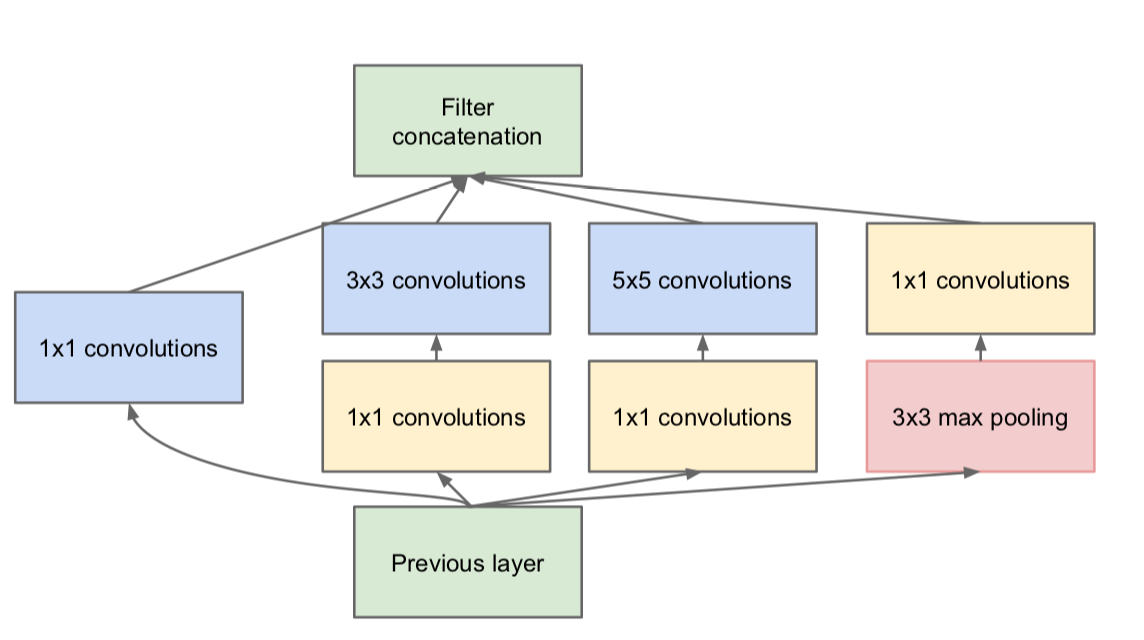}
            \caption{Inception Module \cite{szegedy2015going}.}
            \label{fig:inception}
        \end{figure}

        At the same year of the release of VGGNet, the Google Brain team developed InceptionNet \cite{szegedy2015going}. For each layer, this network uses the inception modules (Figure \ref{fig:inception}). Those modules have different convolutions with different kernel sizes. Thus, the InceptionNet is a wide network which achieve 95\% top-5 accuracy on ImageNet.
        
\begin{figure}
            \centering
            \includegraphics[width=0.6\textwidth]{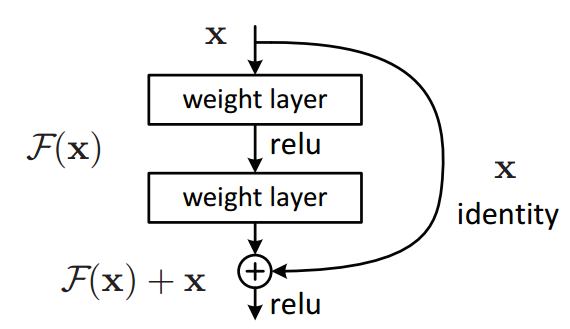}
            \caption{Residual Block \cite{he2016deep}.}
            \label{fig:resnet}
        \end{figure}

        ResNets \cite{he2016deep} developed in residual blocks. The residual blocks introduces the skip connections in each layer (Figure \ref{fig:resnet}). The intuition is that the network should at least learn the identity. By doing it, very deep networks were able to be trained, being robust to the vanishing gradient problem \cite{hochreiter2001gradient}. Other networks like AlexNet cannot be very deep because of the vanishing gradient problem. ResNets achieved 95.5\% top-5 accuracy on ImageNet.
        
        Table \ref{tab:comparison_lr} shows a comparison between the networks in terms of accuracy on ImageNet and number of parameters used to train.
        
        \begin{table}[ht]
        \centering
        \caption{Comparison between neural networks.}\label{tab:comparison_lr}
        \begin{tabular}{l|ccc}
        Network       & Year & Top-5 Accuracy on ImageNet & Parameters \\ \hline
        AlexNet \cite{krizhevsky2012imagenet}     & 2012 & 84.70 \%                   & 62M        \\
        VGGNet \cite{simonyan2014very}       & 2014 & 92.30 \%                   & 138M       \\
        InceptionNet \cite{szegedy2015going} & 2014 & 93.30 \%                   & 6.4M       \\
        ResNet \cite{he2016deep}       & 2015 & 95.51 \%                   & 60.3M     
        \end{tabular}
        \end{table}

        The networks described in the table were used to classify the whole image. In many applications, we also need to localise the object in the image. This is usually described as an object detection task. In the object detection we localise each object in the scene and classify it accordingly. 
        
        We can classify object detection methods based on deep neural networks into two categories: {\it one-stage detector} and {\it two-stage detector}.
        
        \begin{itemize}
            \item  {\bf Two-stage detectors}, like the name suggests, use two stages to detect objects. In the first stage, they detect potential regions, and in the second stage they classify each region. Methods like Overfeat \cite{sermanet2014overfeat}, R-CNN \cite{girshick2014rich}, Fast R-CNN \cite{girshick2015fast} and Faster R-CNN \cite{ren2017faster} are some examples of {\it two-stage detectors}.
            
            \item {\bf One-stage detectors}, train an end-to-end network that detects and classifies in a single pass. The main advantage of these methods is their speed. Since it is a single pass, it does not have any bottleneck with a region proposal algorithm. Methods like SSD \cite{liu2016ssd}, RetinaNet \cite{lin2017focal}, and YOLO \cite{redmonyolo2016} are some examples of {\it one-stage detectors}.
        \end{itemize}
        
        We use object detection methods based on neural networks in Chapters \ref{chapter:pol_lwir} and \ref{ch:navtech2}. We used Faster R-CNN \cite{ren2017faster} and SSD \cite{liu2016ssd} for the development of our methodology, so we gave more emphasis for these methods. We decided to use Faster R-CNN and SSD because they were the networks giving state-of-the-art results on the PASCAL dataset \cite{everingham2010pascal} during the time of our research papers were being developed.
        
\begin{figure}[ht]
            \centering
            \includegraphics[width=1.0\textwidth]{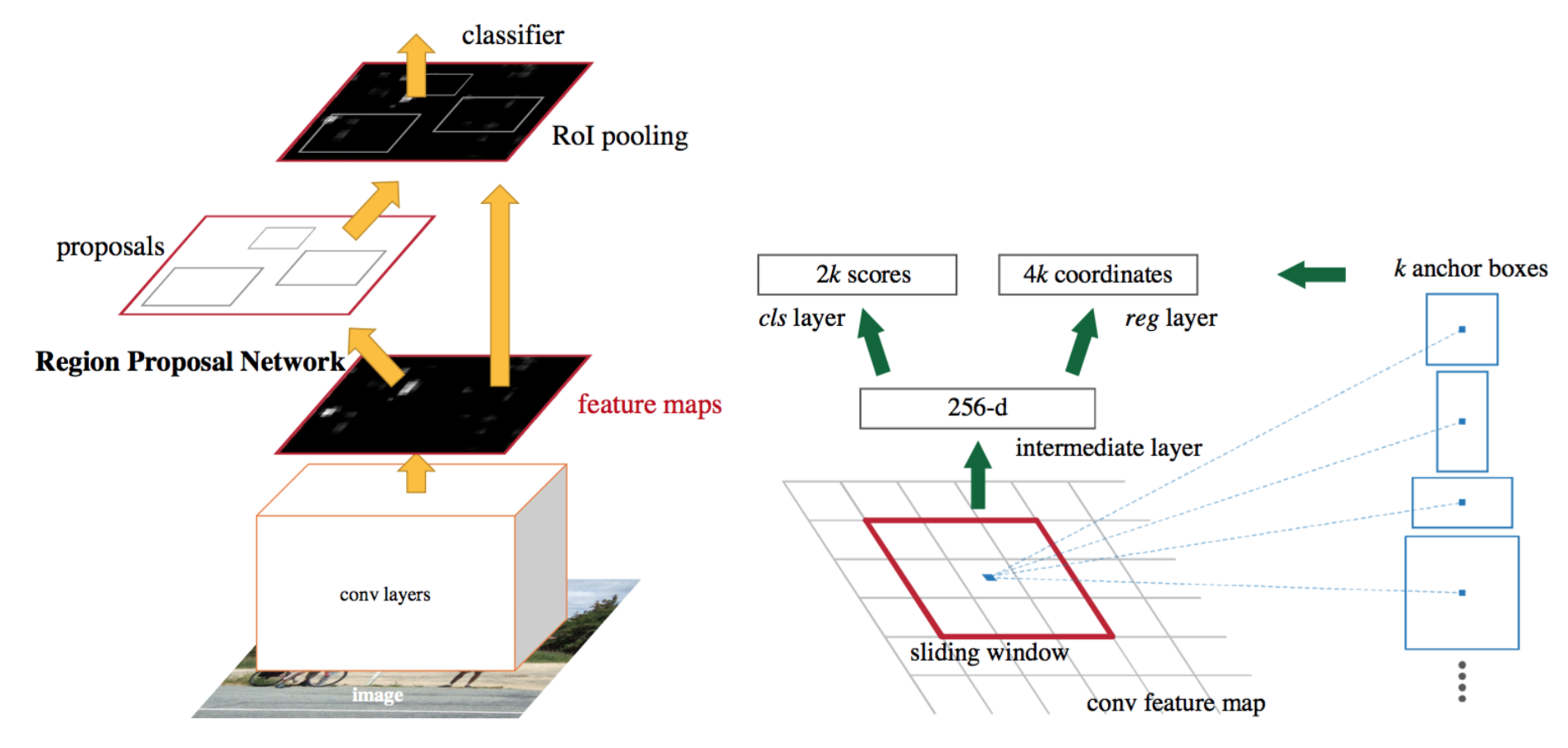}
            \caption{Faster R-CNN \cite{ren2017faster}.}
            \label{fig:faster_rcnn_lr}
        \end{figure}

        Faster R-CNN \cite{ren2015faster} introduces a region proposal network (RPN) to localize possible regions before classifying each region. Previous object detection methods were usually based on a segmentation method, like {\it Selective Search} \cite{uijlings2013selective}. The RPN learns how to propose potential regions, which is faster and more accurate than previous methods. Faster R-CNN also uses anchor boxes (Figure \ref{fig:faster_rcnn_lr}), which split the image into a grid, and for each center of this grid, it generates k-anchor boxes of different sizes. The anchor boxes subdivides the regions for better generalising between small and large objects.
        
        In order to learn how to detect the objects, the RPN has a loss function that takes into account the class of the object and the location as well. The loss functions developed for RPN are:

        \begin{equation}\label{eq:loss_faster}
        \begin{aligned}
        L(\{p_i\}, \{t_i\}) = \frac{1}{N_{cls}}\sum_i L_{cls}(p_i, p^{*}_i) + \lambda\frac{1}{N_{loc}}\sum_i  p^{*}_i L_{loc}(t_i, t^{*}_i).
        \end{aligned}
        \end{equation}
        
        \noindent where, $i$ is the index of the anchor, $p_i$ is the probability of being a object in the anchor location. $p^{*}_i$ is the ground truth, where 1 represents when there is an object, 0 otherwise. $t_i$ represents the predicted location. $t^{*}_i$ the ground-truth location. $t_i$ and $t_{i}^*$ represents a rectangular bounding box with the top left corner in the $x,y$ position with width ($w$) and height ($h$) dimensions.
        
        $L_{cls}$ is the categorical cross-entropy,
        
        \begin{equation}\label{eq:loss_cls_faster}
            L_{cls}(p_i, p^{*}_i) = - \sum_i p^{*}_i log(\hat p_i)
        \end{equation}
        
        \noindent and $L_{loc}$ is the localisation error.
        
        \begin{equation}
        L_\textrm{loc}(t^u, v) = \sum_{i \in \{\textrm{x},\textrm{y},\textrm{w},\textrm{h}\}} \textrm{smooth}_{L_1}(t^u_i - v_i),
        \end{equation}

        \begin{equation}
        \label{eq:smoothL1}
          \textrm{smooth}_{L_1}(x) =
          \begin{cases}
            0.5x^2& \text{if } |x| < 1\\
            |x| - 0.5& \text{otherwise},
          \end{cases}
        \end{equation}

        A illustrative diagram describing Faster R-CNN can be visualised in the Figure \ref{fig:faster_rcnn_lr}. 
        
        An example of {\it one-stage detector} is the {\it You Only Look Once} (YOLO) method. YOLO uses a single neural network to detect and classify objects in an image. YOLO, like Faster R-CNN, uses the concept of anchor boxes.

        Another similar method is the {\it Single Shot MultiBox Detector}, also called SSD \cite{liu2016ssd}. This method uses a single neural network, like YOLO. However, SSD generates the output at several scales of the feature map produced by the convolution.
        
        The overall objective loss function designed by SSD uses the location (loc) and the class confidence loss (conf):
\begin{equation}
L(x, c, l, g) = \frac{1}{N}(L_{conf}(x, c) + \alpha L_{loc}(x, l, g))
\label{eq:loss}
\end{equation}
\noindent where $x_{ij}^{p} = {1,0}$ is matrix for matching the $i$-th default box ($d$) to the $j$-th ground truth box ($g$) of category $p$, $N$ is the number of bounding boxes in the image, $l$ is the predicted box, and $c$ is the class labels. The default bounding ($d$) is formed by a ($c_x$, $c_y$) center and a width ($w$) and height ($h$) dimensions.

$L_{loc}$ is given as,
\begin{equation}
\begin{split}
L_{loc}(x,l,g) = \sum_{i\in Pos}^N \sum_{m\in\{cx, cy, w, h\}}& x_{ij}^k\text{smooth}_{\text{L1}}(l_i^m - \hat{g}_j^m)\\
\hat{g}_j^{cx} = (g_j^{cx} - d_i^{cx}) / d_i^w\quad\quad&
\hat{g}_j^{cy} = (g_j^{cy} - d_i^{cy}) / d_i^h\\
\hat{g}_j^{w} = \log\Big(\frac{g_j^{w}}{d_i^w}\Big)\quad\quad&
\hat{g}_j^{h} = \log\Big(\frac{g_j^{h}}{d_i^h}\Big)
\end{split}
\label{eq:locloss}
\end{equation}

and $L_{conf}$ is given as,
\begin{equation}
L_{conf}(x, c) = - \sum_{i\in Pos}^N x_{ij}^p log(\hat{c}_i^p) - \sum_{i\in Neg} log(\hat{c}_i^0)\quad\text{where}\quad\hat{c}_i^p = \frac{\exp(c_i^p)}{\sum_p \exp(c_i^p)}.
\label{eq:confloss}
\end{equation}

\begin{figure}[ht]
    \centering
    \includegraphics[width=1.0\textwidth]{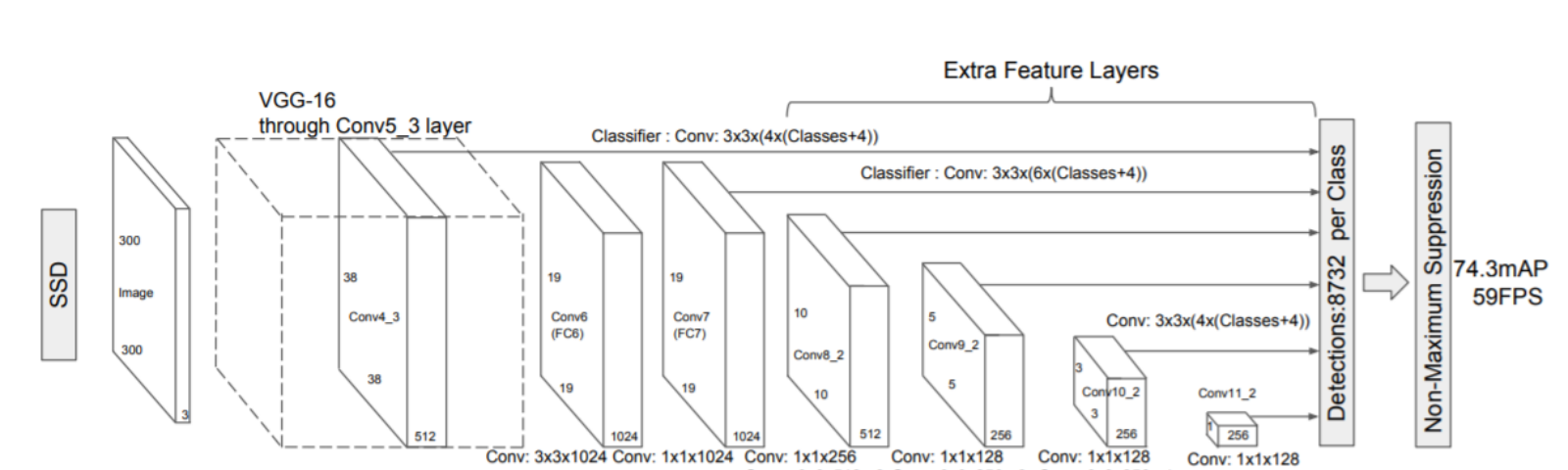}
    \caption{SSD network architecture \cite{liu2016ssd}.}
    \label{fig:ssd_net}
\end{figure}

The overall architecture designed by SSD is shown in Figure \ref{fig:ssd_net}.
        
 Most of the current approaches are following the {\it two-stage method}. Tian, \etal developed {\it Fully Convolutional One-Stage Object Detection} (FCOS) \cite{Tian_2019_ICCV}, this method developed an approach being anchor free. They use a multi-scale feature pyramid and a {\it non-maximum suppression} (nms) is applied as the last processing step.

Most of the datasets have many examples in "easy" scenarios, which means just the object isolated without much clutter or occlusion. So the "hard" scenarios are usually not learned since all samples have the same weight. It means that "easy" objects have the same importance during the learning process, and the algorithm is biased by the large number of "easy" samples. RetinaNet \cite{lin2017focal} developed a novel \textit{focal loss} which gives more weight for the hard cases. By doing so, it can learn the easy cases and also the hard ones.

From the object detection methods described, MS-COCO \cite{mscoco} is a dataset used for benchmarking. MS-COCO isa  large dataset developed by Microsoft for object detection and semantic segmentation research. In Table \ref{tab:od_coco}, we show the Average Precision (AP) (Equation \ref{eq:ap_ir}) for each method. We can see that Faster R-CNN is giving the best results. Decoupling the classification from detection, the network is able to learn robust features, without need to learn features related to location. The {\it two-stage} approach developed by Faster R-CNN can usually detect hard cases, especially small objects. On the other hand, FCOS have comparable AP, being faster and simpler. {\it One-stage methods} are progressively achieving better results over the {\it two-stage} ones.

\begin{table}[ht]
\centering
\caption{Comparison between object detection methods using MS-COCO \cite{mscoco} as benchmark}\label{tab:od_coco}

\begin{tabular}{l|c}
Method             & AP                          \\ \hline
Faster R-CNN & \textbf{39.4 \%}                     \\
SSD          & 31.2 \%                     \\
YOLO         & 33.0 \%                     \\
FCOS         & 39.1 \%                     \\
RetinaNet    & 37.7 \% \\ \hline
\end{tabular}
\end{table}

        \subsubsection{LiDAR}\label{sec:ls}

        Object recognition using LiDAR is widely researched. With the introduction of the KITTI dataset \cite{geiger2013vision}, computer vision tasks using LiDAR sensors (Velodyne HDL-64) became a popularly researched topic.
        
\begin{figure}[ht]
            \centering
            \includegraphics[width=0.7\textwidth]{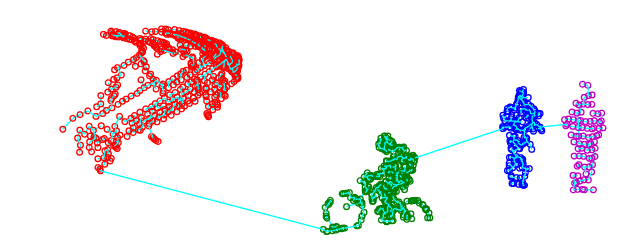}
            \caption{Example of point cloud segmentation result from the EMST-RANSAC algorithm \cite{wang2012could}.}
            \label{fig:wang2012could}
        \end{figure}

        Before the introduction of deep learning methods, to detect objects using LiDAR the first step was to create clusters from the point cloud that belong to the same object. Techniques based on euclidean distance such as k-means \cite{kanungo2002efficient} and {\it Density-based spatial clustering of applications with noise} (DBSCAN) \cite{ester1996density} are common ways of clustering. We can model the point cloud as a graph where the weight of a connection is the distance between the closest corresponding points. Graph-based clustering algorithms (\cite{shi2000normalized, kolmogorov2004energy, isack2012energy}) work effectively on LiDAR. Wang, \etal \cite{wang2012could} created a clustering algorithm based on the minimum spanning tree algorithm and using {\it Random Sample Consensus} (RANSAC) to remove outliers (Figure \ref{fig:wang2012could}).
        
\begin{figure}[ht]
            \centering
            \includegraphics[width=0.7\textwidth]{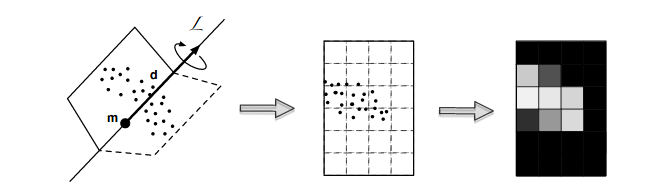}
            \caption{Spin Image Computation \cite{johnson1997spin}.}
            \label{fig:spin_image}
        \end{figure}

        After defining our cluster there are several techniques to extract the feature from the LiDAR. Probably one of the most used methods to represent a surface is the {\it Spin Images} methods developed by Johnson, \etal \cite{johnson1997spin, johnson1999using}. To compute the spin image a plane is computed based on the collection of points, and those points are projected to a discrete image plane. A 2D histogram is created where the intensity represents the number of points in the correspondent grid cell of the histogram. (Figure \ref{fig:spin_image}).
        
        Osada, \textit{et al} \cite{osada2002shape} developed a representation of shapes based on shape distributions. From a pair of points, a shape function based on euclidean distance computes the scalar for each pair. From the scalar a histogram is created which represents the shape.
        
        \begin{figure}[ht]
            \centering
            \includegraphics[width=0.7\textwidth]{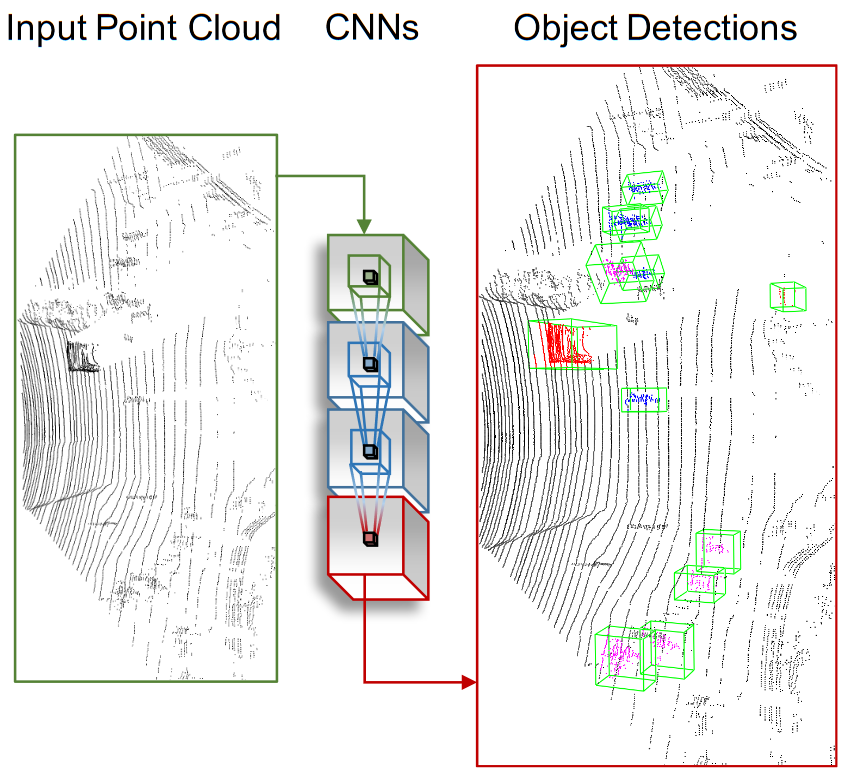}
            \caption{Convolutional Neural Network method designed for LiDAR \cite{engelcke2016vote3deep}.}
            \label{fig:cnn_lidar}
        \end{figure}
        
\begin{figure}[ht]
            \centering
            \includegraphics[width=\textwidth]{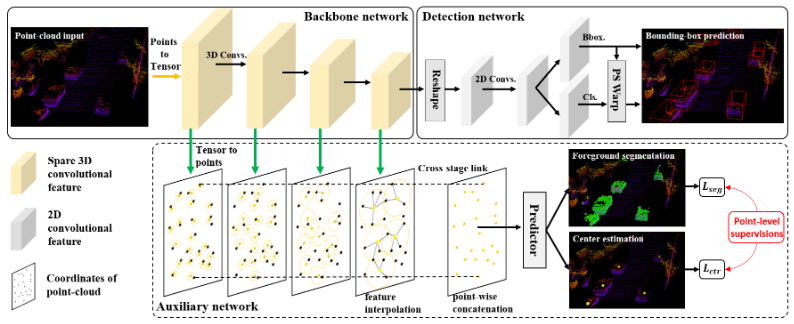}
            \caption{SA-SSD3D Methodology \cite{he2020structure}.}
            \label{fig:sassd}
        \end{figure}

        The current state-of-the-art algorithms for object detection and recognition for LiDAR in automotive scenario are based on deep neural networks. Papers like \cite{li2016vehicle, engelcke2016vote3deep} (Figure \ref{fig:cnn_lidar}) designed convolutional neural networks which detect and recognize vehicles. He, \etal developed SA-SSD3D \cite{he2020structure}. This method uses the SSD adapted to a 3D point cloud. It also uses an auxiliary network to convert the voxel used to a 3D point-cloud to improve the location of the predicted bounding box (Figure \ref{fig:sassd}). 
        
        The previously mentioned methods on LiDAR using neural networks create a voxel grid as input to a neural network, which may not be the best representation for LiDAR. To overcome these problems, methods that use the raw point-cloud were developed. The point-cloud can be seen as a set of points, or a graph structure. The PointNet method \cite{qi2017pointnet, qi2017pointnet++} gives a list of points as input. Since the order of the points matters to a neural network, PointNet used a neural network to reorder the list of points, which gives invariant order output.
        
\begin{figure}[ht]
            \centering
            \includegraphics[width=\textwidth]{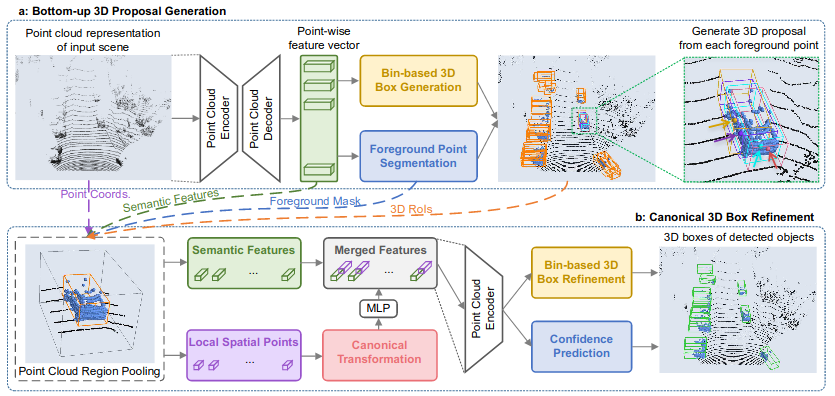}
            \caption{PointRCNN Methodology \cite{shi2019pointrcnn}.}
            \label{fig:pointrcnn}
        \end{figure}

        Shi, \etal \cite{shi2019pointrcnn} also uses the PointNet approach of giving the point cloud as input. It uses the {\it two-stage} approach from Faster-RCNN \cite{ren2017faster}, by using a proposal generation network together with the list of points strategy from PointNet \cite{qi2017pointnet} to be used as input (Figure \ref{fig:pointrcnn}). 
        
        \subsubsection{Infrared}\label{sec:ir_lr}
        
\begin{figure}[h!]
            \centering
            \includegraphics[width=\textwidth]{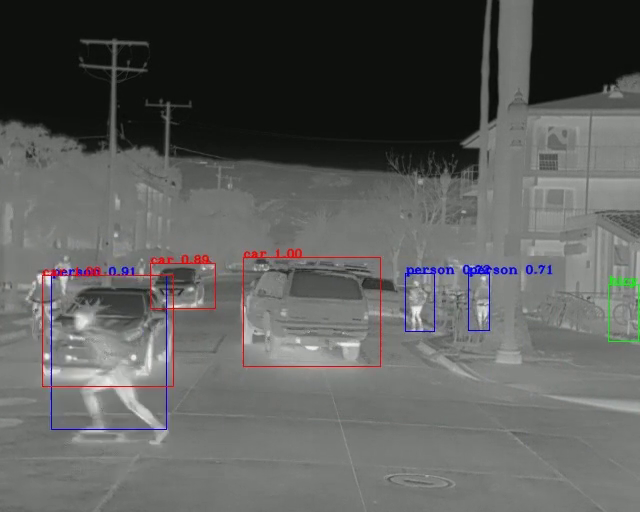}
            \caption{Example of Infrared image \cite{flir_dataset}. }
            \label{fig:flirdataset}
        \end{figure}

        Object recognition using infrared sensors is widely researched due to its capabilities of being robust when used in day and night scenarios. An object recognition system using infrared cameras can provide reliable perception, resilient to any kind of illumination. An example of infrared image can be seen in Figure \ref{fig:flirdataset}.
        
        Many features used to recognise objects in infrared sensors are the same as in RGB cameras, one of which is the shape. Infrared sensors will always get responses independent of the object color or texture, this being another advantage of infrared over RGB cameras.
        
         A 2D matrix with the scene temperature is the usual representation used in infrared cameras, however the raw sensor response is also used. Davis, \etal \cite{Davis2005twostage} developed a two-stage approach to detect pedestrians in infrared imagery. Potential regions are proposed by using contour saliency maps. In the potential regions, adaptive filters are used to extract contours. Those regions then are given to an AdaBoost classifier.
         
         Baek, \etal  \cite{baek2017efficient} developed an efficient algorithm to detect pedestrians based on a novel thermal-position-intensity histogram of gradients (TPIHOG). This model is classified by a {\it Support Vector Machine} (SVM).
        
\begin{figure}[h!]
            \centering
            \includegraphics[width=\textwidth]{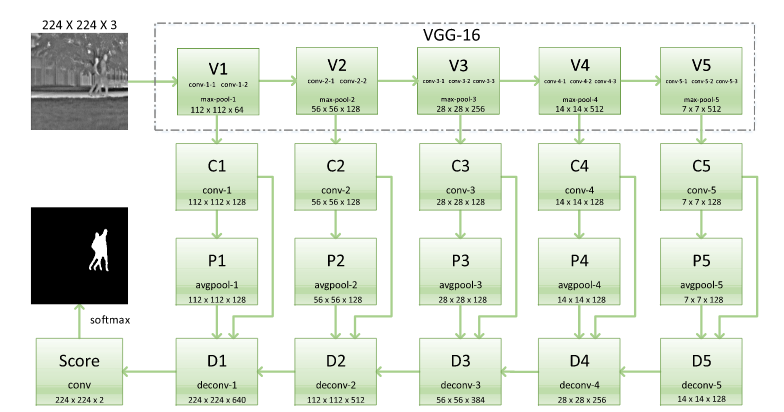}
            \caption{Diagram of the method developed by Zeng \etal \cite{zeng2018multiscale}. Each column of represents a network performing the detection in different scales. All detections are combined in the last row.}
            \label{fig:thermal}
        \end{figure}

        Zeng, \etal \cite{zeng2018multiscale} developed a new object detection system for infrared images, using the multiscale output from VGG-16. By exploiting the multiscale properties of the network, it achieves state-of-the-art foreground segmentation (Figure \ref{fig:thermal}).
        
        Abbott, \etal \cite{abbott2017deep} developed an object detection on infrared images based  on the YOLO method \cite{redmonyolo2016}, and transfer learning from a high resolution IR to a lower resolution IR (LWIR) dataset to detect vehicles and pedestrians.
        
\begin{figure}[h!]
            \centering
            \includegraphics[width=\textwidth]{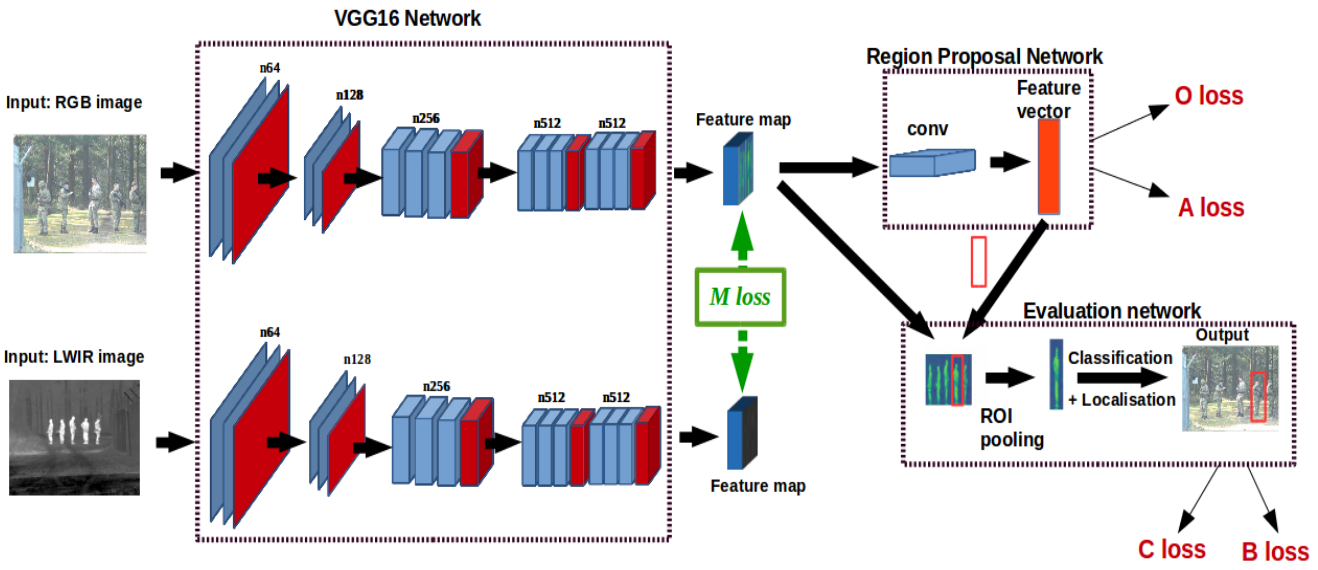}
            \caption{Methodology developed by Abbott \etal \cite{abbott2019multi}.}
            \label{fig:abbott2019multi}
        \end{figure}

        Abbott, \etal \cite{abbott2019multi} uses RGB and infrared cameras to recognise pedestrians (Figure \ref{fig:abbott2019multi}). It employs a loss function which uses output from two neural network from both sensors, thus leading to a better detection rate.
        
        \subsubsection{Radar}\label{sec:rs}
        
        Radar uses radio electromagnetic field to detect objects. It was widely used during the Second World War to detect enemy airplanes, missiles, boats and tanks.

                 During the Second World War, the received signals were converted into sound, so trained people recognized these sounds in order to classify the target. In military applications, the radar is either pointed at the sky to detect different airplanes and missiles, or pointed to the ground to detect ground-based targets.
                
                Radar object recognition is mostly researched for military applications due to the long range capabilities of radar sensors. With the increased development of the radar technology, radar was began to be used in automotive scenarios. Due to unique radar capabilities of detecting objects and its speed, radar is now widely used for automatic cruise control and collision avoidance.  
                
                We are going to review some of the classic target recognition methods in military scenarios and then explore literature on radar for automotive purposes.
                       
\begin{figure}[ht]
                    \centering
                    \includegraphics[width=0.7\textwidth]{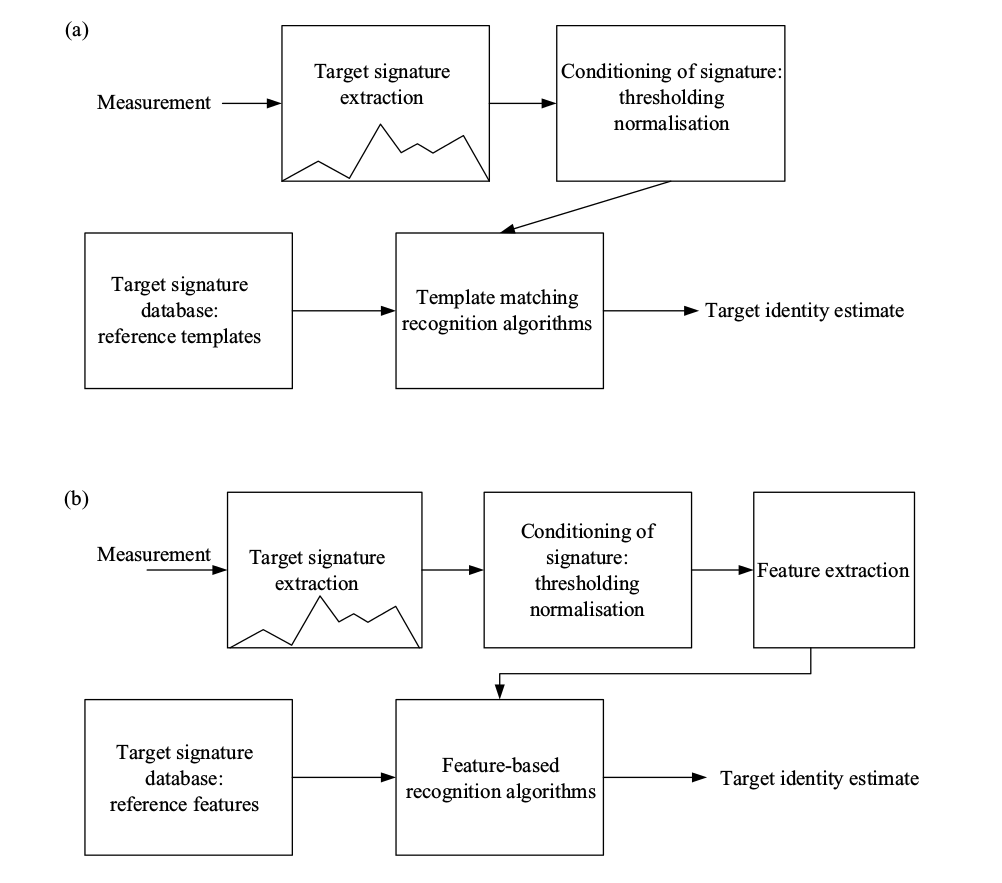}
                    \caption{Radar target recognition approaches. (a) template matching approach (b) feature-based approach \cite{tait2005introduction}.}
                    \label{fig:target_recognition_methodologies}
                \end{figure}

                The classic literature of radar target recognition defines two main methodologies (\cite{tait2005introduction}) (Figure \ref{fig:target_recognition_methodologies}) :
                
                \begin{itemize}
                    \item \textbf{Template matching}: It uses a database of signal measurements of many objects in order to determine the classification of the object. The classification is based on cross-correlation between the received signal and each measured signal in the database. The signal that is used can be the range profile, cross-range profile or the 2D image of range and cross-range.
                    
                    \item \textbf{Feature-based classification}: This approach extracts features from the signal, such as edges, distance between largest peaks, width of highest peak. After extracting the features, the decision process is based on some classification algorithm, such as Naive Bayes, k-Nearest Neighbours and Neural Networks.
                \end{itemize}
                
                The literature of automatic target recognition using radar can be divided into two categories:
                
                \begin{itemize}
                    \item {\bf Off-line Recognition}: This category uses a post-processing technique which uses several timestamps from radar sensing to build a high resolution image. {\it Synthetic Aperture Radar} (SAR) and {\it Inverse Synthetic Aperture Radar} (ISAR) techniques are the most common ones. This is the case for most remote sensing applications.

                    \item {\bf Online Recognition}: This category recognises the object with no future data provided. This category is usually used for the automotive scenario, since it needs to recognise objects in real-time. Forward looking SAR techniques are still an ongoing research topic. When performing online recognition, micro-Doppler is usually the main information that can lead to robust recognition \cite{angelov2019, fioranelli2015classification, fioranelli2015multistatic}. A pedestrian can produce micro-Doppler information by swinging their arms which is different from a car engine micro-Doppler signature.
                \end{itemize}

                Since the measured range profile is different, and influenced by different factors, such as distance of the object and aspect angle, there are some algorithms that try to minimize that influence. Cepstral analysis is usually used to address this problem - it reduces the dynamic range for the signal allowing it to be more consistent for better target signal signature \cite{bilik2006gmm}.
                
                For classifying ground targets in a military scenario, Bilik, \etal \cite{bilik2006gmm} uses Linear Prediction Coding (LPC) and Cepstrum in the Doppler frequency and extracts features from ground targets in synthetic aperture radar (SAR) images. They model the features in a Gaussian Mixture Model (GMM) and classify using Maximum Likelihood (ML).
                
                For air targets Kim, \etal \cite{kim2002efficient} uses Multiple Signal Classification (MUSIC) to detect point targets from airplanes using an inverse synthetic aperture radar (ISAR). The MUSIC algorithm gives a spectrum from the signal and it is resilient to some noise. The signal given by MUSIC is reduced using {\it Principal Components Analysis} (PCA) and the classification is given by a Bayes Classifier.
                
                SAR images are similar to natural images, so many techniques based on natural images are also used to classify targets on SAR. There are many papers on SAR that use shape information to extract features in order to classify targets \cite{wei2011efficient, saidi2008automatic, park2013new}.  To compare between SAR methods the MSTAR dataset \cite{ross1998standard} has a set $\approx 3000$ SAR images of 10 different ground targets rotated in 300 angles.
                
                Park, \etal \cite{park2014modified} generated features based on polar mapping which uses the object shape to generate a histogram based on locations. It has the second best result for the MSTAR dataset.
                
\begin{figure}[ht]
                \centering
                \includegraphics[width=0.7\textwidth]{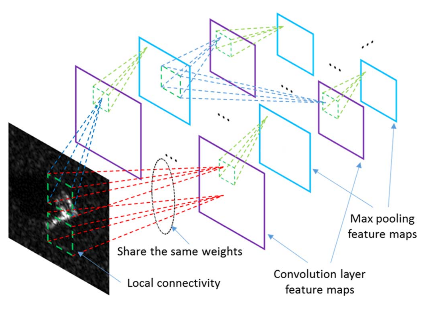}
                \caption{ Convolutional Neural Network developed to classify SAR images \cite{chen2016target}.}
                \label{fig:radar_cnn}
                \end{figure}

                The current state-of-the-art for SAR object recognition using the MSTAR dataset as reference is a paper based on Deep Convolutional Neural Networks \cite{chen2016target}. Chen, \textit{et al.} \cite{chen2016target} created an  all-convolutional neural network to classify targets (Figure \ref{fig:radar_cnn}). The all-convolution means that the neural network created consists only of convolutional layers. 
                
                As we can see, many approaches were used to classify targets in remote sensing scenarios. In these scenarios, especially for air targets, the received signal is sparse and the targets are isolated without influence of other objects. In an automotive scenario, however, there are several challenges. The beams reflect on the ground and other objects creating multi-path effects. In automotive scenarios, there are many types of different objects the radar is going to sense. Even extracting invariant features for some type of object, unwanted objects are going to be sensed all the time. So it is challenging to create an automatic target recognition from radars that will cover all types of scenarios for a reliable autonomous car system.
                
                Most current generation autonomous car systems are based on video and LiDAR sensors - these sensors can give us reliable recognition rates. However in adverse weather, these sensors show poor performance. Radar has a well known advantage of penetrating fog, rain and snow. Research focused on creating autonomous cars systems based on radar can lead car systems to achieve full autonomy in all weather scenarios.
                
                Automotive radar applications is not a new topic. There are a lot of publications that try to use radar in cars for intelligent systems. 
    
                Grimes, \etal \cite{grimes1974automotive} survey the field of autonomous radar. They discuss both the problems and promises of the radar technology. The paper focused on applications involving speed sensing, predictive crash sensing and obstacle detection. It also addressed the weather influence on the radar system showing its attenuation depending on the type of weather. Lastly, it showed that doing target recognition on radar is a promising goal.
                
\begin{figure}[ht]
                    \centering
                    \includegraphics[width=1\textwidth]{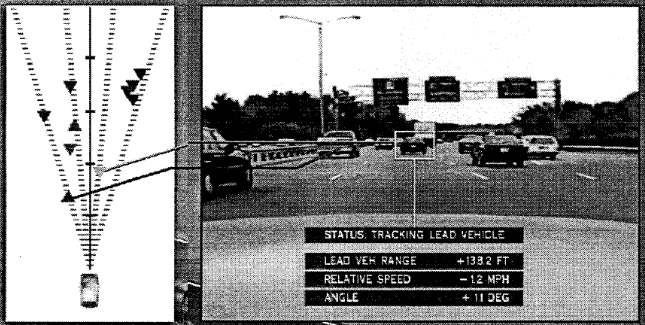}
                    \caption{Intelligent Cruise Control system developed by \cite{ganci1995forward}.}
                    \label{fig:ganci1995forward}
                \end{figure}

                Ganci, \etal \cite{ganci1995forward} created an  intelligent cruise control system which detects and tracks objects thus creating a collision warning system. It uses a Millimetre-wave (MMW) Frequency-Modulated Continuous Wave (FMCW) 77 GHz radar which both senses and does the computation in real-time (Figure \ref{fig:ganci1995forward}).
                
                Rasshofer, \etal \cite{rasshofer2007functional} investigates the functional requirements for radar in automotive scenarios. The paper addresses the requirements related to the cost, high-performance and system architecture. First it explores the possibility of radar imaging within the scope of current technology. The authors argue that the main issue is the poor the angle resolution, and the improvement of this technology would significantly increase the possibility of high resolution radars. They also take into consideration the exploration of low-THz bands to be used in an automotive scenario. It is described as promising, however the influence of rain, fog and snow have to be carefully considered during the development of such sensors.
                
                There is also research addressing the problem of target recognition in automotive scenarios by solely using radar sensors. For instance, \cite{rohling2010pedestrian} used a 24GHz radar to classify pedestrians. They analysed the Doppler spectrum and range profile from radar. By this analysis they made a classification by looking at the signal pattern.
                
                Bartsch, \etal \cite{bartsch2012pedestrian} used a 24 GHz Radar to classify pedestrians using the area of the object, shape of the object, and Doppler spectrum features to recognize pedestrians. They classified the pedestrians by analysing the probability of each feature and they use a simple decision model based on the features. They achieved 95\% classification rates for optimal scenarios, but this dropped to 29.4\% when the pedestrian appeared in a gap between cars due to low resolution from the radar sensors.
                
                Nordenmark, \etal \cite{Nordenmark2015} classified 4 classes (pedestrian, bicyclist, car and truck) using 4 short-range mono-pulse Doppler radars operating at 77 GHz. Their dataset is composed of one target per image. The detection was made using the Density-based spatial clustering of applications with noise algorithm (DBSCAN). Their feature extraction was based on the number of detections, minimum object length, object area, density, mean Doppler velocity, variance Doppler velocity, amplitude per distance and variance of amplitude. The classification was done by Support Vector Machines (SVM). They achieved an average of 95\% of accuracy.
                
                The lack of range and azimuth resolution of current radar technology makes it challenging to develop reliable target recognition rates for autonomous cars in all kinds of scenario. The current technology uses sensor fusion from LiDAR and video to achieve reliable recognition. However in adverse weather, LiDAR and video have bad performance. The development of high resolution radar sensors will lead to more reliable recognition system which can be used in all weather conditions. 
                
                The paper by Patole, \textit{et al.} \cite{patole2017automotive} reviewed the current state of automotive radar and its requirements. They described the current technology and the radar signal processing techniques to develop better resolution imaging. They proposed a 2D super resolution algorithm based on MUSIC.
                
                The papers \cite{8828004} and \cite{8792451} shows the current challenges on developing radar systems for autonomous cars. They showed how radar is emerging from a detection only system to a high resolution perception sensor. They describe how current radar systems are finally achieving high resolution.
                
                Baselice, \textit{et al}, \cite{baselice2016new} described a technique on 3D simulated radar data to produce a point cloud of the scattered points. This technique of imaging is based on compressive sensing.

\begin{figure}[ht]
                    \centering
                    \includegraphics[width=0.7\textwidth]{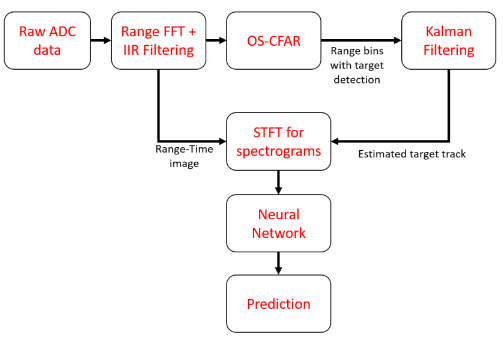}
                    \caption{Methodology developed by Angelov, \etal \cite{angelov2019}.}
                    \label{fig:angelov}
                \end{figure}

                More recently, many deep learning methods are being applied to radar. Angelov, \etal used a Doppler-time spectogram on different types of convolutional neural networks to recognise between pedestrian, bikes and cars. They also explored the use of temporal information by using LSTM, thus improving its recognition. An illustrative image from their methodology can be seen in  Figure \ref{fig:angelov}
                
                Palffy, \etal \cite{radarcube} employed a range-azimuth-Doppler tensor as input to CNN to classify cars, cyclists and pedestrians. They used a real world dataset and achieved 70\% F1-score on average.
                
                In the previous works location of the object was not considered. Major, \etal considered the location of vehicles and applied a range-azimuth-Doppler tensor as input to a convolutional neural network. The CNN used is SSD \cite{liu2016ssd}, which is the backbone for their methodology. They have a highway scenario and achieved 88\% AP.

                \begin{figure}[h]
                    \centering
                    \includegraphics[width=1\textwidth]{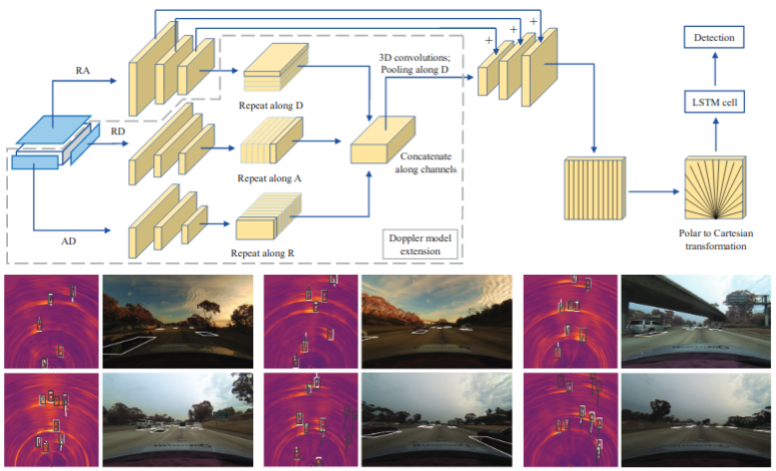}
                    \caption{Illustrative methodology developed by Major, \etal \cite{Major_2019_ICCV}.}
                    \label{fig:major}
                \end{figure}

                \begin{figure}[!h]
                    \centering
                    \includegraphics[width=0.45\textwidth]{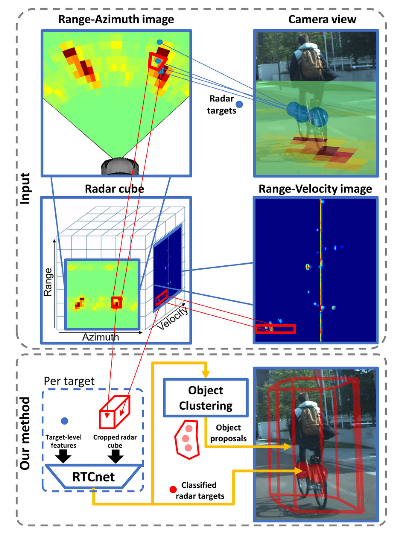}
                    \caption{Methodology employed by Palffy, \etal \cite{radarcube}.}
                    \label{fig:radarcube}
                \end{figure}

                As we stated before, all these previous methods on object recognition in automotive radar use Doppler information, which is an advantage over LiDAR. The lack of resolution is usually compensated through the use of Doppler features.
                
                In this thesis we explore object recognition in 150 GHz, 300 GHz and 79 GHz radar systems without using Doppler. These new sensors provide high resolution range-azimuth images which provide better spatial information.

                \subsection{Sensor Comparison}\label{sec:sensor_comparison}
                
                This chapter presented several aspects of sensing in the automotive context - such as radar, LiDAR, infrared and camera - and its applications on object recognition. Table \ref{tab:sensors_comparison} compares between those in different contexts. In this table we used specific sensor manufacturer models to represent the various sensor categories in different scenarios.
                \begin{itemize}
                    \item  For LiDAR we decided to use the Velodyne LiDAR HDL-64 as reference \cite{velodyne_lidar64}, which is the sensor used in KITTI \cite{geiger2012we}.
                    \item Texas Instruments AWR1443 79 GHz MIMO Radar  \cite{awr1443}, which is a common commercial type of radar system used in the automotive setting.
                    \item Navtech CTS350-X 79 GHz Scanning Radar \cite{navtech_radar}, which is a high resolution commercial radar. This sensor is used for vehicle detection  in Chapter \ref{ch:navtech2}.
                    \item ELVA-1 300 GHz radar, is used to represent current prototypical low-THz radar systems. This sensor is used in Chapters \ref{ch:300dl} and \ref{ch:radio} of this thesis.
                    \item ZED Camera is used to represent stereo cameras. This sensor is used as a reference camera image in Chapters \ref{ch:300dl}, \ref{ch:radio}, \ref{ch:navtech} and \ref{ch:navtech2}.
                \end{itemize} 
                
                \subsubsection{Technology}
                
                In terms of technology, LiDAR uses a visible light to measure range and it uses a scanning mirror to provide $360^o$ field of view. In contrast, radar uses radio waves to measure range. In order to measure azimuth information, it can use a MIMO or a scanning system. Infrared and stereo cameras are passive sensors which capture infrared and visible light respectively.
                
                \subsubsection{Maximum Range}
                
                Regarding maximum range, LiDARs and radars can provide up to 100 m. However, the current technology of the low-THz radar can provide range up to 20 meters \cite{daniel2017low}. This happens because it requires more power than commercial 79 GHz radar systems, and the current power given to the low-THz radar is just enough to sense up to 20 m. Stereo cameras provide range by finding correspondent pixels between the 2 images, and it provides range of up to 25 meters. Infrared sensors do not provide range information.
                
                \subsubsection{Range Resolution}
                
                The range resolution on LiDAR is accurate ($\pm 2$ cm) \cite{velodyne_lidar64}. Both 79 GHz and low-THz radar can also provide high range resolution, 17 cm and 0.75 cm respectively. The range resolution of stereo cameras depend on algorithm it is used for depth computation \cite{zed_camera}.
                
                \subsubsection{Azimuth Resolution}
                
                LiDAR provides very high azimuth resolution (horizontal) . This happens because the beamwidth from LiDARs are very narrow \cite{velodyne_lidar64} and its azimuth resolution is $0.1^o$. MIMO radars provide poor azimuth resolution $\approx 15^o$ \cite{awr1443}. Scanning radars can provide higher azimuth resolution compared to MIMO. 79 GHz scanning radar provides $1.8^o$ azimuth resolution and low-THz radar provides $1.1^o$
                
                \subsubsection{Elevation Resolution}
                
                Regarding elevation resolution (vertical), LiDAR provides accurate measurements ($0.4^o$). MIMO radar can provide elevation information ($20^o$) due to an additional antenna in the vertical position. The current scanning radar technology does not provide elevation.
                
                \subsubsection{Performance in darkness}
                
                LiDAR, radar and infrared can work without problems during dark scenarios. Since radar and LiDAR are active sensors, they are not affected by illumination. Infrared captures emitted radiation from objects which is the same for day and night scenarios. Since stereo cameras are passive sensors, which capture visible light - they will have poor performance in dark scenarios.
                
                \subsubsection{Performance in very bright light}
                
                As discussed previously, LiDAR, radar and infrared are robust for day and night situations. In very bright light scenarios, camera sensor can be over saturated, creating unwanted low dynamic range.
                
                \subsubsection{Performance in fog, rain and snow}
                
                Radars are well known for working in adverse weather conditions \cite{skolnik2001introduction}. In Chapter \ref{ch:navtech2} of this thesis we showed qualitatively, that LiDAR and stereo camera signals are very attenuated during fog, rain and snow. Low-THz radars are also very robust in adverse weather. However, since they work in higher frequencies, they are more prone to interference.
                
                \subsubsection{Speed detector}
                
                Radar can provide speed information by using Doppler, being widely used for adaptive cruise control. LiDARs are also capable of providing speed information through Doppler, however the Velodyne HDL-64e does not provide this. It should be pointed out that the manufacturers of the Navtech CTS350-X radar choose not to not provide speed information, as a compromise for a better resolution.
                
                \subsubsection{Cost}
                
                One main aspect to be considered is the sensor cost. LiDAR provides very high resolution 3D range, but at high cost (£60k). MIMO radars can be very cheap (£200), but lack resolution. Stereo cameras are an overall cheap sensor (£200), but they can come with the drawbacks that were previously discussed. Tesla cars use a combination of MIMO radars and stereo cameras, since they are a cheap solution and the fusion of both sensors can provide reliable information.
                
                \subsubsection{Object Recognition}
                
                Object recognition is a crucial part to enable vehicle autonomy. Stereo cameras can provide very high accuracy in recognising road actors. LiDAR can also provide high recognition rates. MIMO radar provides low azimuth resolution, being very challenging when recognising objects. Scanning radar technologies provide better azimuth resolution, being a viable solution for object recognition. Infrared cameras can also recognise objects accurately, since they give detailed shape information and robust infrared signatures in different scenarios.

\newcolumntype{a}{>{\columncolor{lightgray!25}}l}
\newcommand{\mycell}[1]{\begin{tabular}[c]{@{}c@{}}#1\end{tabular}}
\newcommand{\mycelll}[1]{\begin{tabular}[l]{@{}l@{}}#1\end{tabular}}

\begin{table}[ht]
\resizebox{\textwidth}{!}{\begin{tabular}{a|c|c|c|c|c|c}
\hline
\rowcolor{lightgray!55}Features                                                               &  \mycell{LiDAR\\Velodyne HDL64\\\cite{velodyne_lidar64}}                                                       & \mycell{79 GHz \\MIMO Radar\\TI AWR1443\\\cite{awr1443}}                                                                                                                           & \mycell{79 GHz\\Scanning Radar\\Navtech CTS350-X\\\cite{navtech_radar}}                                                                                                                & \mycell{Scanning\\low-THz Radar\\ELVA-1 300 GHz\\\cite{daniel2017low}}                                                                                                               & \begin{tabular}[c]{@{}c@{}}Passive\\Infrared\\Thales Catherine\\\cite{dickson2015long}\end{tabular}                       & \mycell{Stereo Camera\\ZED Camera\\\cite{zed_camera}}                                                                                         \\ \hline \hline
Technology                                                             & \begin{tabular}[c]{@{}c@{}}Uses light to\\ measure range\end{tabular} & \begin{tabular}[c]{@{}c@{}}Uses radio waves \\ (77-79 GHz)\\ to measure range.\\ Multiple antennas\\ to measure azimuth\end{tabular} & \begin{tabular}[c]{@{}c@{}}Uses radio waves \\ (77-79 GHz)\\ to measure range.\\ Uses a scanner\\ to measure azimuth\end{tabular} & \begin{tabular}[c]{@{}c@{}}Uses radio waves\\ (0.15-0.3 THz)\\ to measure range.\\ Uses a scanner \\ to measure azimuth\end{tabular} & \begin{tabular}[c]{@{}c@{}}Uses infrared\\ to capture\\ temperature\end{tabular} & \begin{tabular}[c]{@{}c@{}}Captures visible\\ light.\\ Uses 2 cameras\\ to capture range\end{tabular} \\ \hline
\mycelll{Maximum\\range}               & up to 100 m                                                           & up to 100 m                                                                                                                          & up to 100 m                                                                                                                       & up to 20 m                                                                                                                           & \xmark                                                                                & up to 25 m                                                                                       \\ \hline
\begin{tabular}[c]{@{}l@{}}Range\\ resolution\end{tabular}             & 2 cm                                                              & 17 cm                                                                                                                             & 17 cm                                                                                                                           & 0.75 cm                                                                                                                              & \xmark                                                                            & \mycell{depends\\on the\\algorithm}                                                                                               \\ \hline
\begin{tabular}[c]{@{}l@{}}Azimuth \\ resolution\end{tabular}          & $0.1^o$                                                               & $15^o$                                                                                                                               & $1.8^o$                                                                                                                           & $1.1^o$                                                                                                                             & \xmark                                                                                & \xmark                                                                                                     \\ \hline
\begin{tabular}[c]{@{}l@{}}Elevation \\ resolution\end{tabular}        & $0.4^o$                                                               & $20^o$                                                                                                                             & \xmark                                                                                                                                 & \xmark                                                                                                                                    & \xmark                                                                                & \xmark                                                                                                     \\ \hline
\mycelll{Performance\\in darkness}                                                          & Good                                                               & Good                                                                                                                              & Good                                                                                                                         & Good                                                                                                                              & Good                                                                          & Poor                                                                                               \\ \hline
\begin{tabular}[c]{@{}l@{}}Performance\\in very\\ bright light\end{tabular}   & Good                                                              & Good                                                                                                                         &Good                                                                                                                        & Good                                                                                                                              & Good                                                                        & Satisfactory                                                                                              \\ \hline
\begin{tabular}[c]{@{}l@{}}Performance\\in fog, rain\\and snow\end{tabular} & Poor                                                              & Good                                                                                                                             & Good                                                                                                                          & Good                                                                                                                              & Poor                                                                    & Poor                                                                                              \\ \hline
\begin{tabular}[c]{@{}l@{}}Speed \\ detector\end{tabular}              & \xmark                                                                    & \cmark                                                                                                                          & \xmark                                                                                                                         & \cmark                                                                                                                             & \xmark                                                                                & \xmark                                                                                                    \\ \hline
Cost & £60k & £200 & £15k& £100k & £30k & £300 \\ \hline 
\mycelll{Object\\Recognition}                                                     & Good                                                              &Poor                                                                                                                             & Satisfactory                                                                                                                        & Satisfactory                                                                                                                           & Good                                                                          & Good                                                                                               \\ \hline
\end{tabular}}
\caption{Comparing LiDAR, radar, infrared and stereo camera in different scenarios. For qualitative measurements, we used the terms \textit{poor, satisfactory} and \textit{good} to evaluate the performance (refer to in-text descriptions of performance for more details).}
\label{tab:sensors_comparison}
\end{table}

%% file: Chapters/Chapter3_infrared.tex
\chapter{Polarised Long Wave Infrared Vehicle Detection}\label{chapter:pol_lwir} %

\lhead{Chapter 3. \emph{Polarised Long Wave Infrared Vehicle Detection}} %

Long wave infrared (LWIR) sensors that receive predominantly emitted radiation have the capability to operate at night as well as during the day.
In this chapter, we employ a polarised LWIR (POL-LWIR) camera to acquire data from a mobile vehicle, to compare and contrast four different convolutional neural network (CNN) configurations to detect other vehicles in video sequences.
We evaluate two distinct and promising approaches, two-stage detection (Faster R-CNN) and one-stage detection (SSD), in
four different configurations.
We also employ two different image decompositions: the first based on the polarisation ellipse and the second on the Stokes parameters themselves.
To evaluate our approach, the experimental trials were quantified by mean average precision (mAP) and processing time, showing a clear trade-off between the two factors. 
For example, the best mAP result of 80.94 \% was achieved using Faster-RCNN, but at a frame rate of 6.4 fps.
In contrast, MobileNet SSD achieved only 64.51 \% mAP, but at 53.4 fps.

\begin{figure}[ht]
\centering
\includegraphics[width=1\textwidth]{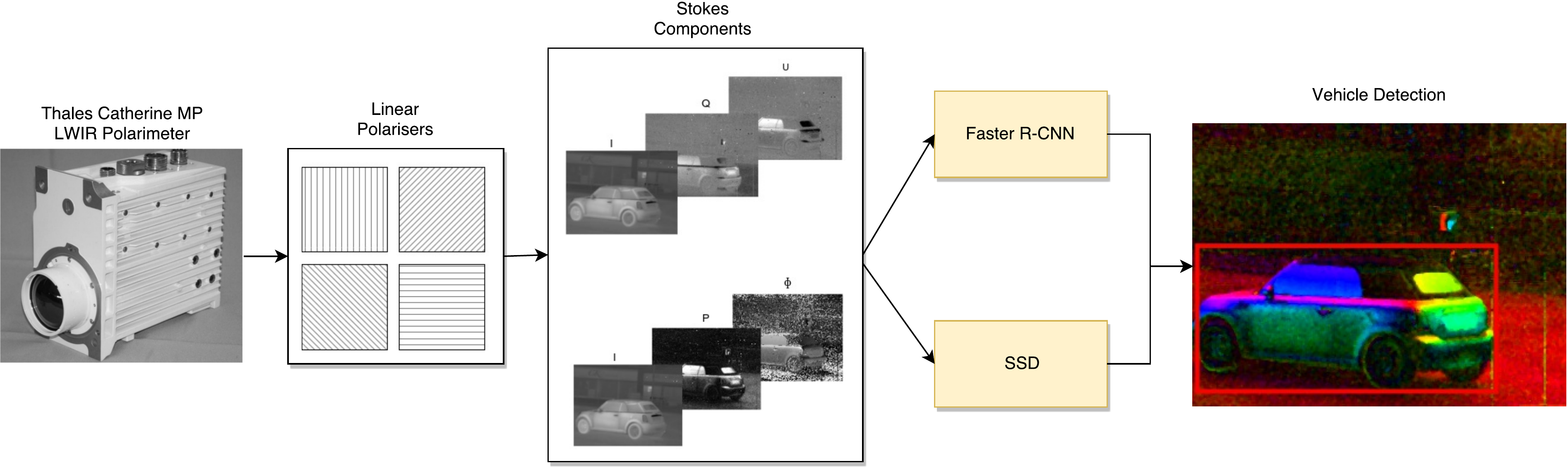}
\caption{POL-LWIR Vehicle detection. In this chapter we use a Thales Catherine MP LWIR sensor, which is based on long wave polarised infrared technology. It contains 4 linear polarisers ($i_{0}$, $i_{45}$, $i_{90}$, $i_{135}$). From the linear polarisers we can compute the Stokes components $I$,$Q$,$U$,$P$ and $\phi$.   Two configurations are created ($I$,$Q$,$U$ and $I$,$P$,$\phi$), which are passed to 2 types of neural networks: Faster R-CNN \cite{ren2017faster} and SSD \cite{liu2016ssd}. The networks are trained to detect vehicles in both day and night conditions.}
\label{fig:pol-ir}
\end{figure}

\section{Introduction}

A necessary capability for autonomy is sensory perception,
but the vast majority of research is based on publicly available video benchmarks like KITTI \cite{geiger2013vision} and CityScapes \cite{cordts2016cityscapes}.
These datasets are acquired during daytime, in good weather conditions, using video cameras.
For full autonomy and situational awareness in all weather conditions, sensors and perceptual algorithms workable continuously in 24 hours capability are required.
Infrared sensors are capable of sensing beyond the visible spectrum and are robust to poor illumination.

The majority of previous works, e.g. \cite{lin2015ATR_IR} \cite {khan2014ATR}, have used IR sensors in a military or surveillance context to detect \emph{hot} objects, especially during night.
From the perspective of a commercial road vehicle, IR sensors have also been used at night to detect other actors such as vehicles and pedestrians \cite{kwakIRPed2017}.
However, it has long been recognised that additional analysis of polarisation state,
governed by the material refractive index, surface orientation and angle of observation,
can lead to better discrimination.
False colour representations of the polarimetric data can visually reveal hidden targets \cite{wong2012novel} and to some extent, 3D structure \cite{lavigne2011target}. 
By exploiting knowledge about the dependence of polarisation state on surface and viewing angle, and the
fixed vehicle to road geometry, Connor \textit{et al.} \cite{connor2011scene} constructed a road segmentation
algorithm.
Bartlett \textit{et al.} \cite{bartlett2011anomaly} used polarisation to improve anomaly detection.
Using a nearest neighbour detection based on Euclidean distance, foreground versus background clustering is performed. 
The points whose distance from background components are higher than a threshold are labeled anomalous.
Romano \textit{et al.} \cite{romano2012day} also used polarisation for anomaly detection.
A data cube was formed from the primary ($i_0$,$i_{90}$) imagery and the sample covariance within a local (sliding) window is compared to the sample covariance of the entire image.
They found this method is better at discriminating between target and background  pixels at different times of the day than using Stokes components.

With its dramatic increase in popularity, there have been a number of recent studies based on deep neural networks to recognise objects in IR images. 
For example, Rodger \textit{et al.} \cite{rodger2016classifying} used CNNs to classify pedestrians, vehicles, helicopters, airplanes and drones using a LWIR sensor.
Abbott, \textit{et al.} \cite{abbott2017deep} used the YOLO method \cite{redmonyolo2016} and transfer learning from a high resolution IR to a lower resolution IR (LWIR) dataset to detect vehicles and pedestrians.
Lie \textit{et al.} \cite{liu2016multispectral} used the KAIST dataset \cite{hwang2015multispectral} to fuse RGB and thermal information in a Faster R-CNN architecture, again to detect pedestrians.
Gundog, \etal \cite{Gundogdu2017ATR} combined a CNN detection stage with a long term correlation tracker to detect
tanks in cluttered backgrounds.
However, all of these studies only used intensity data.

Dickson \textit{et al.} \cite{dickson2015long} exploited a polarised LWIR sensor to detect vehicles in both still images and to a lesser extent, video sequences.
In rural settings, LWIR emissions from man-made objects appear more strongly polarised than
vegetation. However in urban settings, most of the environment is man-made.
Therefore, in their work on vehicle recognition \cite{dickson2015long}, the key observation was that although
there are many other man-made structures in urban scenes, there is a distinct, differential spatial arrangement of surface signatures in LWIR polarimetric images of vehicles due to their
regular structure and size.
This leads to a regular pattern of pixel clusters in a 2D space encoding the degree and angle of polarisation.

In this chapter, our contribution is to evaluate the effectiveness of CNNs to detect vehicles in polarised LWIR data.
We evaluate the two main research directions in deep learning for object detection: two-stage detection, which first proposes the bounding boxes then performs classification in each bounding box based on Faster R-CNN \cite{ren2017faster}, and one-stage detection which detects and classifies in a single network,
based on Single-Shot Multibox Detection (SSD) \cite{liu2016ssd}.
To the best of our knowledge, this is the first work to exploit the use of polarised infrared together with neural networks for object detection.
  
\section{Methodology}

\subsection{Sensing and the Polarisation Parameters}
\label{sec:polir}

Figure~\ref{fig:pol-ir} illustrates a schematic diagram of our approach.
We use the Thales Catherine MP LWIR Polarimeter \cite{dickson2015long}, operating in the range of
$8\mu m$ to $12\mu m$ to record video images.
Each pixel of a $ 320 \times 256 $ image frame has $2 \times 2$ sensing sites that contains linear polarisers oriented at $0$, $45$, $90$ and $135$ degrees.
The data capture rate is 100 frames per second (fps).
The dataset was collected in Glasgow, UK on $14^{th}$ and $15^{th}$ of March, 2013 \cite{dickson2015long}.
Seven sequences are recorded and the bounding boxes of the vehicles are annotated, from which we use 4 sequences for training and 3 for testing.
In total, we have 10,659 annotated frames for training and 4,453 annotated frames for testing.

 The polarisation  state  of  the emitted LWIR radiation can be expressed in terms of the Stokes vector \cite{azzam1977ellipsometry}, which is given as

\begin{equation}
    S = \begin{bmatrix}
    I\\Q\\U\\V
    \end{bmatrix}.
\end{equation}

The Thales Catherine has a set of 4 polarisers with $0^o,45^o,90^o$ and $135^o$ ($i_0,i_{45},i_{90},i_{135}$). The $I$ component measures the total  intensity;  the $Q$ and $U$ components describe the radiation polarised in the horizontal direction and in a plane rotated $45^0$ from the horizontal direction, respectively; the $V$ component describes the amount of right-circularly polarised radiation. $I,Q,U$ and $V \in \mathbf{R}$. 
To measure the $V$ component we require an additional quarter wave-plate. As a result we can only measure $I,Q$ and $U$.
Therefore, with respect to the measured intensities at each pixel site, we deduce that

\begin{gather}
I = \frac{1}{2}(i_{0} + i_{45} + i_{90} + i_{135})\\
Q = i_{0} - i_{90}\\
U = i_{45} - i_{135}
\end{gather}

\noindent The degree of linear polarisation, $P$, is the intensity of the polarised light, given as

\begin{gather}
P = \frac{\sqrt{Q^2 + U^2}}{I}
\end{gather}

And the angle of polarisation, $\phi$, can also be calculated as follows

\begin{gather}
\phi = \frac{1}{2}tan^{-1}\Bigg(\frac{U}{Q}\Bigg).
\end{gather}

$i_0,i_{45},i_{90},i_{135}$ represents an 2D array of sensors, which $I,Q,U,P$ and $\phi$ are computed forming a 2D image which serves as input to the neural networks defined in Sections \ref{sec:faster} and \ref{sec:ssd}.

\subsection{Faster R-CNN}\label{sec:faster}

The Faster R-CNN method \cite{ren2017faster} relies on a two-stage object detection procedure.
First, a sub-network is used to propose the bounding boxes; second, a separate sub-network
is used to classify objects within each bounding box. 
The idea of Faster R-CNN evolved from R-CNN \cite{girshick2014rich}, which proposes several bounding boxes based on the selective search algorithm \cite{uijlings2013selective}.
Selective search applies a segmentation algorithm in many stages to under and over segment the image.
Bounding boxes are proposed at each region segment; these are inputs to a CNN to classify the type of object.
This CNN can be chosen from popular successful architectures, such as VGG \cite{Simonyan14c},
InceptionNet \cite{szegedy2015going} or ResNet \cite{he2016deep}.

Since selective search usually outputs a large number of regions ($\sim$2,000 regions), it is computationally expensive.
Fast R-CNN \cite{girshick2015fast} reduces this complexity by running a CNN over the whole image.
Proposed regions are transformed to the last feature map before the fully connected layers and the regions in the feature map are classified in a simple neural network, resulting in a significant computational complexity reduction. 
Despite such improvement, the authors realised that the selective search is indeed a bottleneck preventing faster execution of the overall algorithm.
As a solution, Faster R-CNN \cite{ren2017faster} created a network to learn how to generate bounding boxes.
This pipeline is also known as region proposal network (RPN).
The RPN creates a grid in the original image which anchors the bounding box annotations to the map.
Using the previous annotations of the bounding box in the original image, the RPN learns how to propose bounding boxes. RPN can also reduce the number of proposals compared to selective search.
Replacing selective search with the RPN improved both speed and accuracy (using the PASCAL VOC 2012 dataset).

Since it is easy to plug any CNN into the Faster R-CNN method, we decided to use ResNet-50 and ResNet-101 \cite{he2016deep}, where the number attached to the name of the network relates to the number of layers used.
ResNet uses residual layers that are CNNs with “shortcut connections”.
Those connections skip the current layer and the skipped output is added to the output after the convolution is applied.
ResNet has a trade-off between accuracy and depth of the network: the smaller the network is, the faster it performs.

\subsection{Single Shot Multi-box Detection}\label{sec:ssd}

Single Shot Multi-box Detection (SSD) \cite{liu2016ssd} uses one-stage detection,
in which the output of a single network is a set of bounding boxes with the respective classes.
This is different from Faster R-CNN which has two stages, the region proposal and the classification stages.

The use of one-stage detection attracted the attention of many researchers in the field.
Sermane \textit{et al.} \cite{sermanet2014overfeat} used a CNN over the whole image, where each cell of the last feature map corresponds to a region in the original image.
The regions are always uniformly distributed over the image, which constitutes a major disadvantage as there is
no {\emph{a priori}} reason why this should be the case.
The YOLO network \cite{redmonyolo2016} created a CNN which is trained using anchor boxes from the annotation (similar to RPN from Faster R-CNN) to output the region location plus its classification.
SSD is similar to YOLO in the sense that it uses a CNN to output the region's location and its classification result.
However, SSD generates the output at several scales of the feature map produced by the convolution.
The output maps are based on a grid of anchor boxes, as with Faster R-CNN. 
All the results at several scales are then combined, followed by a non-maximum suppression step to remove
multiple detections of the same object.

The architecture of SSD is based on convolutional stages of other networks,
such as VGG \cite{Simonyan14c} and InceptionNet \cite{szegedy2015going} and MobileNet \cite{howard2017mobilenets}.
InceptionNet (GoogleNet) was the winner of the ImageNet 2014 competition for image classification.
This network applies convolutions of different sizes ( $1 \times 1$, $3 \times 3$, $5 \times 5$) in the same layer and concatenates them into a single feature map.
These convolutions are called "Inception modules", and when done several times create a deep network of inception modules. 
This showed that applying several convolutions to the same feature map can retrieve more robust features.
MobileNet \cite{howard2017mobilenets} was designed to be a fast and small CNN to run on low powered devices.
It is a convolutional layer approximator;
Instead of applying $N \times M \times C$ convolutions, it first applies a $1 \times 1 \times C$ convolution to reduce the size of the input to $W \times H \times 1$, then applies a $N \times M \times 1$ convolution (where $N$ and $M$ are the convolution mask sizes, $W$ and $H$ are width and height of the image and $C$ is the number of channels).
This strategy reduces the number of weights to be learned by the network and the complexity of the convolution.
SSD with MobileNet is used in our experiments to evaluate how a small network can learn the polarised infrared features.

\begin{figure}[ht]
\begin{center}
\begin{subfigure}{0.8\textwidth}
\includegraphics[width=\textwidth]{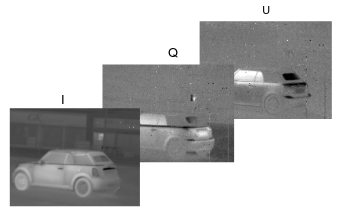}
   \caption{$I,Q,U$ configuration.}
\label{fig:iqu}
\end{subfigure}%
\end{center}

\begin{center}
\begin{subfigure}{0.8\textwidth}
\centering
\includegraphics[width=\textwidth]{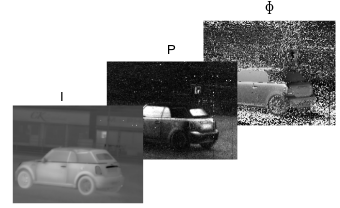}
   \caption{$I,P,\phi$ configuration.}
\label{fig:ipphi}
\end{subfigure}%
\end{center}
\caption{Visualisation of each configuration based on the Stokes components.}
\end{figure}

\subsection{Experiments, Training and Evaluation}\label{sec:ir_eval}

We have evaluated two configurations of the measured polarised image data to train and test
our several CNN architectures.
The first configuration uses the $I,Q,U$ parameters as the input image planes. 
In Figure~\ref{fig:iqu}, we can visualise the image of each component $I,Q,U$.
Again, in Figure~\ref{fig:ipphi} we can visualise the image of each component $I,P,\phi$.
We use four configurations of neural network for our experiments.

\begin{enumerate}
\item 
The SSD network using the InceptionV2 network \cite{szegedy2015going} to extract features.
\item
The SSD network using the MobileNet network \cite{howard2017mobilenets} to extract features.
\item
Faster R-CNN using ResNet-50 \cite{he2016deep} to extract features.
\item
Faster R-CNN using ResNet-101 \cite{he2016deep} to extract features.
\end{enumerate}

We trained our 4 different configurations on both ($I,Q,U)$ and $(I,P,\phi$) data, and
for comparison with previous work that has applied CNNs to intensity data alone,
on the $I$ data in isolation.
The networks are trained using a i7-7700HQ, 32 GB ram, NVIDIA Titan X and developed using the Tensorflow Object Detection API \cite{abadi2016tensorflow}.
The network weights for both SSD and Faster R-CNN are initialised from the MS-COCO object
detection dataset \cite{lin2014microsoft}.
The parameters for the Faster R-CNN networks are: batch size $1$, learning rate $0.0003$, momentum $0.9$.
The parameters for the SSD networks were: batch size $24$, learning rate $0.004$, momentum $0.9$.
Equation \ref{eq:sgd} shows the gradient descent formula.

\begin{equation}\label{eq:sgd}
W_{t+1} = W_{t} - \alpha \nabla f(x;W_t) + \eta \Delta W 
\end{equation}

\noindent
where $\eta$ is the momentum, $\alpha$ is the learning rate, $t$ is the time current time step, $W$ is all weights of the network and $\nabla f(x;W)$ is the gradient of the function that represents the network and $x$ is our dataset.

\begin{figure*}[ht]
    \centering
    \begin{subfigure}[t]{0.5\textwidth}
        \centering
        \includegraphics[width=\textwidth]{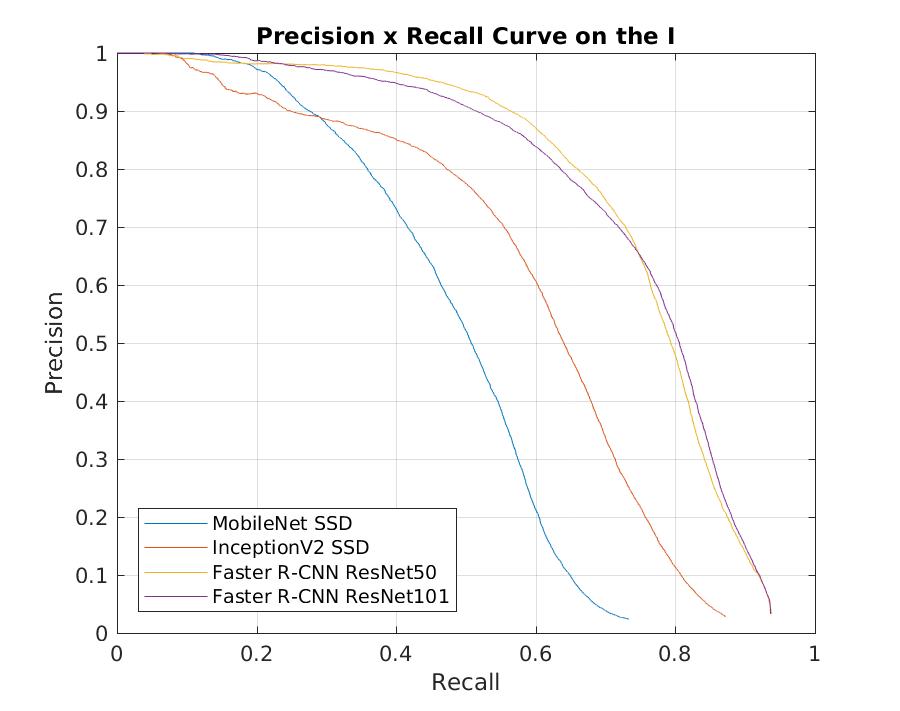}
        \caption{Precision-Recall curve for the $I$ configuration.}\label{fig:pr_i}
    \end{subfigure}%
    ~ 
    \centering
    \begin{subfigure}[t]{0.5\textwidth}
        \centering
        \includegraphics[width=\textwidth]{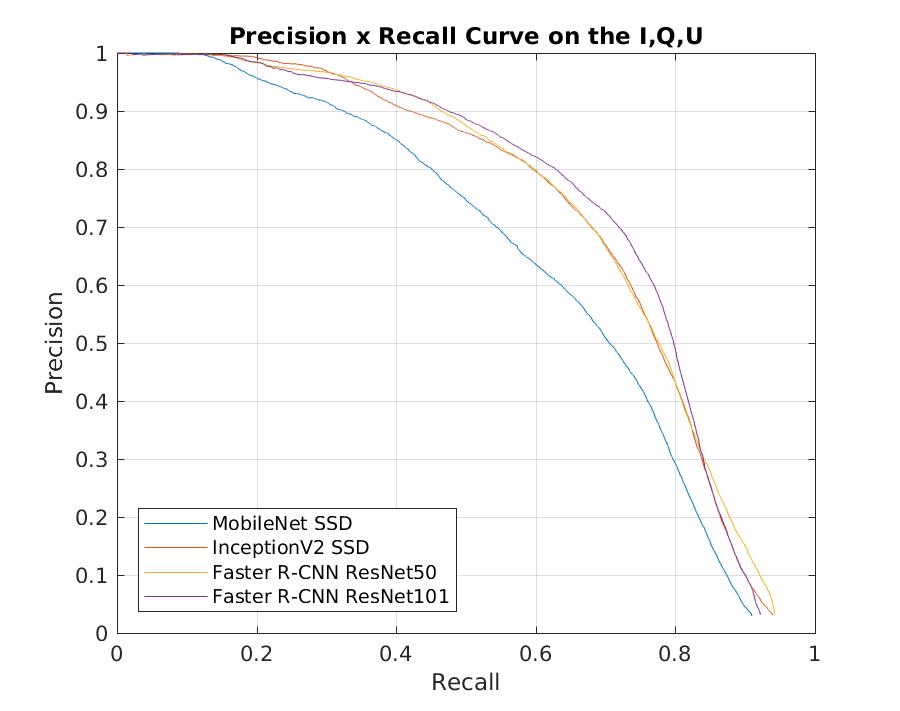}
        \caption{Precision-Recall curve for the $I,Q,U$ configuration.}\label{fig:pr_iqu}
    \end{subfigure}%
    
    \begin{subfigure}[t]{0.5\textwidth}
        \centering
        \includegraphics[width=\textwidth]{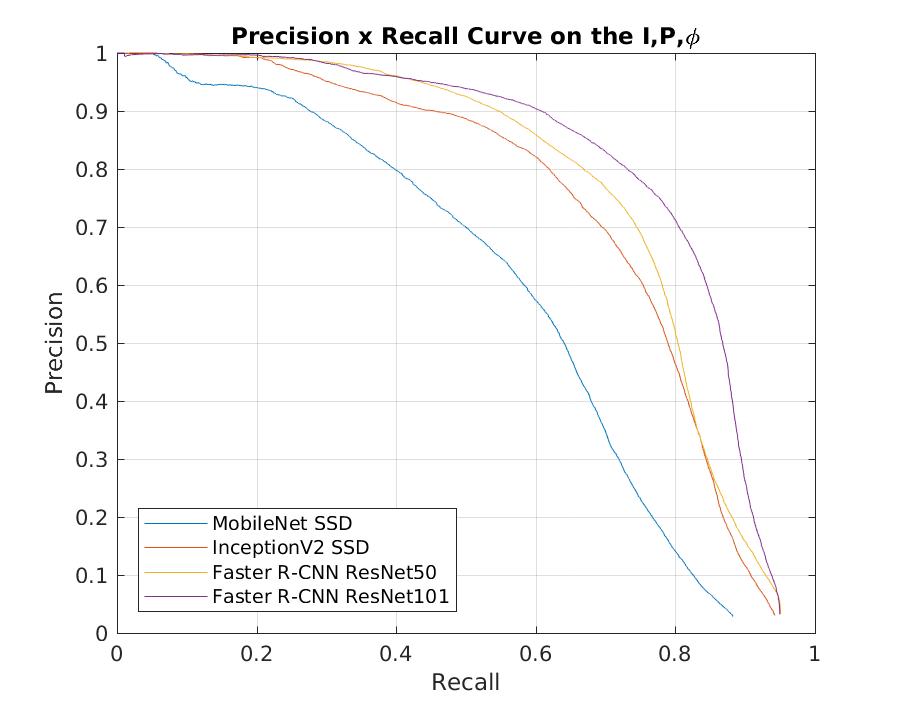}
        \caption{Precision-Recall curve for the $I,P,\phi$ configuration.}\label{fig:pr_ipphi}
    \end{subfigure}
    \caption{Precision-Recall curves}
\end{figure*}

The evaluation metrics used are mean average precision (mAP) and processing time in frames per second (fps).
The mAP classifies correct detection when Intersection over Union (IoU), $ > 0.5$, which follows the PASCAL VOC
protocol \cite{everingham2015pascal}. IoU measures the intersection between the predicted bounding box ($B_{p}$) and the ground truth ($B_{gt}$).  Equations \ref{eq:iou_ir}, \ref{eq:prec_ir}, \ref{eq:rec_ir} and \ref{eq:ap_ir} shows how mAP is computed. 
(The KITTI protocol \cite{geiger2013vision} uses $IoU > 0.7$ for vehicles. However, since unlike KITTI, our annotations are not pixel-level, we followed the PASCAL criteria). In the Equations \ref{eq:prec_ir}, \ref{eq:rec_ir} and \ref{eq:ap_ir}, TP (True Positive) means when a object was correctly detected, FN (False Negative) means that the object was not detected, FP means that the object was detected somewhere else. If there are more than 1 bounding box detecting the object, just one counts as TP, the other bounding boxes count as FP.
To compute the fps we compute the average time over 100 frames. 
Tables \ref{tab:results} and \ref{tab:speed} show the results for each configuration.
Precision-recall curves are also generated for each results and can be visualised
in Figure~\ref{fig:pr_i} for $I$ alone, Figure~\ref{fig:pr_iqu} for $I,Q,U$ and Figure~\ref{fig:pr_ipphi} for $I,P,\phi$.

\resizebox{\columnwidth}{!}{\begin{tabularx}{\textwidth}{XXX}
\begin{equation}\label{eq:iou_ir}
  IoU = \frac{B_p \cap B_{gt}}{B_p \cup B_{gt}}
\end{equation}
  &
\begin{equation}\label{eq:prec_ir}
      Prec = \frac{TP}{TP+FP}
\end{equation} 
&
\begin{equation}\label{eq:rec_ir}
     Rec = \frac{TP}{TP + FN} 
\end{equation}
\label{eq:prec_recall_ir}
\end{tabularx}}

\begin{equation}\label{eq:ap_ir}
     AP = \frac{1}{11} \mathlarger{\mathlarger{\sum}}_{r \in\{ 0,0.1,...1\}} p_{interp}(r) \qquad p_{interp}(r) = \max_{\tilde{r}:\tilde{r}\geq r}(p(\tilde{r}))
\end{equation}

\begin{table}[ht]
\centering
\caption{Results for each configuration}\label{tab:results}
\begin{tabular}{lccc}
                       & mAP [I] & mAP [I,Q,U] & mAP [I,P,$\phi$]   \\ \hline \hline
MobileNet SSD          & 48.50 \%  & 64.51 \% &58.56 \%     \\       \hline    
InceptionV2 SSD        & 59.79 \%  & 72.17 \%&73.24 \%      \\     \hline  
Faster R-CNN Resnet-50 & 75.63 \% & 72.82 \% &76.43 \%      \\ \hline
Faster R-CNN Resnet-101 & 75.21 \% & 73.67  \%   &\textbf{80.94} \%        
\end{tabular}
\end{table}

\begin{table}[h]
\centering
\caption{Computational speed (fps) for each network configuration.}\label{tab:results_speed}\label{tab:speed}
\begin{tabular}{l|c}
                       & fps \\ \hline
MobileNet SSD          & \textbf{53.4}    \\
InceptionV2 SSD        & 37.2     \\
Faster R-CNN Resnet-50 &  7.8    \\
Faster R-CNN Resnet-101 & 6.4
\end{tabular}
\end{table}

\begin{figure*}[ht]
\begin{center}
\includegraphics[width=\textwidth]{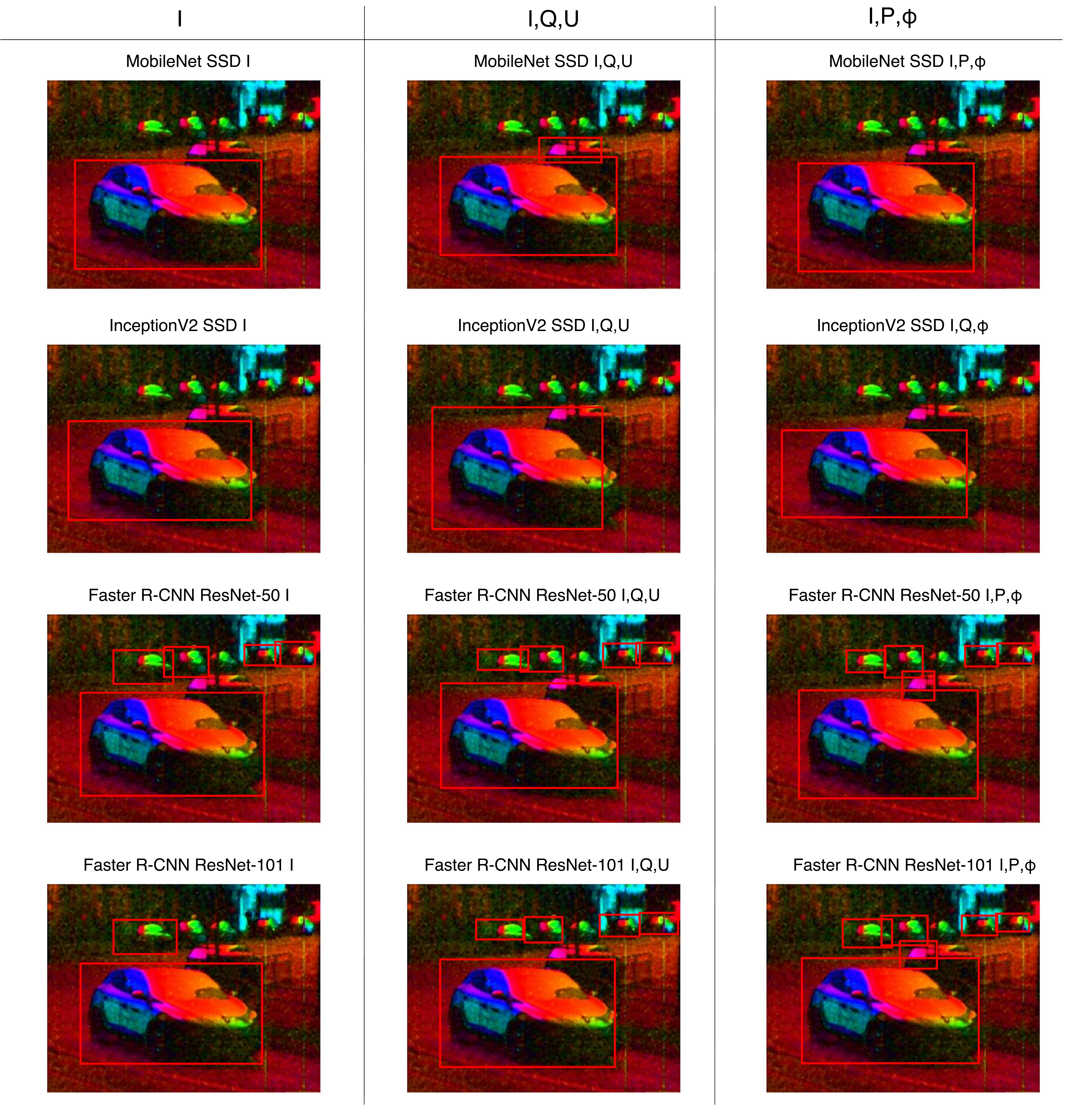}
\end{center}
   \caption{Qualitative results for each configuration with pseudo-probability threshold $>$ 0.7.}
\label{fig:quali}
\end{figure*}

Qualitative examples of images can be seen in Figure~\ref{fig:quali}.
This uses a pseudo-colour display, converting $P$ and $\phi$ to HSV colour space.
Based on qualitative results we can see that the main problem of the SSD lies with small objects.
The network needs to learn the location and features at the same time, which affects the detection of small objects.

Considering the results of Table~\ref{tab:results} and the precision-recall curves, a key question
is whether the use of polarised data improves detection when compared to the intensity data alone,
as used by most previous authors.
We believe that it is indeed the case, particularly for the SSD examples,
although we qualify that statement by noting that the difference in the R-CNN results
is less definitive, and that the dataset is limited and many more trials are needed
for full statistical verification.
In general, for all datasets, Faster R-CNN performs better for this limited trial, as measured by mAP, although this comes with the penalty of a much slower frame rate.
However, this latter result is consistent with the published results on video sequences, where splitting the
tasks of region proposal and classification arguably makes the network more robust in learning
directly the object features.
In contrast, SSD needs to learn the localisation together with classification, and hence
both location and object characteristics influence the weights of the network which can degrade performance.
Nevertheless, in our trial, SSD-InceptionV2 achieves similar performance to Faster R-CNN ResNet-50,
at a much increased speed, since it just needs one CNN for both region proposal and classification. Regarding the computational speed, 6.4 fps achieved by Faster R-CNN Resnet-101 can be potentially used for autonomous cars in urban scenarios. An ego-vehicle moving at 10 m/s will recognise other vehicles after moving for $\approx 1.5 $ meters, which is a reasonable distance for most situations. InceptionV2 SSD should be considered for high speed situations, even with lower accuracy, it is $\approx 6$ faster than Faster R-CNN ResNet-101. Regarding mAP, $> 70\%$ is enough for most autonomous cars situations. By looking at the qualitative results, we visualised that most of the mistakes are in hard scenarios, such as high occlusion, and when the vehicle is very far. 

Comparing $I,Q,U$ and $I,P,\phi$, the best result is obtained with the $I,P,\phi$ parametrisation. Although
the differences are not shown as significant, such that much more extensive characterisation is required.
At this stage, given the complexity of these neural networks, it is hard to define what type of feature
is being learned in each case, although from Dickson \textit{et al.} \cite{dickson2015long} the authors claim that material,
shape and surface and viewing angles influence the underlying polarisation patterns. 
As a specific example, one can see that the $I,P,\phi$ configuration does detect an occluded car that the $I,Q,U$ does not in the Faster R-CNN example. However, although this occurs more often than the converse, much better understanding of the network and more extensive trials are necessary to draw reliable conclusions.
For a necessary perception by an autonomous car, computational time is clearly quite a crucial factor.
As expected, the one-stage SSD architectures shows higher frame rates.
MobileNet SSD is the fastest and can process on average at 53.4 fps, but it has the lowest mAP.

\section{Conclusions}

In this chapter, we evaluated and compared a series of different CNN architectures for vehicle detection
in polarised long wave infrared image sequences, using two different image decompositions.
\begin{itemize}
    \item We showed that the use of polarised infrared data was effective for vehicle detection, and appeared to
perform better when CNNs are used for detection in infrared intensity data alone, confirmed also by previous researches \cite{dickson2013improving, dickson2015long}.
    \item Faster R-CNN based networks achieved better results in terms of detection accuracy, splitting the tasks of
region proposal and classification to  make the network more robust. 
However, it should be mentioned that improving the accuracy of one-stage detection network is quite an active field of research, providing much higher frame rates.
    \item We could reach no firm conclusion on which image decomposition was preferable, although anecdotally the $\{I, P, \phi\}$ parametrisation is both more intuitive in describing the polarisation ellipse and achieved the best overall result
with Faster R-CNN ResNet 101.
    \item  Our detection rates are similar to those of KITTI dataset for vehicle detection from simple video data in daylight using the same networks. Overall, our work shows that polarising LWIR data is a relatively robust
option for day and night operation.
\end{itemize}

In the next chapters we will focus on radar sensors for the development of object recognition system. Firstly using a 300 GHz radar, and develop an object recognition system based on deep neural networks and transfer learning.

%% file: Chapters/Chapter5_300ghz_DL.tex
\chapter{300 GHz Radar Object Recognition based on Deep Neural Networks and Transfer Learning}\label{ch:300dl}

\lhead{Chapter 4. \emph{300 GHz Radar Object Recognition based on Deep Neural Networks, Transfer Learning}} %

The main idea in developing an autonomous car perception system in bad weather relies on the use of radar sensors. Radar is capable of penetrating through rain, snow and fog, however it gives us poor resolution comparing to video and LiDAR.

In this chapter, we describe a methodology based on deep neural networks to recognise objects in 300 GHz radar images using the returned power spectra only, investigating robustness to changes in range, orientation and different receivers in a laboratory environment. As the training data is limited, we have also investigated the effects of transfer learning. As a necessary first step before road trials, we have also considered detection and classification in multiple object scenes.

\section{Introduction}

	Deep Neural Networks (DNNs) have proven to be a powerful technique for image recognition on natural images \cite{krizhevsky2012imagenet, szegedy2015going, Simonyan15}. 
	In contrast to manual selection of suitable features followed by statistical classification, DNNs optimise the learning process to find a wider range of patterns, achieving better results than before on quite complicated scenarios.
	For example, this includes the ImageNet challenge first introduced in 2009 \cite{deng2009imagenet}, which has at the time of writing more than 2000 object categories and 14 million images.
	
	In this chapter, we investigate the capacity of DNNs to recognise prototypical objects in azimuth-range power spectral images by a prospective 300 GHz automotive radar with an operating bandwidth of 20 GHz.
	The first contribution of our work is to assess the robustness of these DNNs to variations in viewing angle, range and the specific receiver operational characteristics, using a simple database of six isolated objects.
	For greater realism, our second contribution is to evaluate the performance of the trained neural networks in a more challenging scenario with multiple objects in the same scene, including detection and classification in the presence of both uniform an cluttered backgrounds.
	Third, since we have limited data, we have also investigated how transfer learning can improve the results. 
	\newtext{This 300 GHz prototype has limited range and scanning speed; therefore our experiments are conducted in a laboratory setting rather than from a mobile vehicle.
		Further, we avoid the use of range-Doppler spectra to classify images, but perform experiments using the radar power data alone.
		This is justifiable because future automotive technology must have the capability to classify traffic participants even when static, e.g. at traffic lights, although motion may be used as an ancillary variable to good effect. Also, we didn't consider use any temporal information in this work, this initial work only consider static scenarios without movement from objects.}
	
	\begin{figure}[t]
		\centering
		\includegraphics[width=\linewidth]{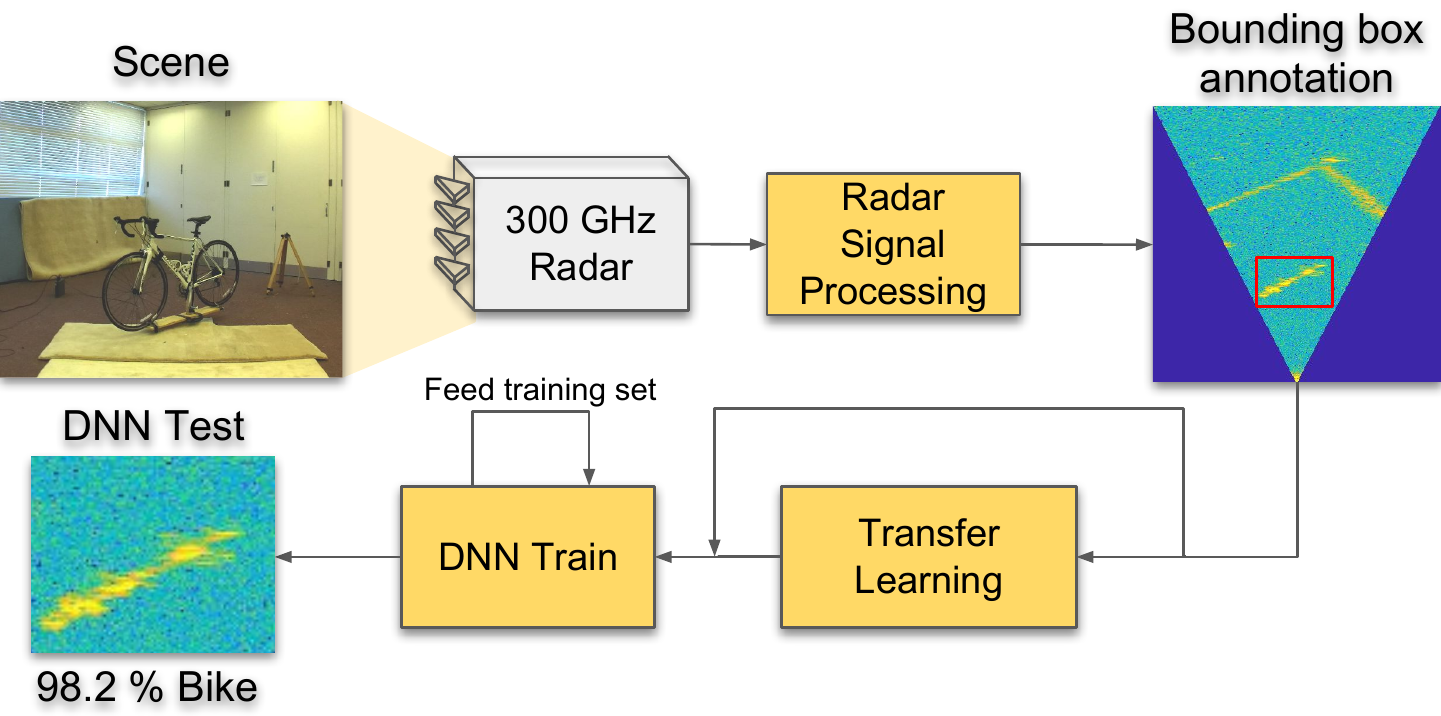}
		\caption{\textbf{300 GHz FMCW Radar Object Recognition:} Methodology developed using deep convolutional neural networks to process data acquired by a prototype high resolution 300 GHz short range radar \cite{Daniel2018}. Steps: 1. Radar signal processing and cartesian radar image generation. 2. Bounding box annotation to crop object region. 3. Deep neural network and transfer learning for radar based recognition.}
		\label{fig:methodology_300}
	\end{figure}

	The use of deep convolutional neural networks (DCNNs) \cite{lecun1995convolutional,krizhevsky2012imagenet} for large scale image recognition has changed significantly the field of computer vision. Although questions remain on verifiability \cite{huang2017safety}, confidence in the results \cite{gal2016dropout}, and on the effects of adversarial examples  \cite{nguyen2015deep}, the best results for correct identifications applied to large image datasets have been dominated by DCNN algorithms. The development of GPU's and large annotated datasets has helped the popularity of deep learning methods in computer vision.
	
    We wish to examine the potential of radar data for reliable recognition.
	This is especially challenging; most automotive radars sense in two dimensions only, azimuth and range, although research is underway to develop full 3D radar \cite{phi300_2017}.
	Although range resolution can be of the orders of $cm$, azimuth resolution is poor, typically $1-2$ degrees although again there is active research to improve this \cite{Daniel2018}.
	Natural image recognition relies to a great extent on surface detail, but the radar imaging of surfaces is much less well understood, is variable, and full electromagnetic modelling of complex scenes is extremely difficult.

	\section{Applying Deep Neural Networks to 300 GHz Radar Data}
	
	\subsection{Objective}
	
	The main objective of the first part of our study is to design and evaluate a methodology for object classification in 300 GHz radar data using DCNNs, as illustrated schematically in Figure \ref{fig:methodology_300}. 
	This is a prototype radar system; we have limited data so we have employed data augmentation and transfer learning to examine
	whether this improves our recognition success.
	To verify the robustness of our approach,
	we have assessed recognition rates using different receivers at different positions, and objects at different orientations and range. We also evaluated the performance of the method in a more challenging scenario with multiple objects per scene.
	
	\subsection{300 GHz FMCW Radar}
	
\begin{figure*}
    \centering
    \includegraphics[width=\textwidth]{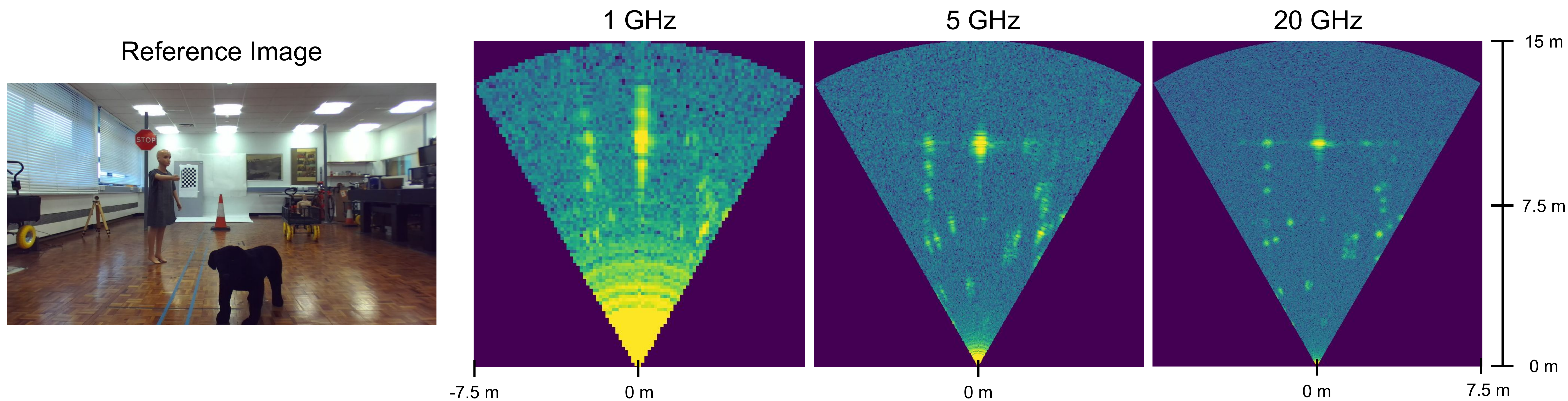}
    \caption{Images from the 300 GHz radar with different radar bandwidths.}
    \label{fig:radar_resolution}
\end{figure*}

\begin{figure*}
    \centering
    \includegraphics[width=\textwidth]{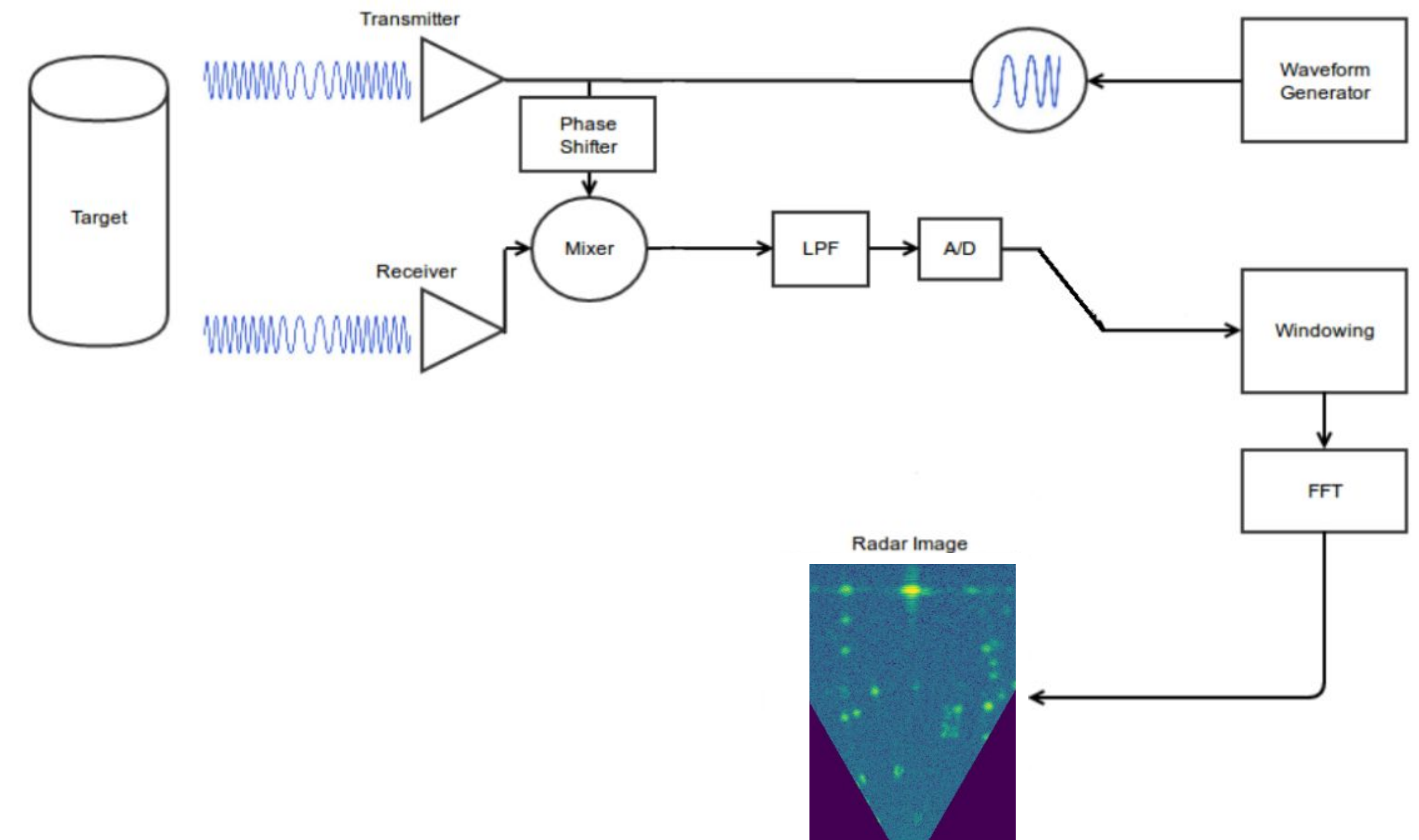}
    \caption{300 GHz radar circuit diagram.}
    \label{fig:radar_300ghz_diagram}
\end{figure*}

	A current, typical commercial vehicle radar uses  MIMO technology at 77-79 GHz with up to 4GHz IF bandwidth, and a range resolution of 4.3 cm, and an azimuth resolution of 15 degrees \cite{tiradar}. This equates to a cross range resolution of $\approx 4m$ at $15m$ such that a car will just occupy one cell in the radar image. This clearly makes object recognition very challenging on the basis of radar cross section. Rather, in this work, we collected data using a FMCW 300 GHz scanning radar designed at the University of Birmingham \cite{phi300_2017}. The main advantage of the increased frequency and bandwidth is a better resolved radar image which may lead to more reliable object classification. The 300 GHz radar used in this work has a bandwidth of 20 GHz which equates to 0.75 cm range resolution. The azimuth resolution is $1.2^o$ which corresponds to $20$ cm at 10 meters.  Figure \ref{fig:radar_resolution} shows 300 GHz radar images with different bandwidths. The 1GHz radar could be used for detection but not classification due to its lack of resolution. Even at 5GHz, classification is still quite challenging. The 20 GHz bandwidth provides much better resolution compared to current commercial sensors, and as such is considered for classification in the rest of this paper.
	The parameters for the 300 GHz sensor used in this work can be seen in Table \ref{tab:300ghz_radar} and the Figure \ref{fig:radar_300ghz_diagram} shows the circuit diagram designed used for the 300 GHz radar developed by the University of Birmingham.
	
	\begin{table}[h!]
		\caption{300 GHz FMCW Radar parameters for the system described in \cite{phi300_2017}.} \label{tab:300ghz_radar}
		\centering
		\begin{tabular}{r|c}
			\hline
			Sweep Bandwidth               & $20$ GHz                    \\ 
			H-Plane (Azimuth) beamwidth   & $1.2^o$                    \\ 
			E-Plane (Elevation) beamwidth & $7.0^o$            \\  
			Antenna gain                  & $39$ dBi                   \\ 
			Range resolution              & $0.75$ cm                     \\ 
			Azimuth resolution (at 10 m)  & $20$ cm                    \\ 
			Transmitter height & 0.695 m \\
			Receiver 1 height & 0.945 m \\
			Receiver 2 height & 0.785 m \\
			Receiver 3 height & 0.410 m \\
			
			\hline
		\end{tabular}

	\end{table}
	
\begin{figure}[t]
		\centering
		\includegraphics[width=\linewidth]{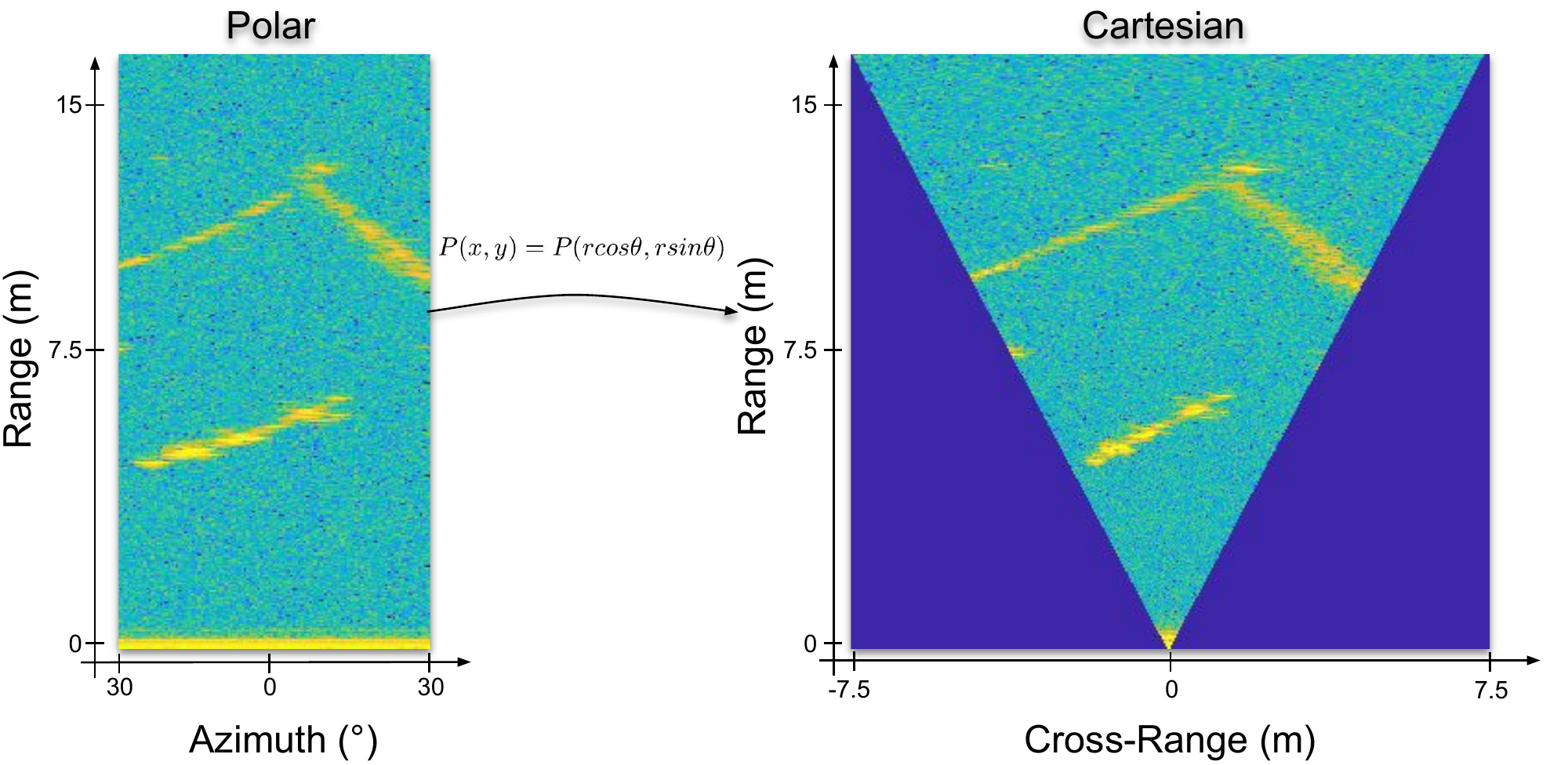}
		\caption{Polar to Cartesian radar image.}
		\label{fig:pol2cart}
	\end{figure}

	The raw data captured by the 300 GHz radar is a time-domain signal at each azimuth direction, where each azimuth is measured by a mechanical scanner. To transform the raw signal into an image two steps were performed. The first step is to apply Fast Fourier Transform (FFT) to each azimuth signal to create a range profile. The original polar image is converted to cartesian coordinates as shown in Figure \ref{fig:pol2cart}. This ensures equal dimensions in the x and y planes over all distances. \newtext{ 
		Before training the neural network with this data, we applied whitening by subtracting the mean value of the image data, as this helps the stochastic gradient descent (SGD) to converge faster. The convergence happens faster because the weight initialisation of neural networks is based on a Gaussian distribution with zero mean \cite{glorot2010understanding}. It means that the bias term will have less influence during the learning process.}
	
	\subsection{Experimental Design and Data Collection}\label{sec:data_collected}
	
	The main objective is to establish whether the proposed methodology has the potential to discriminate between a limited set of prototypical objects in a laboratory scenario, prior to collecting wild data in a scaled down or alternate radar system. \newtext{In the wild, by which we mean outside the laboratory and as a vehicle mounted sensor navigating the road network, we anticipate even more problems due to overall object density and proximity of targets to other scene objects.}
	In the laboratory, we wanted to gain knowledge of what features were important in 300 GHz radar data, and whether such features were invariant to the several possible transformations. The objects we decided to use were a bike, trolley, mannequin, sign, stuffed dog and cone. Those objects contain a varieties of shapes and materials which to some extent typify the expected, roadside  radar images that we might acquire from a vehicle.
	
\begin{figure}[!]
		\centering
		\includegraphics[width=0.9\linewidth]{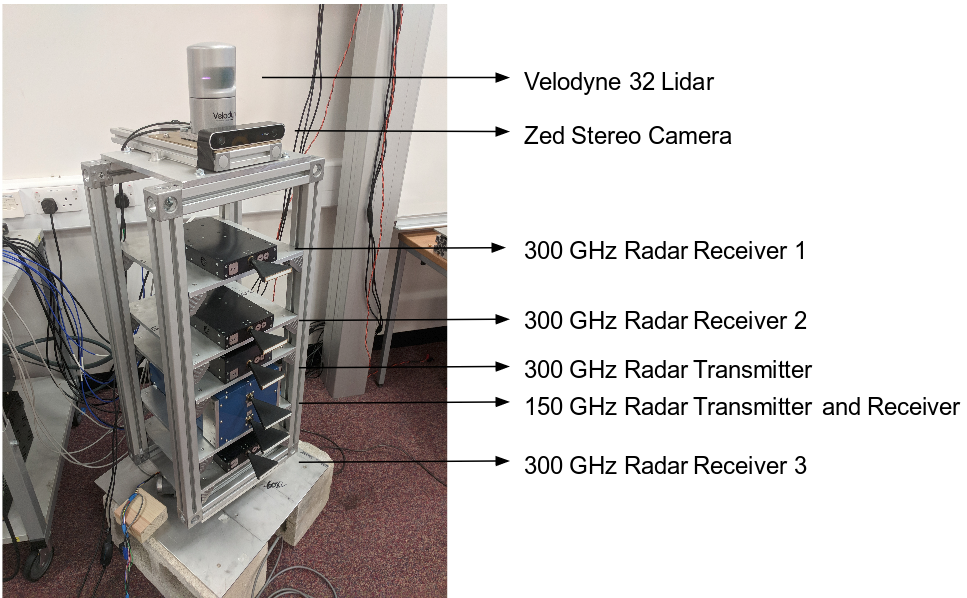}
		\caption{Experimental sensor setup.}
		\label{fig:exp_config}
	\end{figure}

\begin{figure}[!]
		\centering
		\includegraphics[width=1.0\linewidth]{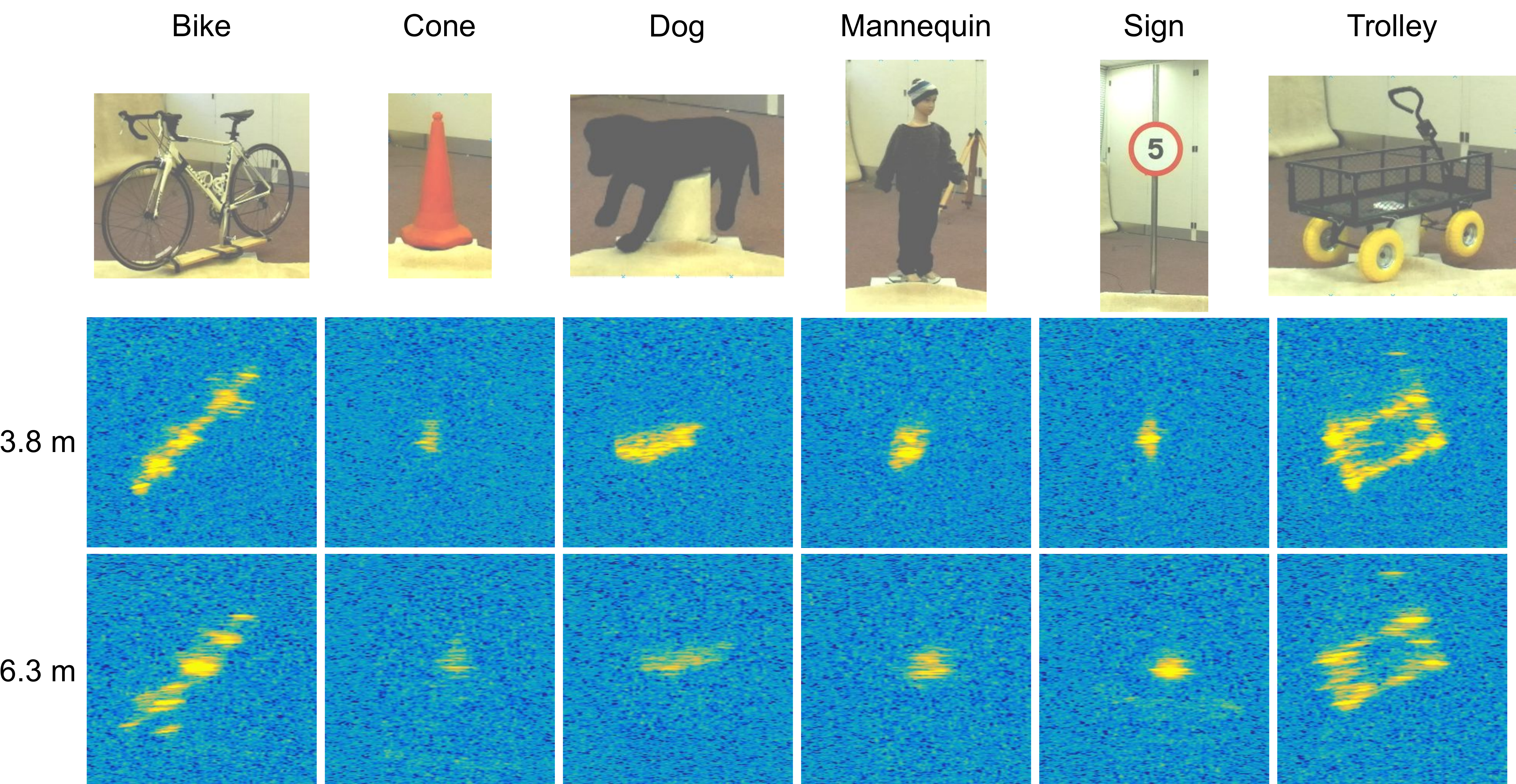}
		\caption{Sample images from each object from the dataset collected using the 300 GHz radar.}
		\label{fig:samples}
	\end{figure}

	The equipment for automatic data collection included a turntable to acquire samples every 4 degrees, covering all aspect angles, and at two stand-off distances, 3.8 m and 6.3 m. The sensors are shown in Figure \ref{fig:exp_config}. In collecting data, we used 300 GHz and 150 GHz radars, a Stereo Zed camera and a Velodyne HDL-32e LiDAR, but in this thesis only data from the 300 GHz radar is considered. The 300 GHz radar has 1 transmitter and 3 receivers. The 3 receivers were used to compare the object signatures at different heights, and to a lesser extent whether the different receivers had different operational characteristics. We used a carpet below the objects to avoid multi-path and ground reflections.
	Table \ref{tab:tab_dataset_300} summarises how many samples were captured from each object at each range. The object cone contains less samples because the signature for each rotation was very similar, so it was decided to collect less cone samples. Since we have 3 receivers, we have 1425 images from each range and 2850 images in total. In Figure \ref{fig:samples} we can see sample images from all objects at different ranges using receiver 3.
	
	\begin{table}[!]
		\caption{Data set collection showing number of different raw images collected\newline at each range.}\label{tab:tab_dataset_300}
		\begin{center}
			\begin{tabular}{lcc}
				\hline
				& \textbf{3.8 m}	& \textbf{6.3 m}\\
				\hline
				Bike         & 90                                             & 90                                             \\
				Trolley      & 90                                             & 90                                             \\
				Mannequin    & 90                                             & 90                                             \\
				Cone         & 25                                             & 25                                             \\
				Traffic Sign & 90                                             & 90                                             \\
				Stuffed Dog  & 90                                             & 90                                             \\ \hline
				Total        & \multicolumn{1}{l}{475 $\times$ (3 rec.) = 1425} & \multicolumn{1}{l}{475 $\times$ (3 rec.) = 1425} \\ \hline
				
			\end{tabular}

		\end{center}
		
	\end{table}
	
	All images were labelled with the correct object identity, irrespective of viewing range, angle and receiver height. A fixed size bounding box of $400 \times 400$ cells, which corresponds to $3m \times 3m$,  was cropped with the object in the middle of the box.
	
	\subsection{Neural Network Architecture}
	
\begin{figure*}[t]
		\centering
		\includegraphics[width=0.7\linewidth]{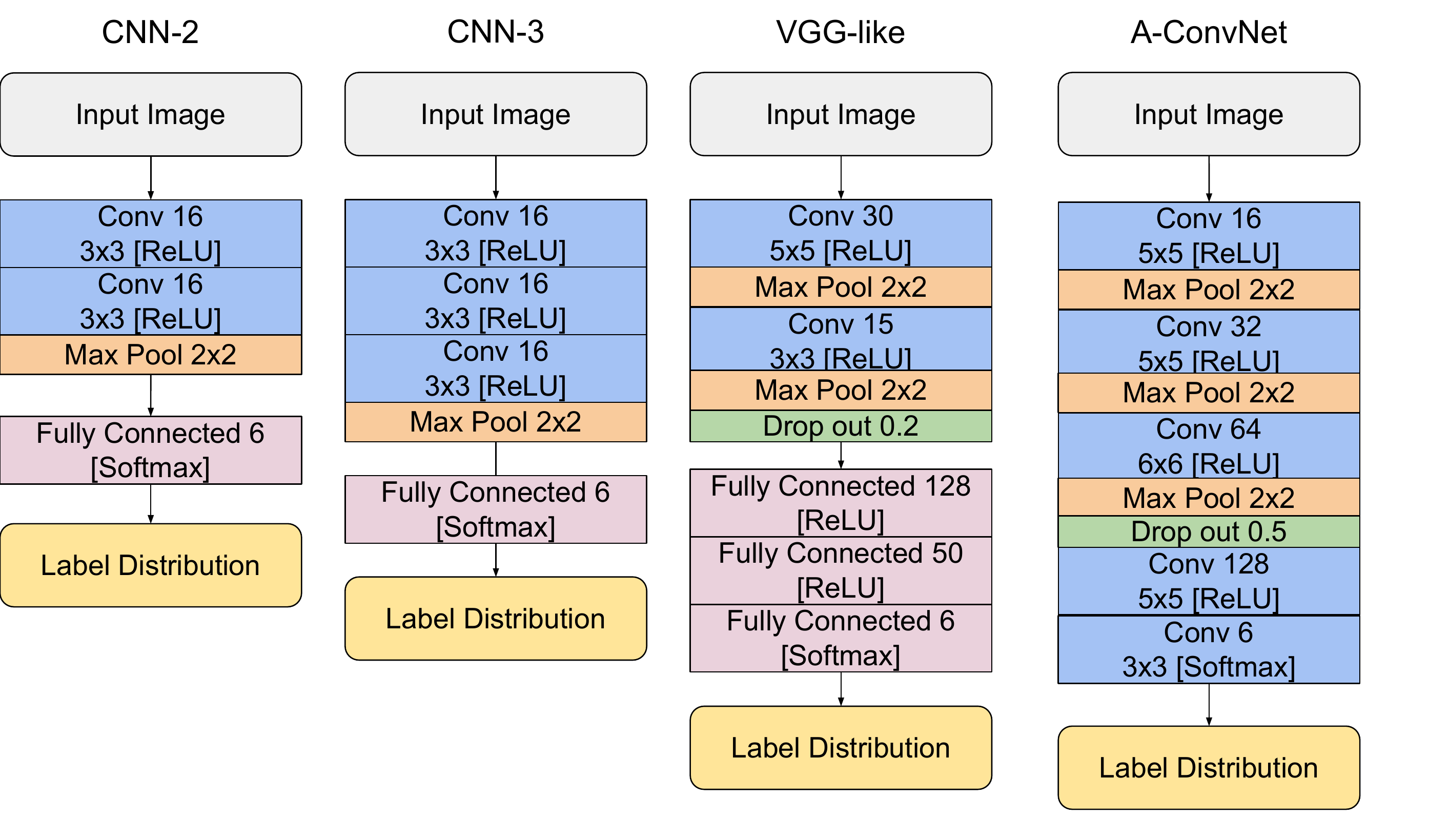}
		\caption{Networks architectures used.}
		\label{fig:iet_networks}
	\end{figure*}

	\newtext{In this work we used four networks which are illustrated in the Figure \ref{fig:iet_networks}}.
	
	\begin{itemize}
		\item \newtext{\textbf{CNN-2:} This network is a vanilla CNN with 2 convolutional layers and a fully connected layer in the end to classify the objects}
		\item \newtext{\textbf{CNN-3:} This network is the same as CNN-2 with an aditional convolutional layer.}
		\item \newtext{\textbf{VGG-like:} The VGG-like network was developed by Angelov \cite{angelov2019} for time-Doppler radar object recognition.}
		\item \newtext{\textbf{A-ConvNet:} This network developed by \cite{chen2016target} achieved state-of-the-art recognition on SAR target recognition.}
	\end{itemize}
	
	\newtext{All networks contains standard layers such as a convolutional layer, rectified linear unit (ReLU), max pooling, dropout, fully connected and softmax layers.
		A description of the properties of all these layers can be found in \cite{Goodfellow:2016:DL:3086952}.
		The CNN-2 and -3 networks provide a baseline solution of minimal complexity.
		The VGG-like network was chosen as it provided a very recent point of comparison on a similar problem, of course with the significant difference that it was designed for time-Doppler data.
		Finally, we chose the A-ConvNet architecture because it was also employed to recognise static objects in radar images, albeit synthetic aperture radar (SAR) images.
		This also allowed us to investigate transfer learning using this same network, trained on the SAR data and sharing the initial weights.
		For all networks we decided to use the original input layer of A-ConvNet ($88 \times 88$), so our input data was down-sampled using bilinear interpolation.}
	
	To train our neural network, we used Stochastic Gradient Descent (SGD). SGD updates the weights of the network depending on the gradient of the function that represents the current layer, as in Equation \ref{eq:sgd_300}.
	
	\begin{equation}\label{eq:sgd_300}
		W_{t+1} = W_{t} - \alpha \nabla f(x;W_t) + \eta \Delta W 
	\end{equation}
	
	In Equation \ref{eq:sgd_300}, $\eta$ is the momentum, $\alpha$ is the learning rate, $t$ is the current time step, $W$ defines the weights of the network and $\nabla f(x;W)$ is the derivative of the function that represents the network. To compute the derivative for all layers, we need to apply the chain rule, so we can compute the gradient through the whole network. The loss function used to minimise was the categorical cross-entropy (Equation \ref{eq:loss_300}). The parameters used in all experiments in all training procedures are given in Table \ref{tab:nn_params}. For all experiments we used 20\% of the training data as validation, and we used the best results from the validation set to evaluate the performance.
	In Equation \ref{eq:loss_300}, $\hat y$ is the predicted vector from softmax output and $y$ is the ground truth.
	
	\begin{equation}\label{eq:loss_300}
		L(\hat y,y) = - \sum_i y_i log(\hat y_i)
	\end{equation}
	
	\begin{table}[!]
		\caption{Neural Network Parameters.}\label{tab:nn_params}
		\centering
		
		\begin{tabular}{l|l}
			\hline
			Learning rate ($\alpha$) & 0.001 \\
			Momentum ($\eta$)      & 0.9   \\
			Epochs        & 100   \\
			Batch size    & 100  \\
			\hline
		\end{tabular}
		
	\end{table}
	
	\subsection{Data Augmentation}
	
	As shown in Table \ref{tab:tab_dataset_300}, we have limited training data.
	Using a restricted dataset, the DCNNs will easily overfit and be biased towards specific artifacts in the dataset. To help overcome this problem, we generated new samples to create a better generalisation. The simple technique of random cropping takes as input the image data of size $128 \times 128$ and creates a random crop of $88 \times 88$.
	This random crop ensures that the target is not always fixed at the same location, so that the location of object should not be a feature. We cropped each sample 8 times and also flipped all the images left to right to increase the size of the dataset and remove positional bias.
	
	\section{Experiments: Classification of Isolated Objects}
	\label{sec:experiments}
	
	As described in Section \ref{sec:data_collected}, we used six objects imaged from ninety viewpoints with three receivers at two different ranges (3.8 m and 6.3 m).
	Four different experiments were performed, shown in Table \ref{tab:experiments}. 
	The metric used to evaluate the results is accuracy, i.e. the number of correct divided by the total number of classifications in the test data. 
	
	\begin{table}[htbp]
		\caption{Set of experiments performed to verify the effectiveness in different configurations.}\label{tab:experiments}
		\centering
		\begin{tabular}{lll}
			\hline
			& \textbf{Train}	& \textbf{Test}\\
			\hline
			Experiment 1		& Random (70 \%)			& Random (30 \%)\\
			Experiment 2		& 2 Receivers			& 1 Receiver\\
			Experiment 3		& One range		& Other range \\
			Experiment 4		& Quadrants 1,3	& Quadrants 2,4\\
			
			\hline
		\end{tabular}
		
	\end{table}
	
	\subsection*{Experiment 1: Random selection from the entire data set}
	
	This is the often used, best case scenario, with random selection from all available data to form training and test sets. Intuitively, the assumption is that the dataset contains representative samples of all possible cases. To perform this experiment we randomly selected 70 \% of the data as training and 30 \% as test data. The results are summarised in Table \ref{tab:exp1}.

	\begin{table}[htbp]
		\centering
		\caption{\newtext{Accuracy for experiment 1: Random selection from all data}}
		\label{tab:exp1}
		\begin{tabular}{l|cccc}
				\hline
				& CNN-2 & CNN-3 & VGG-Like & A-ConvNet \\ \hline
				Random Selection  & 92.3\% & 94.9 \% & 96.4 \% & \textbf{99.7 \%} \\
				\hline
		\end{tabular}
		
	\end{table}
	
	From Table \ref{tab:exp1} we conclude that the results are very high across the board, so it is possible to recognize objects in the 300 GHz radar images, with the considerable caveats that the object set is limited, they are at short range in an uncluttered environment, and as all samples are used to train, then any test image will have many near neighbours included in the training data with a high statistical probability.

	\subsection*{Experiment 2: Receiver/Height Influence}
	
	The second experiment was designed to investigate the influence of the receiver antenna characteristics and height (see Figure \ref{fig:exp_config}).
	The potential problem is that the DCNNs may effectively overfit the training data to learn partly the antenna pattern from a specific receiver or a specific reflection from a certain height.
	All available possibilities were tried, i.e. 
	
	\begin{itemize}
		\item  \textbf{Experiment 2.1 :} Receivers 2 and 3 to train and receiver 1 to test
		\item  \textbf{Experiment 2.2 :} Receivers 1 and 3 to train and receiver 2 to test
		\item \textbf{Experiment 2.3 :} Receivers 1 and 2 to train and receiver 3 to test
	\end{itemize}
	
	Table \ref{tab:exp2} shows the results for Experiment 2.
	In comparison with Experiment 1, the results are poorer, but not by an extent that we can determine as significant on a limited trial. This was expected from examination of the raw radar data, since there is not much difference in the signal signatures from the receivers at different heights. If anything, receiver 3, which was closest to the floor and so received more intense reflections, gave poorer results when used as the test case which implied that the DCNNs did include some measure of receiver or view-dependent characteristics from the learnt data. In this instance, the drop in performance is markedly less severe in the preferred A-ConvNet architecture. 
	
	\begin{table}[htbp]
		\caption{\newtext{Accuracy for Experiment 2: Receiver influence}}\label{tab:exp2}
		\centering
		\begin{tabular}{l|cccc}
				\hline
				& CNN-2 & CNN-3 & VGG-Like & A-ConvNet \\ \hline
				Receiver 1 Test Experiment & 82.6 \% & 85.9 \% & 81.1 \% & \textbf{98.9 \%} \\ 
				Receiver 2 Test Experiment & 90.0 \% & 93.2 \% & 93.5 \% & \textbf{98.4 \%}\\
				Receiver 3 Test Experiment & 60.4 \% & 65.9 \% & 65.6 \% & \textbf{87.7 \%}\\
				\hline
		\end{tabular}
		
	\end{table}
	
	\subsection*{Experiment 3: Range Influence}
	
	Clearly, the range of the object influences the return signature to the radar as the received power will be less due to attenuation, and less cells are occupied by the target in the polar radar image due to degrading resolution over azimuth. Therefore, if the training data set is selected only at range 3.8m, for example, to what extent are the features learnt representative of the expected data at 6.8m (and vice versa)? Table \ref{tab:exp3} summarises the results achieved when we used one range to train the network, and the other range to test performance.
	
	\begin{itemize}
		\item \textbf{Experiment 3.1 :} Train with object on 3.8 m. Test with object on 6.3 m.
		\item \textbf{Experiment 3.2 :} Train with object on 6.3 m. Test with object on 3.8 m.
	\end{itemize}
	
	\begin{table}[htbp]
		\caption{\newtext{Accuracy for Experiment 3: Range influence}}\label{tab:exp3}
		\begin{center}
			\centering
			\begin{tabular}{l|cccc}
					\hline
					& CNN-2 & CNN-3 & VGG-Like & A-ConvNet \\ \hline
					Object at 6.3 m Test Experiment & 58.3 \% & 60.2 \% & 59.6 \% & \textbf{82.5 \%}\\
					Object at 3.8 m Test Experiment & 58.5 \% & 69.5 \% & 37.7 \% & \textbf{91.1 \%}\\
					
					\hline
			\end{tabular}
		\end{center}
		
	\end{table}

	The key observation from Table 7 is that if we train the DCNNs at one specific range which has a given cell structure and received power distribution, and then test at a different range, the performance is not as accurate as in the base case as this drops from over 99\% to 82.5\% and 91.1\% respectively in the case of AConvNet.
	Again, the other networks do not perform as well. 
	
	\subsection*{Experiment 4: Orientation Influence}
	
\begin{figure}[h]
		\centering
		\includegraphics[width=0.3\linewidth]{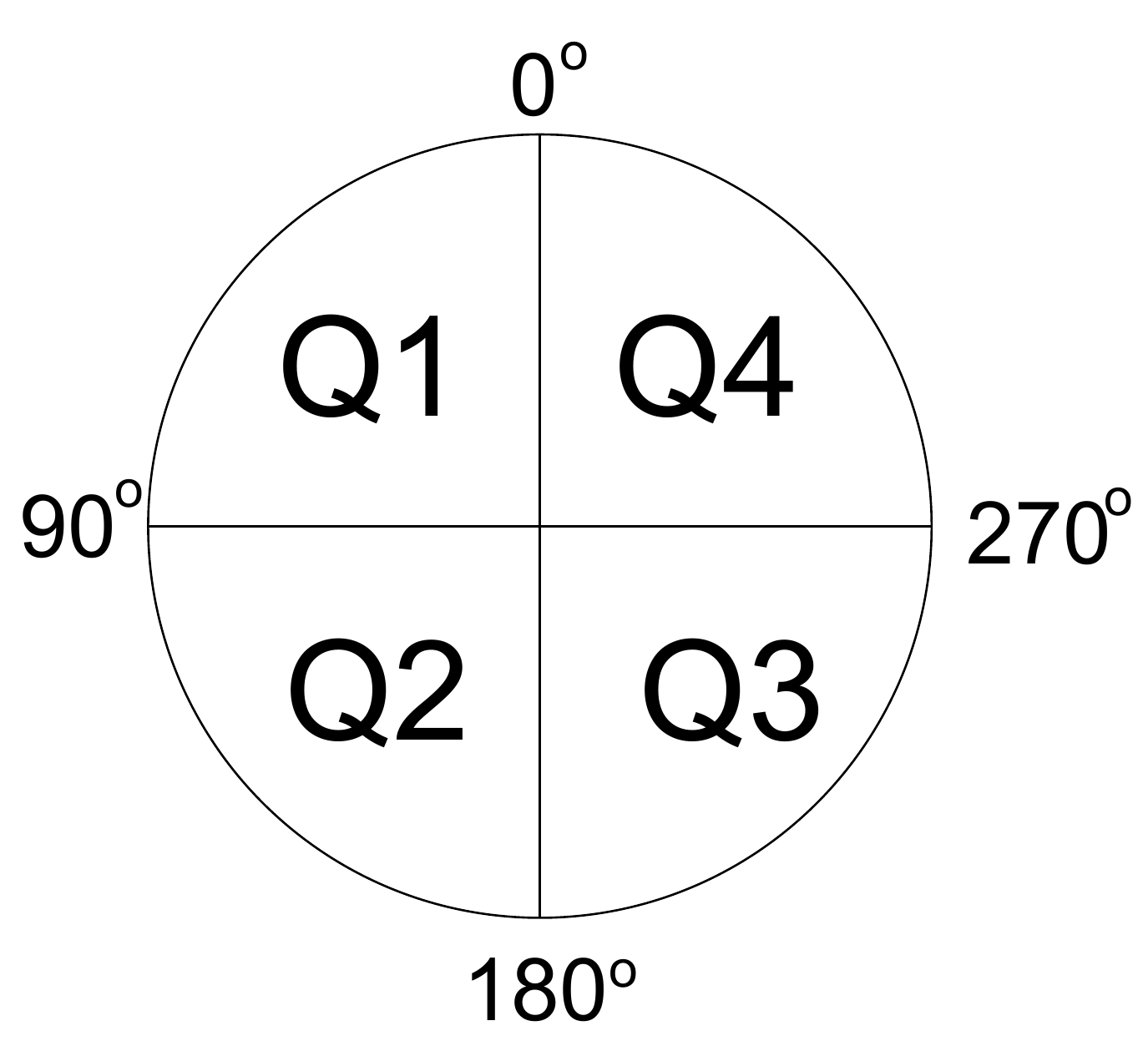}
		\caption{Quadrants}
		\label{fig:quadrant}
	\end{figure}

	The final experiment was designed to examine whether the neural network was robust to change of viewing orientation. Here, we used as training sets the objects in quadrants 1 and 3, and as test sets the objects in  quadrants 2 and 4. Quadrant 1 means orientation from $0^o$ to $89^o$, quadrant 2 means orientation from $90^o$ to $179^o$, quadrant 3 means orientation from $180^o$ to $269^o$ and quadrant 4 means orientation from $270^o$ to $359^o$, as seen in the Figure \ref{fig:quadrant}
	
	\begin{table}[htbp]
		\caption{\newtext{Accuracy for experiment 4: Orientation influence}}\label{tab:exp4}
		\centering
		\begin{tabular}{l|cccc}
				\hline
				& CNN-2 & CNN-3 & VGG-Like & A-ConvNet \\ \hline
				Q2,Q4 Test Experiment & 76.5\% & 78.3 \% & 92.2 \% & \textbf{92.5}\%\\
				\hline
				
		\end{tabular}
		
	\end{table}
	
	In the Table \ref{tab:exp4}, the DCNNs do not perform as well compared to Experiments 1 and 2, for example dropping to 92.5\% for A-ConvNet.
	However, since we flipped the images left to right as a data augmentation strategy, the network was capable of learning the orientation features, as the objects exhibit near mirror symmetry, and in one case, the cone, is identical from all angles. 
	Therefore, we have to be hesitant in drawing conclusions about any viewpoint invariance within the network as the experiments are limited and all objects have an axis or axes of symmetry (as do many objects in practice).
	
	Together with Experiments 2 and 3, this experiment shows that it is necessary to take into account the differences in the acquisition process using different receivers at different ranges and orientation in training the network.
	While this is to some extent obvious and equally true for natural images, we would observe that the artefacts introduced by different radar receivers are much less standardised that those introduced by standard video cameras, so the results obtained in future may be far less easy to generalise. Although Experiment 2 only showed limited variation in such a careful context, we would speculate that the effects of multipath and clutter would be far more damaging than in the natural image case, as highlighted in \cite{bartsch2012pedestrian}.
	
	\subsection{Comparison Between the Networks}
	
	\newtext{As seen in Tables \ref{tab:exp1}, \ref{tab:exp2}, \ref{tab:exp3} and \ref{tab:exp4}, in all scenarios, A-ConvNet was superior. The CNN-2 and CNN-3 networks are the baseline, and it shows that without much engineering we manage to have networks with suitable results, however they are not as good as a network designed to SAR target recognition. Angelov, \etal \cite{angelov2019} designed a network for a time-Doppler image, which in our scenario (just using the power spectra) did not manage to work as well as A-ConvNet. A-ConvNet architecture was designed for a more similar problem, and for these experiments achieved very good results. Since A-ConvNet was shown to be the superior network of the ones presented here, succeeding experiments in this paper use this architecture.}
	
	\subsection{Transfer Learning}
	
\begin{figure}[t!]
		\centering
		\includegraphics[width=\linewidth]{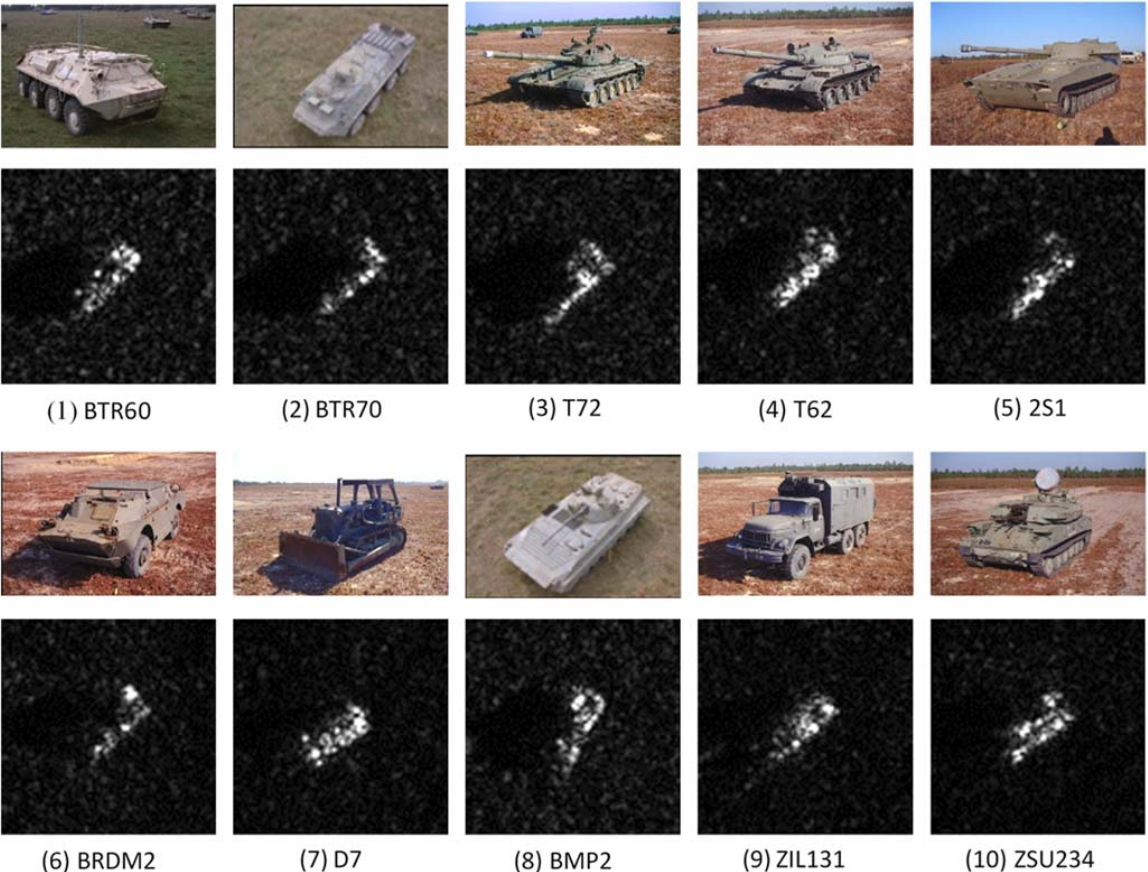}
		\caption{10 military targets used in MSTAR Dataset \cite{keydel1996mstar}}
		\label{fig:mstar_dataset}
	\end{figure}

	As summarised in Table \ref{tab:tab_dataset_300}, we have a small dataset and there is the potential to learn image-specific characteristics rather than features of the objects themselves.
	Therefore, we have investigated the use of transfer learning to help capture more robust features using a pre-existing dataset, i.e. to use prior knowledge from one domain and transfer it to another \cite{yosinski2014transferable}.
	To apply transfer learning, we first trained the DCNNs on the MSTAR (source) data, then the weights from the network were used as initial weights for a the DCNNs trained on our own 300 GHz (target) data.
	The MSTAR data is different in viewing angle and range compared to our own data as shown in Figure \ref{fig:mstar_dataset}.
	It was developed to recognise military targets using SAR images. The data contains 10 different military targets and around 300 images per target with similar elevation viewing angles of \ang{15} and \ang{17}.
	In total MSTAR has around 6000 images and is used widely by the radar community in order to verify classification algorithms.
	
	The DCNNs function in the source domain (MSTAR in our case) is defined by Equation \ref{eq:dnn}.
	
	\begin{equation}\label{eq:dnn}
		y_s = f(W_s,x_s)
	\end{equation}
	
	\noindent 
	where $W_s$ are the weights of a network, $x_s$ and $y_s$ are the input and and output from the source domain. To learn the representation, an optimizer must be used, again stochastic gradient descent (SGD), expressed in Equation \ref{eq:optimizer}.
	
	\begin{equation}\label{eq:optimizer}
		W_{s_{i+1}} = SGD(W_{s_i}, x_s, y_s)
	\end{equation}
	
	\noindent 
	where $SGD$ is an algorithm which updates the weights of the neural network, as expressed in Equation \ref{eq:sgd}
	Hence, using the trained weights from our source domain as the initial weights, this is expressed as Equation \ref{eq:tf}.
	It is intended that the initial weights give a better initial robust representation which can be adapted to the smaller dataset. \todo{ $W_{t_1}$ represents the first step of the SGD before we start to train and $W_s$ is the trained weights from the source dataset.}
	
	\begin{equation}\label{eq:tf}
		W_{t_1} = SGD(W_s,x_t, y_t)
	\end{equation}
	
	We repeated experiments 1,2,3 and 4 using transfer learning. The results are summarised in Table \ref{tab:exp_tl}.
	To gain further insight, we also show the confusion matrix from the orientation experiments without and with transfer learning in Tables. \ref{tab:conf_mat} and \ref{tab:conf_mat_tl}. The main confusion is between the dog and mannequin, since both have similar clothed material; and cone and sign, since they have similar shape.
	
	\begin{table}[htbp]
		\caption{Accuracy after applying transfer learning}\label{tab:exp_tl}
		\centering
		\begin{tabular}{ll|cc}
			\hline
			
			&  & without TL & with TL \\
			\hline
			Exp 1: & Random Split Exp. & \textbf{99.7\%} & 99.1\% \\
			Exp 2.1: & Rec. 1 Test Exp. & \textbf{98.9\%} & 95.8\% \\
			Exp 2.2: & Rec. 2 Test Exp. & 98.4\% & \textbf{98.8\%} \\
			Exp 2.3: & Rec. 3 Test Exp. & 87.7\% & \textbf{94.1\%} \\
			Exp 3.1: & 6.3 m Test Exp. & 82.5\% & \textbf{85.2\%} \\
			Exp 3.2: & 3.8 m Test Exp. & 91.1\% & \textbf{93.5\%} \\
			Exp 4: & Q2,Q4 Test Exp. & 92.5\% &\textbf{98.5\%}  \\
			\hline
		\end{tabular}
		
	\end{table}

	\subsection{The Effect of Transfer Learning}

	\begin{figure*}[t]
		\centering
		\begin{subfigure}[b]{0.3\textwidth}
			\includegraphics[width=\textwidth]{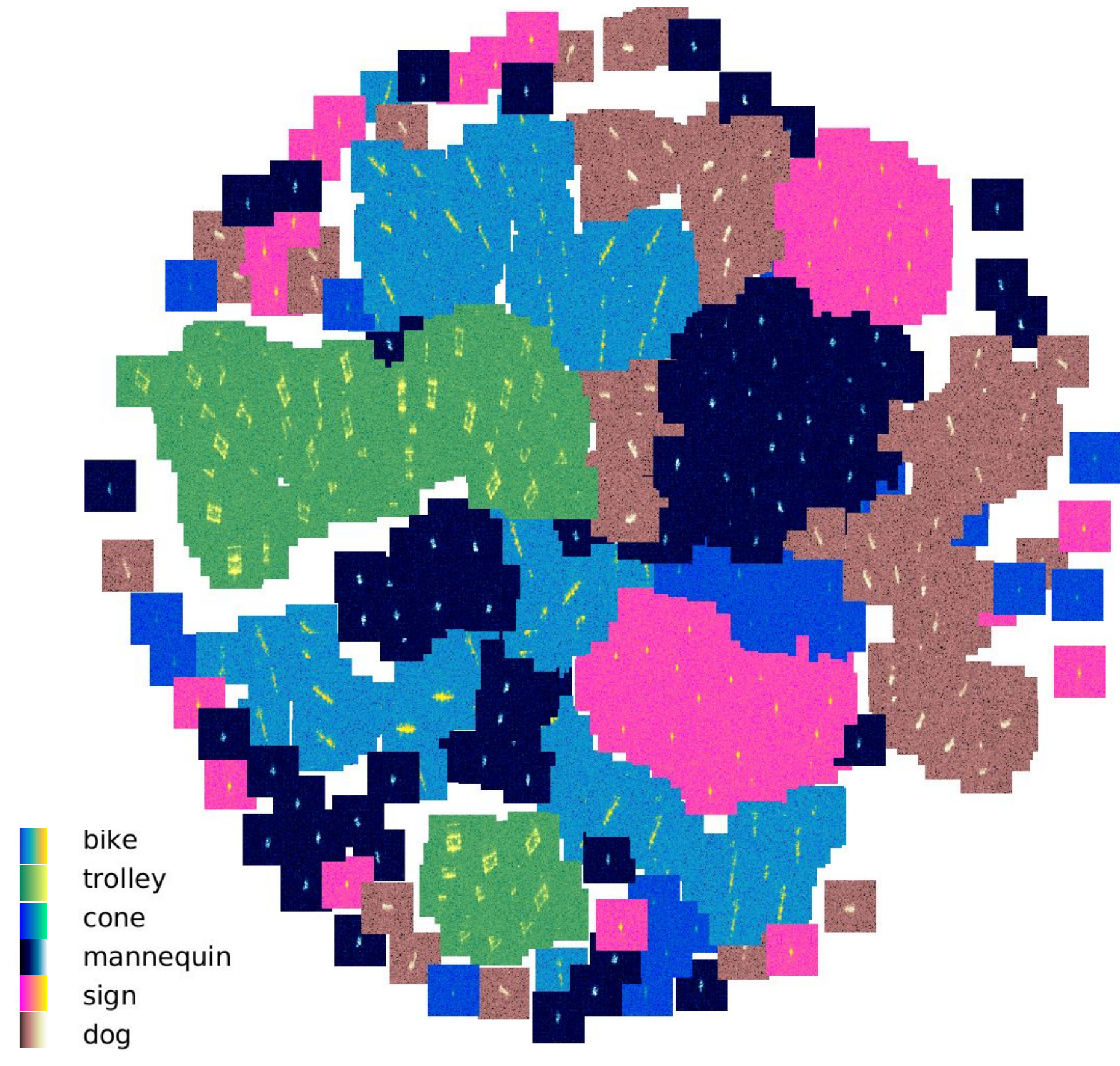}
			\caption{Raw features}
			\label{fig:tsne_raw}
		\end{subfigure}%
		    \rulesep
		\begin{subfigure}[b]{0.3\textwidth}
			\includegraphics[width=\textwidth]{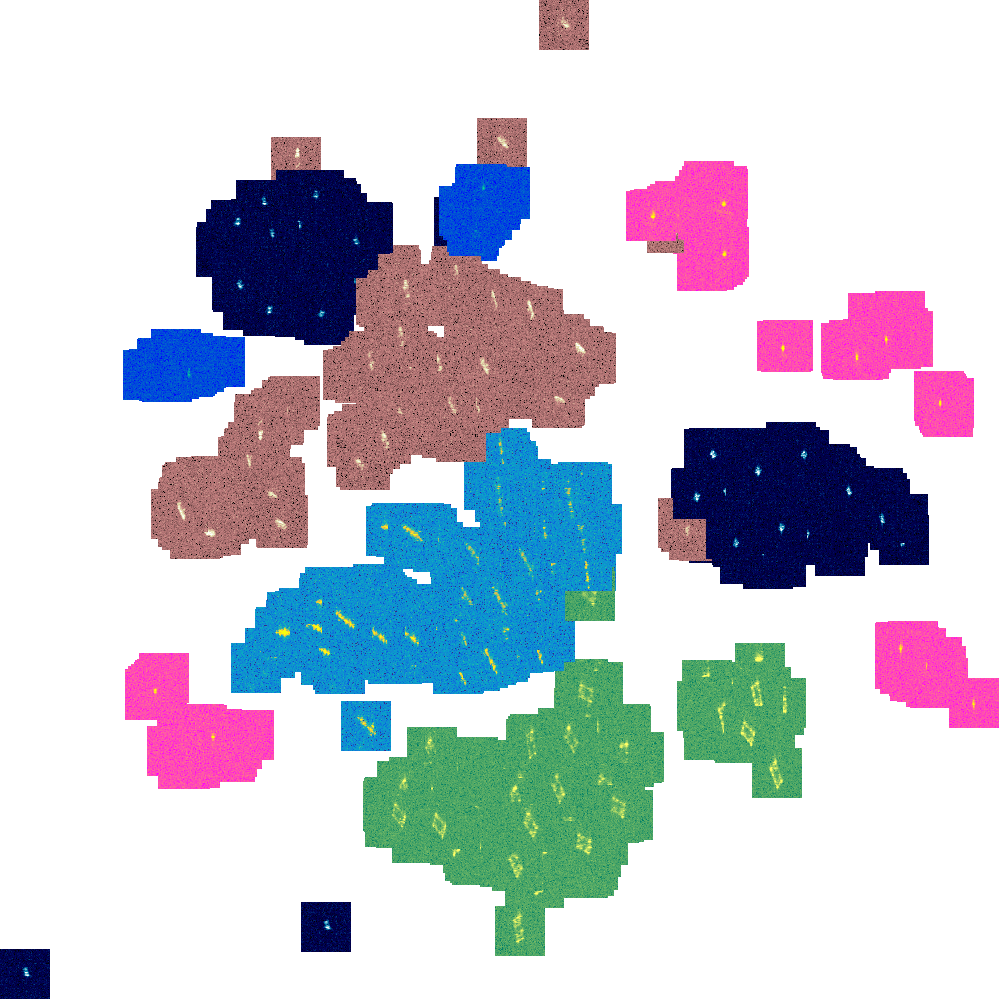}
			\caption{Without Transfer learning}
			\label{fig:tsne_da}
		\end{subfigure}%
		   \rulesep
		\begin{subfigure}[b]{0.3\textwidth}
			\includegraphics[width=\textwidth]{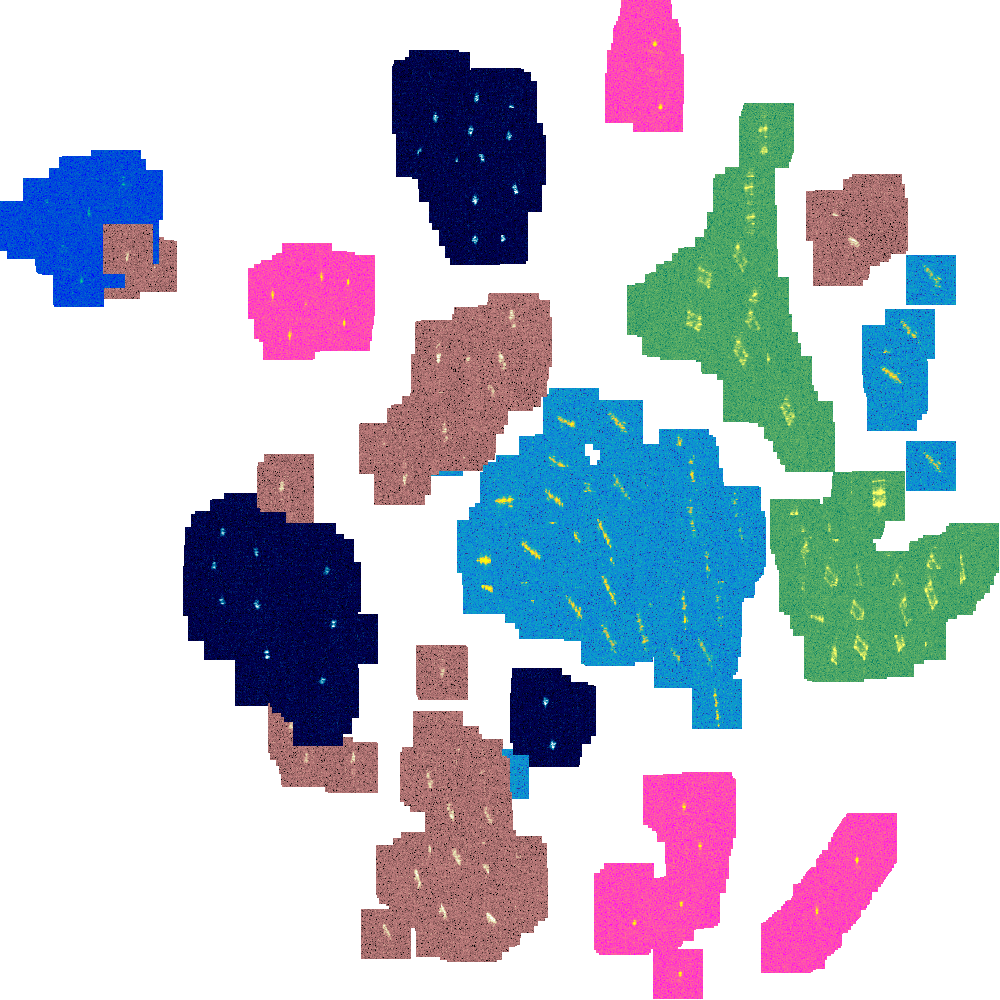}
			\caption{With Transfer learning}
			\label{fig:tsne_da_tl}
		\end{subfigure}
		\caption{t-SNE plots from the orientation experiment}\label{fig:tsne_plots}
	\end{figure*}
	
	As can be seen, transfer learning gives higher values for accuracy in the majority but not all cases. The MSTAR dataset is a much bigger dataset, and although it exhibits some characteristics in common with our own data, it uses a synthetic aperture technique, and there is no significant variation in elevation angle during data collection. \newtext{However, there are 2 distinguishable strong features, the shape and reflected power, and like our data, the objects are viewed at all possible rotations in the ground plane.
		As these characteristics have much in common with our own data, it is possible that the network is able to better generalise to cope with new situations as shown for example in the Receiver 3 and different range experiments. 
		To draw any firmer conclusion requires much more extensive evaluation.}
	
	Nevertheless, in these experiments, we can conclude that the neural network approach is robust in maintaining accuracy with respect to sensor hardware, height, range and orientation.

	\begin{table}[!htb]
		\caption{Orientation Experiment trained on A-ConvNet without Transfer Learning}\label{tab:conf_mat}
		\centering
		\begin{tabular}{ll|cccccc}
			\multicolumn{2}{c|}{\multirow{2}{*}{Acc: 0.925}} &\multicolumn{6}{c}{Predicted Label} \\
			&           & Bike                 & Trolley              & Cone                 & Mannequin            & Sign                 & Dog                  \\ \hline
			\parbox[t]{2mm}{\multirow{6}{*}{\rotatebox[origin=c]{90}{True Label}}}& Bike      & 1.00                 &   0.00                   &     0.00                 &         0.00             &        0.00              &         0.00             \\
			& Trolley   &    0.03                  & 0.97                 &    0.00                  &       0.00               &   0.00                   &     0.00                 \\
			& Cone      &     0.00                 &       0.00               & 1.00                 &      0.00                &    0.00                  &    0.00                  \\
			& Mannequin &      0.00                &   0.00                   &     0.00                 & 0.86                 &   0.00                   & 0.14                 \\
			& Sign      &       0.00               &    0.00                  &    0.00                  &       0.00               & 1.00                 &   0.00                   \\
			& Dog       & 0.03                 &              0.00        & 0.02                 & 0.10                 &    0.00                  & 0.86                
		\end{tabular}
	\end{table}
	
	\begin{table}[!htb]
		\caption{Orientation Experiment trained on A-ConvNet with Transfer Learning \newline from MSTAR}\label{tab:conf_mat_tl}
		\centering
		\begin{tabular}{ll|cccccc}
			\multicolumn{2}{c|}{\multirow{2}{*}{Acc: 0.985}} &\multicolumn{6}{c}{Predicted Label} \\
			&           & Bike                 & Trolley              & Cone                 & Mannequin            & Sign                 & Dog                  \\ \hline
			\parbox[t]{2mm}{\multirow{6}{*}{\rotatebox[origin=c]{90}{True Label}}}& Bike      & 1.00                 &   0.00                   &     0.00                 &         0.00             &        0.00              &         0.00             \\
			& Trolley   &    0.00                  & 1.00                 &    0.00                  &       0.00               &   0.00                   &     0.00                 \\
			& Cone      &     0.00                 &       0.00               & 1.00                 &      0.00                &    0.00                  &    0.00                  \\
			& Mannequin &      0.00                &   0.00                   &     0.00                 & 0.96                 &   0.00                   & 0.04                 \\
			& Sign      &       0.00               &    0.00                  &    0.00                  &       0.00               & 1.00                 &   0.00                   \\
			& Dog       & 0.00                 &              0.00        & 0.00                 & 0.03                 &    0.00                  & 0.97                
		\end{tabular}
	\end{table}

	\subsection{Visualisation of Feature Clusters}
	
	To better understand what is being learned by our network, the \newtext{t-Stochastic Neighbour Embedding technique (t-SNE)} \cite{maaten2008visualizing} was used to visualise the feature clusters. t-SNE employs nonlinear dimensionality reduction to build a probability distribution by comparing the similarity of all pairs of data, then transformed to a lower dimension. Then it uses \newtext{Kullback-Leibler (KL)-divergence} to minimise with respect to the locations in the cluster space.
	
	Figure \ref{fig:tsne_plots} shows the result from t-SNE clustering of samples using raw image features, in this case the orientation experiment. Figures \ref{fig:tsne_da} and \ref{fig:tsne_da_tl} show the t-SNE clusters from the features extracted from the penultimate layer of the trained neural network with and without transfer learning, using different colormaps for each object for better visualisation. First, we can see that the trained neural network was able to cluster similar classes and similar features in each case.
	Second, transfer learning shows slight improvement by creating larger, better separated clusters of objects of the same class.
	Although it is hard to give actual interpretability of neural networks, the t-SNE framework can give some insights into the type of features that have been learned.
	
\begin{figure}[ht]
    \centering
    \includegraphics[width=\columnwidth]{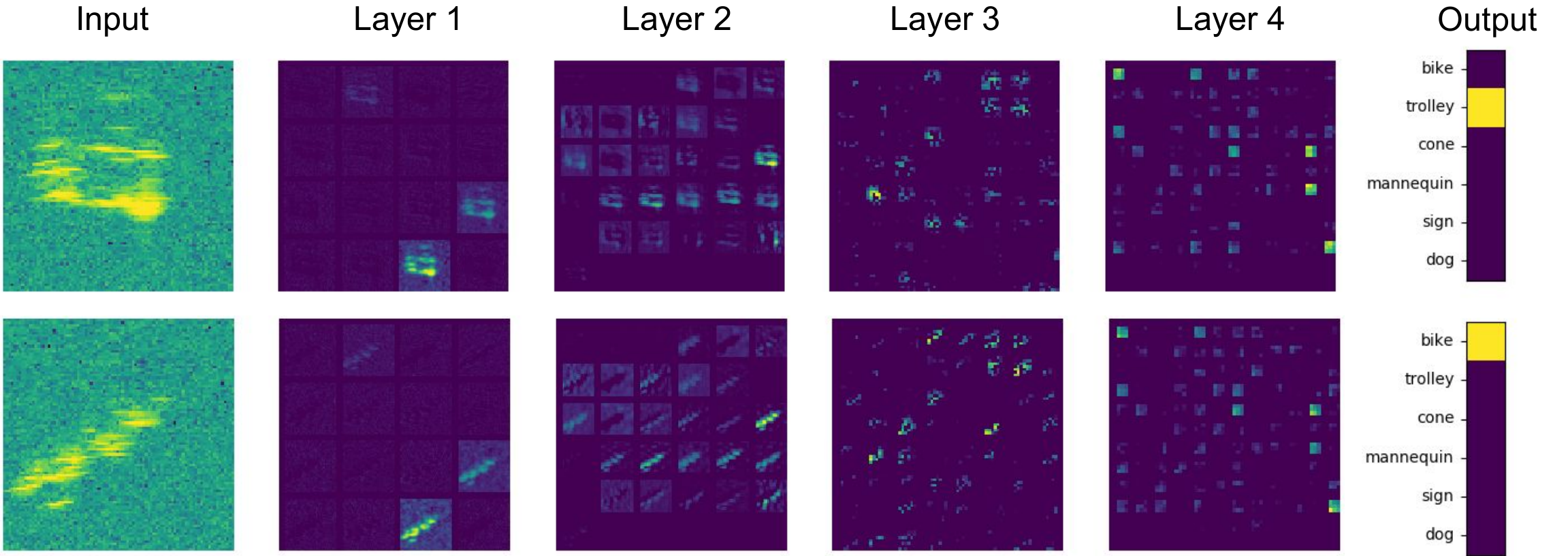}
    \caption{Visualisation of feature maps: the trolley is on top and the bike on the bottom.}
    \label{fig:network_vis}
\end{figure}

	\newtext{To further understand what is being learned, we plotted the feature maps from each layer of A-ConvNet trained using transfer learning (Figure \ref{fig:network_vis}). 
Visualising the response of each convolutional layer in this way, we can see that for both trolley and bike we have neurons responsible for activation on the object region, finding shape and intensity patterns.}

	\section{Experiments: Detection and Classification within a Multiple Object Scenario}
	\label{sec:detection_300}
	
\begin{figure*}[h]
		\centering
		\includegraphics[width=\linewidth]{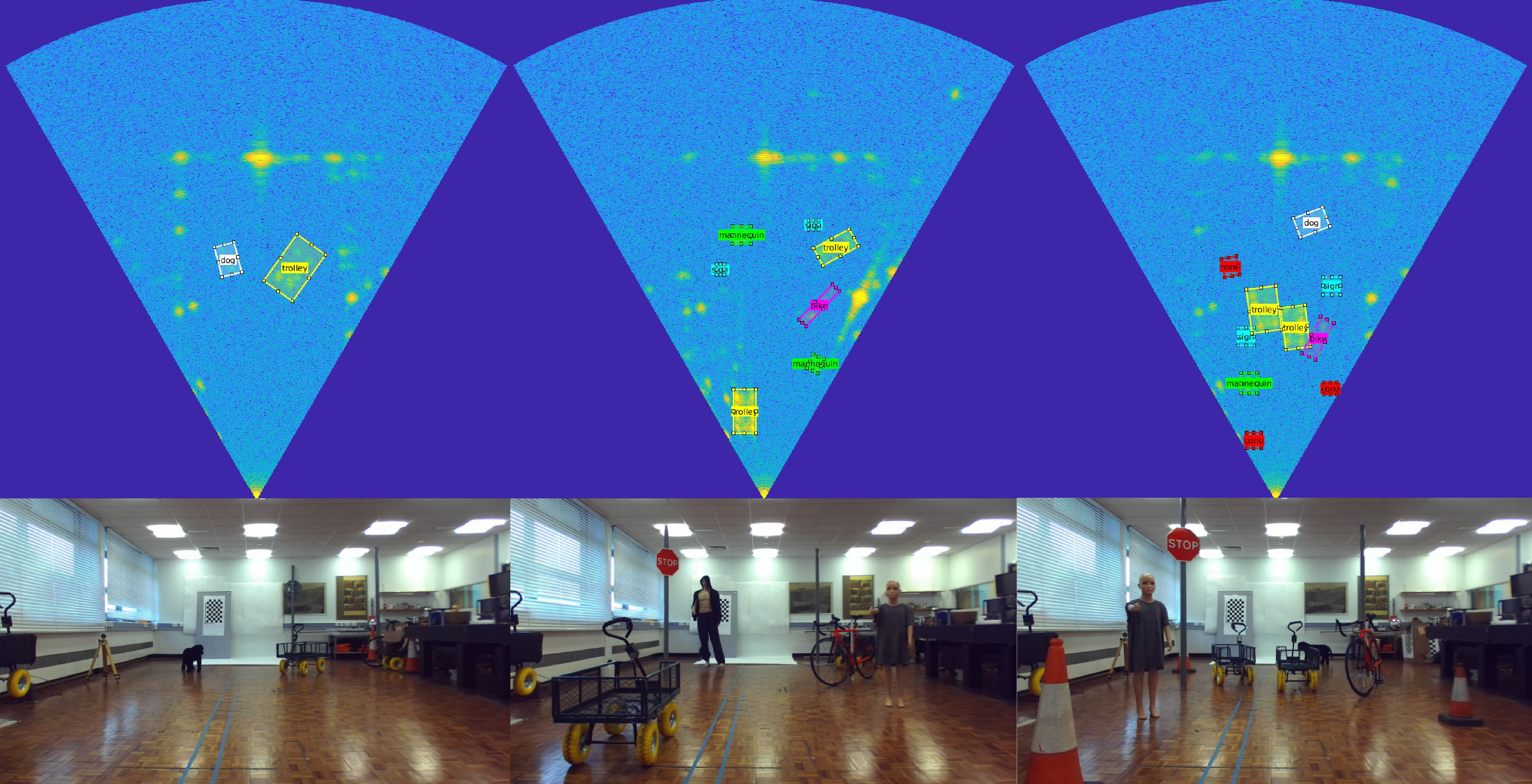}
		\caption{Multiple Object Dataset. Above: 300 GHz radar image. Below: Reference RGB image.}
		\label{fig:multi_object}
	\end{figure*}

	The previous dataset contains one windowed object in each image. In an automotive or more general radar scenario we must both detect and classify road actors in a scene with many pre-learnt and unknown objects which is much more challenging. We use the same object dataset (bike, trolley, cone, mannequin, sign, dog) in different parts of the room with arbitrary rotations and ranges, and the network is trained by viewing the objects in isolation, as before. We also include some within-object variation, using for example different mannequins, trolleys and bikes. The unknown, laboratory walls are also very evident in the radar images. Figure \ref{fig:multi_object} shows examples of 3 scenes in the multiple object dataset.
	
	Hence, in the next set of experiments we include multiple objects, and this has several additional phenomena including:
	\begin{itemize}
	    \item \textbf{Occlusion}: Objects can create shadow, which creates attenuation, affecting the signature captured by the receiver.
	    \item \textbf{Multi-path}: Many objects in the scene can cause multi-path, which can create ghost objects.
	    \item \textbf{Interference between objects} : Objects near by can be clustered as a single object.
	    \item \textbf{Objects which are not included as a learnt object of interest}: Objects which are not in the training set will interfere in the overall result, since they were never sensed in the training set.
	\end{itemize}
	
\begin{figure*}[h]
		\centering
		\includegraphics[width=1.0\linewidth]{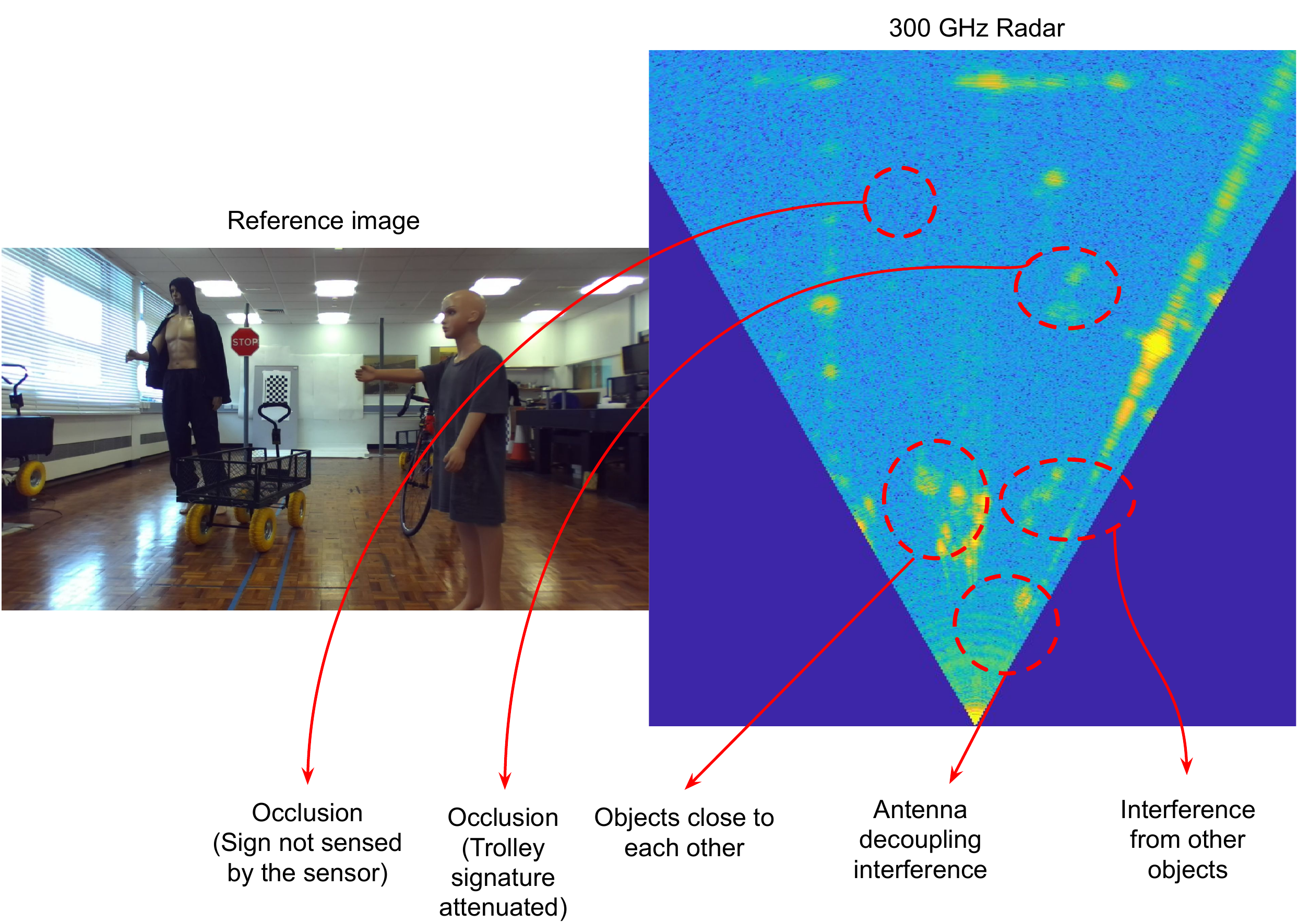}
		\caption{Possible unwanted effects in the multiple object dataset}
		\label{fig:multi_object_problems}
	\end{figure*}

	 Figure \ref{fig:multi_object_problems} illustrates possible problems that can occur in the multiple objects dataset.
	
\begin{figure*}[t]
		\centering
		\begin{subfigure}{.5\textwidth}
			\centering
			\includegraphics[width=\linewidth]{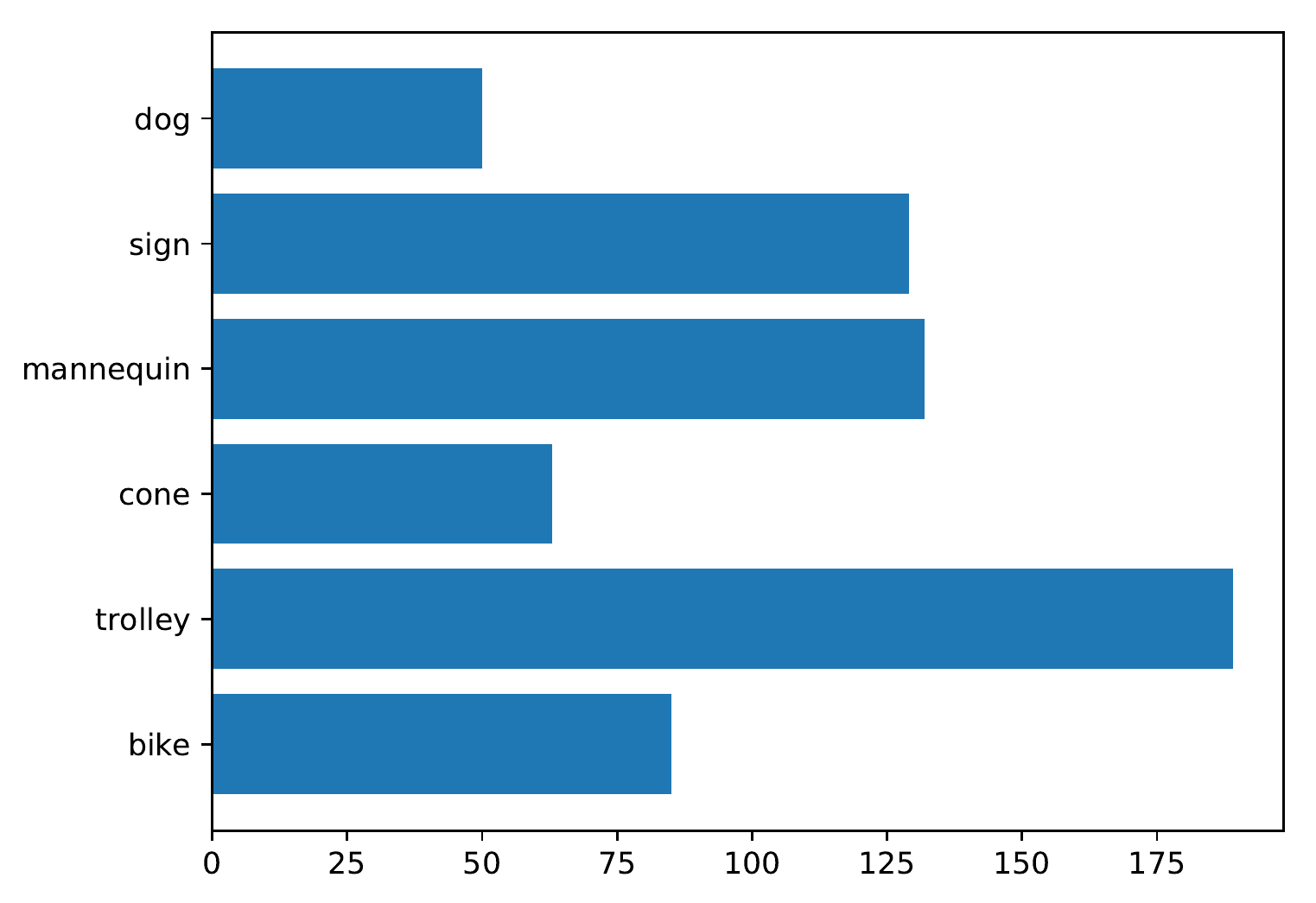}
			\caption{Number of object instances per class}
			\label{fig:nb_obj}
		\end{subfigure}%
		\begin{subfigure}{.5\textwidth}
			\centering
			\includegraphics[width=\linewidth]{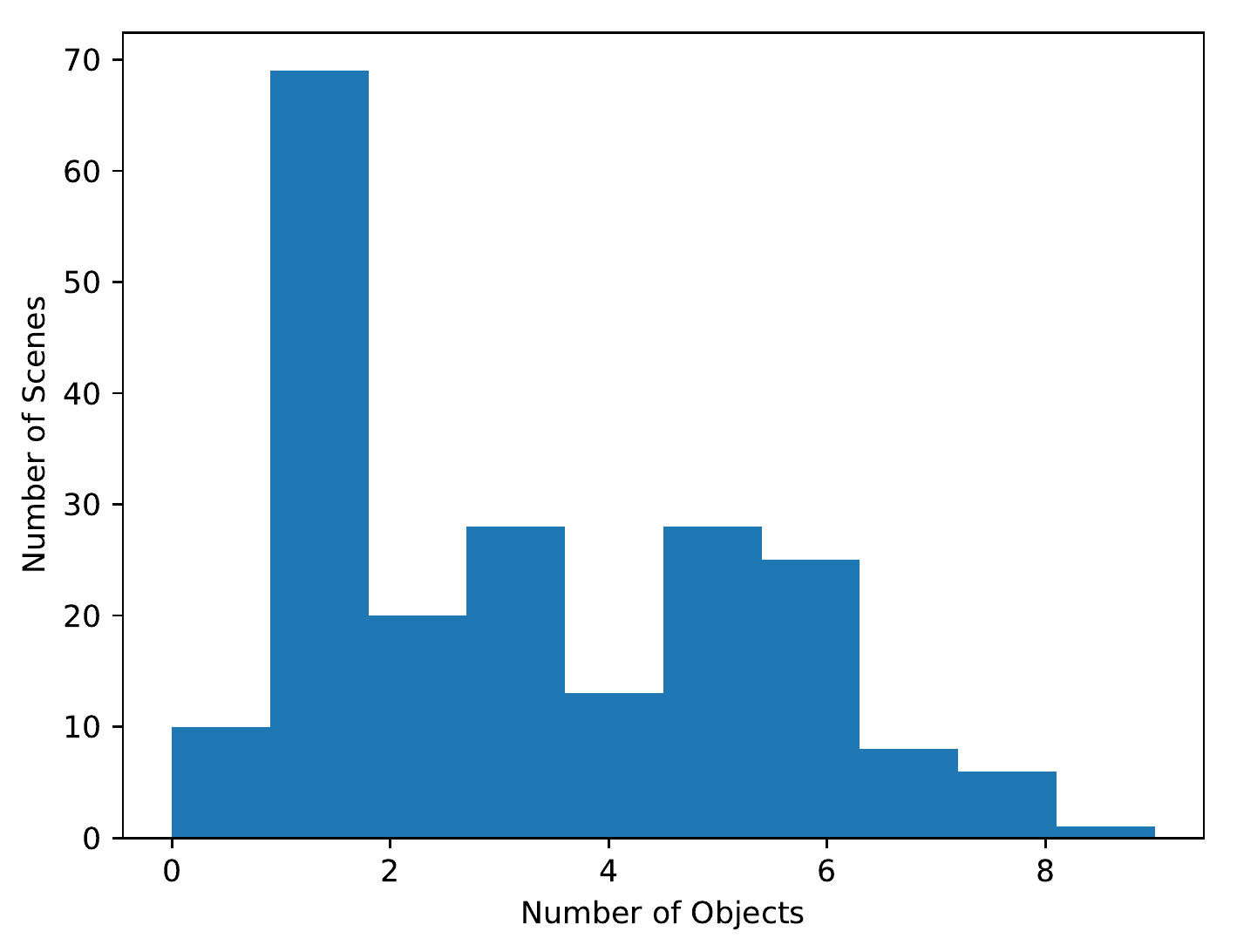}
			\caption{Histogram of number of objects per scene}
			\label{fig:hist_obj}
		\end{subfigure}
		\begin{subfigure}{.5\textwidth}
			\centering
			\includegraphics[width=\linewidth]{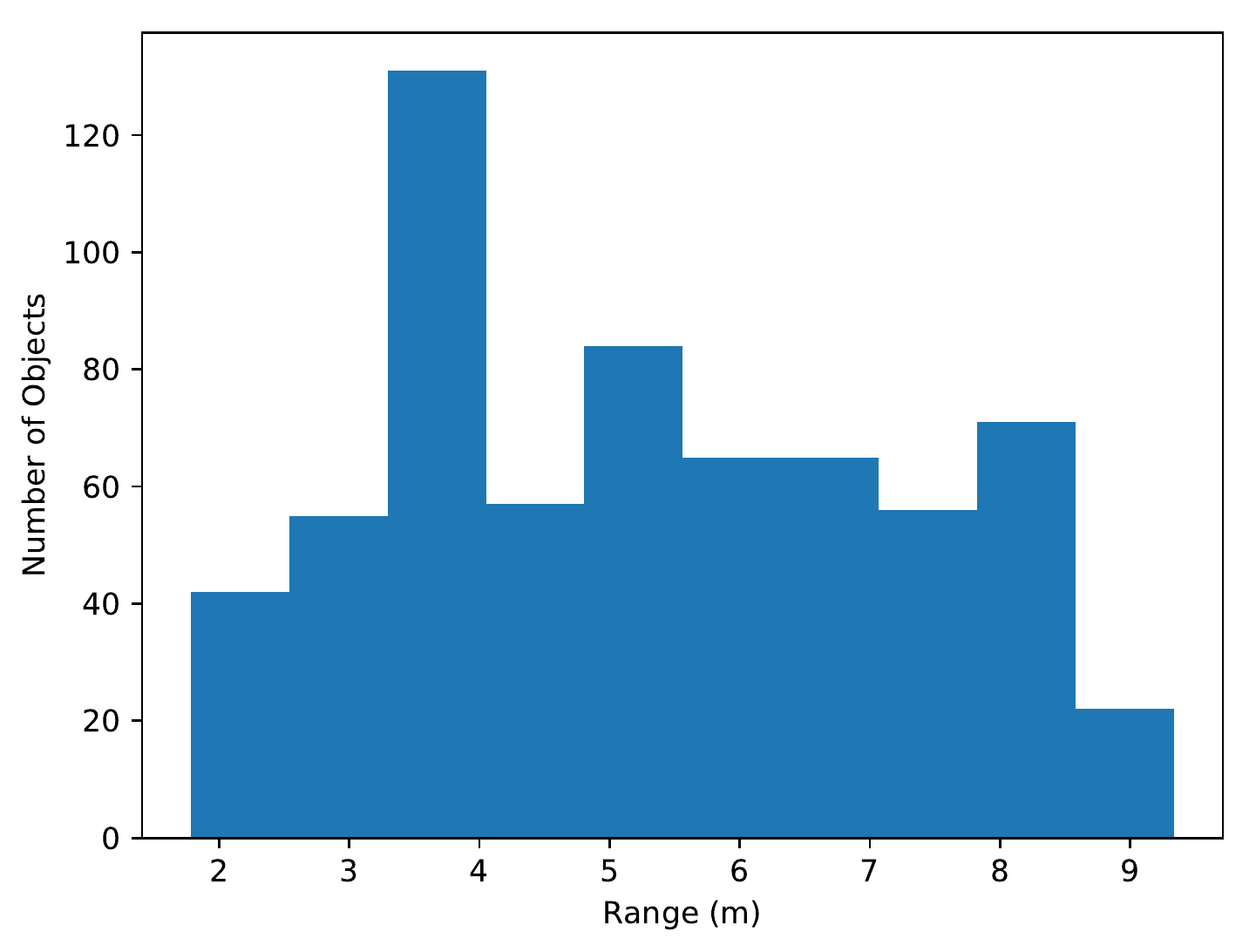}
			\caption{Histogram of distribution of objects over range}
			\label{fig:hist_range}
		\end{subfigure}
		\caption{Multi-object dataset statistics}
		\label{fig:multiobj_stats}
	\end{figure*}

	 This new dataset contains 198 scenes, 648 objects, an average of 3.27 movable objects per scene. Figure \ref{fig:multiobj_stats} shows statistical data explaining the number of instances of each learnt object, the number of objects in each scene, and the distribution of ranges of the objects.

	\subsection{Methodology}\label{sec:det_method}
	
	In classical radar terminology, detection is described as "determining whether the receiver output at a given time represents the echo from a reflecting object or only noise" \cite{richards2005fundamentals}.
	Conversely, in computer vision, using visible camera imagery to which the vast majority of CNN methods have been applied, detection is the precise location of an object in an image (assuming it is present) containing many other objects, as for example in the pedestrian detection survey of Dollar et al. \cite{dollar2012}. Although the image may be noisy, this is generally not the major cause of false alarms.
	
	The extensive literature on object detection and classification using cameras, e.g. \cite{ren2017faster, liu2016ssd, lin2017focal, redmon2016yolo}, can be grouped into \textit{one-stage} and \textit{two-stage} approaches. In the \textit{one-stage} approach localisation and classification is done within a single step, as with the YOLO \cite{redmon2016yolo}, RetinaNet \cite{lin2017focal} and SSD \cite{liu2016ssd} methods. Using a \textit{Two-stage} approach first localizes objects, proposing bounding boxes and then performs classification in those boxes. R-CNN \cite{girshick2014rich}, Fast R-CNN \cite{girshick2015fast} and Faster R-CNN \cite{ren2017faster} are examples of the \textit{two-stage} approach.
	
\begin{figure*}[t]
    \centering
    \includegraphics[width=\linewidth]{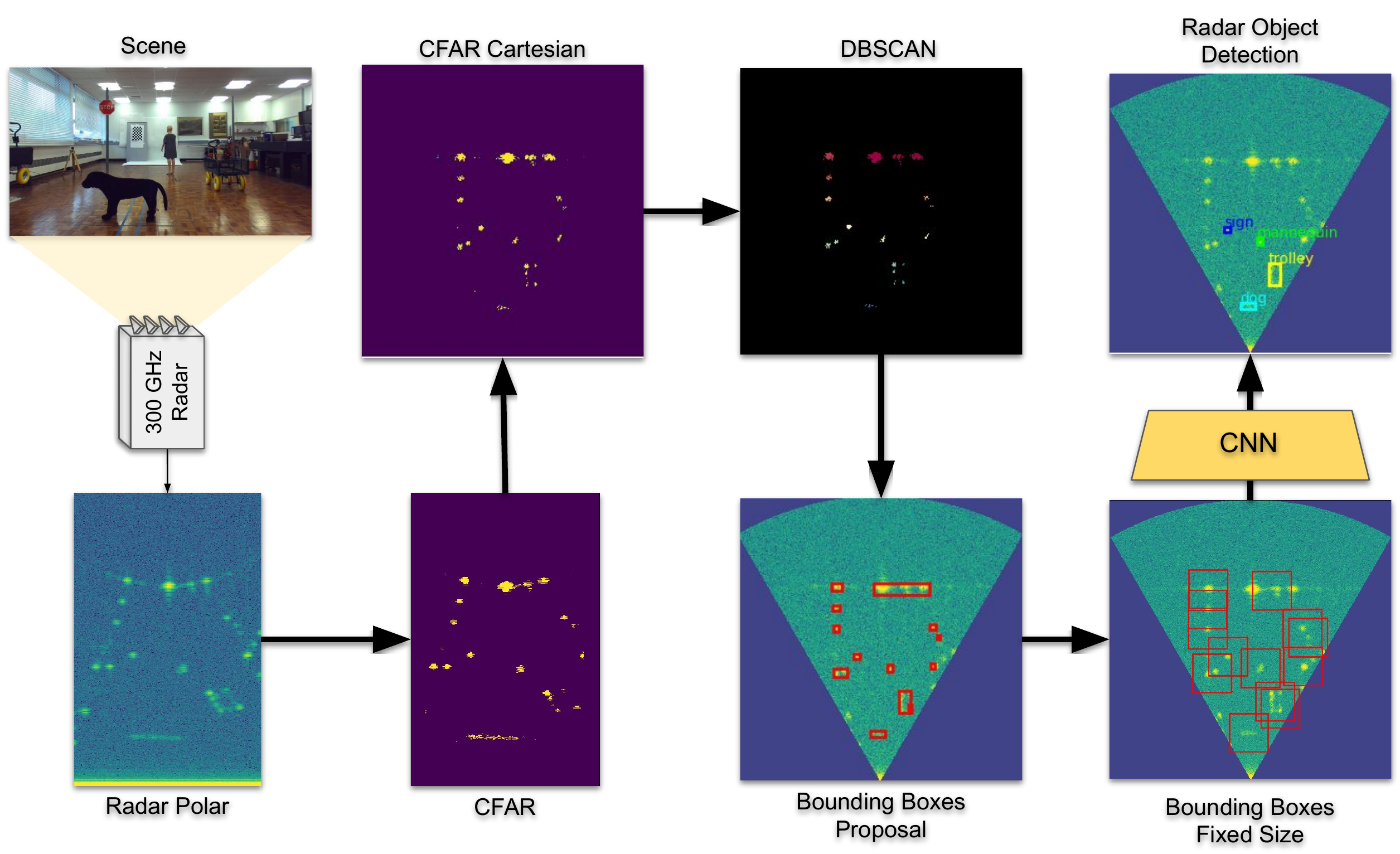}
    \caption{Methodology developed for the detection task.}
    \label{fig:detection_method}
\end{figure*}

	For this work we developed a \textit{two-stage} technique. We first generate bounding boxes based on the physical properties of the radar signal, then the image within each bounding box is classified, similar to the R-CNN \cite{girshick2014rich}.
	Figure \ref{fig:detection_method} shows the pipeline of the detection methodology developed.
	For radar echo detection, we use simply \textit{Constant False Alarm Rate} (CFAR) \cite{richards2005fundamentals} detection.
	There are many variations including \textit{Cell Averaging Constant False Alarm Rate} (CA-CFAR) and \textit{Order Statistics Constant False Alarm Rate} (OS-CFAR). In this work we used the CA-CFAR algorithm to detect potential radar targets. In order to compute the false alarm rate, we measured the background noise level, and the power level from the objects (using training images), setting a CFAR level of $0.22$. 
	After detecting potential cells, we form clusters using the common \textit{Density-based spatial clustering of applications with noise} (DBSCAN) algorithm \cite{Ester:1996:DAD:3001460.3001507} which forms clusters from proximal points and removes outliers.
	For each cluster created we use the maximum and minimum points to create a bounding box of the detected area. The parameters for DBSCAN used were selected empirically; $\epsilon = 0.3 m$ which is the maximum distance of separation between 2 detected points, and $S = 40$, were $S$ is the minimum number of points to form a cluster.
	
	To compute the proposed bounding boxes with DBSCAN, we use the center of the clusters to generate fixed size bounding boxes of known dimensions, since, in contrast to the application of CNNs to camera data, the radar images are metric and of known size. Hence, the boxes are of size $275 \times 275$, the same size as the data used to train the neural network for the classification task. The image is resized to $88 \times 88$ and each box is classified. As with the isolated objects experiments, we used the A-ConvNet architecture.
	
	To consider the background we randomly cropped 4 boxes which do not intersect with the ground truth bounding boxes containing objects in each  scene image from the multiple object dataset and incorporated these in our training set. However, as there are effectively two types of background, that which contains other unknown objects such as the wall, and the floor areas which have low reflected power, we ensured that the random cropping contained a significant number of unknown object boxes. This is not ideal, but we are limited to collect data in a relatively small laboratory area due to the restricted range of the radar sensor and cannot fully model all possible cluttering scenarios.
	
	\begin{table*}[h]
		\caption{Perfect Detector}\label{tab:pd_300}
		\resizebox{\textwidth}{!}{%
			\begin{tabular}{l|c|c:ccc|c:ccc|c:ccc|c|c|c}
				\multicolumn{1}{c|}{\multirow{2}{*}{AP}}& \multirow{2}{*}{Overall} & \multicolumn{4}{c|}{$\# Objects < 4$} & \multicolumn{4}{c|}{$4 \leq \# Objects < 7$} & \multicolumn{4}{c|}{$\# Objects \geq 7$} & \multirow{2}{*}{Short} & \multirow{2}{*}{Mid} & \multirow{2}{*}{Long} \\
				& & Overall & Short & Mid & Long & Overall & Short & Mid & Long & Overall & Short & Mid & Long & & & \\ \hline

				bike & 64.88 & 79.17 & 50.00 & 83.33 & 75.00 & 48.42 & 25.00 & 56.77 & 33.33 & 76.26 & N/A & 67.05 & N/A & 35.00 & 66.19 & 57.14 \\ 
				cone & 46.87 & 50.00 & 50.00 & 50.00 & 50.00 & 58.29 & 55.56 & 83.33 & N/A & 42.49 & 68.75 & 26.67 & N/A & 62.07 & 43.30 & 3.57 \\ 
				dog & 51.34 & 77.62 & 77.78 & 87.72 & 47.62 & 49.13 & 77.78 & 55.45 & 33.33 & 26.40 & 60.0 & N/A & 12.50 & 70.95 & 65.02 & 20.19 \\ 
				mannequin & 37.73 & 70.53 & 53.33 & 85.71 & 33.33 & 25.57 & 36.36 & 30.00 & 8.00 & 37.78 & 14.29 & 50.00 & 22.35 & 33.08 & 48.72 & 13.61 \\ 
				sign & 85.64 & 81.86 & 0.00 & 89.47 & 66.67 & 86.60 & N/A & 90.10 & 81.08 & 86.44 & N/A & 88.89 & 85.46 & 0.00 & 89.65 & 81.94 \\ 
				trolley & 81.68 & 87.75 & 79.17 & 97.06 & 82.35 & 85.35 & 100.00 & 87.61 & 70.13 & 75.45 & 92.67 & 83.65 & 10.00 & 93.53 & 88.76 & 60.41 \\ 
				\hline 
				mAP & 61.36 & 74.49 & 51.71 & 82.22 & 59.16 & 58.89 & 58.94 & 67.21 & 37.65 & 57.47 & 58.93 & 63.25 & 26.06 & 49.1 & 66.94 & 39.48 \\ 
				
		\end{tabular}}

		\caption{CFAR+DBSCAN Detector Easy}\label{tab:dbscan_easy_300}
		\resizebox{\textwidth}{!}{%
			\begin{tabular}{l|c|c:ccc|c:ccc|c:ccc|c|c|c}
				\multicolumn{1}{c|}{\multirow{2}{*}{AP}}& \multirow{2}{*}{Overall} & \multicolumn{4}{c|}{$\# Objects < 4$} & \multicolumn{4}{c|}{$4 \leq \# Objects < 7$} & \multicolumn{4}{c|}{$\# Objects \geq 7$} & \multirow{2}{*}{Short} & \multirow{2}{*}{Mid} & \multirow{2}{*}{Long} \\
				& & Overall & Short & Mid & Long & Overall & Short & Mid & Long & Overall & Short & Mid & Long & & & \\ \hline

				bike & 53.97 & 79.17 & 50.00 & 66.67 & 100.0 & 43.79 & 50.00 & 36.28 & 6.67 & 43.80 & N/A & 31.82 & N/A & 50.00 & 42.39 & 65.08 \\ 
				cone & 19.49 & 36.36 & 50.0 & 16.67 & 0.00 & 47.60 & 55.56 & 60.00 & N/A & 2.12 & 0.00 & 3.81 & N/A & 23.71 & 18.16 & 0.00 \\ 
				dog & 34.32 & 53.36 & 77.78 & 77.35 & 12.12 & 31.09 & 50.00 & 37.01 & 0.00 & 18.18 & 60.0 & N/A & 0.00 & 64.00 & 47.33 & 3.33 \\ 
				mannequin & 36.91 & 70.57 & 64.00 & 85.71 & 16.67 & 21.51 & 32.73 & 26.67 & 5.83 & 39.66 & 0.00 & 52.94 & 29.41 & 29.57 & 48.72 & 14.22 \\ 
				sign & 81.84 & 81.86 & 0.00 & 89.47 & 66.67 & 81.65 & N/A & 84.88 & 77.17 & 83.89 & N/A & 83.33 & 84.5 & 0.00 & 85.02 & 79.31 \\ 
				trolley & 75.55 & 77.30 & 67.42 & 87.72 & 71.56 & 81.56 & 97.44 & 87.32 & 56.73 & 71.33 & 79.56 & 74.49 & 13.33 & 82.46 & 80.78 & 51.62 \\ 
				\hline 
				mAP & 50.35 & 66.44 & 51.53 & 70.60 & 44.50 & 51.20 & 57.14 & 55.36 & 24.40 & 43.16 & 34.89 & 49.28 & 25.45 & 41.62 & 53.73 & 35.60 \\ 
				
		\end{tabular}}

	\end{table*}

	\subsection{Results for Multiple Objects}\label{sec:det_results_300}
	
	In order to evaluate performance, we have considered 2 different scenarios. 
	In particular, we wish to ascertain how performance is affected by failures in classification, using mean averange precision (mAP), assuming a perfect CFAR+DBSCAN pipeline, and to what extent failures in the box detection process lead to mis-classification. Further, we make a distinction between confusing objects (mainly the lab wall) and due to system noise from the floor area.
	
	\begin{itemize}
		\item \textbf{Perfect Detector} : In this scenario we do not use the CFAR + DBSCAN pipeline, we use the ground truth as the detected bounding boxes. Each bounding box is fed to the trained neural network.
		
		\item \textbf{Easy} : In this scenario we manually crop the walls and focus on the potential area containing objects of interest. This includes the CFAR + DBSCAN in a easy scenario, in which removal of static objects is analogous to background subtraction.
		
	\end{itemize}

	We also decided to label our scene data depending on the density of objects, since a highly cluttered scene should increase the likelihood of unwanted radar sensing effects, such as multi-path, occlusion, and multiple objects in the same bounding box.
	
	\begin{itemize}
		\item \textbf{\#Objects $<$ 4} : At low density of objects, it is likely that the scene will suffer less from these effects.
		\item \textbf{4 $\leq$ \#Objects $<$ 7} : At mid density, we will encounter some of the unwanted effects.
		\item \textbf{\#Objects $\geq$ 7} : At high density, many of these effects occur.
	\end{itemize}
	
	We also have decided to evaluate performance at different ranges.
	
	\begin{itemize}
		\item \textbf{Short Range (Objects $<$ 3.5 m)}: This scenario is not necessarily the easiest since coupling between the the transmitter and receiver happens at this range \cite{8468328}.
		\item \textbf{Mid Range (3.5 m $<$ Objects  7 m)}: This is the ideal scenario, as the objects were learnt within these ranges, and the antenna coupling interference is reduced.
		\item \textbf{Long Range (Objects $>$ 7 m)}: This is the most challenging scenario. At more than 7 meters, most of the objects have low power of return, close to the systemic background.
	\end{itemize}
	
	The metric we use for evaluation is average-precision (AP) which is a commonly used standard in the computer vision literature for object detection, classification and localisation \cite{everingham2015pascal} in which the Intersection over Union (IoU) measures the overlap between 2 bounding boxes. We used the same metric previously in the Section \ref{sec:ir_eval}, where it is defined.

	For these experiments we retrained the neural network from the single object dataset using the orientation experiments. For the \textit{Easy} and \textit{Perfect Detector} cases, we do not include the background data.
	Extensive results for all these scenarios are shown in Tables \ref{tab:pd_300} and \ref{tab:dbscan_easy_300}, where \textit{overall} means the metric computed for all samples in all ranges.

	As expected, the results from a scene containing many known objects and confusing artefacts are much poorer than when the objects are classified from images of isolated objects.
	Nevertheless, the results show promise.  For example, considering the mid range, \textit{Perfect Detector} case, there is an overall \newtext{mean average precision (mAP)} of $61.36\%$,
	and for specific easily distinguishable objects such as the trolley it is as high as $97.06\%$ in one instance. Other objects are more confusing, for example cones usually have low return power and can be easily confused with other small objects. As also expected the results degrade at long range and in scenes with a higher density of objects.
	
	The \textit{Easy} case shows performance comparable but not as good as the \textit{Perfect Detector}, for example dropping to $50.35\%$ mAP. The CFAR + DBSCAN method is a standard option to detect objects in radar, but it does introduce some mistakes where, for example, the bounding box is misplaced with respect to the learnt radar patterns.

	Finally, we observe that the trolley is the easiest object to recognise in both cases. The trolley has a very characteristic shape, and strongly reflecting metal corner sections that create a distinguishable signature from all other objects. In interpreting true and false results in non-standardised datasets, which is the case in radar as opposed to visible camera imagery, one should be careful when comparing diverse published material.

	\section{Conclusions}
	
	In this chapter we evaluated the use of DCNNs applied to images from a 300 GHz radar system to recognise objects in a laboratory setting. Four types of experiments were performed to assess the robustness of the network.
	These included the optimal scenario when all data is available for training and testing at different ranges, different viewing angles, and using different receivers. As expected, this performs best when all the training and test data are drawn from the same set.
	This is a valuable experiment as it sets an optimal benchmark, but this is not a likely scenario for any radar system applied in the wild, first because radar data is far less ubiquitous or consistent than camera data, and second because the influence of clutter and multipath effects are potentially more serious than for optical technology.
	
	Regarding the single object scene data, we should be encouraged by two principal results, first that the performance was so high for the optimal case, and second that transfer learning may lead to improvements in other cases,
	Transfer learning from MSTAR using A-ConvNet can prevent overfitting to the 300 GHz source data, by generalizing using more samples from a different radar data set, e.g. increasing from $92.5\%$ to $98.5\%$ in the experiment using Q1 and Q3 to train and Q2 and Q4 to test. This leads to more robust classification.

	The multiple object dataset is a very challenging scenario, but we achieved mean average precision rates in the easy case $> 60\% (< 4 objects)$, but much less, $35.18\%$, in a high cluttered scenario. However, the pipeline we have adopted is probably subject to improvement, in particular using the classification results to feed back to the detection and clustering. To avoid problems with occlusion, object adjacency, and multi-path, further research on high resolution radar images is necessary.
	We also note that we have not made use of Doppler processing, as this implies motion of the scene, the sensor or both. For automotive radar, there are many stationary objects (e.g a car at a traffic light), and many different motion trajectories in the same scene, so this too requires further research.

%% file: Chapters/Chapter6_300ghz_DA_mdpi_version.tex
	\chapter{300 GHz Radar Object Recognition Based on Effective Data Augmentation}\label{ch:radio}
	
	\lhead{Chapter 5. \emph{300 GHz Radar Object Recognition Based on Effective Data Augmentation}} %

\begin{figure*}[t]
\centering
\includegraphics[width=0.7\linewidth]{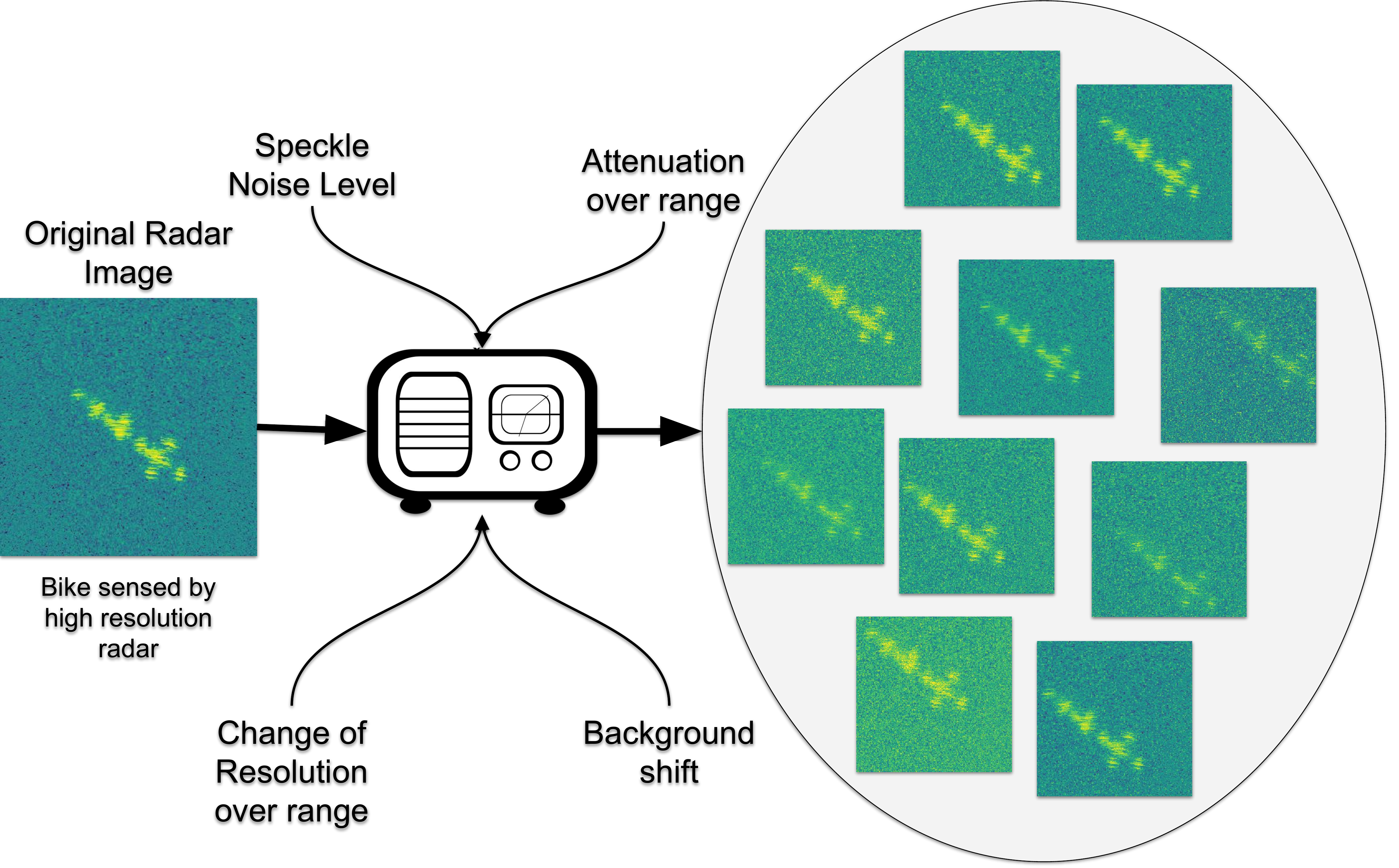}
\caption{\textbf{RADIO:} \textbf{RA}dar \textbf{D}ata Augmentat\textbf{IO}n. This method uses domain specific knowledge about the radar sensor to generate realistic radar data. It can simulate attenuation, change of resolution, speckle noise and background shift.}
\label{fig:radio_methodology1}
\end{figure*}

The previous chapter showed the effectiveness of using deep neural networks together with transfer learning. Transfer learning can improve generalizability by capturing a larger number of features from a larger dataset. Another way of creating generalizability is to artificially augment the dataset. We present in this chapter a novel, parameterised RAdar Data augmentatIOn (RADIO) technique to generate realistic radar samples from small datasets for the development of radar related deep learning models. RADIO leverages the physical properties of radar signals, such as attenuation, azimuthal beam divergence and speckle noise, for data generation and augmentation (Figure \ref{fig:radio_methodology1}). Exemplary applications on radar based classification and detection demonstrate that RADIO can generate meaningful radar samples that effectively boost the accuracy of classification and generalizability of deep models trained with a small dataset.
	 
\section{Introduction}\label{sec:intro}

We have studied in the previous chapter the performance of a number of neural network architectures applied to 300 GHz data for object detection and classification using a prototypical object set of six different objects in both isolated and multiple object settings.
Our results were very much dependent on the different scenarios and objects, achieving accuracy ratings in excess of $90\%$ in the easier cases, but much lower success rates in a detection and classification pipeline with many confusing, objects in the same scene.
In our work, we do not rely on Doppler signatures \cite{rohling2010pedestrian,bartsch2012pedestrian,angelov2019} as we cannot consider only moving objects, nor do we use multiple views \cite{lombacher2017semantic} to build a more detailed radar power map, as these are not wholly suited to the automotive requirement to classify both static and moving actors from a single view.

\begin{figure*}[t]
\centering
\includegraphics[width=\linewidth]{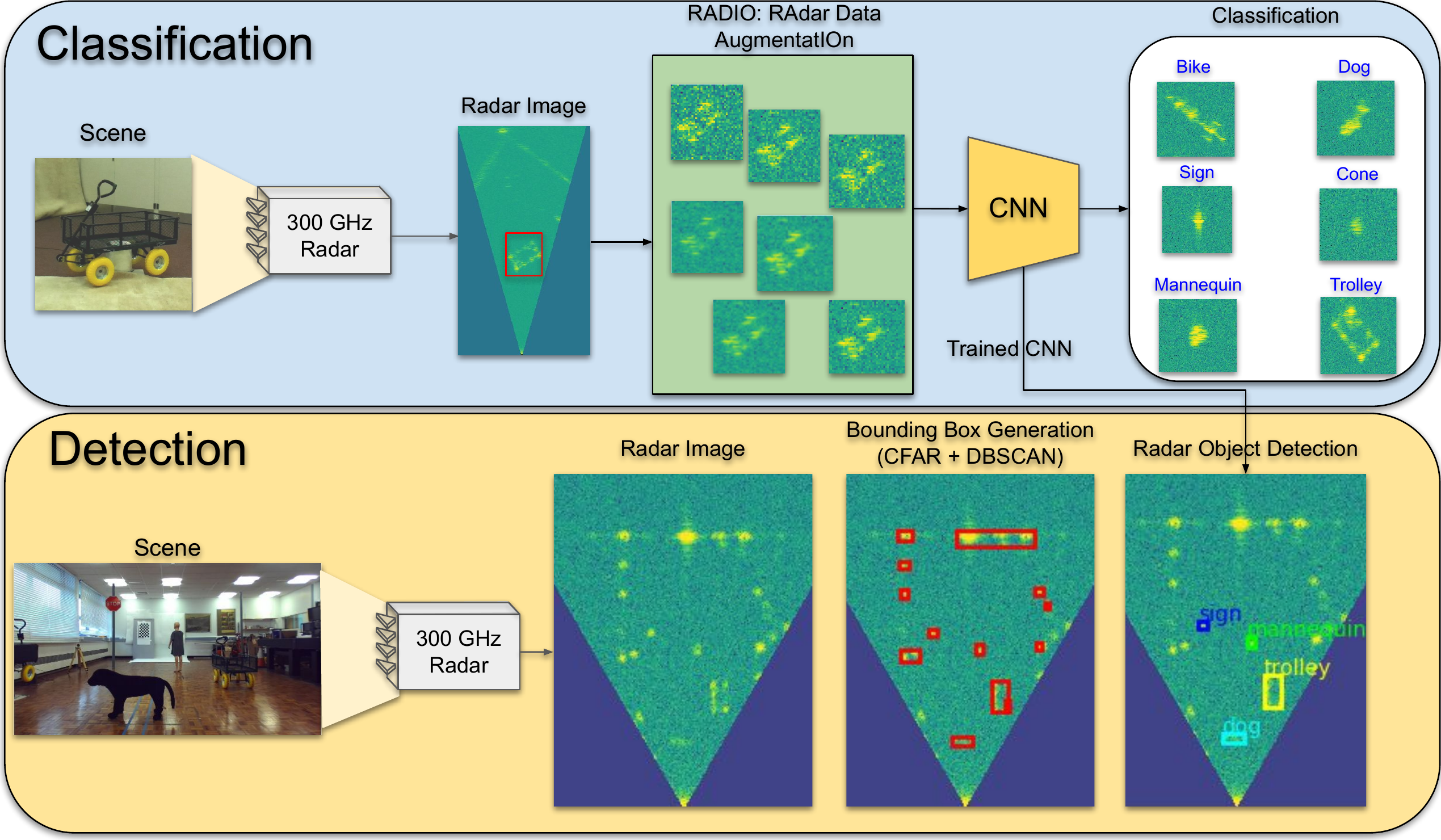}
\caption{RADIO is a data augmentation technique based on the physical properties of the radar. The methodology is developed for both object detection and classification.}
\label{fig:methodology}
\end{figure*}

In our work and indeed in the general literature, radar datasets tend to be small due to the difficulty of data collection and human annotation. While we endeavour to collect and label larger datasets in the wild, this is challenging due to a lack of advanced, high resolution automotive radars for high frame rate imaging, and the difficulty of data labelling in comparison with video sequences.  Therefore, to create larger datasets to train deep models for such tasks, we present in this chapter a novel radar data augmentation technique based on physical properties which can be used for both classification and detection (Fig. \ref{fig:methodology}).
Our experimental hypothesis is that such augmentation will improve the accuracy over standard camera based data augmentation source data. Hence, the main contributions of this chapter are:
\begin{itemize}
    \item A novel radar data augmentation technique (RADIO) based on the physical properties of the radar signal. This models signal attenuation and resolution change over range, speckle noise and background shift for radar image generation.
    \item We demonstrate that such data augmentation can boost the accuracy and generalizability of deep models for object classification and detection, trained only with a small amount of source radar data. This verifies the effectiveness of RADIO.
  
\end{itemize}

\section{RADIO: Radar Data Augmentation}\label{sec:data_aug}

Using a restricted dataset, a DNN will easily overfit and be biased towards specific artifacts in the few examples the dataset provides. Our original experiments to recognise objects in 300 GHz radar data had a database of only 950 examples at short range.
In this chapter, we report a radar data augmentation technique to generate realistic data in which the neural network can learn patterns of data not present in the restricted source data, and so avoid overfitting.
Given the nature of radar signal acquisition and processing, the  received power and azimuthal resolution vary with range and viewing angle. We also consider the effects of speckle noise and background shift during augmentation.

\subsection{Attenuation}
\label{sec:attenuation}

\begin{figure}[t]
\centering
\includegraphics[width=0.8\linewidth]{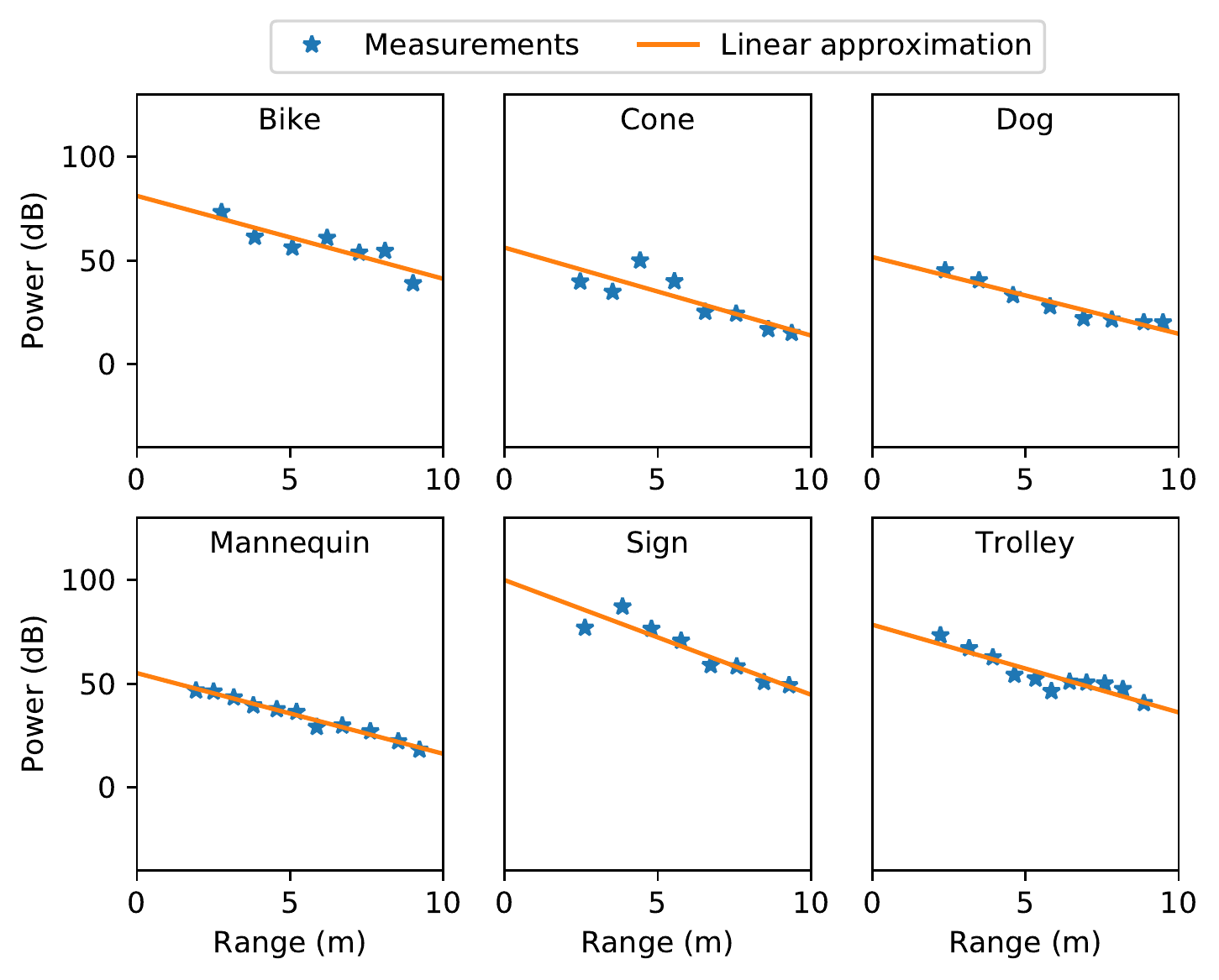}
\caption{Attenuation graphs}
\label{fig:attenuation}
\end{figure}

In the basic form of the radar equation, the power received by the radar transceiver \cite{radartutorial} can be computed by
\begin{equation}\label{eq:power1}
    P_r = \frac{P_t G^2 \sigma \lambda^2}{(4 \pi)^3  R^4 L},
\end{equation}
\noindent where $P_r$ is the received power, $P_t$ is the transmitted power, $G$ is the antenna gain, $\sigma$ is the radar cross section (RCS), $R$ is the range between the radar and the object, $\lambda$ is the radar wavelength and $L$ is the loss factor. $L$ can be atmospheric loss, fluctuation loss and/or internal attenuation from the sensor. $P_r$ and $P_t$ are measured in Watts (W), for our experiments, we used $P_r$ in Decibel (dB), where its value in dB computed by

\begin{equation}
    \hat{P}_r = log_{10}(P_r).
\end{equation}

However, this assumes an idealised model in which a point, isotropic radiator modified by the directional gain, $G$, propagates a radio signal that is reflected by a set of point scatterers in the far field at the same range, $R$.
In practice, in our experiments we use antennae of the order of 6 cm lateral dimension propagating at 300 GHz to extended targets in the range 3-12 m, and simulate radar images in the range 2-15 m, as our operating range was constrained by the limited radar power and lab dimensions, such that we operated in the near field below the crossover point.
Further, for precise theoretical computation of the received power, we would need information regarding the shape of the objects, placement of scatterers, surface material reflectivity and of multiple reflections caused by the surrounding walls, floor, ceiling and furniture \cite{knott2004radar}. Characterising the nature of point scatterers, for real traffic actors such as vehicles is a very daunting and error prone task; simulation and measurement rarely correspond \cite{SchBecWie2008}. 
Our own data set is restricted, as shown in Fig. \ref{fig:samples}, but there are still noticeable differences between say, the cone which is a symmetrical plastic object,
and the trolley which is a collection of linear metallic struts and corners that act almost as ideal corner reflectors.
Hence, as complete modelling is intractable, to augment our radar data we have taken an empirical approach that measures the received power over pixels in the radar images as a function of range for all objects in our target set, and hence generates augmented radar image data at ranges not included in the real data using a regression fit to the actual data.  

Fig. \ref{fig:attenuation} shows the measurements and the linear regression model for the six different objects (trolley, bike, stuffed dog, mannequin, sign and cone).
To collect this data, we manually placed objects at different ranges and retrieved the mean received power intensity ($P_r$) from each object as a function of range.
To compute $P_r$, the area of the image containing the object was cropped manually, then a simple threshold was applied to remove background pixels.
Simple thresholding is applicable to the training data as the object power levels are considerably higher than the background, which was at a consistent level at all ranges.
Looking at the collected data points, it appears that a reasonable approximation for variation of received power (dB) over range can be made by a linear regression between the received power and range, which is not predicted by Eq. \ref{eq:power1}, which would suggest that the linear relationship should be over the logarithm of range.
However, as we have noted the basic equation makes many inapplicable assumptions, so we employ the empirical approximation in this scenario. An algorithm to describe the change of attenuation can be seen in the Algorithm \ref{alg:attenuation}. In the Algorithm \ref{alg:attenuation}, $I_ori$ is the original image, $c$ is the object class, $r_{curr}$ is the current range and $r_new$ is the new range which will be simulated. The function \textit{segment} is segmentation method which uses a simple threshold that separates the object pixels from the background pixels. The function \textit{getSlope} gets the slope depending on the class of the object. The Algorithm \ref{alg:attenuation} gets the original pixels and attenuate each pixel according the the simulated attenuation, while the background keeps the same value.

\begin{figure}[h]
\centering
\includegraphics[width=0.8\linewidth]{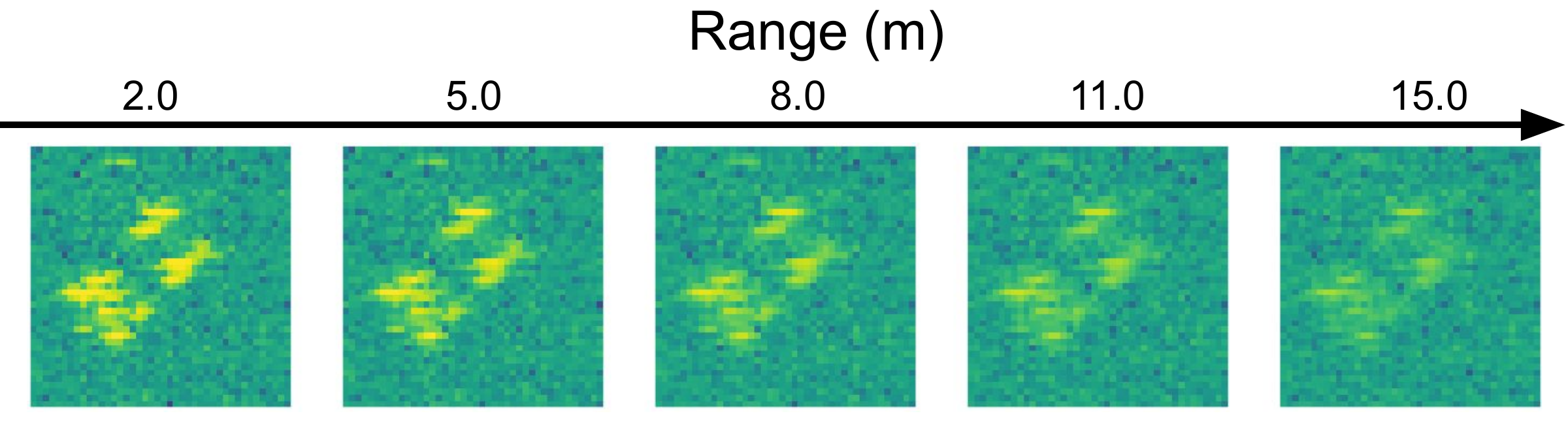}
\caption{Range data augmentation}
\label{fig:range_image}
\end{figure} 

\begin{algorithm}
\setlength\baselineskip{20.5pt}
\caption{Attenuation procedure}\label{alg:attenuation}
\begin{algorithmic}[1]
\Procedure{Attenuation}{$I_{ori},c,r_{curr}, r_{new}$}

\State $I_{seg} = segment(I_{ori})$
\State $r_{diff} = r_{new} - r_{curr}$
\State $a = getSlope(c)$
\State $I_{seg_{att}} = I_{seg} * I_{ori} + r_{diff} * a$
\State $I_{att} = \neg I_{seg} * I_{ori} + I_{seg_{att}}$
\State \textbf{return} $I_{att}$
\EndProcedure
\end{algorithmic}
\end{algorithm}

\subsection{Change of Resolution}

The next effect we consider is the change of resolution over range. The raw radar signal is in a polar coordinate system; when converting to Cartesian, interpolation is necessary to convert cells of irregular size as a function of range to uniformly spaced square cells (or pixels). The resolution over range can be measured by

\begin{equation}
    a = 2R\sin\Bigg(\frac{\theta}{2}\Bigg),
\end{equation}\label{eq:angres}

\noindent where $R$ is the range and $\theta$ is the antenna angular resolution. Again, this means that the distribution of assumed point scatterers to cells might be quite different and this is not measurable, as stated in Section \ref{sec:attenuation}.
To account for the effect of changing Cartesian resolution with range, we determine the size of the polar cell at the range at which we wish to augment the radar data, and use nearest neighbor interpolation to generate the new Cartesian image. Fig. \ref{fig:range_image} shows examples of trolley images generated at ranges from $2-15$ m using this methodology and as source data the real range image at $6.3$ m.

\begin{figure}[ht]
\centering
\includegraphics[width=0.8\linewidth]{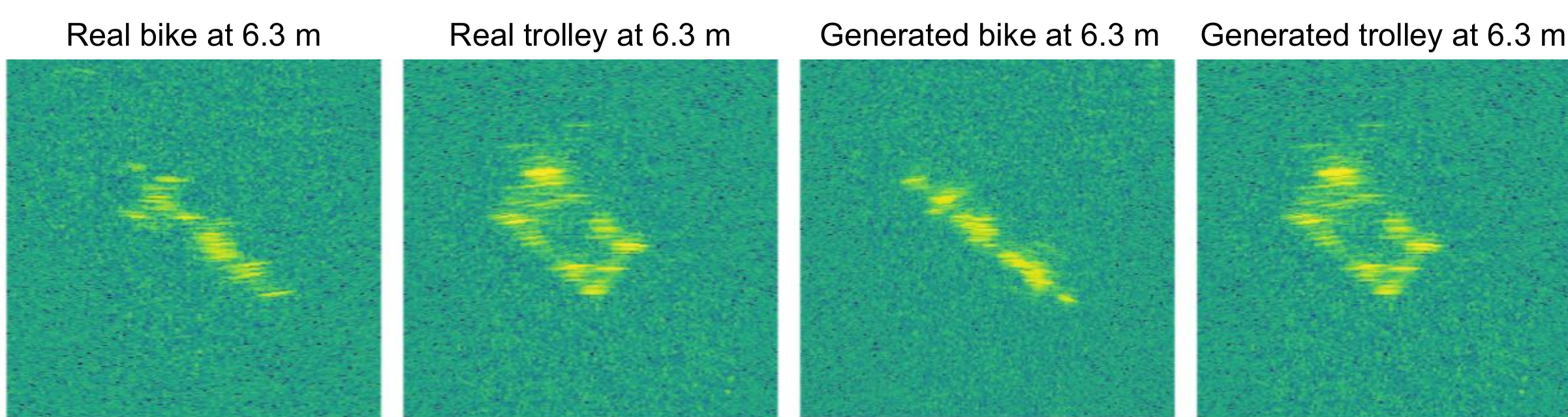}
\caption{Comparing the real data at 6.3 m with the RADIO augmented data.}
\label{fig:sanity_check}
\end{figure}

To check whether the augmented data has similarity to real data, and has thus some prospect of improving the training data, we compared the mean sum of absolute differences (MSAD) between real and augmented $3.8$ m and $6.3$ m images. As source data, we used the real $3.8$ m image, but in this case we generated both a $3.8$ m augmented image and a $6.3$ m image from the source data and the regression curves in Fig. \ref{fig:attenuation}.
We used all samples with the same rotation for each class.
The differences between the real and augmented images can be compared visually in Fig. \ref{fig:sanity_check}, and are also shown in Tab. \ref{tab:sanity_check}.
The low values for MSAD between the real and augmented images give us some confidence that the coupled attenuation and resolution procedures may be effective as an augmentation process throughout the range for training a neural network.

\begin{table}
\centering
\caption{Mean sum of absolute differences (MSAD) between the real and augmented images at 3.8m and 6.3m, generated from real data at 3.8m only.}\label{tab:sanity_check}
\begin{tabular}{l|cccccc}\hline
{} &   Trolley &      Bike &      Cone &  Mannequin &      Sign &       Dog \\
\hline
MSAD (3.8m) &  2.593 &  1.444 &  0.073 &   0.551 &  0.426 &  0.964 \\

MSAD (6.3m)    &  \textbf{2.374} &  \textbf{1.298} &  \textbf{0.056} &   \textbf{0.506} &  \textbf{0.359} &  \textbf{0.881} \\
\hline
\end{tabular}
\end{table}

\subsection{Speckle Noise and Background Shift}
\label{sec:speckle_noise}

\begin{figure}[h]
\centering
\includegraphics[width=0.8\linewidth]{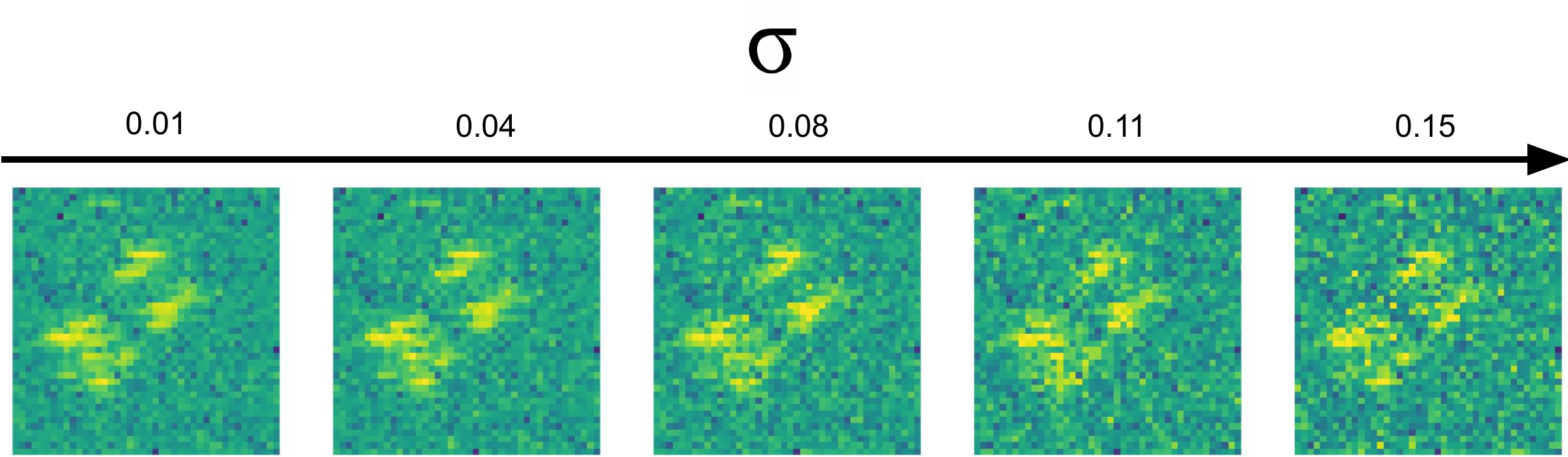}
\caption{Speckle noise data augmentation}
\label{fig:speckle_image}
\end{figure}

\begin{figure}[h]
\centering
\includegraphics[width=0.8\linewidth]{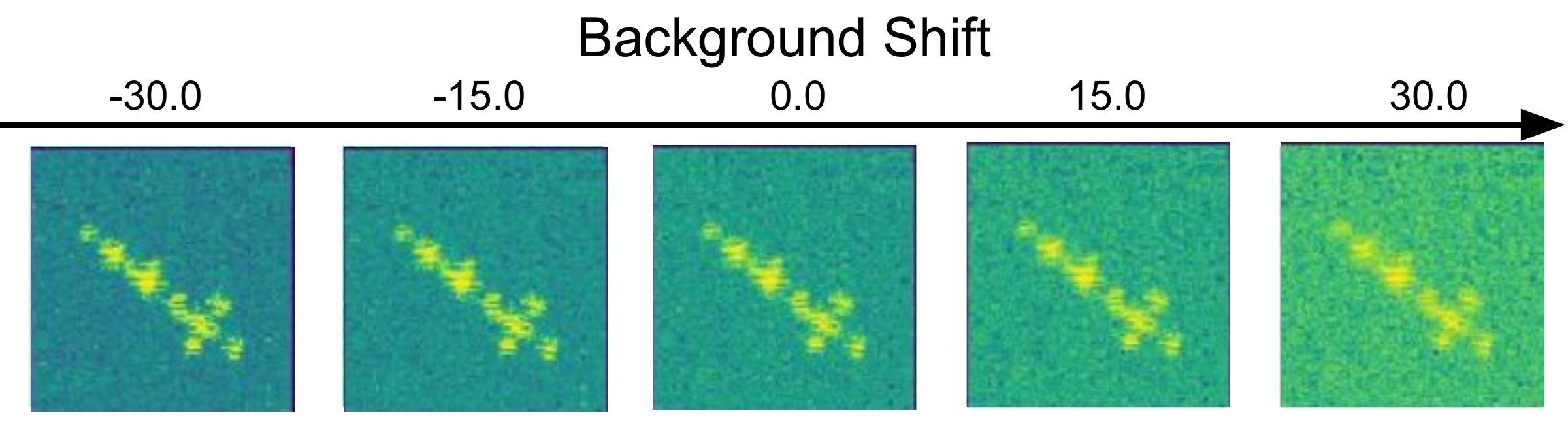}
\caption{Background shift data augmentation}
\label{fig:bg_shift_images}
\end{figure}

To further simulate real radar data we included the capability to approximate the effects of different levels of speckle noise and background shift. 
Full modelling of speckle noise requires detailed knowledge of the properties of the radar system and antennae, but drawing on previous work \cite{ding2016convolutional} we performed an approximation by imposing multiplicative Gaussian noise, $I = I \times \mathcal{N}(1,\,\sigma^{2})$ where $\sigma^{2}$ is sampled uniformly from 0.01 to 0.15. 
Examples of speckle noise augmentation are given in Fig. \ref{fig:speckle_image} with different multiplicative noise levels. 
As a final process, we also added background shift. This changes the brightness creating scenes with objects and different background levels. It is a simple technique where CFAR is applied to the image classifying the areas of the image as either object or background. A constant value is added/subtracted only to/from the background. Fig. \ref{fig:bg_shift_images} shows examples of images applying background shift.

\section{Experimental Results}

The purpose of the experimental study is to verify the hypothesis that radar data augmentation will lead to improved recognition of objects in radar images when compared to either the raw data alone, or to data augmented by the ''vanilla'' techniques \cite{Goodfellow:2016:DL:3086952} commonly applied to visual image databases when training neural networks. By using RADIO we can simulate more realistic radar images which is more appropriate for radar datasets. The neural network used in our work is  A-ConvNet \cite{chen2016target}, same network from Chapter \ref{ch:300dl}.

We have sub-divided these experiments into two main sets.
In the first case, which we term \textit{Classification}, we consider images of isolated objects as both training and validation/test set data. This is a 'per window' task.
In the second case we consider \textit{Detection} and \textit{Classification} in which the training data is as before, but the test/validation data is a scene or image containing one or more instances of the known objects as well as additional background.
This is a 'per image' task.

\subsection{Classification}

For classification, we used the single object dataset which was used in the Chapter \ref{ch:300dl}.
All the collected images were labelled with the correct object identity, irrespective of viewing range, angle and receiver height. A fixed size bounding box of $400 \times 400$ cells, which corresponds to $3m^2$,  was cropped with the object in the middle. As stated above, the original images were resized to the $88 \times 88$ input layer, and we used stochastic gradient descent (SGD) with the  parameters shown in Tab. \ref{tab:nn_params_da}.

\begin{table}[h]
\caption{Neural Networks Parameters.}\label{tab:nn_params_da}
\centering

\begin{tabular}{l|c}
\hline
Learning rate ($\alpha$) & 0.001 \\
Momentum ($\eta$)      & 0.9   \\
Epochs        & 100   \\
Batch size    & 100  \\
\hline
\end{tabular}

\end{table}

To test the hypothesis that principled data augmentation by RADIO improves on commonly used, image inspired, data augmentation techniques, we conducted a series of experiments, as shown in  Tab. \ref{tab:sod}. The standard data augmentation (SDA) techniques for image data that we employed are random translations and image mirroring \cite{imgaug}.  Rotations are not used because we already have data at all angles in the image plane. Although, RGB image augmentation may include colour changes, i.e. modifying the colour map, this has no equivalent in the radar data which is monochromatic, so this type of augmentation is not applied.  To ensure comparability of results we applied the same number of camera based data augmentations to each sample such that each dataset is of similar size as shown in Tab. \ref{tab:sod}. We have 475 samples for training, and we decided to employ the data augmentation methods used 41 times for each sample, resulting in total 19,475 artificially generated samples and 19,950 in total.
In Tab. \ref{tab:sod}, range data augmentation includes both attenuation and change of resolution ranges between 1 and 12 meters, the addition of speckle noise with $\sigma$ [0.01 0.15], and 4 levels of background shift, as described in Section \ref{sec:data_aug}. To verify the use of RADIO augmentation at different ranges, and to avoid overfitting to range specific features, we used the images from 3.8 meters to train and the images from 6.3 meters to test.

We used accuracy (${TP+FN} \over {TP+TN+FP+FN}$) as a metric, where $T$ is true, $F$ is false, $P$ is positive and $N$ is negative. 

The key observation from Tab. \ref{tab:sod} is that the progressive augmentation of the training data by the RADIO methods, leads to progressively improved scores according to all the commonly used metrics.
The key comparison is between the second and fifth row of the table, in which equally sized training datasets were used, but the use of the radar adapted augmentation leads to an improvement from $\approx 80\%$
to $99.79\%$. The only mistake is between the sign and the mannequin, as both have a very similar shape signature. Admittedly, this improvement is in a relatively easy scenario as the object location is pre-defined in a classification window. In Section \ref{sec:detection_da} we address a more challenging scenario in which multiple objects are seen in a laboratory  setting with potential occlusion and mutual interference, and there is additional background clutter caused by unknown artefacts.

\begin{table*}[h]
\centering
\caption{Results for the single object, classification task. SDA = Standard (RGB) Data Augmentation, TL = Transfer Learning, RDA = Range Data Augmentation, SN = Addition of Speckle Noise, BS = Background Shift.}\label{tab:sod}
\begin{tabular}{l|cc}
\hline
~  & Samples&  Accuracy (\%) \\ 
\hline
Original Data    & 450& 39.15       \\ 
Original Data + SDA & 19,950 & 82.13       \\
Original Data + SDA + TL (Chapter \ref{ch:300dl}) & 19,950 & 85.20       \\
\hline
Original Data + SDA + SN + BS  (Ours)  & 19,950& 83.83        \\ 
Original Data + SDA + RDA + BS (Ours) & 19,950  & 94.89        \\ 
Original Data + SDA + RDA + SN + BS (Ours)  & 19,950  &  \textbf{99.79}      \\ 
\hline
\end{tabular}
\end{table*}

\subsection{Detection and Classification}
\label{sec:detection_da}

\begin{figure}[t]
    \centering
    \includegraphics[width=1.0\textwidth]{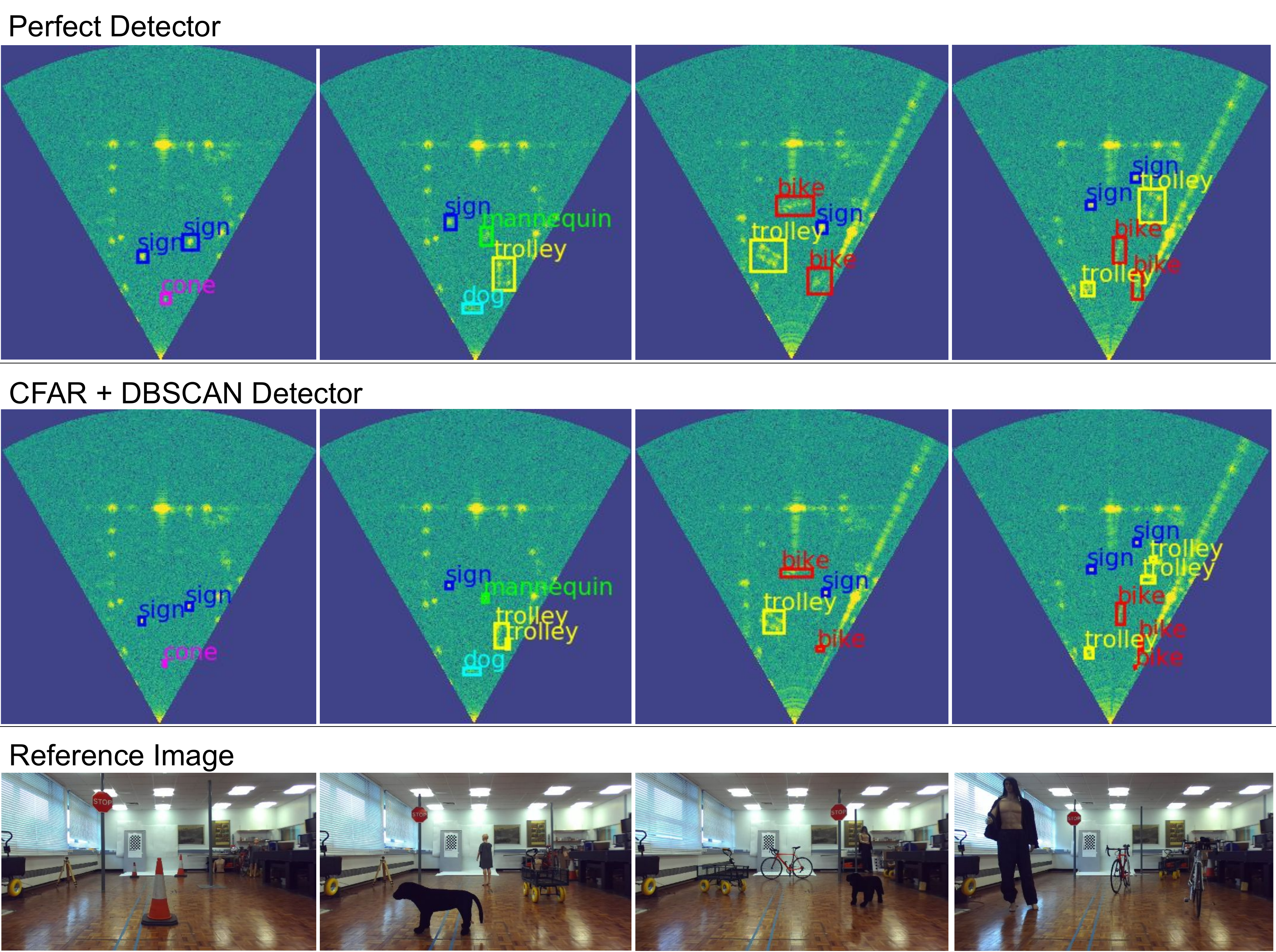}
    \caption{Qualitative results using RADIO for perfect detection, the CFAR+DBSCAN+Network pipeline, and the corresponding video images}
    \label{fig:detection}
\end{figure}

In this more realistic scenario, we also used the same dataset from Chapter \ref{ch:300dl}. This dataset contains several different scenes with several objects viewed against a background containing  walls as a significant feature.
Hence, the scene data includes examples of occluded objects, multi-path reflection and interference between objects.
This dataset contains the same objects (bike, trolley, cone, mannequin, sign, dog) in different parts of the room with arbitrary rotations and ranges. We also include different types of mannequin, cone, trolley, traffic sign, and bikes, so those objects can have different radar signature. Some examples of scene data are shown in Fig. \ref{fig:detection}

To apply detection and classification, we developed a two stage methodology similar to R-CNN \cite{girshick2014rich}, shown in Fig \ref{fig:methodology}.
First, we detect potential regions of interest and then classify each detected region.
The CA-CFAR (Constant False Alarm Rate) detection method \cite{richards2005fundamentals} employs an adaptive threshold based on the statistics of the region to decide whether the return is of sufficient power to be of interest. In our implementation, this was applied to the original polar image at each azimuth direction. The empirically chosen parameters used for CFAR were 500 for cell size, 30 for guard cells and 0.22 for false alarm rate.
Following detection, the standard  \textit{Density-based spatial clustering of applications with noise} (DBSCAN) \cite{Ester:1996:DAD:3001460.3001507} technique was used to form a cluster of cells and remove outliers.
The parameters for DBSCAN used were also selected empirically; $\epsilon = 0.3$ m which is the maximum distance of separation between 2 detected points and $S = 40$, were $S$ is the minimum number of points to form a cluster.
We then generate fixed size bounding boxes of size $275 \times 275$, the same size as the input from the single object images. Then the image is resized to $88 \times 88$ as before.

We aim to separate the effects of detection (by CFAR and clustering) from object classification (using the neural network). Therefore, we perform two types of experiment, first assuming \textit{perfect detection}, second including the whole detection and classification pipeline.
Hence, in the perfect detection case, the exact object location is known, and the question is whether the classifier can still determine the correct object identity in the presence of the additional artefacts described above, bearing in mind that the network has been trained on the isolated objects, not the objects in the multiple object scenarios, and hence cannot learn any features that arise in that instance.
As in the single object classification case, the key hypothesis is that the radar augmentation technique will give improved results when compared to the standard, optical image derived methods.

In our experiments, we classified our scenes with regard to the number of objects and at  short, mid and long ranges. 
Extensive results are shown in Tabs. \ref{tab:pd_da}, \ref{tab:dbscan_easy_da}.
We anticipated that, with more objects in the scene, and at a greater distance, the effects of attenuation and mutual interference will make the classification rates fall.
The metric for evaluation is average-precision (AP) which is used commonly in an object detection scenario. We use the PASCAL VOC implementation \cite{everingham2015pascal} with Intersection over Union (IoU) equals $0.5$. 
AP is computed as $  AP = \int_{0}^{1} p(r)dr$ where $p$ is precision and $r$ is recall.

We retrained the neural network from the single object dataset using both ranges (3.8 m and 6.3 m) and tested on this new more realistic scenario. All the other parameters were the same. We applied the same data augmentation methods as before. We compared the standard data augmentation (SDA) with RADIO (SDA + RDA + SN + BS).
There are several key observations to be made from these experiments.

\begin{itemize}
    \item First, the results in all cases are significantly poorer than in the isolated object case, even for perfect detection, as we would expect as the data is much more challenging and realistic. For example, the overall mAP for perfect detection is only $66.28\%$.
    \item The second conclusion is that there are significant object differences. As ever, the easy task is to recognise the trolley having mAP of $90.08\%$ in the perfect case, as opposed to the mannequin at $41.15\%$ which is often confused with the cone for example.
    \item Third, we observe that, as expected, the type of scene has a significant effect. For example, in the perfect case, the overall mAP for less than $4$ objects is $85.13\%$ but this drops to $52.74\%$ for more than $7$ objects.
    \item Similarly, for $4 < 7$ objects, the mAP drops from $79.09\%$ to $66.58\%$ as we move from short to long range, but the optimal scenario is often at middle range. This is probably attributable to stronger multipath  effects at the shorter range but this would require further examination.
    \item Regarding the cone object, we realised that the reflection of cone, especially for long ranges, can get completely attenuated with values close to the background level, achieving the worse results between all or sensed objects.
\end{itemize}

\begin{table*}[!]
\caption{Perfect Detector. In white standard data augmentation, in yellow Transfer Learning (Chapter \ref{ch:300dl}) and in gray our RADIO technique. The values in {\bf bold} are the best for each scenario.}\label{tab:pd_da}
\resizebox{\textwidth}{!}{%
\begin{tabular}{l||c||c:ccc|c:ccc|c:ccc|c|c|c}
\multicolumn{1}{c|}{\multirow{2}{*}{AP}}& \multirow{2}{*}{Overall} & \multicolumn{4}{c|}{$\# Objects < 4$} & \multicolumn{4}{c|}{$4 \leq \# Objects < 7$} & \multicolumn{4}{c|}{$\# Objects \geq 7$} & \multirow{2}{*}{Short} & \multirow{2}{*}{Mid} & \multirow{2}{*}{Long} \\
& & Overall & Short & Mid & Long & Overall & Short & Mid & Long & Overall & Short & Mid & Long & & & \\ \hline \hline

bike & 65.78 & 81.7 & 50.0 & 88.89 & 75.0 & 53.82 & 0.0 & 59.92 & 66.67 & 66.36 & N/A & 67.2 & N/A & 25.0 & 69.31 & 71.43 \\
\rowcolor{yellow!25}bike (TL) & {\bf77.62} & 86.98 & 100.0 & 88.89 & 75.0 & {\bf73.36} & 50.0 & 70.21 & 86.67 & {\bf69.21} & N/A & 62.63 & N/A & {\bf75.0} & {\bf72.91} & 76.47\\ 
\rowcolor{Gray}
bike (RADIO) & 71.94 & {\bf90.98} & 50.0 & 94.44 & 100.0 & 64.55 & 100.0 & 59.99 & 100.0 & 53.43 & N/A & 48.92 & N/A & {\bf75.0} & 67.17 & {\bf100.0} \\  \hline

cone & 42.29 & 51.25 & 50.0 & 61.11 & 50.0 & 66.44 & 65.08 & 83.33 & 28.57 & 34.35 & 62.5 & 42.5 & 23.19 & 60.44 & {\bf52.71} & 24.22 \\
\rowcolor{yellow!25}cone (TL) & 32.46 & 4.17 & 0.0 & 0.0 & 25.0 & 37.58 & 42.22 & 16.67 & 66.67 & {\bf37.22} & 62.5 & 40.28 & 0.0 & 48.05 & 29.7 & 6.04 \\ 
\rowcolor{Gray}
cone (RADIO) & {\bf49.96} & {\bf66.67} & 50.0 & 66.67 & 100.0 & {\bf69.23} & 66.67 & 66.67 & 83.33 & 35.07 & 62.5 & 13.33 & 0.0 & {\bf62.07} & 37.04 & {\bf45.31} \\  \hline

dog & 26.72 & 45.07 & 55.56 & 47.32 & 30.95 & 25.22 & 33.33 & 36.67 & 16.67 & 13.68 & 50.0 & N/A & 7.5 & 48.0 & 38.33 & 9.44 \\ 
\rowcolor{yellow!25}dog (TL) & 53.09 & 79.48 & 55.56 & 98.57 & 44.64 & 58.57 & 66.67 & 53.33 & 50.0 & 13.49 & 10.0 & N/A & 16.67 & 40.0 & 83.67 & {\bf25.14} \\
\rowcolor{Gray}
dog (RADIO) & {\bf67.26} & {\bf86.43} & 88.89 & 99.05 & 11.11 & {\bf64.74} & 83.33 & 60.42 & 50.0 & {\bf30.91} & 40.0 & N/A & 0.0 & {\bf68.0} & {\bf85.02} & 6.67 \\  \hline

mannequin & 42.05 & 71.28 & 53.33 & 87.02 & 55.56 & 28.3 & 42.42 & 34.18 & 7.5 & {\bf44.47} & 14.29 & 58.82 & 19.25 & 34.72 & {\bf54.16} & {\bf14.94} \\
\rowcolor{yellow!25}mannequin (TL) & 28.27 & 56.52 & 50.0 & 71.43 & 0.0 & 20.48 & 36.36 & 22.58 & 5.0 & 24.14 & 14.29 & 26.47 & 23.53 & 33.33 & 32.91 & 11.9 \\ 
\rowcolor{Gray}
mannequin (RADIO) & {\bf41.15} & {\bf82.61} & 83.33 & 92.86 & 33.33 & {\bf28.56} & 45.45 & 26.67 & 14.55 & 36.28 & 14.29 & 52.94 & 10.92 & {\bf45.83} & 49.94 & 10.62 \\  \hline

sign & 49.36 & 41.3 & 0.0 & 44.71 & 40.0 & 59.77 & N/A & 62.5 & 57.89 & 40.35 & N/A & 41.94 & 38.46 & 0.0 & 50.57 & 49.28 \\ 
\rowcolor{yellow!25}sign (TL) & {\bf84.92} & 83.87 & 100.0 & 92.97 & 60.0 & {\bf90.11} & N/A & 89.81 & 89.47 & {\bf79.63} & N/A & 80.2 & 79.43 & {\bf100.0} & {\bf86.83} & 83.5 \\ 
\rowcolor{Gray}
sign (RADIO) & 77.26 & {\bf87.01} & 100.0 & 90.09 & 66.67 & 74.49 & N/A & 80.19 & 70.27 & 76.92 & N/A & 77.78 & 76.19 & {\bf100.0} & 82.52 &{\bf 72.13} \\  \hline

trolley & 84.61 & 90.23 & 84.72 & 99.92 & 76.47 & 87.35 & 100.0 & 91.66 & 68.81 & 78.51 & 95.99 & 85.78 & 26.67 & 96.29 & 91.69 & 61.94 \\ 
\rowcolor{yellow!25}trolley (TL) & {\bf93.2} & 96.22 & 100.0 & 99.6 & 88.24 & {\bf95.35} & 100.0 & 97.07 & 89.26 & {\bf89.06} & 100.0 & 94.01 & 49.34 & {\bf100.0} & {\bf96.08} & {\bf79.78} \\ 
\rowcolor{Gray}
trolley (RADIO) & 90.08 & 97.08 & 97.62 & 99.44 & 94.12 & 94.16 & 100.0 & 96.42 & 81.36 & 83.82 & 100.0 & 87.79 & 44.37 & 99.01 & 92.91 & 75.11 \\ 
\hline \hline

mAP & 51.8 & 63.47 & 48.94 & 71.49 & 54.66 & 53.48 & 48.17 & 61.38 & 41.02 & 46.29 & 55.69 & 59.25 & 23.01 & 44.07 & 59.46 & 38.54 \\
\rowcolor{yellow!25}mAP (TL) & 61.59 & 67.87 & 67.59 & 75.24 & 48.81 & 62.58 & 59.05 & 58.28 & 64.51 & 52.12 & 46.7 & 60.72 & 33.79 & 66.06 & 67.02 & 47.14 \\ 
\rowcolor{Gray}
mAP (RADIO) & {\bf66.28} & {\bf85.13} & 78.31 & 90.43 & 67.54 & {\bf65.96} & 79.09 & 65.06 & 66.58 & {\bf52.74} & 54.2 & 56.15 & 26.3 & {\bf74.99} & {\bf69.1} & {\bf51.64} \\ 

\end{tabular}}
\\ \\
\caption{CFAR+DBSCAN Detector}\label{tab:dbscan_easy_da}
\resizebox{\textwidth}{!}{%
\begin{tabular}{l||c||c:ccc|c:ccc|c:ccc|c|c|c}
\multicolumn{1}{c|}{\multirow{2}{*}{AP}}& \multirow{2}{*}{Overall} & \multicolumn{4}{c|}{$\# Objects < 4$} & \multicolumn{4}{c|}{$4 \leq \# Objects < 7$} & \multicolumn{4}{c|}{$\# Objects \geq 7$} & \multirow{2}{*}{Short} & \multirow{2}{*}{Mid} & \multirow{2}{*}{Long} \\
& & Overall & Short & Mid & Long & Overall & Short & Mid & Long & Overall & Short & Mid & Long & & & \\ \hline \hline

bike & 54.97 & 81.96 & 50.0 & 71.83 & 100.0 & 57.49 & 50.0 & 59.69 & 33.33 & 29.82 & N/A & 32.47 & N/A & 50.0 & 52.07 & 71.43 \\ 
\rowcolor{yellow!25}bike (TL) & 70.55 & {\bf95.32} & 66.67 & 82.68 & 100.0 & {\bf66.11} & 100.0 & 55.8 & 66.67 & 55.02 & N/A & 40.62 & N/A & {\bf80.0} & 55.43 & {\bf85.71} \\ 
\rowcolor{Gray}
bike (RADIO) & {\bf70.60} & 88.55 & 70.0 & 74.16 & 100.0 & 63.75 & 100.0 & 57.3 & 66.67 & {\bf61.13} & N/A & 58.92 & N/A & 78.85 & {\bf61.66} & 83.67 \\  \hline

cone & 12.95 & {\bf45.45} & 62.5 & 25.0 & 0.0 & 41.72 & 55.56 & 66.67 & 0.0 & 1.07 & 0.0 & 1.39 & 0.0 & {\bf25.17} & 13.06 & 0.0 \\ 
\rowcolor{yellow!25}cone (TL) & 5.83 & 0.0 & 0.0 & 0.0 & 0.0 & 27.45 & 33.33 & 33.33 & 0.0 & 0.0 & 0.0 & 0.0 & 0.0 & 10.34 & 5.56 & 0.0 \\ 
\rowcolor{Gray}
cone (RADIO) & {\bf18.41} & 27.27 & 25.0 & 16.67 & 0.0 & {\bf45.1} & 55.56 & 50.0 & 0.0 & {\bf2.94} & 0.0 & 6.67 & 0.0 & 18.72 & {\bf17.9} & 0.0 \\  \hline

dog & 18.76 & 36.83 & 66.67 & 43.17 & 4.17 & 17.86 & 33.33 & 21.43 & 0.0 & 8.68 & 40.0 & N/A & 1.56 & 48.0 & 27.29 & 1.28 \\ 
\rowcolor{yellow!25}dog (TL) & 44.18 & 72.96 & 53.7 & 95.51 & 8.33 & {\bf51.59} & 66.67 & 53.33 & 0.0 & 8.33 & 10.0 & N/A & 6.25 & 39.64 & {\bf78.36} & {\bf5.56}  \\ 
\rowcolor{Gray}
dog (RADIO) & {\bf58.74} & {\bf77.88} & 77.78 & 98.08 & 5.56 & 50.51 & 66.67 & 44.44 & 0.0 & {\bf33.75} & 48.33 & N/A & 0.0 & {\bf63.53} & 74.81 & 1.54 \\  \hline

mannequin & {\bf41.14} & 72.6 & 64.0 & 85.87 & 44.44 & 21.64 & 29.09 & 28.39 & 9.41 & {\bf48.54} & 0.0 & 61.76 & 34.81 & 27.83 & {\bf52.93} & {\bf18.67} \\ 
\rowcolor{yellow!25}mannequin (TL) & 27.83 & 60.05 & 60.0 & 78.57 & 0.0 & 17.83 & 27.27 & 19.35 & 5.0 & 25.75 & 0.0 & 26.47 & 34.45 & 26.09 & 32.82 & 17.33 \\
\rowcolor{Gray}
mannequin (RADIO) & 39.49 & {\bf87.18} & 100.0 & 92.86 & 44.44 & {\bf26.0} & 36.36 & 23.33 & 19.23 & 33.65 & 0.0 & 50.0 & 13.33 & {\bf39.13} & 47.27 & 15.3 \\  \hline

sign & 46.26 & 43.43 & 0.0 & 44.71 & 50.0 & 50.87 & N/A & 54.11 & 45.6 & 43.32 & N/A & 37.72 & 50.0 & 0.0 & 44.78 & 47.81 \\
\rowcolor{yellow!25}sign (TL) & 80.56 & 85.63 & 100.0 & 94.54 & 50.0 & {\bf86.73} & N/A & 83.93 & 86.53 & {\bf73.07} & N/A & 75.6 & 71.03 & {\bf50.0} & {\bf82.43} & {\bf77.77} \\ 
\rowcolor{Gray}
sign (RADIO) & {\bf72.67} & {\bf88.22} & 100.0 & 91.93 & 66.67 & 71.94 & N/A & 78.05 & 65.25 & 69.51 & N/A & 65.24 & 78.36 & {\bf50.0} & 76.82 & 68.43 \\  \hline

trolley & 80.03 & {\bf83.94} & 68.33 & 94.36 & 73.12 & 84.18 & 97.44 & 88.26 & 64.48 & 74.47 & 79.56 & 75.54 & 34.0 & 82.82 & 83.87 & 59.2 \\ 
\rowcolor{yellow!25}trolley (TL) & 82.35 & 83.35 & 90.48 & 81.45 & 83.0 & 86.83 & 95.6 & 91.56 & 73.46 & {\bf82.14} & 79.56 & 80.06 & 47.14 & 87.88 & 81.29 & {\bf68.91} \\ 
\rowcolor{Gray}
trolley (RADIO) & {\bf82.75} & 81.06 & 84.72 & 88.37 & 64.46 & {\bf88.66} & 95.6 & 92.89 & 68.35 & 82.1 & 87.41 & 80.15 & 45.24 & {\bf88.19} & {\bf84.48} & 58.0 \\ 
\hline \hline

mAP & 42.35 & 60.7 & 51.92 & 60.82 & 45.29 & 45.63 & 53.08 & 53.09 & 25.47 & 34.32 & 29.89 & 41.78 & 24.08 & 38.97 & 45.67 & 33.06 \\ 
\rowcolor{yellow!25}mAP (TL) & 51.88 & 66.22 & 61.81 & 72.13 & 40.22 & 56.09 & 64.57 & 56.22 & 38.61 & 40.72 & 22.39 & 44.55 & 31.77 & 48.99 & 55.98 & {\bf42.55} \\ 
\rowcolor{Gray}
mAP (RADIO) & {\bf57.11} &  {\bf75.02} &   76.25 &  77.01 &   46.85 &  {\bf57.66} &   59.03 &  57.66 &  36.58 &  {\bf47.18} &  22.62 &  43.49 &  22.82 & {\bf 56.40} &  {\bf60.49} &  37.82 \\ 

\end{tabular}}

\end{table*}

As one might expect the results of the detection classification pipeline are worse across the board in all cases, culminating in a headline, overall figure of $57.11\%$ as opposed to $66.28\%$. However, bearing in mind the key hypothesis, the most important results is in comparing the results with radar (in gray) as opposed to standard argumentation techniques. 
Although there is the occasional anomaly, in almost all cases the RADIO technique is shown to be an improvement over standard augmentation techniques.
Again looking at the overall figures, the mAP improves from $51.80\%$ to $66.28\%$ for perfect detection, and from $42.35\%$ to $57.11\%$ in the pipelined case.

\section{Conclusions}

In this chapter, we proposed and investigated a radar data augmentation method (RADIO) and compared it to standard augmentation methods to see if this would lead to higher object classification rates in both isolated object and multi-object scenes. 
We were able to show that the use of radar data augmentation did indeed give improved results, and this offers a useful avenue to train a neural network when the source data is limited, and the network cannot be trained under all the eventual test and validation conditions. In the absence of augmentation,
using only the original data, overfitting is probable and the network cannot generalise. 

In the isolated object case, after data augmentation, an accuracy of $99.79\%$ was achieved. RADIO created realistic samples which helped the neural network to achieve near perfect recognition. In the more challenging multiple object dataset, we achieved $66.28\%$ mAP for perfect detection and $57.11\%$ mAP for the easy detection and classification scenario, in which the possible effects of multi-path, objects in close proximity, antenna coupling, and clutter from unwanted objects are more challenging. Compared to standard camera based data augmentation, the RADIO technique gives improved results.

To develop further, it is necessary to improve detection, as this gives rise to additional errors in comparison with perfect detection, and extend the work to the more challenging scenarios of data collected  "in the wild".
In the latter case, it is probable that both the training and test/validation data will be collected in wild conditions, as the differences between lab-based and outside data would be too great.

This Chapter concluded the use of low-THz radar, the next chapters work with a 79 GHz radar. We developed a novel dataset in the wild with different weather scenarios (sunny, overcast, night, rain, fog and snow) and also developed a vehicle detection method as baseline for further research. 

%% file: Chapters/Chapter7_navtech_data_bmvc.tex
\chapter{A Multi-Modal Object Recognition Dataset for Autonomous Cars in Adverse Weather}\label{ch:navtech}

\lhead{Chapter 6. \emph{A Multi-Modal Object Recognition Dataset for Autonomous Cars in Adverse Weather}} %

In Chapters \ref{ch:300dl} and \ref{ch:radio} we used prototypical sensors in a controlled environment. In this chapter we created a new dataset with commercial sensors in different weather scenarios (sunny, overcast, night, fog, rain and snow).

Datasets for autonomous cars are essential for the development and benchmarking of perception systems. However, most existing datasets are captured with camera and LiDAR sensors in good weather conditions. In this paper, we present the RAdar Dataset In Adverse weaThEr (RADIATE)\footnote{\url{http://pro.hw.ac.uk/radiate/}}, aiming to facilitate research on object detection, tracking and scene understanding using radar sensing for safe autonomous driving. RADIATE includes 3 hours of annotated radar images with more than 200K labelled road actors in total, on average about 4.6 instances per radar image. It covers 8 different categories of actors in a variety of weather conditions (e.g., sun, night, rain, fog and snow) and driving scenarios (e.g., parked, urban, motorway and suburban), representing different levels of challenge. To the best of our knowledge, this is the first public radar dataset which provides high-resolution radar images on public roads with a large amount of road actors labelled. The data collected in adverse weather, e.g., fog and snowfall, is unique. Some baseline results of radar based object detection and recognition are given to show that the use of radar data is promising for automotive applications in bad weather, where vision and LiDAR can fail. RADIATE also has stereo images, 32-channel LiDAR and GPS data, making it applicable for other applications, like fusion, localisation and mapping. The website to download the dataset can be found at \url{http://pro.hw.ac.uk/radiate/}.

\begin{figure*}[t]
\begin{center}
\includegraphics[width=1.0\textwidth]{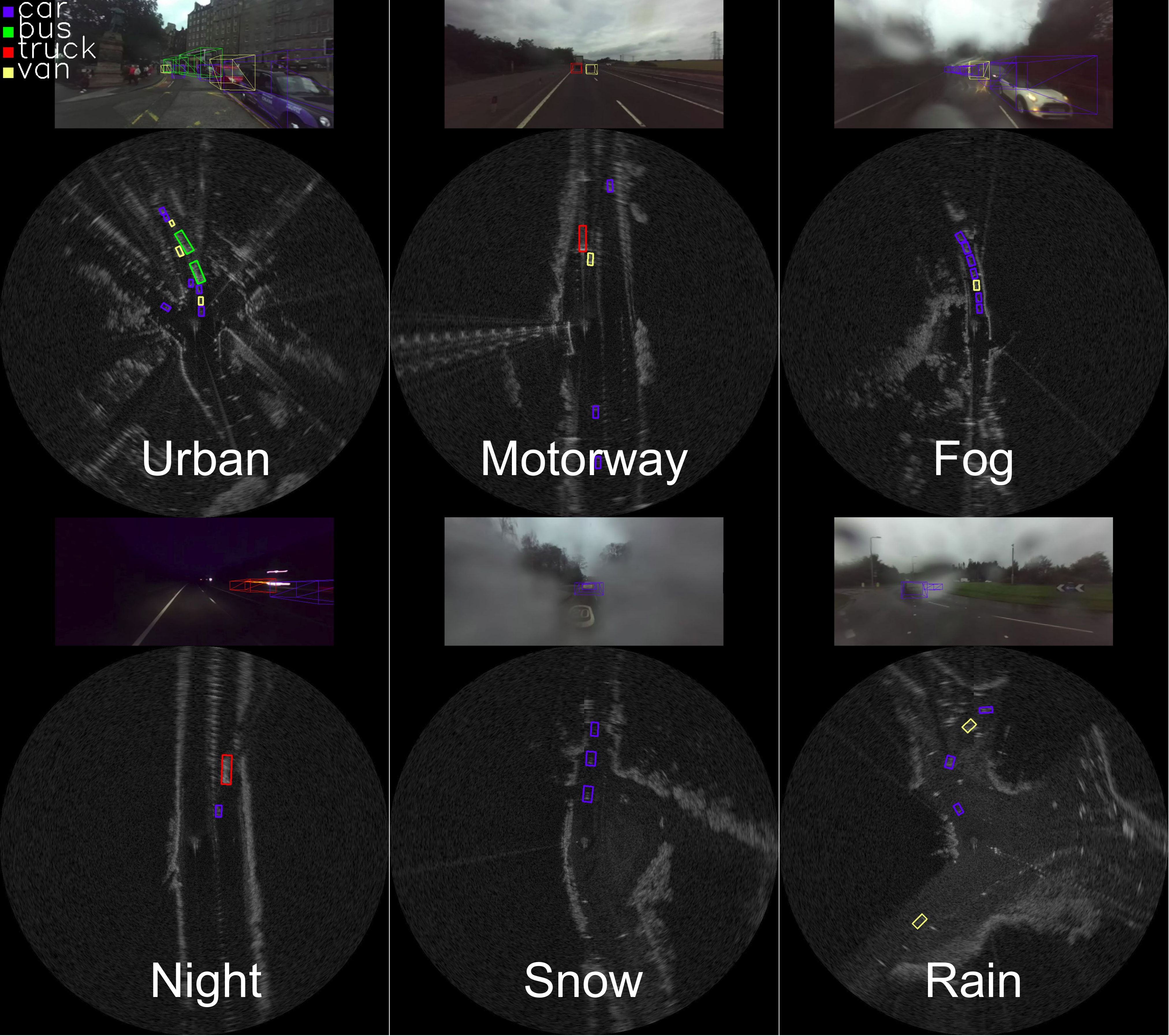}
\end{center}
\caption{Examples from \textbf{RADIATE}. This dataset contains radar, stereo camera, LiDAR and GPS data. It was collected in various weather conditions and driving scenarios with 8 categories of annotated objects.}\label{fig:radiate_data}
\end{figure*}

\section{Introduction}\label{intro}

 Autonomous driving research and development rely heavily on the use of public datasets in the computer vision and robotics communities \cite{Geiger2013IJRR, nuscenes2019, sun2019scalability}. Camera and LiDAR are the two primary perceptual sensors that are usually adopted. However, since these are visible spectrum sensors, the data be affected dramatically by bad weather due to attenuation, multiple scattering and turbulence \cite{pfennigbauer2014online, SDV18, bijelic2018benchmark, kutila2018automotive,WalHalBul2020}, most existing datasets were captured in good weather. On the other hand, a radar sensor is known to be more robust in many adverse conditions \cite{daniel2017low, norouzian2019experimental, norouzian2019rain}. However, there are few public radar datasets for automotive applications, especially with object annotation and in bad weather.

In this paper, we present the RAdar Dataset In Adverse weaThEr (RADIATE) for perception tasks in autonomous driving. It includes a mixture of weather conditions and driving scenarios, representing different levels of challenge. A high-resolution 79 GHz $360^o$ radar is chosen as the main sensor for object annotation. RADIATE includes 5 hours of radar images in total, 3 hours of which are fully annotated. This gives RADIATE more than 200K labelled object instances with 8 categories of road actors (i.e., car, van, bus, truck, motorbike, bicycle, pedestrian and a group of pedestrians). RADIATE also has stereo camera, LiDAR and GPS data collected simultaneously. Some examples from RADIATE are shown in Figure \ref{fig:radiate_data}.
\vspace{2em}

This chapter presents the following contributions:
\begin{itemize}
    \setlength\itemsep{-0.4em}
    \item To the best of our knowledge, RADIATE is the first public radar dataset which includes a large number of labelled road actors on public roads.
    \item It includes multi-modal sensor data collected in challenging weather conditions, such as dense fog and heavy snowfall. Camera, LiDAR and GPS data are also provided for all sequences.
\end{itemize}

\section{Comparison with other relevant datasets}

There are many publicly available datasets for research into perception for autonomous and assisted driving. The most popular is the KITTI dataset \cite{Geiger2013IJRR}, using cameras and a Velodyne HDL-64e LiDAR to provide data for several tasks, such as object detection and tracking, odometry and semantic segmentation. Although widely used as a benchmark, data was captured only in good weather and appears now as rather small scale.
Waymo \cite{sun2019scalability} and Argo \cite{Chang_2019_CVPR} are automotive datasets which are larger than KITTI and provide more data variability. Some data was collected in the rain and at night although adverse weather is not their main research focus. Foggy Cityscapes \cite{SDV18} developed a synthetic foggy dataset aimed at object recognition tasks but only images are provided.
All these datasets use only optical sensors.

\begin{table}[t]
    \begin{center}
    \resizebox{\columnwidth}{!}{
    \begin{tabular}{l|c|ccc|cccc|ccc|c}
    \hline
    Dataset          & Size  & Radar                      & LiDAR & Camera & Night &Fog & Rain & Snow & \makecell{Object\\ Detection} & \makecell{Object\\ Tracking}  & Odometry                               & \makecell{3D\\ Annotations} \\
    \hline \hline
    nuScenes \cite{nuscenes2019}        & Large & \makecell{\cmark \\(Sparse Point Cloud)}   & \cmark   & \cmark    & \cmark  & \xmark& \cmark  & \xmark   & \cmark              & \cmark                             & \xmark                                     & \cmark            \\ \hline
    Oxford Radar RobotCar \cite{RadarRobotCarDatasetICRA2020}   & Large & \makecell{\cmark \\ (High-Res Radar Image)} & \cmark   & \cmark    &\cmark &\xmark  & \cmark   & \xmark   & \xmark               & \xmark                                & \cmark                                    & \xmark            \\ \hline
    MulRan \cite{gskim-2020-mulran}           & Large & \makecell{ \cmark  \\(High-Res Radar Image)}       & \cmark    & \xmark & \xmark&\xmark & \xmark   & \xmark   & \xmark               & \xmark                            & \cmark                                    & \xmark          \\ \hline
    Astyx \cite{meyer2019automotive}            & Small & \makecell{\cmark\\ (Sparse Radar Point Cloud)} & \cmark   & \cmark & \xmark   & \xmark  & \xmark   & \xmark   & \cmark              & \xmark                                & \xmark                                     & \cmark           \\ \hline
    RADIATE (Ours)          & Large &\makecell{ \cmark\\ (High-Res Radar Image)} & \cmark   & \cmark    & \cmark & \cmark & \cmark  & \cmark  & \cmark              & \cmark                             & \cmark
    & \makecell{\xmark\\ (Pseudo-3D)} \\
    \hline
    \end{tabular}}
    \end{center}
    \caption{Comparison of RADIATE with public automotive datasets that use radar sensing.}\label{tab:comparison}
    \end{table}

Radar, on the other hand, provides a sensing solution that is more resilient to fog, rain and snow. It usually provides low-resolution images, which makes it very challenging for object recognition or semantic segmentation. Current automotive radar technology relies on the Multiple Input Multiple Output (MIMO) technique, which uses several transmitters and receivers to measure the direction of arrival (DOA) \cite{tiradar}. Although this is inexpensive, it lacks azimuth resolution, e.g. the cross-range image of a commercial radar with $15^o$ angular resolution is around 10 meters at 20-meter distance. This means that a radar image provide insufficient detail for object recognition. Scanning radar measures at each azimuth using a moving antenna, providing better azimuth resolution. Thus, this type of sensor has recently been developed to tackle radar based perception tasks for automotive applications \cite{navtech_radar, daniel2017low}.

For most datasets which provide radar data for automotive applications, radar is used only as a simple detector, e.g. NuScenes \cite{nuscenes2019} that gives sparse 2D point clouds. Recently, the Oxford Robotcar Radar dataset \cite{RadarRobotCarDatasetICRA2020} and MulRan dataset \cite{gskim-2020-mulran} provided data collected from a scanning Navtech radar in various weather conditions. However, they do not provide object annotations as the data was designed primarily for Simultaneous Localisation and Mapping (SLAM) and place recognition in long-term autonomy. The Astyx dataset \cite{meyer2019automotive} provides denser data (compared to current MIMO technology) but with only about $500$ frames annotated. It is also limited in terms of weather variability. Table \ref{tab:comparison} compares existing public automotive radar datasets with RADIATE.
To summarise, although research into radar perception for autonomous driving has recently been gaining popularity \cite{Major_2019_ICCV, Sless_2019_ICCV,wang2020rodnet}, there is no radar dataset with a large set of actor annotations publicly available. We hope the introduction of RADIATE can boost autonomous driving research in the community.

\begin{figure}[h]
    \begin{center}
    \includegraphics[width=1.0\textwidth]{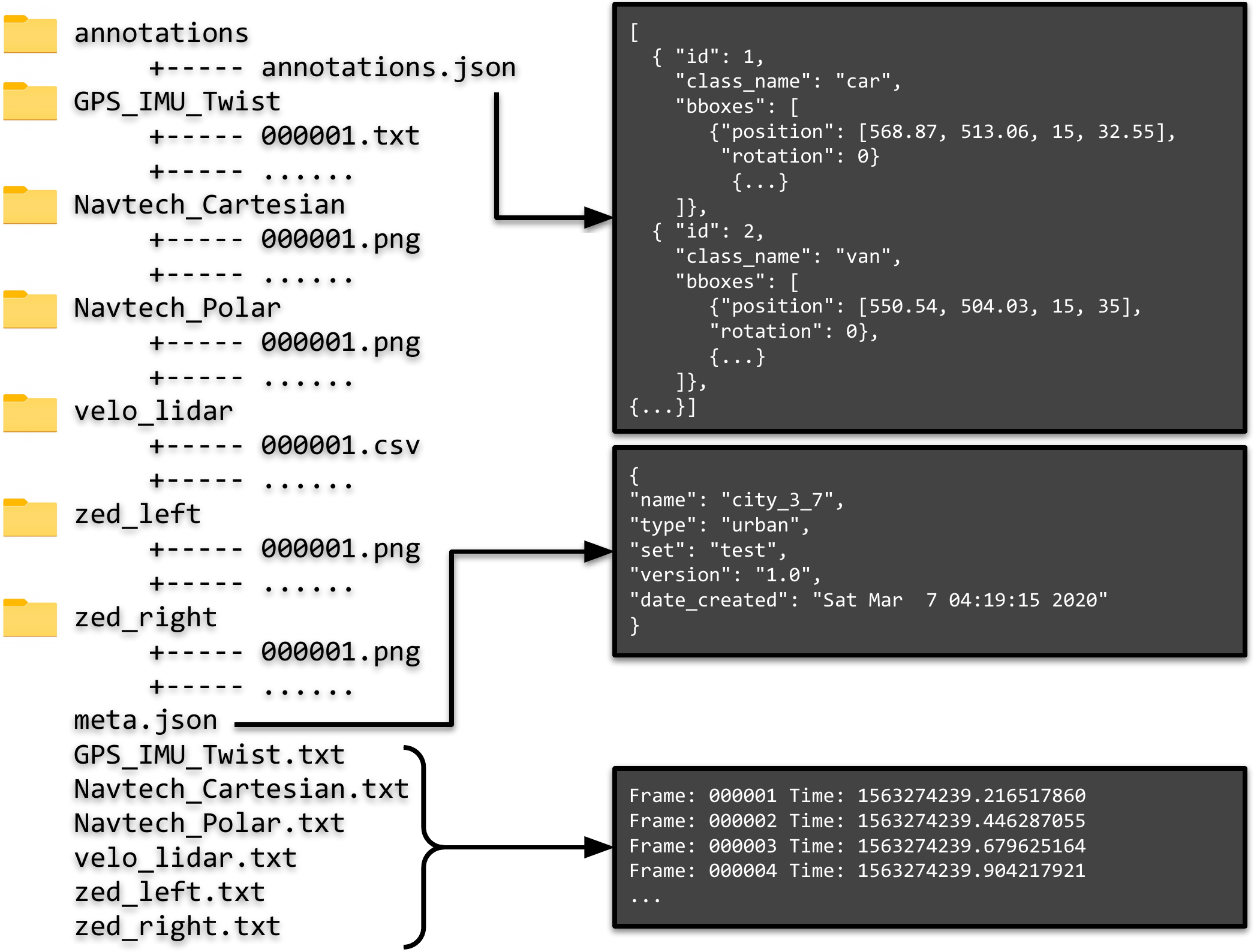}
    \end{center}
    \caption{Folder tree, annotation, metadata and timestamp structure of each sequence}
    \label{fig:folder_tree}
\end{figure}

\section{The RADIATE Dataset}

\begin{figure*}[t]
\begin{center}
        \includegraphics[width=1.0\linewidth]{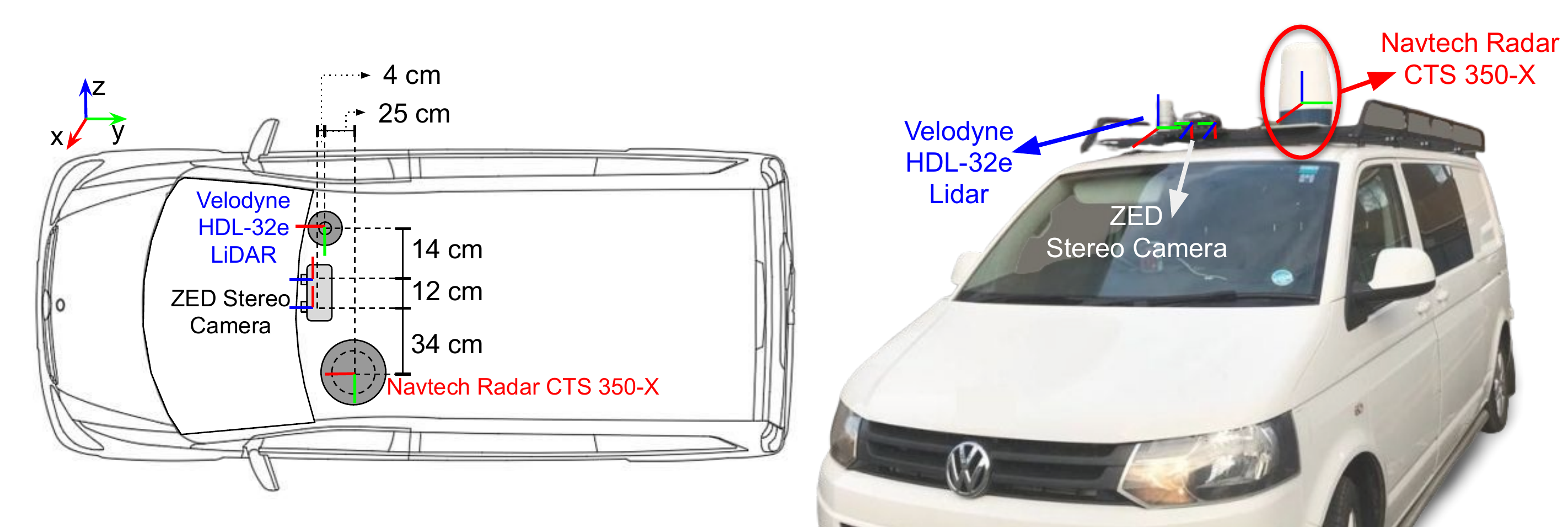}
    \end{center}

    \caption{Sensor setup for data collection.}\label{fig:sensors_setup}
\end{figure*}

The RADIATE dataset was collected between February 2019 and February 2020. The data collection system was created using Robot Operating System (ROS) \cite{ros}. From the rosbag created by ROS, we extracted sensor information, each with its respective timestamp. Figure \ref{fig:folder_tree} shows the folder structure used for the dataset.
To facilitate access, a RADIATE software development kit (SDK) is released for data calibration, visualisation, and pre-processing.

\subsection{Perception Sensors}

RADIATE includes radar, LiDAR and stereo cameras. Figure \ref{fig:sensors_setup} shows the sensor setup and the extrinsic configuration on our vehicle. To support sensor fusion, the sensors are calibrated (details in Section \ref{sec:sensor_calibration}).

\begin{figure}[ht]
    \centering
    \begin{subfigure}[t]{0.48\textwidth}
        \centering
        \includegraphics[height=1.2in]{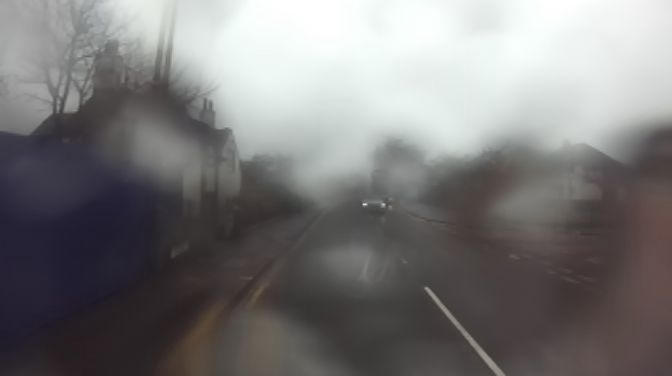}
    \end{subfigure}%
    ~ 
    \begin{subfigure}[t]{0.48\textwidth}
        \centering
        \includegraphics[height=1.2in]{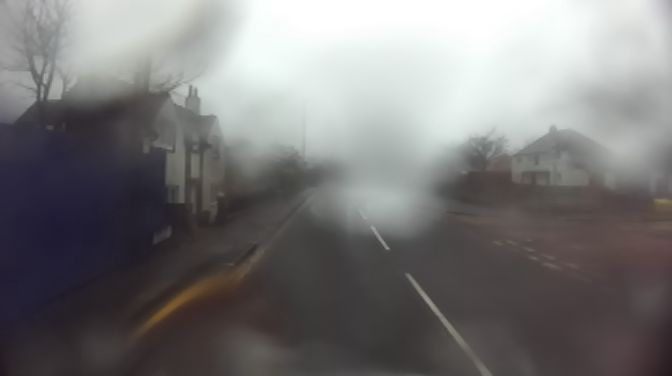}
    \end{subfigure}
    \caption{Example of left and right images using ZED stereo camera.}\label{fig:stereo}
\end{figure}

\paragraph{Stereo Camera} An off-the-shelf ZED stereo camera is used. It is set at $672 \times 376$ image resolution at 15 frames per second for each camera. It is protected by a waterproof housing for extreme weather. The images can be seriously blurred, hazy or fully blocked due to rain drops, dense fog or heavy snow, respectively. Some examples are shown in Figure \ref{fig:stereo}.

\begin{figure}[h]
    \begin{center}
    \includegraphics[width=0.5\textwidth]{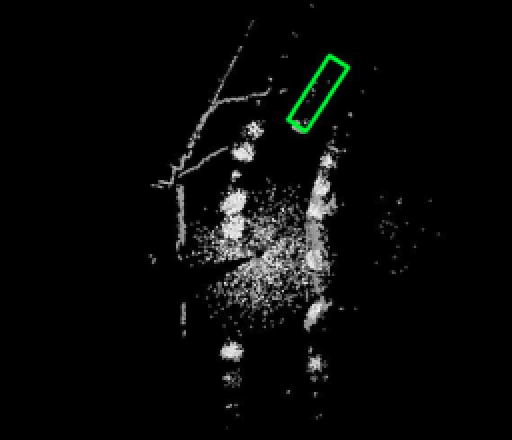}
    \end{center}
    \caption{Example of LiDAR point cloud in snow (the green rectangle is a bus).}
    \label{fig:lidar_snow}
\end{figure}

\paragraph{LiDAR} A 32 channel, 10Hz, Velodyne HDL-32e LiDAR \cite{velodyne} is used to give $360^o$ coverage. Since the LiDAR signal can be severely attenuated and reflected by intervening fog or snow \cite{WalHalBul2020}, the data can be missing, noisy and incorrect. An example of LiDAR in snow can be seen in Figure \ref{fig:lidar_snow}.

\begin{figure}[ht]
    \begin{center}
    \includegraphics[width=0.8\textwidth]{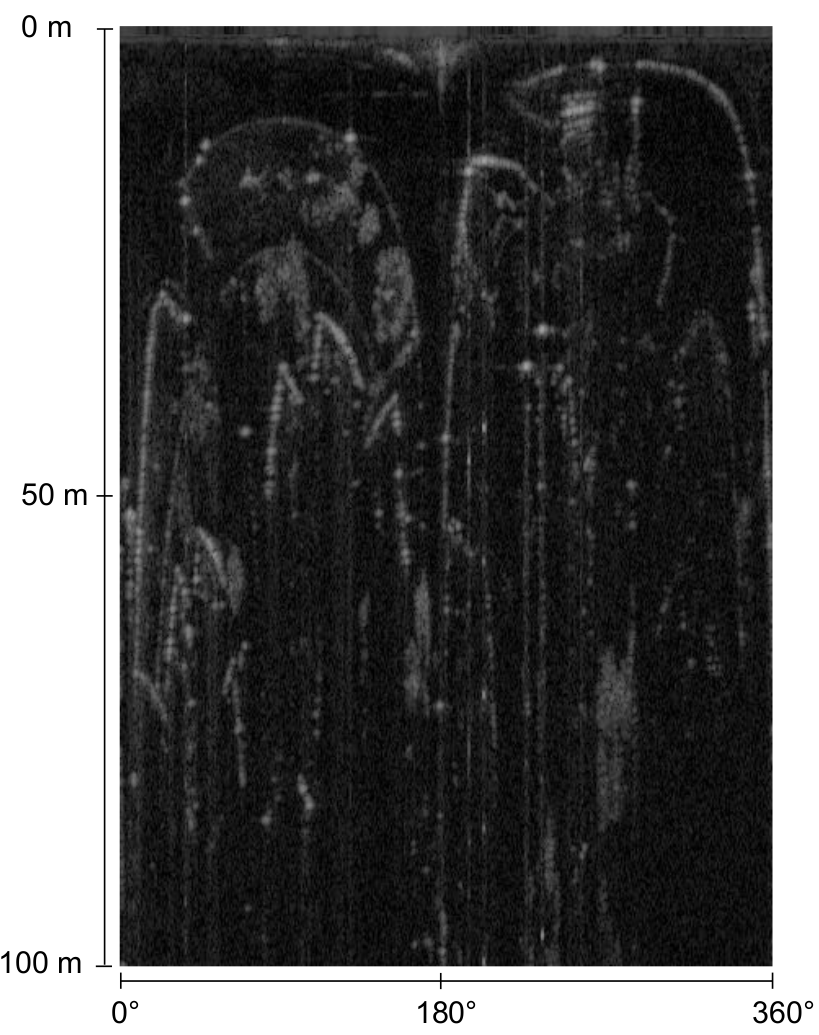}
    \end{center}
    \caption{Example of radar image in polar coordinates.}
    \label{fig:radar_polar}
\end{figure}

\begin{figure}[ht]
    \begin{center}
    \includegraphics[width=1.0\textwidth]{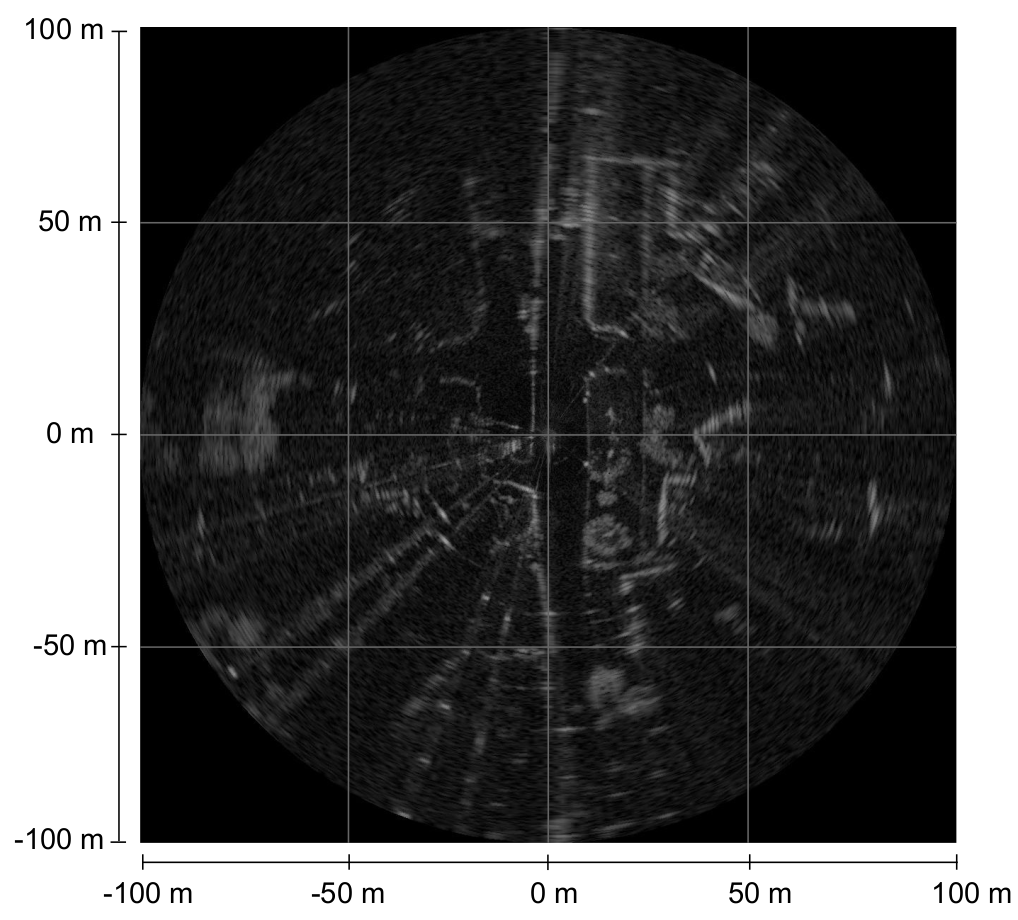}
    \end{center}
    \caption{Example of radar image in cartesian coordinates. This is a typical image captured in a road scene, where the ego car is in the centre.}
    \label{fig:radar_cartesian}
\end{figure}

\paragraph{Radar} RADIATE adopts the Navtech CTS350-X \cite{navtech_radar} radar. It is a scanning radar which provides $360^o$ high-resolution range-azimuth images. It has 100 meters maximum operating range with 0.175m range resolution, $1.8^o$ azimuth resolution and $1.8^o$ elevation resolution, Currently, it does not provide Doppler information. Examples of radar images in polar and cartesian coordinates can be seen in Figures \ref{fig:radar_polar} and \ref{fig:radar_cartesian} respectively.

\subsection{Sensor Calibration} \label{sec:sensor_calibration}

Sensor calibration is required for multi-sensor fusion, feature and actor correspondence. The intrinsic parameters and distortion coefficients of the stereo camera are calibrated using the Matlab camera calibration toolbox \cite{matlabtool}. Then, rectified images can be generated to calculate depths. In terms of extrinsic calibration, the radar sensor is chosen as the origin of the local coordinate frame as it is the main sensor. The extrinsic parameters for the radar, camera and LiDAR are represented as 6 degree-of-freedom transformations (translation and rotation). They are performed by first explicitly measuring the distance between the sensors, and then fine-tuned by aligning measurements between each pair of sensors. The sensor calibration parameters are provided in a {\tt yaml} file. The sensors operate at different frame rates and  we simply adopt each sensor data's time of arrival as the timestamp.

\subsection{Data Collection Scenarios}

We collected data in 7 different scenarios, i.e., sunny (parked), sunny/overcast (urban), overcast (motorway), night (motorway), rain (suburban), fog (suburban) and snow (suburban).

\paragraph{Sunny (Parked)}
In this scenario the vehicle was parked at the side of the road, sensing the surrounding actors passing by. This is intended as the easiest scenario for object detection, target tracking and trajectory prediction. This was collected in sunny weather.

\paragraph{Sunny/Overcast (Urban)}
The urban scenario was collected in the city centre with busy traffic and dense buildings. This is challenging since many road actors appear. The radar collected in this scenario is also cluttered by numerous reflections of non-road actors, such as trees, fences, bins and buildings, and multi-path effects, increasing the challenge. Those sequences were collected in sunny and overcast weather.

\paragraph{Overcast (Motorway)}
The motorway scenario was captured on the city bypass. This can be considered as  relatively easy since most of the surrounding dynamic actors are vehicles and the background is mostly very similar. This scenario was collected in overcast weather.

\paragraph{Night (Motorway)}
We collected data during night in a motorway scenario. Night is an adverse scenario for passive optical sensors due to lack of illumination. LiDAR and radar were expected to behave well since they are active sensors and do not depend on an external light source to work.

\paragraph{Rain (Suburban)}
We collected 18 minutes of data in substantial rain (7 mm). The collection took place in a suburban scenario close to the university campus.

\paragraph{Fog (Suburban)}
We found foggy data challenging to collect. Fog does not happen very often in most places and it is very hard to predict when it will occur and how dense it will be. In practice, we collected foggy data opportunistically when parked and driving in suburban areas.

\begin{figure}[h]
        \begin{center}
        \includegraphics[width=0.5\textwidth]{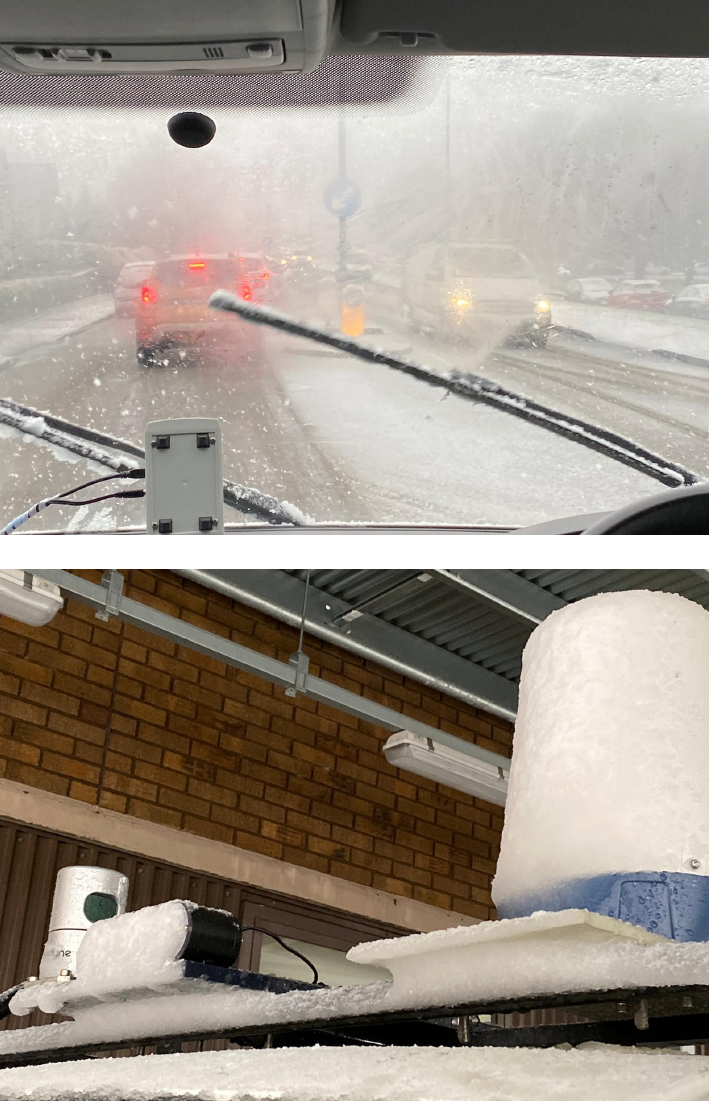} %
    \end{center}
    \begin{center}
    \caption{Sensors covered in snow}\label{fig:sensors_snow}
    \end{center}
\end{figure}

\paragraph{Snow (Suburban)}
RADIATE includes 34 minutes of data in snow, in which 3 minutes are labelled. Snowflakes are expected to interfere with LiDAR and camera data, and can also affect radar images to a lesser extent. Moreover, heavy snowfall can block the sensors within 3 minutes of data collection (see Figure \ref{fig:sensors_snow}). The data in snow was collected in a suburban scenario.

\begin{figure*}[h]
    \begin{center}
        \includegraphics[width=\textwidth]{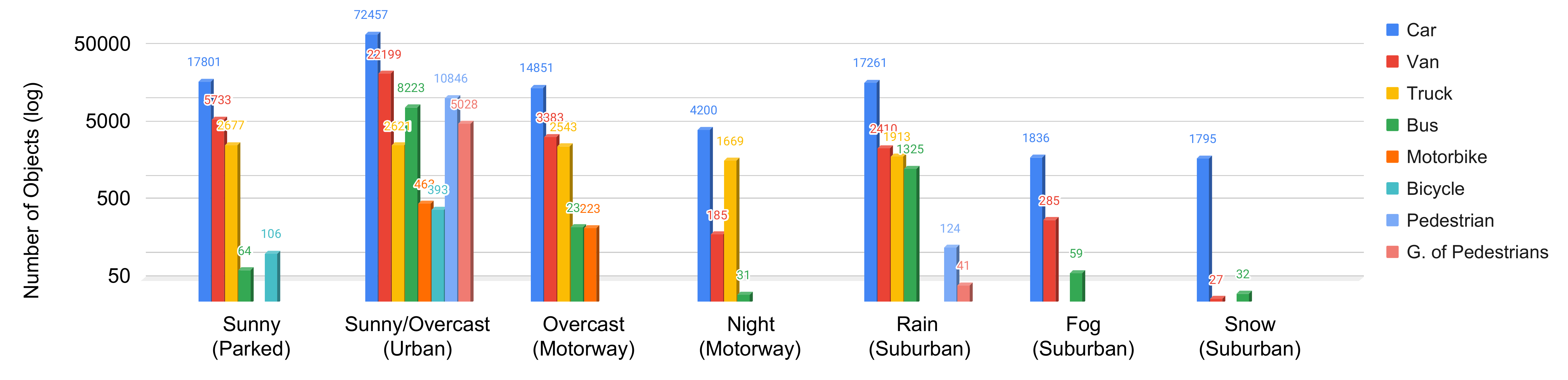}
    \end{center}
    \caption{Category distribution for each scenario.}\label{fig:class_distribution}
\end{figure*}

\begin{figure}[h]
        \begin{center}
        \includegraphics[width=1.0\textwidth]{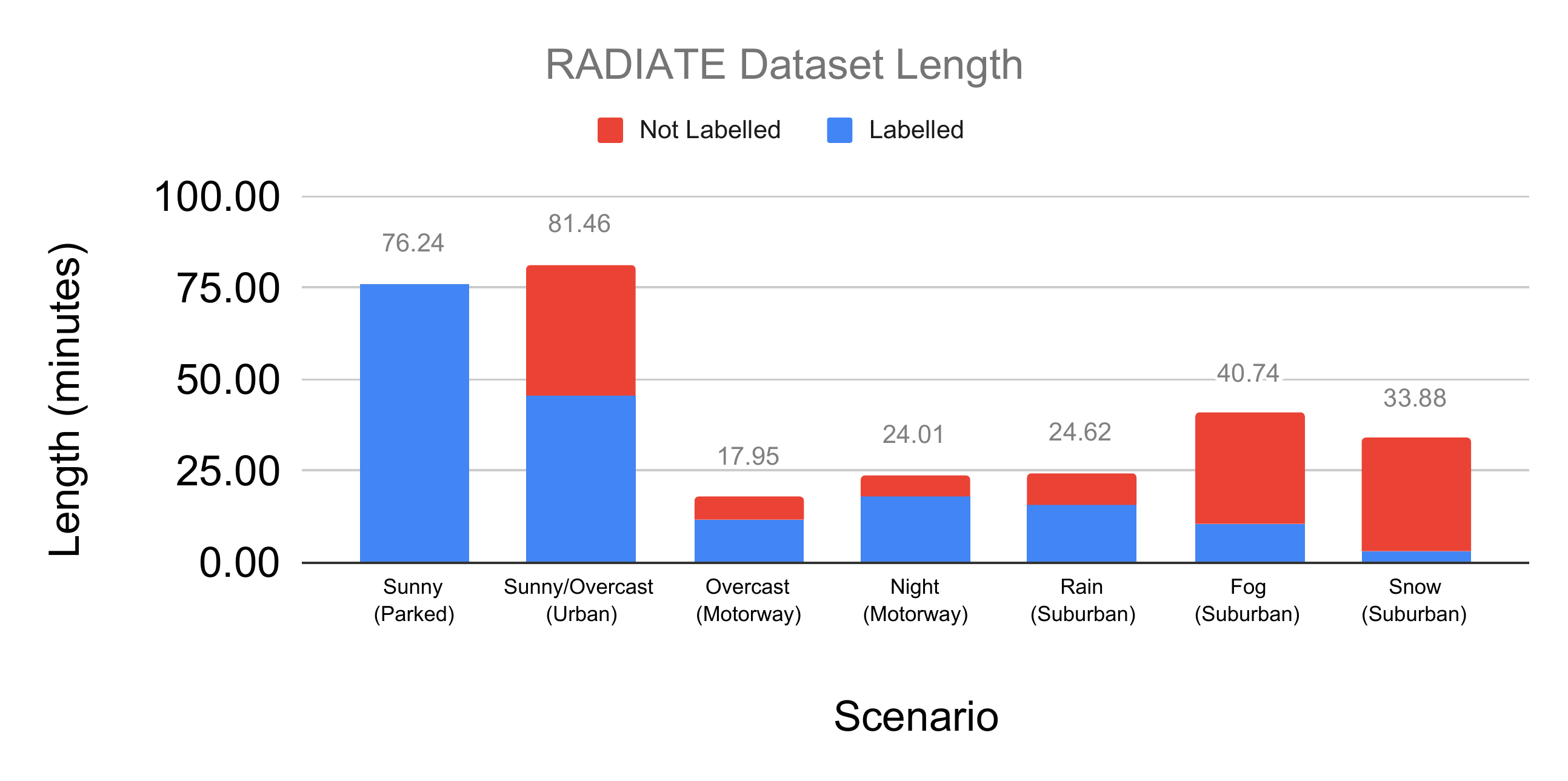} %
    \end{center}
    \caption{Dataset length for driving scenarios and weather conditions (in minutes).}\label{fig:dataset_length}
\end{figure}

\subsection{Labelling}\label{sec:labelling}

Labelling radar images is challenging because it is not easy for a human to recognise objects in radar data. This means most existing annotation tools which are designed for a single sensor \cite{cvat, vatic, labelimg} are inadequate. Therefore, a new annotation tool was developed to automatically correlate and visualise multiple sensors through sensor calibration.

Eight different road actors are labelled in RADIATE: \textit{cars, vans, trucks, buses, motorbikes, bicycle, pedestrian and group of pedestrians}. 2D bounding boxes were annotated on radar images after verifying the corresponding camera images. In total, RADIATE has more than $200$K bounding boxes over $44$K radar images, an average 4.6 bounding boxes per image. The class distribution for each driving scenario is illustrated in Figure \ref{fig:class_distribution}. In Figure \ref{fig:dataset_length}, the length and the total number of objects in each scenario are given. 

\begin{figure}[ht]
        \begin{center}
        \includegraphics[width=1\textwidth]{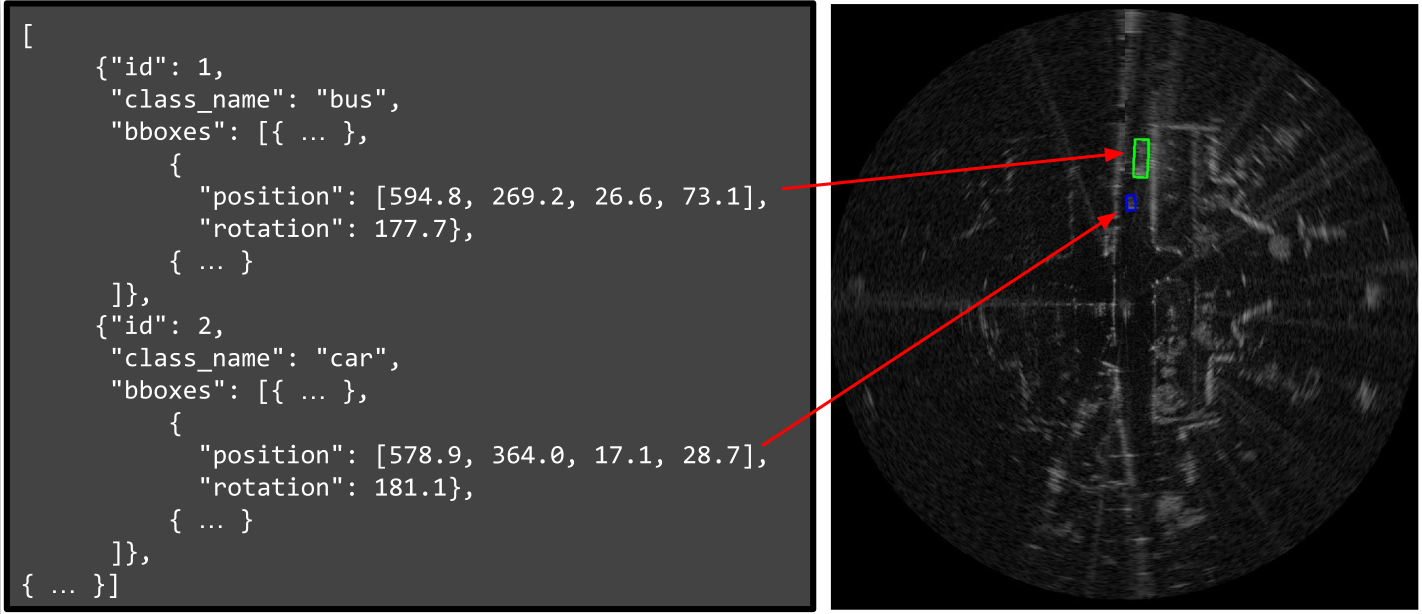} %
    \end{center}
    \begin{center}
    \caption{Example of annotation used in RADIATE}\label{fig:annotation_ex}
    \end{center}
\end{figure}

Each bounding box is represented as (x,y,width,height,angle), where (x,y) is the upper-left pixel locations of the bounding box, of given width and height and counter-clockwise rotation angle. An example of annotation can be visualised in Figure \ref{fig:annotation_ex}. To achieve efficient annotation, the CamShift tracker was used \cite{allen2004object}. %

\section{RADIATE SDK}

\begin{figure}[t!]
    \centering
        \centering
        \includegraphics[width=\textwidth]{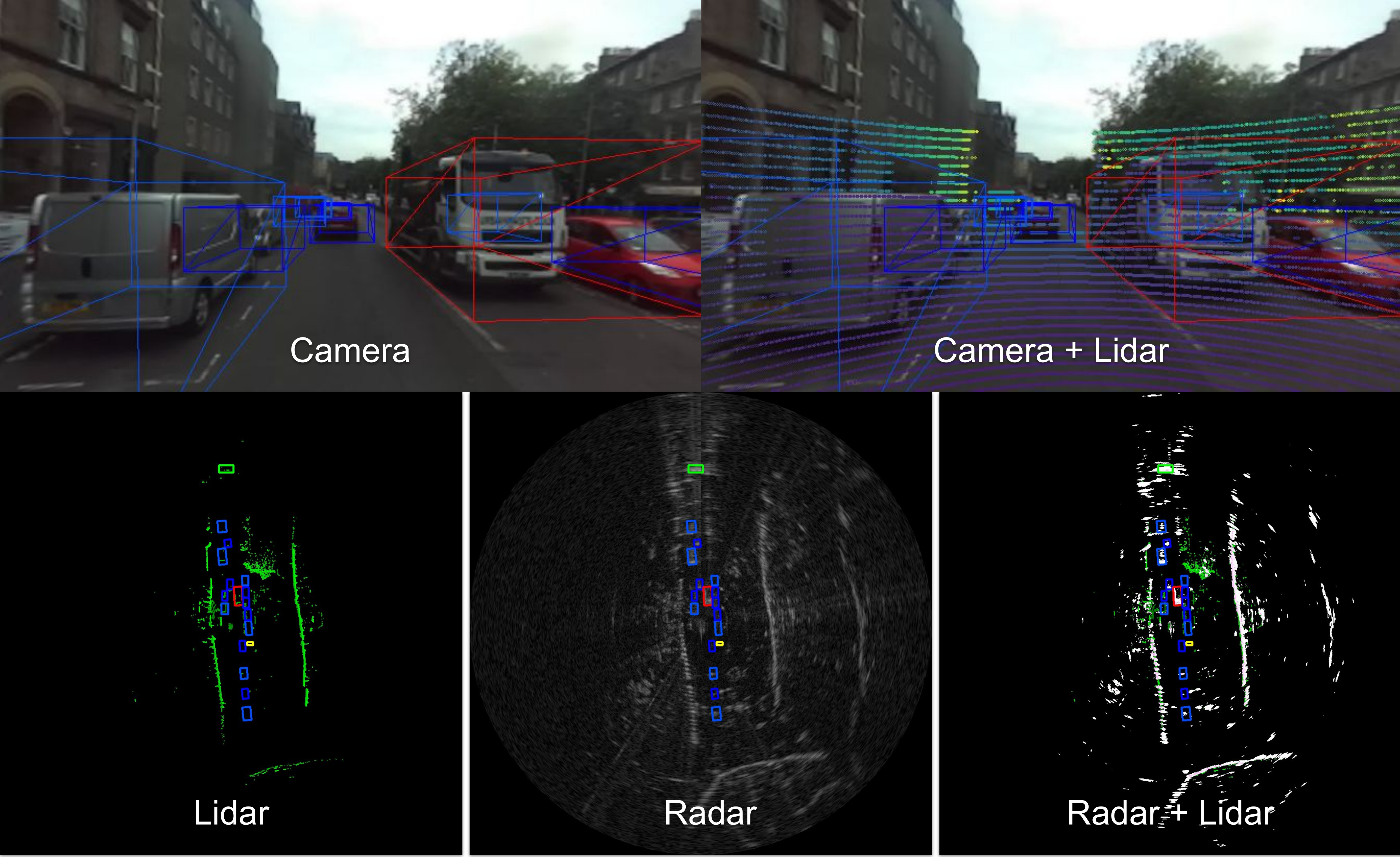}
        \caption{Information which can be retrieved by using RADIATE SDK.}\label{fig:radiate_sdk}
\end{figure}

In order to access the data for research, we provide a software development kit (SDK)\footnote{\url{https://github.com/marcelsheeny/radiate_sdk}} to help researchers retrieve specific information about the data.

We provide a \texttt{Python} code which researchers can use to access the data. In Figure \ref{fig:radiate_sdk} we can visualise an example of data which can be retrieved by using our RADIATE SDK.

This dataset was designed to focus on radar research, and that is why we use the radar image as a coordinate frame for the annotation. Since we know the size of the pixel of the radar image ($0.17$m), we can transform those values to the other sensors from the calibration parameters.

\begin{figure}[H]
    \centering
    \includegraphics[width=\textwidth]{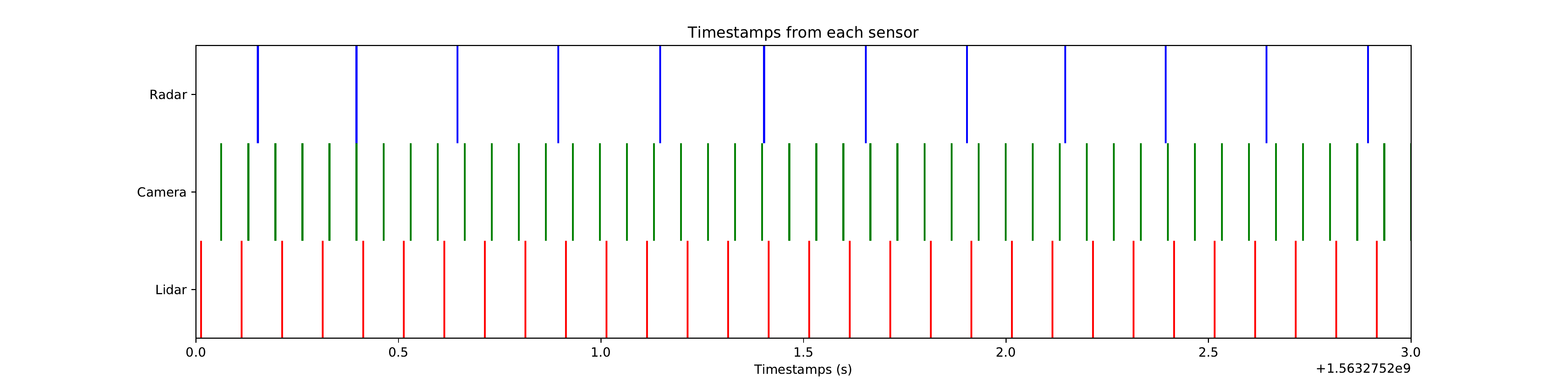}
    \caption{Timestamp given for each sensor}
    \label{fig:timestamps}
\end{figure}

The only issue in achieving accurate annotation in the other sensors (camera and LiDAR) is their different capture frequencies (Figure \ref{fig:timestamps}). At low speed it does not affect the final result a lot, but when vehicles go at high speed this tends to create inaccuracies. We can minimise this problem by interpolating the annotation by using the timestamp information, which can be used for multi-sensor research. In the SDK we provide a linear interpolation when we request the bounding box position in any of the sensors.

\subsection{Using RADIATE SDK}

We designed SDK for easy access to each sensor and its annotation. We wanted the user to get the desired information from the dataset with minimum coding needed. The main class on the SDK is the \texttt{Sequence} class. It handles all the sensor synchronisation, calibration, image rectification. It only requires the root path from the desired sequence as parameter to be initialised. It is possible to access the information by the actual frame number of each sensor, or by the timestamp. When accessing using the timestamp, it gets the closest timestamp from each sensor.

\definecolor{codegreen}{rgb}{0,0.6,0}
\definecolor{codegray}{rgb}{0.5,0.5,0.5}
\definecolor{codepurple}{rgb}{0.58,0,0.82}
\definecolor{backcolour}{rgb}{0.95,0.95,0.92}

\lstdefinestyle{mystyle}{
    backgroundcolor=\color{backcolour},   
    commentstyle=\color{codegreen},
    keywordstyle=\color{magenta},
    numberstyle=\tiny\color{codegray},
    stringstyle=\color{codepurple},
    basicstyle=\ttfamily\footnotesize,
    breakatwhitespace=false,         
    breaklines=true,                 
    captionpos=b,                    
    keepspaces=true,                 
    numbers=left,                    
    numbersep=5pt,                  
    showspaces=false,                
    showstringspaces=false,
    showtabs=false,                  
    tabsize=2
}

\lstset{style=mystyle}
\begin{minipage}{\linewidth}
\begin{lstlisting}[language=Python, floatplacement=H]
import numpy as np
import radiate

# load sequence
seq = radiate.Sequence('/radiate_dataset/urban7')
        
# play sequence
for t in np.arange(seq.init_timestamp, seq.end_timestamp, 0.1):
    output = seq.get_from_timestamp(t)
    seq.vis_all(output)
    
\end{lstlisting}
\end{minipage}

The {\tt output} is a dictionary type variable whose keys are {\tt sensors} and {\tt annotations}. {\tt Sensors} and {\tt annotations} are also dictionaries with the same keys, which are the names of the sensors. With the {\tt output} variable, we can access the sensor data with its respective annotation. The only information we need to give is the frame/timestamp to retrieve the desired sensor/annotation information. 

The explanation for each key is below:

The file {\tt config/config.yaml} controls which sensors to use and configures their parameters.

The file {\tt radiate\_sdk/docs/html/radiate.html} contains a HTML documentation of each method developed for the SDK.

\begin{itemize}
    \item {\tt camera\_(left\textbackslash right)\_raw}: This is the raw (left\textbackslash right) image captured from the ZED camera with the resolution 672 x 376. We do not provide the annotation, since the calibration is based on the rectified version. We provide these images to the user in case they want to apply their own rectification/calibration method.
    
    \item {\tt camera\_(left\textbackslash right)\_rect}: This is the rectified (left\textbackslash right) image from the calibration parameters. Since we calibrated the other sensors related to the rectified version, we provide an approximated 2D annotation. We used the distance to the ground and average height of the object to estimate the 2D bounding box. We suppose the measurement is always done in flat roads. We cannot guarantee that the bounding box projection will always occur accurately. Moreover, since the resolution of radar is low (17 cm), the annotation in the camera images may not be very precise. Also the sensors have different capture frequencies, which can affect the accuracy of the annotation in the camera image.
    \item {\tt radar\_polar}: It accesses the radar image in its raw polar format with resolution 400 x 576 (azimuth x range). The index 0 from the azimuth axis represents the angle $0^o$ and 399 represents the angle $360^o$. Regarding the range axis, index 0 represents 0 meters and index 575 represents 100 meters. This raw format is provided by the sensor manufacturer after applying Fast Fourier Transform (FFT). The manufacturer converts the raw information to decibel (dB), then it is quantised to values between 0 to 255. Therefore, we do not have the raw information in Decibel or Watts. The pixel value represents the power received by the sensor. This value comes mainly from the object material and the shape.     
    \item {\tt radar\_cartesian}: It gives the radar image in cartesian coordinates. We converted the polar image to a cartesian image by projecting each point onto a (x,y) plane. After projecting each point we used bilinear interpolation to fill the holes without values. This gives an image with 1152 x 1152 image resolution.
    \item {\tt lidar\_bev\_image}: It gives an image with the same size as {\tt radar\_cartesian} with a bird's eye view representation. This type of image is created for researchers who want to use the LiDAR in a grid format and also use it together with the radar in a grid format. 
    \item {\tt proj\_lidar\_(left\textbackslash right)}: This gives the projected LiDAR points in a camera coordinate frame. It can be used to improve the stereo reconstruction and also fuse the information from the camera with LiDAR.
\end{itemize}

\section{Conclusions}

RADIATE is the first publicly available object detection and tracking dataset that addresses a wide variety of weather scenarios. With the introduction of radar to the multi-modal setup, we created a dataset which has the potential of recognising objects in all-weather scenarios.

RADIATE annotated around 200k objects, making it a large dataset suitable for deep learning research. With the default training/testing set suggested, domain adaptation between weather scenarios is a potential research aspect.

The RADIATE SDK created a simple {\tt Python} code which can easily be used to retrieve information from the dataset. With the calibration files, research on sensor fusion is also possible.
By releasing the dataset, we encourage researchers to try their methodologies on RADIATE and we hope that this can help the development of new methods which are robust to all-weather scenarios applied to self-driving cars.

In the next chapter, we will detect vehicles using only radar based deep neural network methods on RADIATE.

%% file: Chapters/Chapter8_navtech_results_bmvc.tex
\chapter{Using Radar to Recognise Vehicles in Adverse Weather}\label{ch:navtech2}

\lhead{Chapter 7. \emph Using Radar to Recognise Vehicles in Adverse Weather} %

In the previous chapter, we introduced the RADIATE dataset. In this chapter we use  deep neural networks on radar, and show how radar based vehicle recognition works in diverse weather conditions in the wild. In our evaluation, we trained object detection deep neural networks using Faster R-CNN in good weather only and tested in scenarios which include adverse weather obtaining 45.8 \% mAP. These results show the potential of a radar-only perception system for automotive applications in all-weather scenarios even with the lack of bad weather radar data. We expect these results to be used as baseline, so further development with the findings in this thesis is encouraged.

\begin{figure*}[t!]
        \begin{center}
        \includegraphics[width=\textwidth]{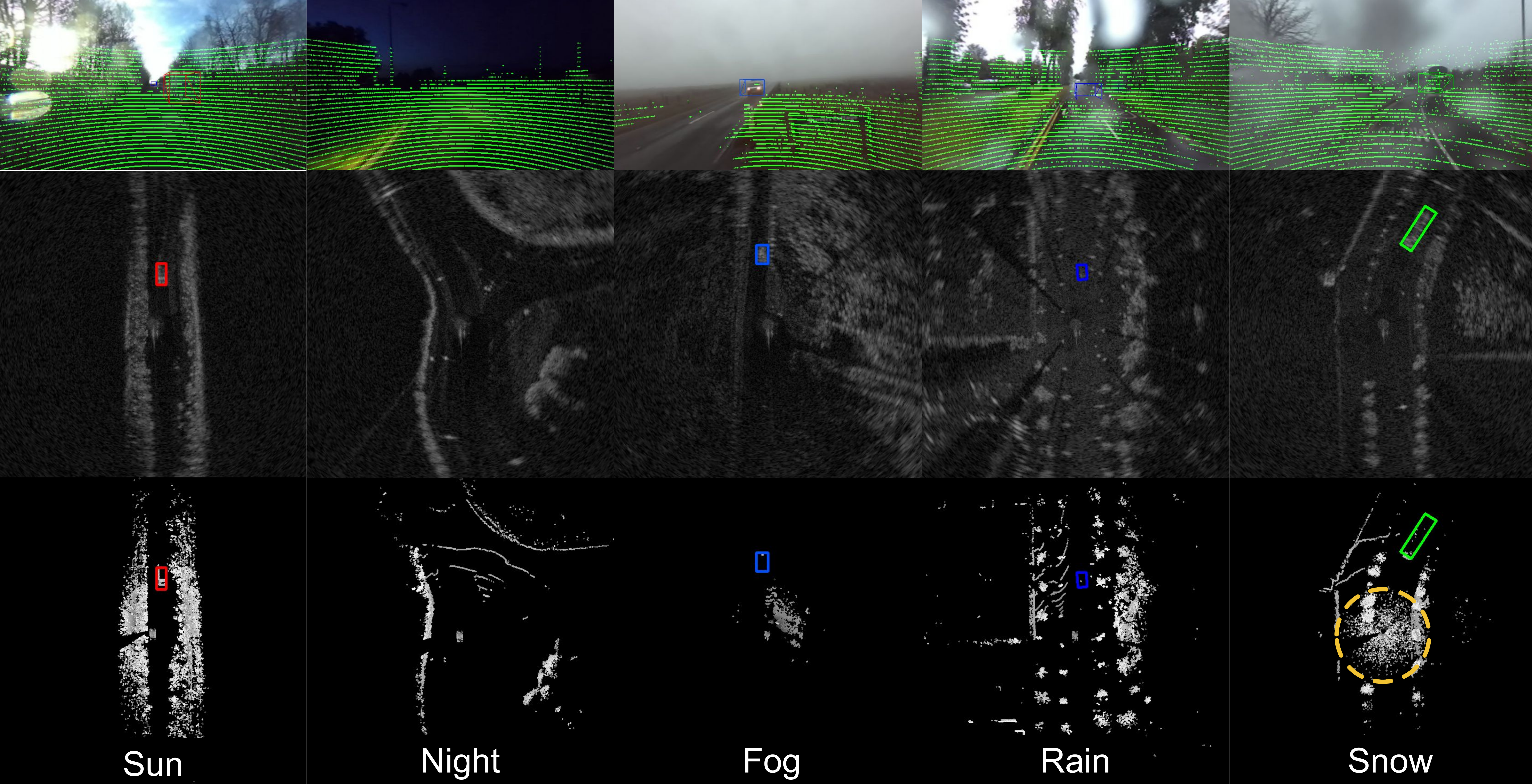}
        \end{center}
        \caption{Data in various weather conditions. \textbf{Top:} Image with LiDAR points projected. \textbf{Middle:} Radar with objects annotated. \textbf{Bottom}: LiDAR with objects projected from radar annotation. Note both image and LiDAR images are degraded in fog, rain and snow. The yellow circles encloses false LiDAR points caused by snow flakes.}\label{fig:lidar_fog}
\end{figure*}

\section{Radar based Vehicle Detection}

One of the key perception capabilities of autonomous vehicles is to detect and recognise road actors for safe navigation and decision making. To the best of our knowledge, there is no existing work on radar based object detection for autonomous driving in extreme weathers. Therefore, we present the first baseline results using RADIATE.
Figure \ref{fig:lidar_fog} shows some examples of our data collected in adverse weather, in which trained networks fail to recognise actors in the optical images. As shown in Figure \ref{fig:lidar_fog}. The radar images are less affected so we would hypothesise that robust object detection is more likely in all conditions.

Drawn from the data acquired in several weather conditions, RADIATE includes 3 exemplary data sets that can be used for evaluation for vehicle detection:
\begin{itemize}
    \item \textbf{Training set in good weather}: This only contains data from the good weather conditions, i.e. sunny or overcast. It was created to validate whether sensors are capable to adapt from good weather to bad weather conditions.
    \item \textbf{Training set in good and bad weather}: This is an additional set with which includes data from both good and bad weather conditions (night, rain, fog and snow). This was used to develop algorithms in all weathers.
    \item \textbf{Test set}: This test set includes data from both good and bad weather conditions and is used for evaluation and benchmarking.
\end{itemize}

As a first baseline, we have performed evaluation of vehicle detection from single images. We defined a vehicle as one of the following classes: car, van, truck, bus, motorbike and bicycle. In Table \ref{tab:set_size} are shown the number of radar images and vehicles used to train and test the network.

\begin{table}[h]
\begin{center}
\resizebox{0.8\textwidth}{!}{\begin{tabular}{l|c|c}
\hline
{} & \#Radar Images&  \#Vehicles   \\
\hline
\hline
Training Set in Good Weather      &     23,091 & 106,931  \\
\hline
Training Set in Good and Bad Weather &      9,760 & 39,647   \\
\hline
Test Set                                              &     11,289 & 42,707  \\
\hline
\hline
Total                                              &     44,140 & 147,005 \\
\hline
\end{tabular}}
\end{center}
\caption{Number of images and number of vehicles for each set defined.}\label{tab:set_size}
\end{table}

\subsection{Radar based Vehicle Detection in the Wild}

We adopted the popular Faster R-CNN \cite{ren2015faster} architecture to demonstrate the use of RADIATE for radar based object detection. Two modifications were made to the original architecture to better suit radar detection:
\begin{itemize}
    \item Pre-defined sizes were used for anchor generation because vehicle volumes are typically well-known and radar images provide metric scales, different from camera images. %
    \item We modified the Region Proposal Network (RPN) from Faster R-CNN to output the bounding box and a rotation angle by which the bounding boxes are represented by {\it x, y, width, height, angle}.
\end{itemize}

\begin{figure}[ht]
        \begin{center}
        \includegraphics[width=0.6\textwidth]{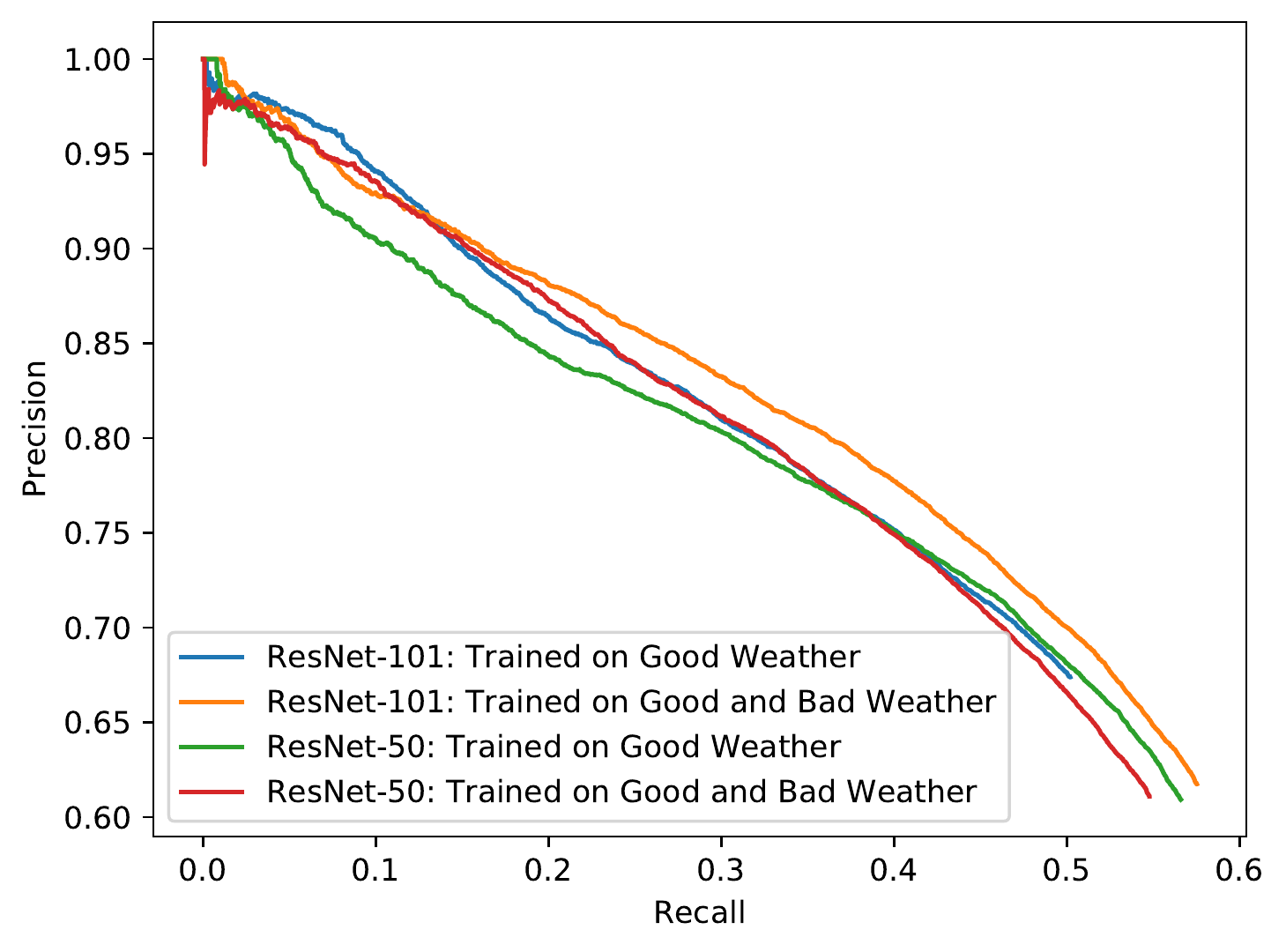}
\end{center}
    \caption{Precision recall curves.}\label{fig:prec_call_rot}
\end{figure}

\begin{figure*}[ht!]
    \begin{center}
    \includegraphics[width=\textwidth]{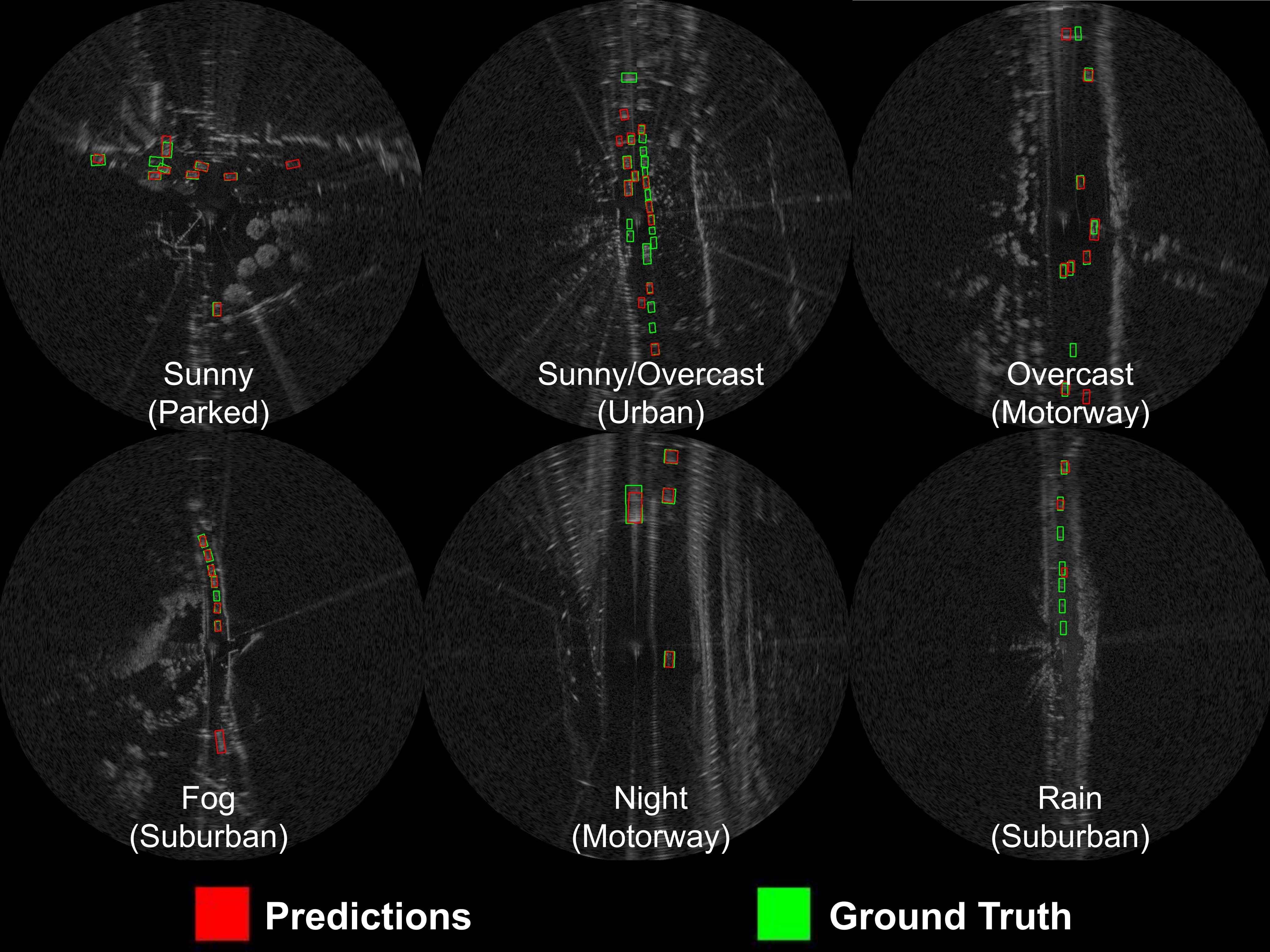}
    \end{center}
    \caption{Qualitative results of radar based vehicle detection.}
    \label{fig:qual_rot}
    
\end{figure*}

\begin{figure}[ht]
        \includegraphics[width=\textwidth]{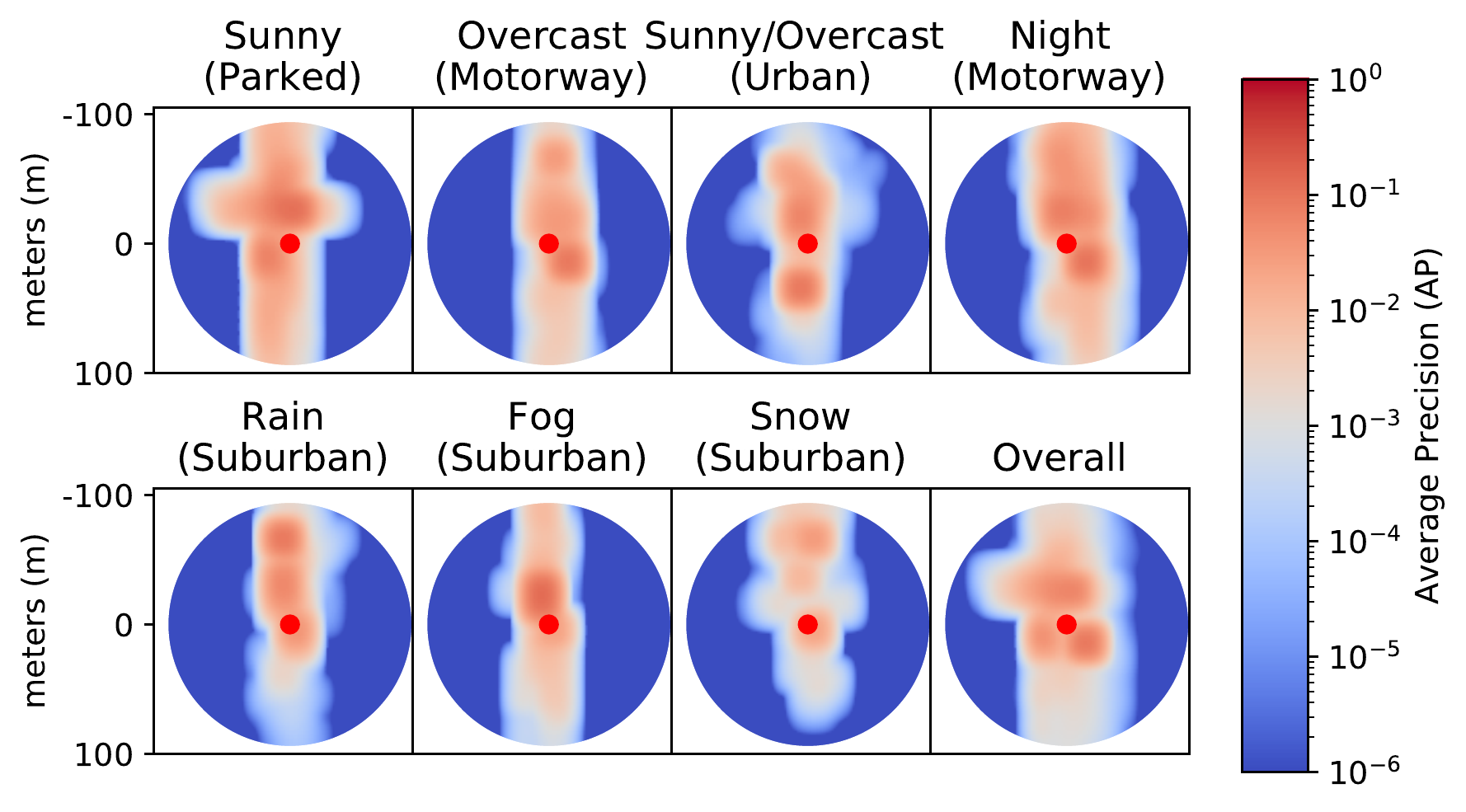}

    \caption{Heatmap of Average Precision trained in both good and bad weather.}\label{fig:heatmap_good_and_bad}
\end{figure}

\begin{figure}[ht!]

        \includegraphics[width=\textwidth]{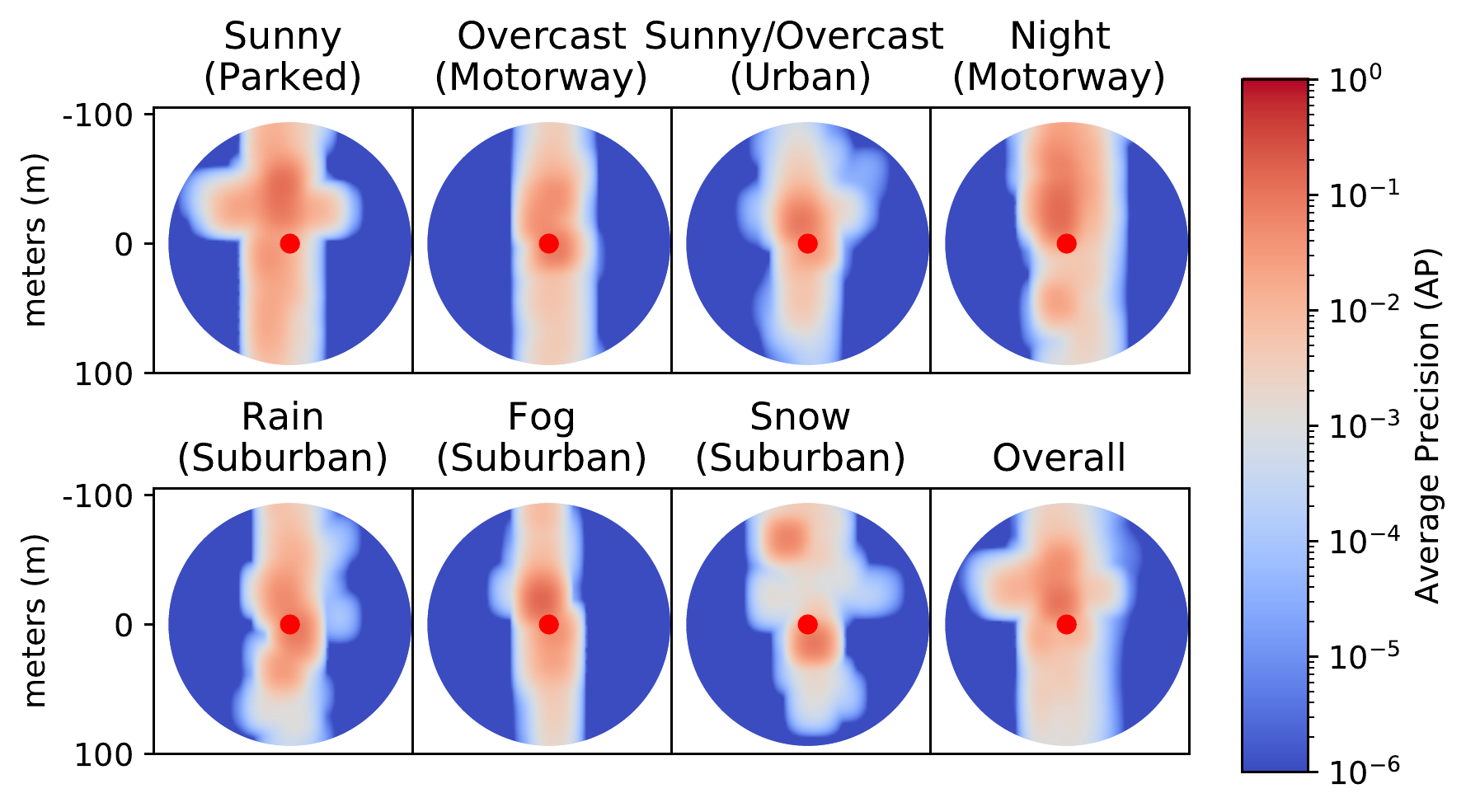}
     \caption{Heatmap of Average Precision trained in good weather only.}\label{fig:heatmap_good}
\end{figure}

To investigate the impact of weather conditions, the models were trained with the 2 different training datasets: data from only good and data from both good and bad weather. ResNet-50 and ResNet-101 were chosen as backbone models \cite{he2016deep}. The trained models were tested on a test set collected from all weather conditions and driving scenarios.  The metric used for evaluation was Average Precision with Intersection over Union (IoU) equal to 0.5, which is the same as the PASCAL VOC \cite{everingham2010pascal} and DOTA \cite{xia2018dota} evaluation metrics. Table \ref{tab:ap_rot} shows the Average Precision (AP) results and Figure \ref{fig:prec_call_rot} shows the precision recall curve. It can be seen that the AP difference between training with good weather and good\&bad weathers is marginal, which suggests that the weather conditions cause no or only a subtle impact on radar based object detection. The heat maps of the AP with respect to the radar image coordinates are also given in Figures \ref{fig:heatmap_good_and_bad} and \ref{fig:heatmap_good}. Since the AP distribution of the model trained only on the good weather data is very similar to the one of the model trained with both good and bad weather data, it further verifies that radar is promising for object detection in all weathers. Regarding the results in each scenario, it is mainly biased by the type of data, rather than the weather itself. The parked scenario is shown to be the easiest, achieving close to 80\% AP, potentially aided by the consistency in the radar return from the environmental surround. Results in snow and rain data performed worse. From Figure \ref{fig:lidar_fog}, the radar sensor used was shown to be affected by rain changing the background pixel values. The size of the snow data set was small and never used in any training set, which may well have affected the result. In the foggy scenario we could achieve considerably better results. Since it is a quite challenging scenario for optical sensors, radar is shown to be a good solution for dense fog perception. Again, the night data was collected in motorway scenarios, and the results were close to the motorway results in daytime. As expected, since radar is an active sensor, it is not affected by the lack of illumination.  Figure \ref{fig:qual_rot} illustrates some qualitative results of radar based vehicle detection in various driving scenarios and weather conditions, using Faster R-CNN ResNet-101 trained in good weather only.

\begin{table}[ht]

\begin{center}
\def\arraystretch{2.5}%
\resizebox{\columnwidth}{!}{\begin{tabular}{l|c|ccccccc}

{} & Overall & \makecell{Sunny \\(Parked)} & \makecell{Overcast\\(Motorway)} & \makecell{Sunny/Overcast\\(Urban)} & \makecell{Night \\ (Motorway)} &  \makecell{Rain \\ (Suburban)} &   \makecell{Fog \\ (Suburban)} &  \makecell{Snow \\ (Suburban)} \\
\hline
ResNet-50 Trained on Good and Bad Weather &   45.77 &  78.99 &    42.06 & 36.12 & 54.71 & 33.53 & 48.24 & 12.81 \\
ResNet-50 Trained on Good Weather        &   45.31 &  78.15 &    47.06 & 37.04 & 51.80 & 26.45 & 47.25 &  5.47 \\
ResNet-101 Trained on Good and Bad Weather &   46.55 &  79.72 &    44.23 & 35.45 & 64.29 & 31.96 & 51.22 &  8.14 \\
ResNet-101 Trained on Good Weather   &   45.84 &  78.88 &    41.91 & 30.36 & 40.49 & 29.18 & 48.30 & 11.16 \\
\end{tabular}}
\end{center}
\captionof{table}{Average Precision results on test set.}\label{tab:ap_rot}
\end{table}

\begin{figure*}[ht!]
    \centering
        \includegraphics[width=0.8\textwidth]{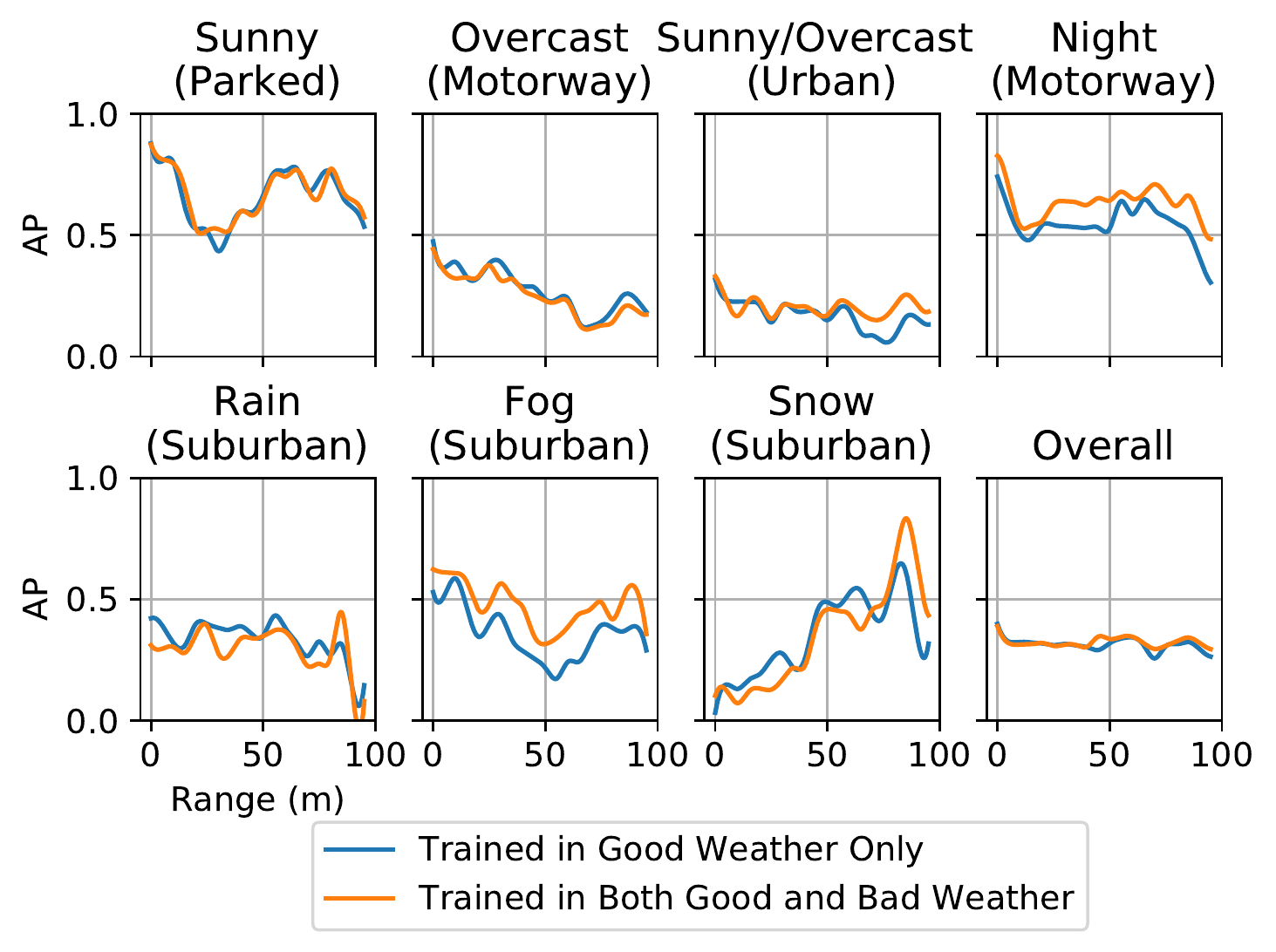}
    \caption{Average Precision (AP) over range for each scenario. Comparing the influence of training in good weather only vs. training in both good and bad weather.}\label{fig:radar_g_b_ap_range}
\end{figure*}
        
Figure \ref{fig:radar_g_b_ap_range} shows the comparison between trained in both good and bad weather vs. trained in good weather only. We can visualise that training in good and bad weather for all scenarios can achieve slightly superior results. As we stated previously, training on just good weather does not have much influence in the overall results.
\clearpage

\section{Conclusions}

We showed that radar is a potential perception sensor for autonomous cars and that it works in all-weather scenarios. We show how it can be used for radar based vehicle detection in the wild. These results show that radar based object detection are less affected by the weather, especially in fog scenarios, where recognition from LiDAR data fails at very short range depending on the fog density. When using sequences in both good and bad weather, we achieved 46.55\% AP. When using only sequences in good weather, the result is 45.84\% AP. This shows that even with a smaller dataset and no sequence in adverse weather, the recognition difference was only 0.71\%. This initial baseline experiment demonstrated promising results, especially in adverse conditions.
In this preliminary study, we emphasise that motion is not used for recognition, in part as actors may well be stationary, and in part because no Doppler data is available, although frame to frame tracking is an obvious avenue for further research provided a long term motion memory model is employed.

%% file: Chapters/Chapter9_conclusions.tex
\chapter{Conclusions}\label{ch:conclusions}

\lhead{Chapter 8. \emph{Conclusions}} %
This thesis investigated the use of long wave infrared and low THz radar sensors which can be used potentially for self-driving cars in all-weather scenarios.

\subsection*{Polarised Infrared}
We firstly investigated the use of polarised infrared sensors as a day-night vehicle detection system. This sensor contains 4 linear polarisers ($0^o$, $45^o$, $90^o$, $135^o$). From the linear polarisers we can compute the Stokes components $I$,$Q$,$U$,$P$ and $\phi$.  Two configurations are created ($I$,$Q$,$U$ and $I$,$P$,$\phi$). The $\{I, P, \phi\}$ parametrisation is intuitive in describing the polarisation ellipse and achieved the best overall result with Faster R-CNN ResNet-101. Our detection rates are similar to those of the KITTI dataset for vehicle detection from simple video data in daylight using the same networks. The polarised infrared data appeared to perform better than when CNNs are used for detection in infrared intensity data alone, confirmed also by previous research. Polarised infrared creates a better material discrimination which helped the CNNs to achieve a high average precision which is robust when we have a lack of illumination.

\subsection*{300 GHz Radar}
Most research on radar object recognition in the automotive scenario relies on Doppler features. Due to the lack of spatial resolution, Doppler is used to recognise moving targets. Can we recognise objects using the power data without Doppler information? The findings of this thesis suggest that we can achieve very high accuracy by increasing the radar sensor resolution.

We created a new dataset using a 300 GHz radar. This new dataset is bigger, and more suitable for deep learning. We did extensive experiments using different receiver locations, ranges and rotations. We also used transfer learning from larger datasets to improve generalizability, improving the accuracy from $92.5\%$ to $98.5\%$. We applied this technique to a more challenging scenario with multiple objects achieving 61.3\% AP using a perfect detector, and 50.3\% AP using a CFAR+DBSCAN detector.

Another way to improve generalizability is by artificially adding more data. We developed RADIO (Radar Data augmentatIOn), a novel data augmentation method targeting low-THz radar based on attenuation, range and speckle noise. This improved generalizability by artificially adding new realistic radar samples to the dataset. Comparing to a standard, camera based, data augmentation technique, RADIO improved AP by 14.76\% using a perfect detector and 14.48\% using CFAR+DBSCAN. Compared with the transfer learning technique developed previously, for the multiple object scenario developed, we improved it by 4.69\% AP using the perfect detector and by 5.23\% using CFAR+DBSCAN. These findings corroborate our hypothesis that specific data augmentation for radar improves the generalisation for both single object and multiple object datasets.

\subsection*{79 GHz Imaging Radar}
There is a lack of public radar datasets targeting autonomous vehicles. To fill that gap, we developed a new, large public dataset which included radar, LiDAR and stereo camera sensors. This data was collected under various weather scenarios. We collected data in sunny weather, overcast weather, night, rain, fog and snow. This is the first public radar dataset which addresses all these scenarios. We also developed an open-source tool which can help the development of new research in object detection, tracking and domain adaptation.

With this new dataset we investigated the robustness of object recognition in different weather scenarios. We compared LiDAR and radar in adverse weather. LiDAR was shown to have many unwanted artifacts, especially under foggy conditions, by having low point detection.
By comparison this had very little effects on the radar data. We also developed a vehicle detection method based on deep neural networks. We trained the networks in both good and bad weather. We achieved 46.55\% AP when training in both good and bad weather and 45.84\% AP when training in good weather data only using Faster R-CNN ResNet-101. These results suggest that training on good weather data only is sufficient to achieve comparable results.

We hope that this thesis will contribute to the development of new sensors which can be applied to autonomous cars in adverse weather. These are the initial steps which can lead to a  safer perception system which in turn can lead to full vehicle autonomy in all-weather scenarios.

\section{Future work}

Within this thesis there is a lot of potential for future work. Firstly, we did not use any temporal information (previous frames or Doppler). An obvious step is to use temporal information, where target tracking would be a research direction. There is a extensive literature on multi-target tracking which can be investigated for radar perception in automotive scenarios \cite{vo2006gaussian, bewley2016simple, 8296962}

We want to address is that the raw signal from our prototypical radar is a complex number. There is currently a research interest from the community in developing complex deep neural networks \cite{trabelsi2018deep, gaudet2018deep, zhu2018quaternion}.
Currently we only use the amplitude information, but maybe the phase can also help to improve the results. Applying the latest complex neural network research may lead to a better recognition by incorporating the phase information into the methodology.

We always treated the received signal as an image. But is it the best signal representation? We can represent the radar signals as a 2D point cloud. There is currently extensive literature on deep learning on point cloud data \cite{pmlr-v87-yang18b, qi2017pointnet, qi2017pointnet++, bronstein2017geometric, zaheer2017deep}. A way to use deep neural networks with point cloud data is to use graph neural networks (GNNs) \cite{bronstein2017geometric}. GNNs use a graph representation which is a natural way to represent a point cloud. We would investigate if the point cloud could be a potential representation.

We also would want to investigate multi-modal recognition in bad weather. We showed that radar is a robust sensor under severe weather conditions, but we can also combine radar with other sensors. An investigation of how to fuse radar with LiDAR and video in adverse weather is the next intuitive step. 

Single sensor and multi-sensor setups can also be investigated in domain adaptation tasks \cite{tzeng2014deep}. Since we have a  high variety of driving scenarios, investigating training in one scenario and adapting to recognise in another scenario is a clear option for future work.

We also contributed significantly to the creation of a publicly available dataset\footnote{\url{http://pro.hw.ac.uk/radiate/}} that has the potential for aiding novel research. We hope this dataset will be used by the community to develop new knowledge.

%% file: Chapters/Appendix_150ghz.tex
\chapter{150 GHz Object Recognition Based on Hand-Crafted Features} %
\label{chapter:150ghz} %

\lhead{Appendix A. \emph{150 GHz Object Recognition Based on Hand-Crafted Features}} %

In this appendix we will show a preliminary work developed during my PhD. We initially used a 150 GHz radar with a small dataset, which is not suitable for deep learning application. We have developed an initial object recognition system using hand-crafted features.
 It means that we will explicitly design the features which will discriminate the object. This is in contrast with deep learning techniques, where the feature extraction is learned by the neural network. In Chapters \ref{ch:300dl}, \ref{ch:radio}, \ref{ch:navtech} and \ref{ch:navtech2}, we used a larger dataset which was suitable for deep learning applications. In this appendix a methodology which combines shape and intensity features was developed. Small datasets with single and multi-objects were developed.
From the methodology developed, an accuracy of 91\% in a single object dataset was achieved and 77\% was achieved on a multiple object dataset.

\section{Introduction}

Research on object recognition based on radar usually exploits the use of Doppler features. Doppler is a natural phenomena which shifts the original frequency sent by the sensor due to the object speed. However by relying on Doppler features, we will not be able to recognise static objects. With the introduction of high resolution imaging radar, we hypothesise that we can recognise objects by only using the range image created by the radar sensor.

\begin{figure}[h]
  \centering
  \includegraphics[width=1\linewidth]{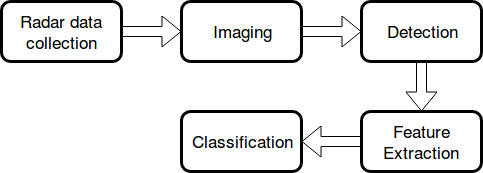}
  \caption{An overview of the methodology developed for this report.}
  \label{fig:methodology_150}
\end{figure}

In this chapter we used a high resolution 150 GHz radar developed by the University of Birmingham. We collected a small dataset with 4 different objects (pedestrian, bicycle, trolley and sign). An initial methodology based on shape and intensity features was developed. An overview of the methodology can be seen in Figure \ref{fig:methodology_150}.

\section{Dataset Collected}

We collected a dataset using a ZED Stereo Camera, a Velodyne HDL-32e LiDAR and 150 GHz radar. Below we describe the characteristics of each sensor which was used for the dataset.

\subsection*{Velodyne HDL-32e}
\label{sec:vel}

The LiDAR sensor used was Velodyne HDL-32e. This sensor is widely used in automotive scenarios for scene mapping and also target classification. It provides a real-time 3D point-cloud. The characteristics of the LiDAR are described in Table \ref{tab:lidar} \cite{velodyne_lidar}.

\begin{table}[ht]
\centering
\caption{Velodyne LiDAR HDL-32E characteristics}
\label{tab:lidar}
\begin{tabular}{rc}
\multicolumn{1}{|r|}{Channels}                & \multicolumn{1}{c|}{32}                                   \\
\multicolumn{1}{|r|}{Wavelength}              & \multicolumn{1}{c|}{903 nm}                                   \\
\multicolumn{1}{|r|}{Range}                   & \multicolumn{1}{c|}{80 - 100 m}                           \\
\multicolumn{1}{|r|}{Accuracy}                & \multicolumn{1}{c|}{+/- 2 cm}                              \\
\multicolumn{1}{|r|}{Data}                    & \multicolumn{1}{c|}{Distance / Calibrated Reflectivities} \\
\multicolumn{1}{|r|}{Data Rate}               & \multicolumn{1}{c|}{700,000 pts/sec}                      \\
\multicolumn{1}{|r|}{Vertical FOV}            & \multicolumn{1}{c|}{$41.33^o$}                                  \\
\multicolumn{1}{|r|}{Horizontal FOV}          & \multicolumn{1}{c|}{$360^o$}                                 \\
\multicolumn{1}{|r|}{Horizontal Resolution}   & \multicolumn{1}{c|}{5 Hz: $0.08^o$, 10 Hz: $0.17^o$, 20 Hz: $0.35^o$}     \\
\multicolumn{1}{|r|}{Power}                   & \multicolumn{1}{c|}{12 W}                                   \\
\multicolumn{1}{|r|}{Size}                    & \multicolumn{1}{c|}{86 mm x 145 mm }  \\
\multicolumn{1}{|r|}{Weight}                  & \multicolumn{1}{c|}{1 kg}                                      
\end{tabular}
\end{table}

\subsection*{ZED Stereo Camera}
\label{sec:zed}

The video sensor used for the data collection was the ZED Stereo Camera. The ZED sensor provides two cameras up to 4416 x 1242 image resolution. The ZED Camera provides dense depth reconstruction computation from 2 images. The algorithms for dense depth reconstruction are based on computing the disparity maps between the 2 images. To compute the disparity, we need to find the correspondent points. Equation \ref{eq:disparity} shows how to compute the disparity.

\begin{equation}\label{eq:disparity}
    disparity = x - x' = \frac{Bf}{Z}
\end{equation}

\begin{figure}[h]
\begin{subfigure}{.5\textwidth}
  \centering
  \includegraphics[width=1\linewidth]{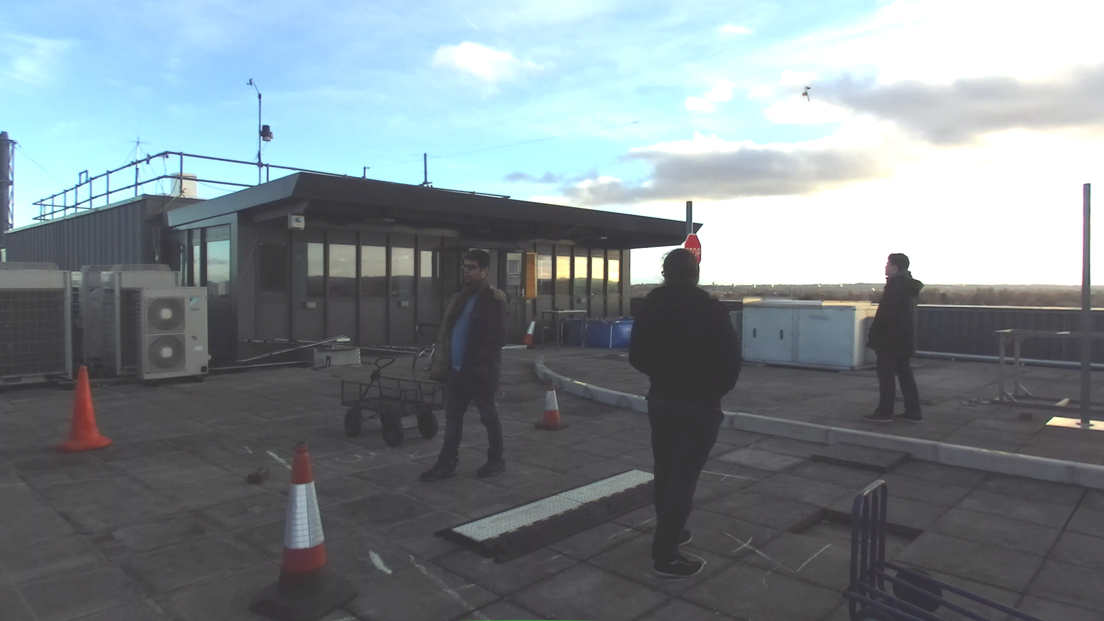}
  \caption{ZED Image Left}
   \label{fig:zed_left}
\end{subfigure}%
\begin{subfigure}{.5\textwidth}
  \centering
  \includegraphics[width=1\linewidth]{chapter4_1_figures/zed_example_right}
  \caption{ZED Image Right}
  \label{fig:zed_right}
\end{subfigure}

\centering
\begin{subfigure}{.5\textwidth}
  \centering
  \includegraphics[width=1\linewidth]{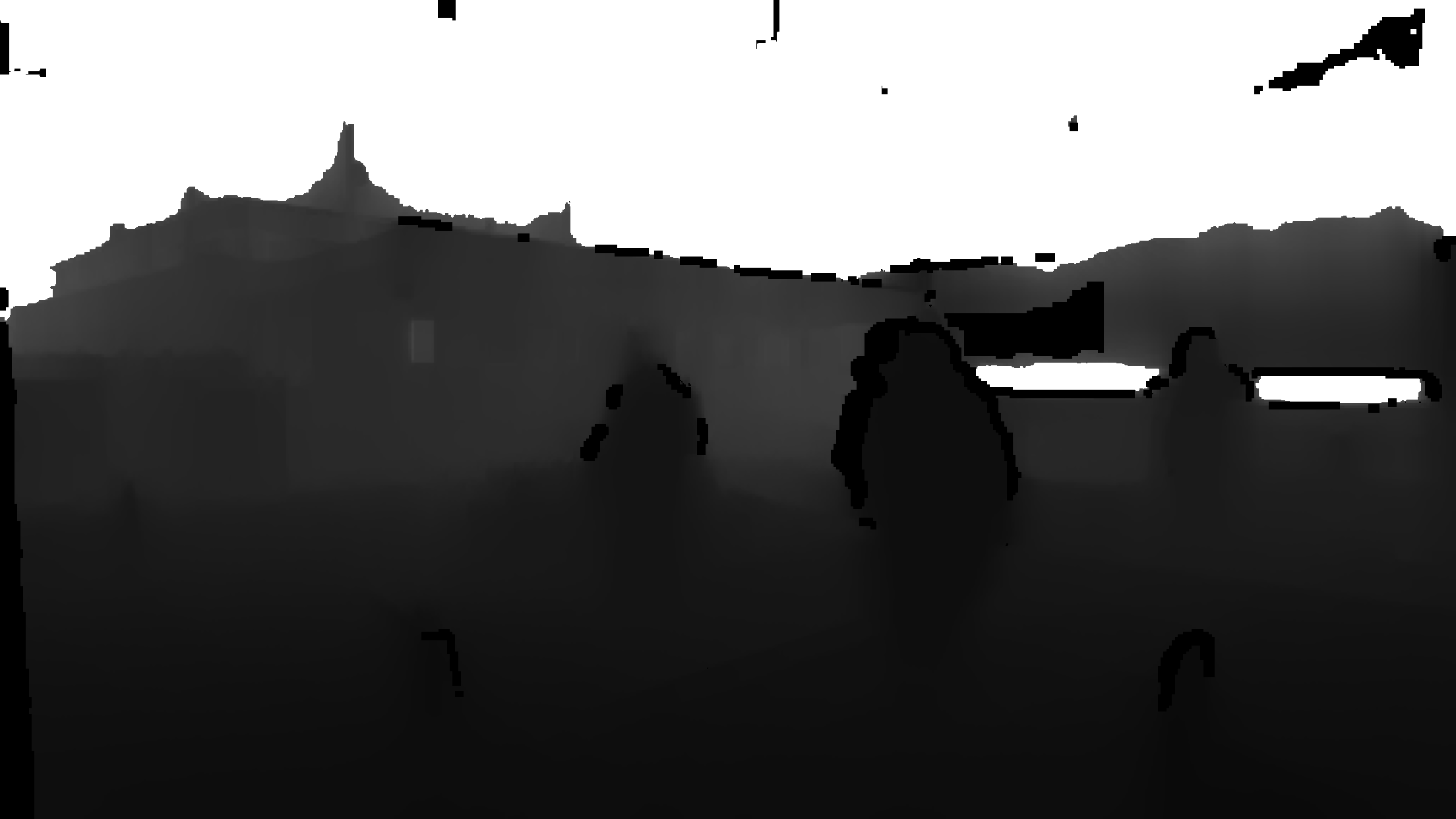}
  \caption{Depth Reconstruction from ZED camera}
  \label{fig:depth_zed}
\end{subfigure}%

\caption{Example of dense depth reconstruction given by Zed Stereo Camera.}
\label{fig:zed_example}
\end{figure}

The values of $x$ and $x'$ are the positions of the correspondent pixels in each camera, $B$ is the baseline distance between both cameras, $f$ is the focal length and Z is the distance of the object from the camera. The main challenge of this technique is to find the correspondent pixels - the classic approach is the block matching algorithm which uses sum of absolute differences \cite{muhlmann2001calculating}. However the algorithm used in the ZED device is not described by the developer. The ZED camera characteristics can be visualized in Table \ref{tab:zed} and an example of depth reconstruction can be seen in Figure \ref{fig:zed_example}.

\begin{table}[ht]
\centering
\caption{ZED Stereo Camera characteristics}
\label{tab:zed}
\begin{tabular}{rc}
\multicolumn{1}{|r|}{Resolution}              & \multicolumn{1}{c|}{4416x1242 at 15 fps, 3840x1080 at 30 fps}\\
\multicolumn{1}{|r|}{Depth Resolution}        & \multicolumn{1}{c|}{Same as selected video resolution}       \\
\multicolumn{1}{|r|}{Depth Range}             & \multicolumn{1}{c|}{0.5 - 20 m}                              \\
\multicolumn{1}{|r|}{Stereo Baseline}         & \multicolumn{1}{c|}{120 mm}                                  \\
\multicolumn{1}{|r|}{6-axis Position Accuracy}& \multicolumn{1}{c|}{+/- 1mm }                                \\
\multicolumn{1}{|r|}{6-axis Orientation Accuracy} & \multicolumn{1}{c|}{$0.1^o$}                             \\
\multicolumn{1}{|r|}{Field of View}           & \multicolumn{1}{c|}{$110^o$}                                 \\
\multicolumn{1}{|r|}{Aperture}                & \multicolumn{1}{c|}{$f/2.0$}                                 \\
\multicolumn{1}{|r|}{Dimensions}              & \multicolumn{1}{c|}{175 x 30 x 33 mm}                        \\
\multicolumn{1}{|r|}{Weight}                  & \multicolumn{1}{c|}{159 g}                                   \\                                   
\end{tabular}
\end{table}

\subsection*{FMCW 150 GHz Radar}
\label{sec:radar}

\begin{figure}[h]
  \centering
  \includegraphics[width=1\linewidth]{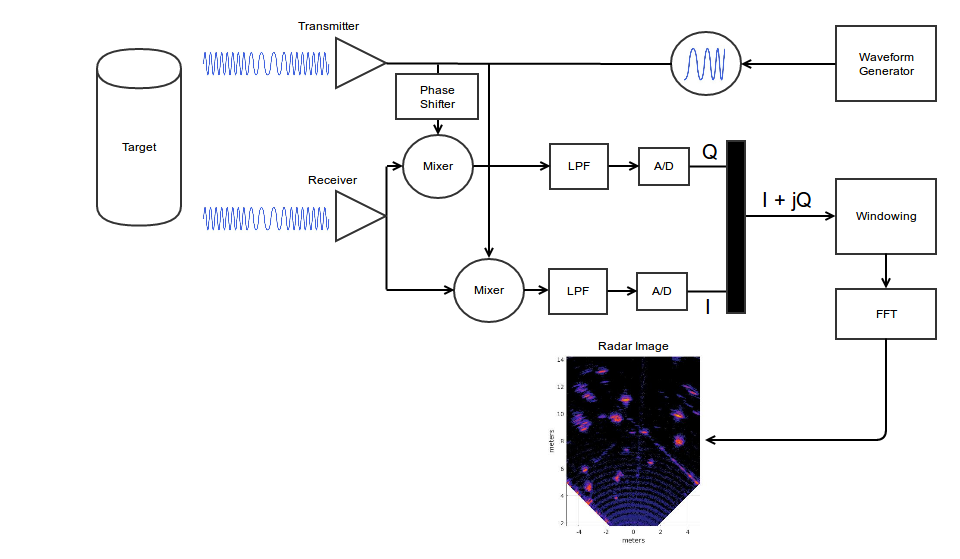}
  \caption{FMCW 150 GHz radar diagram developed by the University of Birmingham.}
  \label{fig:150ghz}
\end{figure}

The radar used was developed in the University of Birmingham in the Microwave Integrated Systems Laboratory (MISL). This radar developed an imaging system which is required for automotive scenarios. The main research question is to investigate the radar parameters in order to optimize general requirements, such as pulse bandwidth, antenna size, vehicle speed and scan rate. The main requirements are range, azimuth and elevation resolution which will be able to deliver reliable radar imaging for autonomous car perception. The antenna designed was compact in order to provide high azimuth resolution ($2.2^o$). To measure the cross-resolution over range we can use the angle resolution equation which is given as Equation \ref{eq:az_resolution}, where $S_A$ is the cross-range resolution given in meters (m), $R$ is the detection distance in meters (m) and $\theta$ is the H-Plane beamwidth in radians. Regarding elevation resolution, the beamwidth is $15^o$, the resolution of which can be found also using Equation \ref{eq:az_resolution}. There is current research trying to improve the elevation resolution using multiple receivers, but it needs to investigate where to place those sensors to retrieve reliable elevation \cite{phi300_2017} and \cite{phippen}. More details about the investigation on antenna parameters of low-THz imaging for automotive applications can be found in \cite{vizard2014antenna}. The diagram of the 150 GHz radar developed by the University of Birmingham can be seen in the Figure \ref{fig:150ghz}

\begin{equation}\label{eq:az_resolution}
    S_A \leq 2 R sin \bigg(\frac{\theta}{2} \bigg)
\end{equation}

The range resolution is based on the bandwidth transmitted - it is given by Equation \ref{eq:range_resolution}, where $S_R$ is the range resolution in meters (m), $c_0$ is the speed of light (m/s) and $B$ is the bandwidth (Hertz).

\begin{equation}\label{eq:range_resolution}
    S_R \geq \frac{c_0}{2 B}
\end{equation}

More details about the radar parameters can be seen in Table \ref{tab:150ghz_radar}.

\begin{table}[ht]
\centering
\caption{150 GHz FMCW Radar parameters}
\label{tab:150ghz_radar}
\begin{tabular}{r|c}
\hline
Bandwidth               & $6$ GHz                    \\ 
H-Plane (Azimuth) beamwidth   & $2.2^o$                    \\ 
E-Plane (Elevation) beamwidth & $+/-7.5^o$ ($15^o$)            \\  
Antenna aperture size         & Lens Horn ($7$ cm, $1.2$ cm) \\ 
Antenna gain                  & $29$ dBi                   \\ 
Range resolution              & $2.5$ cm                     \\ 
Azimuth resolution (at 10 m)  & $38$ cm                    \\ 
Elevation resolution (at 10 m)  & $2.61$ m                    \\ 
Power                         & $15$ mW ($11$ dBm)           \\
\hline
\end{tabular}
\end{table}

The main advantage of this radar over the current commercial radar is its azimuth resolution. A 30 GHz radar, for example, gives $61$ cm of azimuth resolution at 10 meters. According to \cite{vizard2014antenna}, a reliable azimuth resolution would be 20 cm, which is the measurement of the width of a tire.

\subsection{Data Collection}
\label{sec:data}

\begin{figure}[ht]
    \centering
    \includegraphics[width=0.6\textwidth]{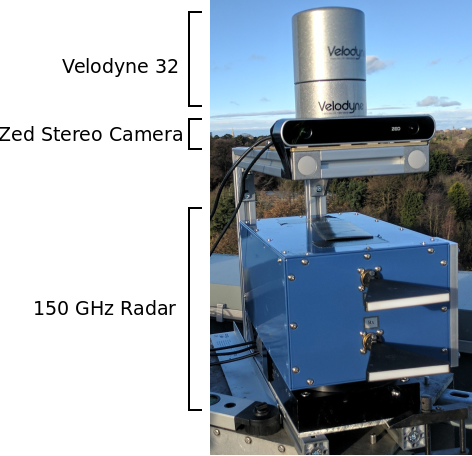}
    \caption{Configuration of the data collection equipment}
    \label{fig:radar_lidar_zed}
\end{figure}

A first batch of data was collected in the University of Birmingham in December of 2016. We developed this dataset with the intention of producing results in three research aspects: object recognition, sensor fusion, and target tracking. All samples were collected using a ZED Stereo Camera, Velodyne HDL-32e and the 150 GHz Radar (Figure \ref{fig:radar_lidar_zed}). Regarding sensor fusion, it can be done for both scenarios. For target classification, usually a fused representation is done after classification for each sensor \cite{chavez2016multiple}. For scene mapping, usually it fuses all data creating a probability occupancy grid, and tasks like classification and tracking are done in the fused representation \cite{chavez2016multiple}.

Two types of data collection scenarios were defined - a target classification set (dataset 1) and a tracking scenario set (dataset 2).
The radar described in the previous section is an experimental radar - it uses a mechanical round table for the scanning process. It takes around 2 minutes for one image scan of 60 degrees.

\begin{figure}[ht]
\begin{subfigure}{.5\textwidth}
  \centering
  \includegraphics[width=0.8\linewidth]{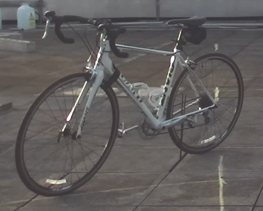}
  \caption{Bike}
\end{subfigure}%
\begin{subfigure}{.5\textwidth}
  \centering
  \includegraphics[width=0.9\linewidth]{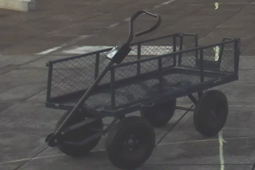}
  \caption{Trolley}
\end{subfigure}

\begin{subfigure}{.5\textwidth}
  \centering
  \includegraphics[width=0.3\linewidth]{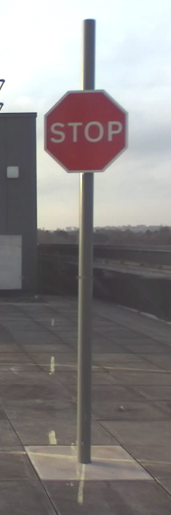}
  \caption{Traffic Sign}
\end{subfigure}%
\begin{subfigure}{.5\textwidth}
  \centering
  \includegraphics[width=0.37\linewidth]{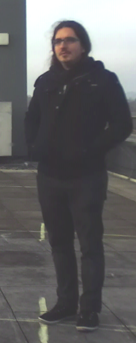}
  \caption{Pedestrian}
\end{subfigure}

\caption{The objects used in the target classification scenario.}
\label{fig:objects}
\end{figure}

\begin{itemize}
    \item \textbf{Dataset 1} : The first dataset collected is in a "easy" scenario, with the objects isolated from each other. 4 objects were chosen that are the most similar to what we see in a real road scenario. The objects were a pedestrian, a trolley, a stop sign and a bike (the objects can be seen in \ref{fig:objects}). The objects were positioned in 3 different ranges from the radar (4 m, 7 m and 11 m), and also 5 different poses for each range ($0^o$, $45^o$, $90^o$, $135^o$ and $180^o$). We considered the objects to be symmetrical, so it has the same target signature for poses larger than $180^o$. In the end we had 15 images for each object, 60 images in total. For this scenario we did not try to add objects to create hard scenarios, such as a cluttered scenario, or close objects. The idea of this dataset is to use it as training and use the training model in more realistic datasets.
    
    \item \textbf{Dataset 2} : This second dataset is in a "hard" scenario. This dataset tried to be the closest scenario to a road environment in order to track and map several objects. It used the same objects of the first scenario (trolley, bicycle, stop sign and pedestrian), plus other objects such as a bump, a tile, cones and kerbs. A scene was created where just the pedestrians were moving. We positioned the sensors in a small rail and moved it every 50 cm. This scenario has more challenging aspects regarding occlusion and close objects. This scenario captured 20 scans.
\end{itemize}

\begin{figure}[H]
\begin{subfigure}{.5\textwidth}
  \centering
  \includegraphics[width=\linewidth]{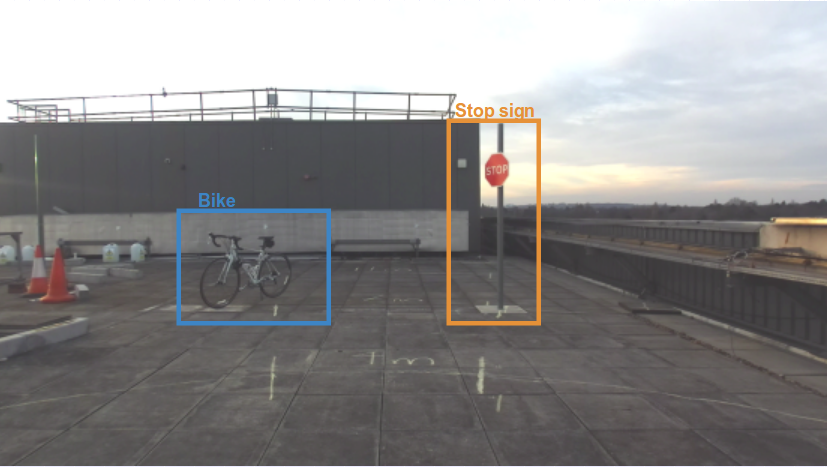}
  \caption{ZED Camera sensing bike and stop sign}
\end{subfigure}%
\begin{subfigure}{.5\textwidth}
  \centering
  \includegraphics[width=\linewidth]{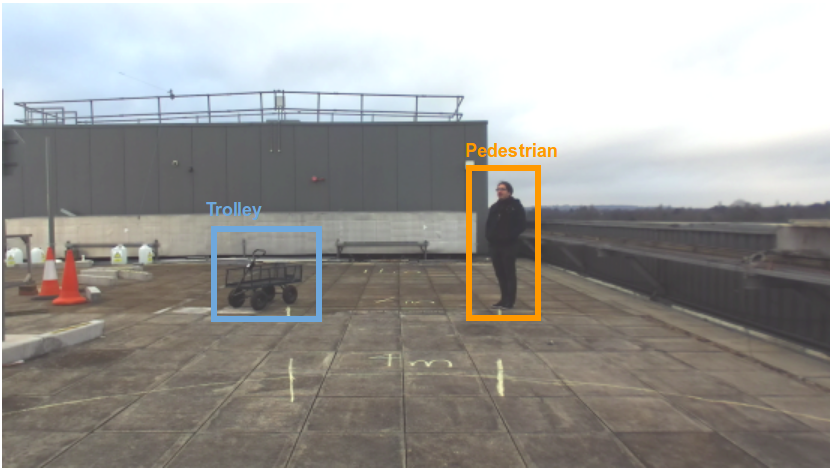}
  \caption{ZED Camera sensing trolley and pedestrian}
\end{subfigure}

\begin{subfigure}{.5\textwidth}
  \centering
  \includegraphics[width=\linewidth]{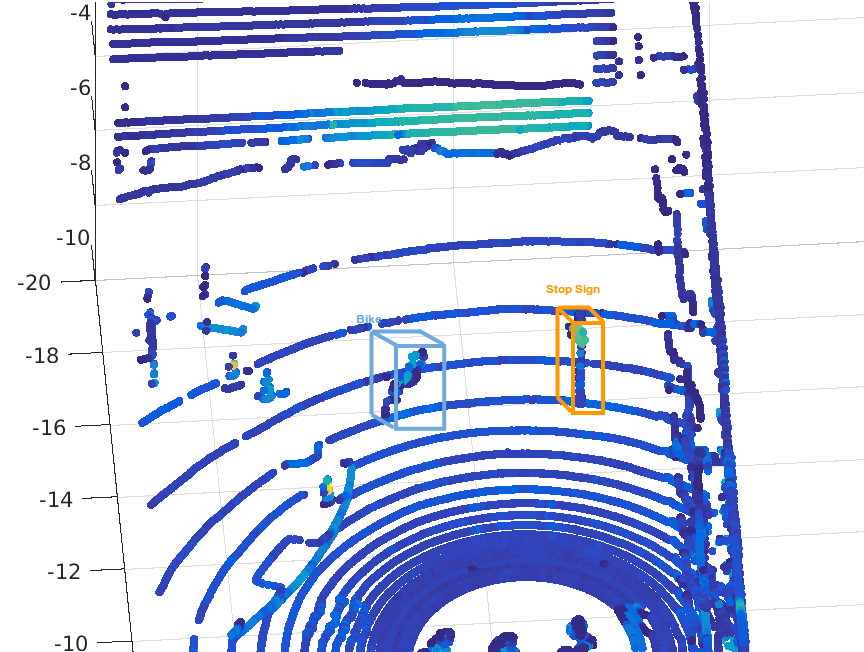}
  \caption{LiDAR sensing bike and stop sign}
\end{subfigure}
\begin{subfigure}{.5\textwidth}
  \centering
  \includegraphics[width=\linewidth]{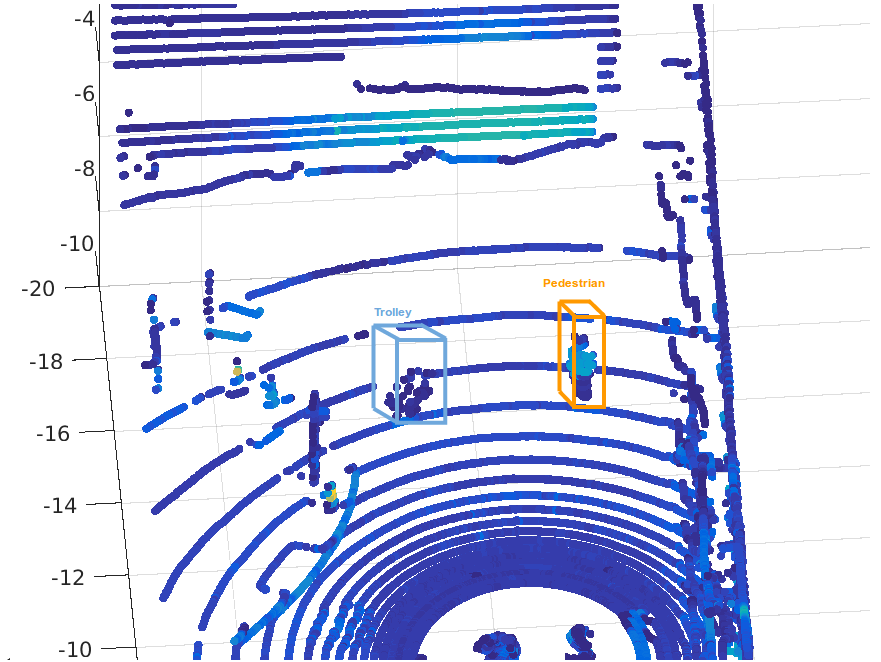}
  \caption{LiDAR sensing trolley and pedestrian}
\end{subfigure}

\begin{subfigure}{.5\textwidth}
  \centering
  \includegraphics[width=\linewidth]{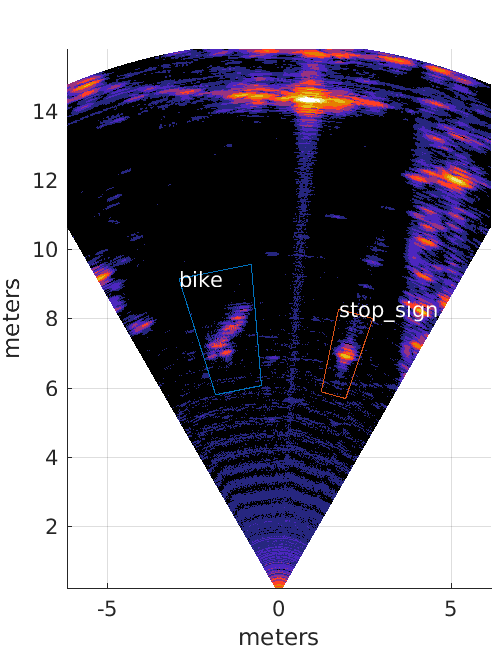}
  \caption{Radar sensing bike and stop sign}
\end{subfigure}
\begin{subfigure}{.5\textwidth}
  \centering
  \includegraphics[width=\linewidth]{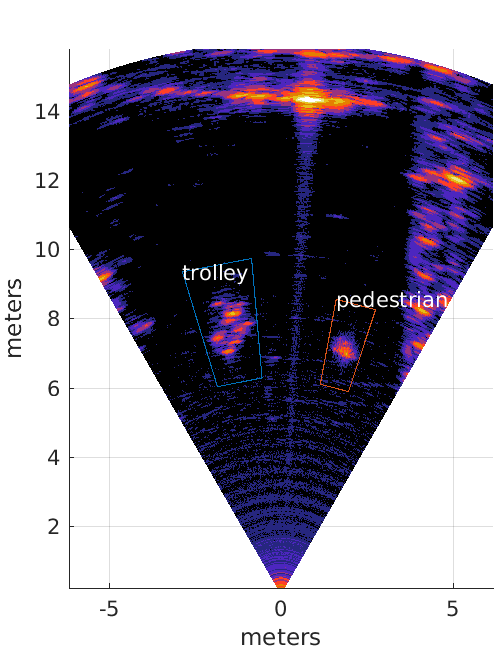}
  \caption{Radar sensing trolley and pedestrian}
\end{subfigure}
\caption{Two scenes showing how the target scenario was sensed. First column shows a bike and a stop sign and the second column shows the trolley and the pedestrian. All objects were at 7 meters and $45^o$.}
\label{fig:cls_example}

\end{figure}

\begin{figure}[H]
\begin{subfigure}{.5\textwidth}
  \centering
  \includegraphics[width=\linewidth]{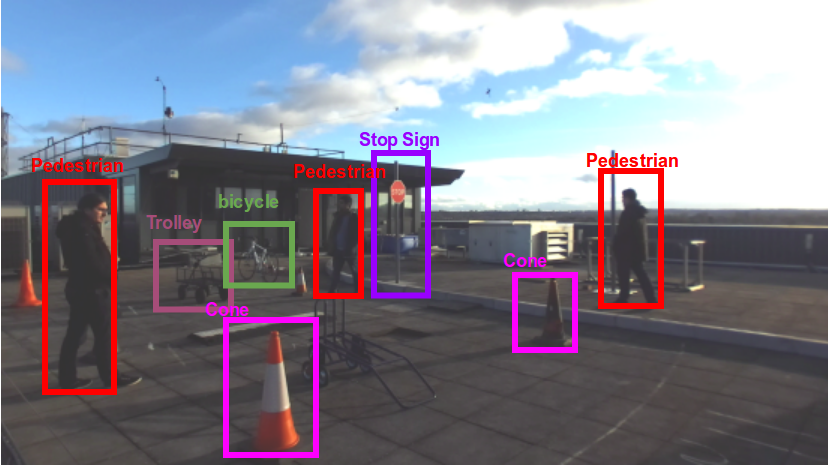}
  \caption{Left ZED Camera Frame 1}
\end{subfigure}%
\begin{subfigure}{.5\textwidth}
  \centering
  \includegraphics[width=\linewidth]{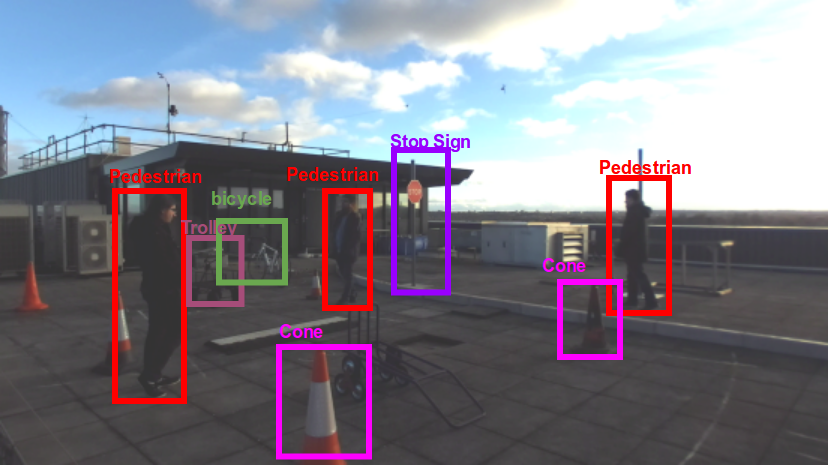}
  \caption{Left ZED Camera Frame 2}
\end{subfigure}

\begin{subfigure}{.5\textwidth}
  \centering
  \includegraphics[width=\linewidth]{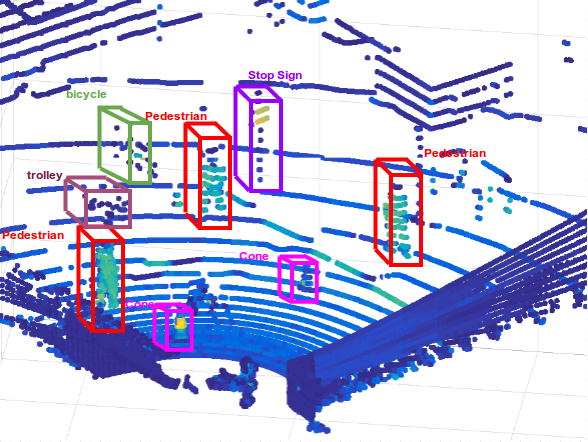}
  \caption{LiDAR Frame 1}
\end{subfigure}
\begin{subfigure}{.5\textwidth}
  \centering
  \includegraphics[width=\linewidth]{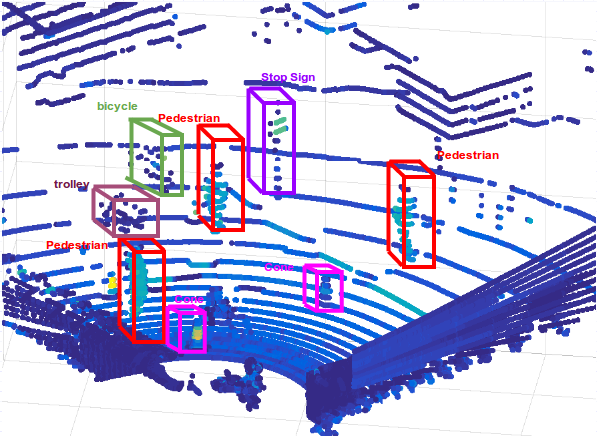}
  \caption{LiDAR Frame 2}
\end{subfigure}

\begin{subfigure}{.5\textwidth}
  \centering
  \includegraphics[width=\linewidth]{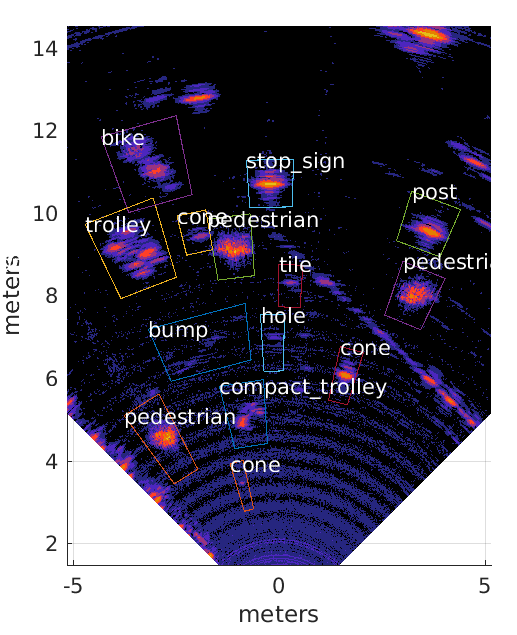}
  \caption{150 GHz Radar Frame 1}
\end{subfigure}
\begin{subfigure}{.5\textwidth}
  \centering
  \includegraphics[width=\linewidth]{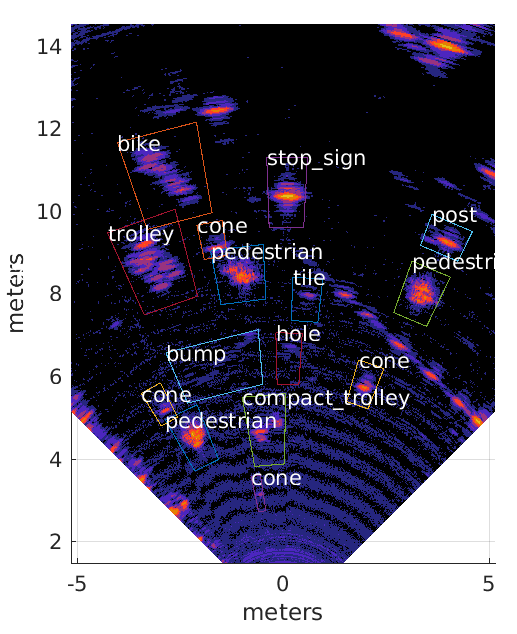}
  \caption{150 GHz Radar Frame 2}
\end{subfigure}
\caption{Two consecutive frames of the scene mapping scenarios. The first column corresponds to the first frame and the second column to the second frame. The ZED Camera, LiDAR and radar images are in each row.}
\label{fig:track_example}

\end{figure}

\section{Data Analysis}
\label{sec:data_analysis}

Doing an analysis of the collected data we can try to see how the signal behaves over angle and over range to identify the best features to extract.

\begin{figure}[H]
  \centering
  \includegraphics[width=1\linewidth]{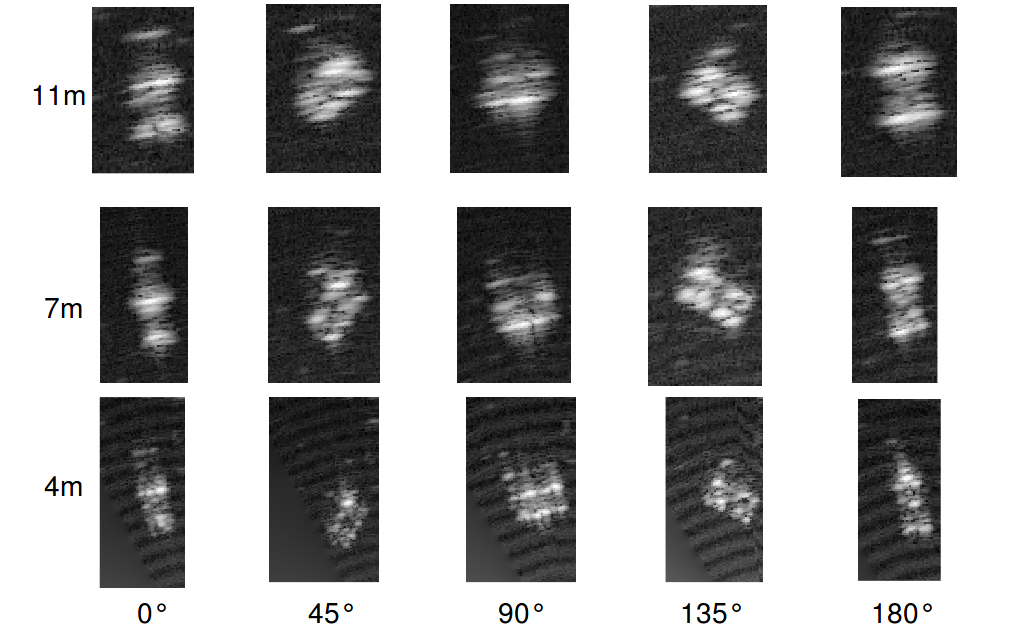}
  \caption{Trolley image at different angles and different ranges. Each row corresponds to 11, 7 and 4 meters. Each column corresponds to $0^o$, $45^o$, $90^o$, $135^o$, $180^o$.}
  \label{fig:trolleys}
\end{figure}

As we can see in Figure \ref{fig:trolleys}, the radar images change over range and over orientation. As expected the images loose cross-range resolution over range. With different orientations the object changes its signature. Multi path phenomena occurs in most of the images. We can observe some reflected signals that happen after the shape of the structure. 

To design features we need to find patterns from the reflected signal that will always occur. We can observe that the shape structure is preserved in most of the images. Shape features from top view can be extracted from this type of images. The features will be more widely discussed in Section \ref{sec:features}.

\subsection{Radar Detection}\label{sec:detection}

There are many approaches to performing radar detection . Radar detection has to be able to remove all of the noise and clutter from the received signal. One of the most popular ways is to use the Constant False Alarm Rate (CFAR) method.

\begin{figure}[ht]
\centering
\includegraphics[width=1\textwidth]{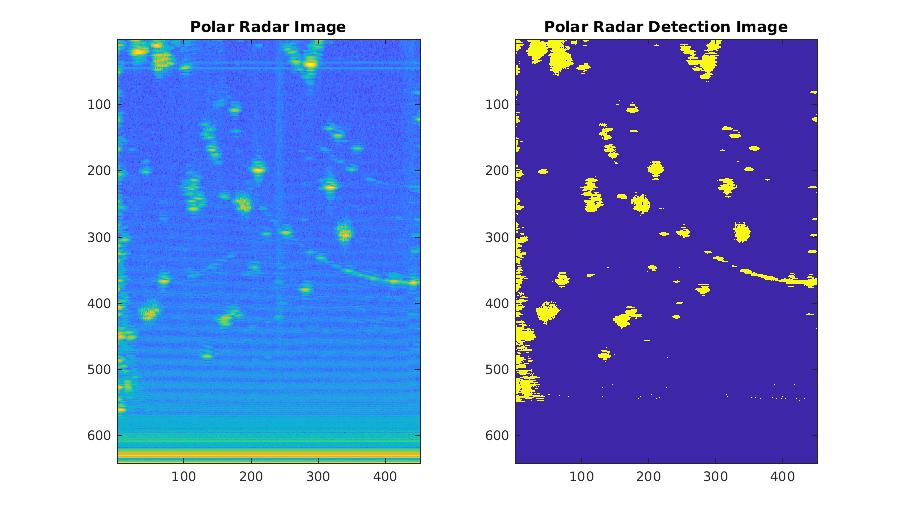}
\caption{Global Threshold Radar Detection.}
\label{fig:thresh}
\end{figure}

However in the first step of this project, since localization is not being performed, a simple global threshold is being used. To decide the best global threshold, an empirical analysis of the values of the background were done. Global threshold is not the best way of doing radar detection, but for our problem it was sufficient. Figure \ref{fig:thresh} shows a radar detection based on global threshold using polar coordinates. Since the first two meters have a high noise ratio, it was manually removed.

\section{Radar Feature Extraction}
\label{sec:features}
The feature extraction is the most crucial part in order to have a good recognition rate. As shown in the literature review chapter (Chapter \ref{chapter:literature_review}), there are many ways of classifying radar targets. Many of the target classification algorithms for automotive scenarios are based on Doppler signatures (\cite{sun2006radar, rohling2010pedestrian, heuel2013pedestrian}). We cannot use Doppler since our scene is always static. Some SAR and ISAR use shape features \cite{wei2011efficient, saidi2008automatic, park2013new} in order to classify the targets. In the radar image that we get, it is possible to develop a set of features to distinguish between different objects using its shape and amplitude features. Below a set of the features used are described. All features were computed in the cartesian coordinates image.

\begin{itemize}

\item \textbf{Area of Object} : This feature measures the area of the object. This feature is crucial in distinguishing between large objects and small objects. To calculate it, we simply sum all the pixels of the region that is detected (Equation \ref{eq:area}).

\begin{equation}\label{eq:area}
A = \sum_i  p(i)
\end{equation}

\item \textbf{Filled Area} : The filled area computes the area of the object including filled holes. This feature captures a better area if there are many holes in the image that can correspond to a better area of the object.

\item \textbf{Convex Area} : It measures the area of the Convex Hull. The Convex Hull is the minimum convex area that surrounds the object. This feature is similar to the Filled Area, however the convex hull can approximate to better areas since most of the objects that we are interested in have a  convex shape. As seen in the data analysis section, the trolley can have non-convex shapes depending on the pose. This feature will reduce the influence of different poses.

\item \textbf{Solidity} : The solidity returns the ratio of the area of the object by its Convex Hull area. Equation \ref{eq:solidity} shows how it is computed where $C$ is the convex area and $A$ is the area of the object. This feature represents how convex is our object of interest.

\begin{equation}
\label{eq:solidity}
S = \frac{A}{C}
\end{equation}

\item \textbf{Minor Axis Length} : The minor axis length approximates the object to an ellipse format and captures the minor axis length. This feature captures ellipse information from the object. Since we can approximate our object as an  ellipse, it can differentiate various objects using this feature.

\item \textbf{Major Axis Length} : This feature is similar to the previous one, but this one captures the major axis length instead of the minor axis length. 

\item \textbf{Eccentricity} : The eccentricity measures the ratio between the focus of the ellipse and the major axis length. It returns a value between 0 and 1, where 0 is a circle and 1 is a line. It approximates the best ellipse into region of the object. This kind of feature can distinguish easily between a trolley and a bike. Figure \ref{fig:eccentricity} shows different types of eccentricity. Equation \ref{eq:eccentricity} shows how to compute the eccentricity where $c$ is the focal point length, $a$ is the major axis length and $b$ is the minor axis length.

\begin{equation}
\label{eq:eccentricity}
\begin{aligned}
c^2 = a^2 - b^2 \\
e = \frac{c}{a}
\end{aligned}
\end{equation}

\begin{figure}[H]
\centering
\includegraphics[width=0.5\textwidth]{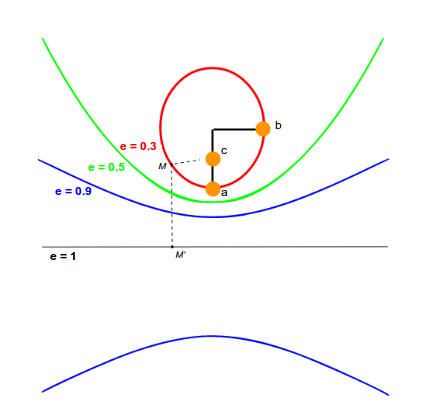}
\caption{Eccentricity.}
\label{fig:eccentricity}
\end{figure}

\item \textbf{Euler Number} : The Euler Number returns a value which is the area of the object minus the number of holes inside. 

\begin{equation}
\label{eq:euler}
E = A - H
\end{equation}

\item \textbf{Equivalent Diameter} : This feature measures the equivalent diameter. Equation \ref{eq:equiv_diameter} shows how it is computed.

\begin{equation}\label{eq:equiv_diameter}
D = \sqrt{\frac{4A}{\pi}}
\end{equation}

\item \textbf{Perimeter} : The perimeter approximates the object to a minimum boundary and measures the perimeter.

\item \textbf{Maximum Amplitude} : The maximum amplitude returns the maximum amplitude value from the object. The maximum value contains information about the shape, material and orientation of the object. The material is the most influential factor, as highly reflective materials have higher returns.

\item \textbf{Mean Amplitude} : The mean amplitude averages all the return values from the object. Getting the mean value can retrieve a better feature since we can minimize the influence of shape and orientation.

\end{itemize}

\section{Target Classification}
\label{sec:classification}

To recognize the objects in the radar image we need to use some classification algorithm. The Support Vector Machine was chosen to be used since it maximizes the margin between features creating a better generalization for the classification.

\subsection{Feature Scaling}

Before classifying, feature scaling is necessary to avoid some features having more influence than others. The rescaling method was used, where the features are normalized between 0 and 1. Equation \ref{eq:rescaling} shows how it is computed where x is one feature vector.

\begin{equation}
\label{eq:rescaling}
x' = \frac{x - min(x)}{max(x) - min(x)}
\end{equation}

\subsection{Support Vector Machine}

The  Support  Vector  Machine  (SVM)  \cite{gunn1998support}  was  used  to classify each target.   The  SVM has interesting properties as a classifier.  It is a linear classifier which uses support
vectors  to  maximize  the  margin  between  the  features. The SVM optimizes  the  margin
between  two  classes  creating  a  better  generalization  of  the  classification.   The  SVM  used  is the C-SVC, it uses a soft-margin approach, which allows some data inside the margin, and the C
is a parameter that controls it.  This important feature helps to create a better generalization in the classification, and also helps to avoid outliers and overfitting.  The kernel trick is also an important property of the SVM for non-linear classification - the kernel trick changes the basis of the features to a higher dimension where it can be linearly separable. The radial basis function was used to transform our dataset.

\section{Implementation details}
\label{sec:implementation}

The implementation of the methodology was developed using MATLAB 2016b using LibSVM \cite{chang2011libsvm}. The computer used is an Intel i7 3610QM, 8GB RAM. Below is the time it took for each step of the development: 

\begin{itemize}
 \item The radar imaging procedure takes 2.874 s
 \item The detection method takes 26.8 ms
  \item To extract features it takes 41.9 ms
  \item To train it it takes 41.9 ms
  \item to test it takes 0.867 ms
\end{itemize}

The bottleneck of this procedure is the radar imaging process. FFT algorithm is done for each azimuth to generate the image. But most of the methodology can be computed in real-time.

\section{Results}\label{sec:results} %

This section will show the results for the initial approach developed for this project. It shows that shape information has a high potential to classify objects in low-THz radar images.

\subsection{Results on Dataset 1}

 The dataset 1 is a set with 60 images with 15 images from each class (pedestrian, trolley, bike and stop sign) was used. Leave-one-out cross validation was used to evaluate the results. To select the best parameters of the SVM, grid search was used. Leave-one-out cross validation was used since we just have 60 samples, and computational power is not an issue for a small dataset, so we can have a good evaluation of our result with low bias between the training and the test set.

\begin{figure}[ht!]
  \centering
  \includegraphics[width=0.7\linewidth]{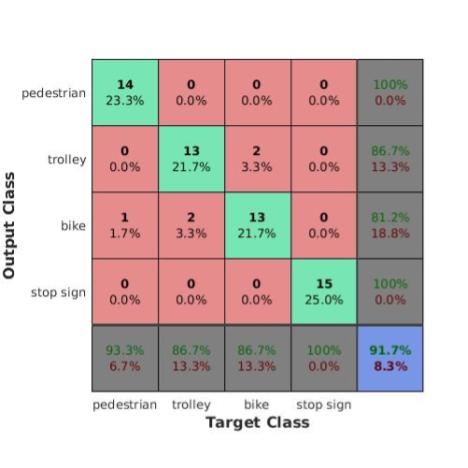}
  \caption{Confusion Matrix for the Dataset 1 test set.}
  \label{fig:conf_train}
\end{figure}

In Figure \ref{fig:conf_train} we can visualize the confusion matrix for the target classification set. As we can see, it generated high accuracy results 91.7\%. Just 5 targets were classified incorrectly.

\begin{figure}[ht!]
\centering
\includegraphics[width=\textwidth]{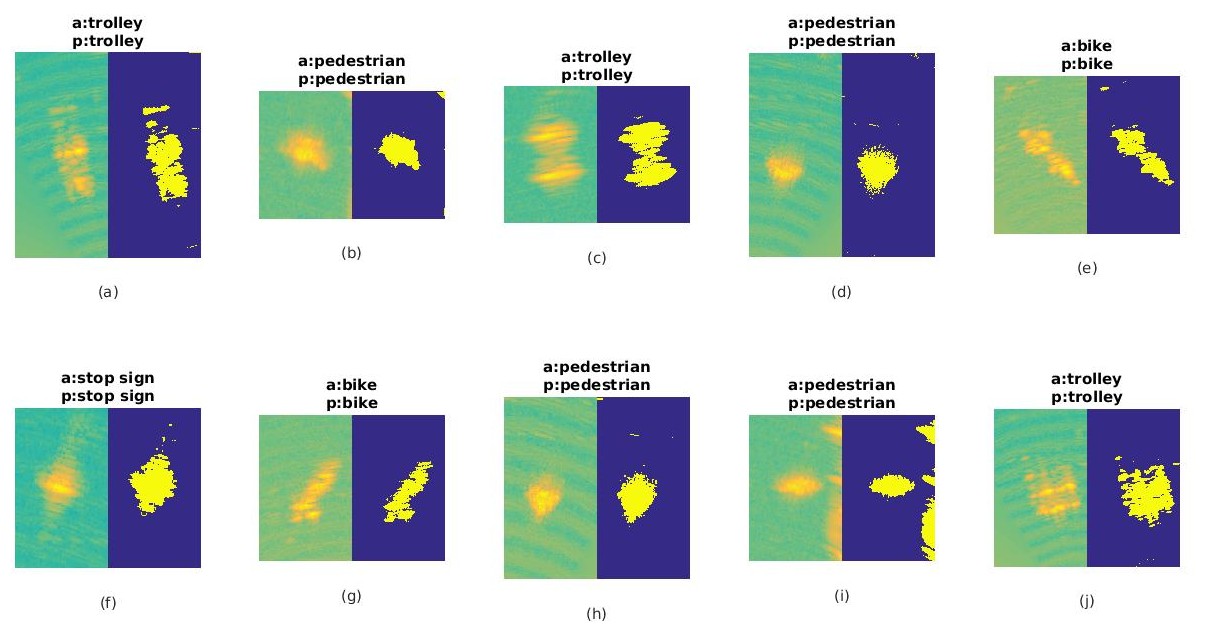}
\caption{Correct recognitions with correspondent detections from the target classification dataset. \textit{a:} means actual labels and \textit{p:} means predicted labels.}
\label{fig:correct_train}
\end{figure}

Figure \ref{fig:correct_train} shows some examples of targets with correspondent detection and its classification, where \textit{a:} means actual and \textit{p:} means predicted. Pedestrians and stop signs have similar shapes, however the material from the stop sign has a much high reflection, so the amplitude can distinguish between the two classes. 

\begin{figure}[ht!]
\centering
\includegraphics[width=\textwidth]{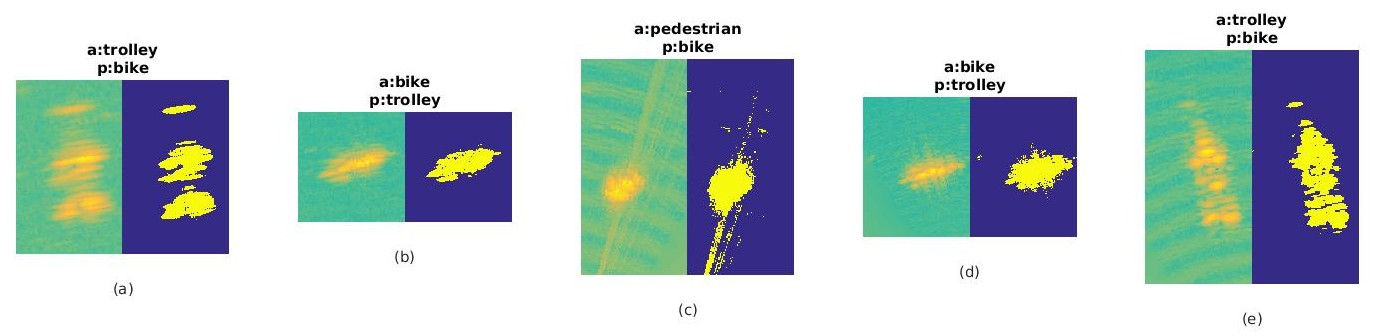}
\caption{Misrecognitions with correspondent detections from the dataset 1. \textit{a:} means actual labels and \textit{p:} means predicted labels.}
\label{fig:miss_train}
\end{figure}

Figure \ref{fig:miss_train} shows all misclassified targets. The target in Figure \ref{fig:miss_train}c has some corrupted signals, generating false detections. On features like minor and major axis this will be much different from the training set, generating misclassification. Figures \ref{fig:miss_train}a and \ref{fig:miss_train}e contain multi-path returns in the bottom of the object, making it larger, being misclassified as a bike. Figures \ref{fig:miss_train}b and \ref{fig:miss_train}d the bike at $90^o$ has a higher reflection than in the other rotations, generating misclassifications.

\subsection{Results on Dataset 2}

The dataset 2 is a more challenging scenario. It was not considered as target localization, the localization used is from the annotation given. The objects that are not present in the training set (cones, bumps and tiles) were not considered for the results. Localization and non-objects are addressed in Chapters \ref{ch:300dl}, \ref{ch:radio} and \ref{ch:navtech2}.

\begin{figure}[ht!]
  \centering
  \includegraphics[width=0.7\linewidth]{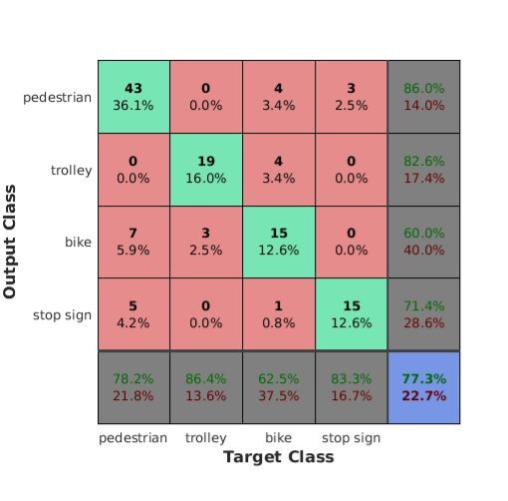}
  \caption{Confusion Matrix for the Dataset 2.}
  \label{fig:conf_test}
\end{figure}

In the dataset 2, 119 objects were used, 55 of which are pedestrians, 22 are trolleys, 24 bikes and 18 stop signs. Figure \ref{fig:conf_test} shows the confusion matrix for the dataset 2 as test set. The training model was generated using the dataset 1.

\begin{figure}[ht!]
\centering
\includegraphics[width=\textwidth]{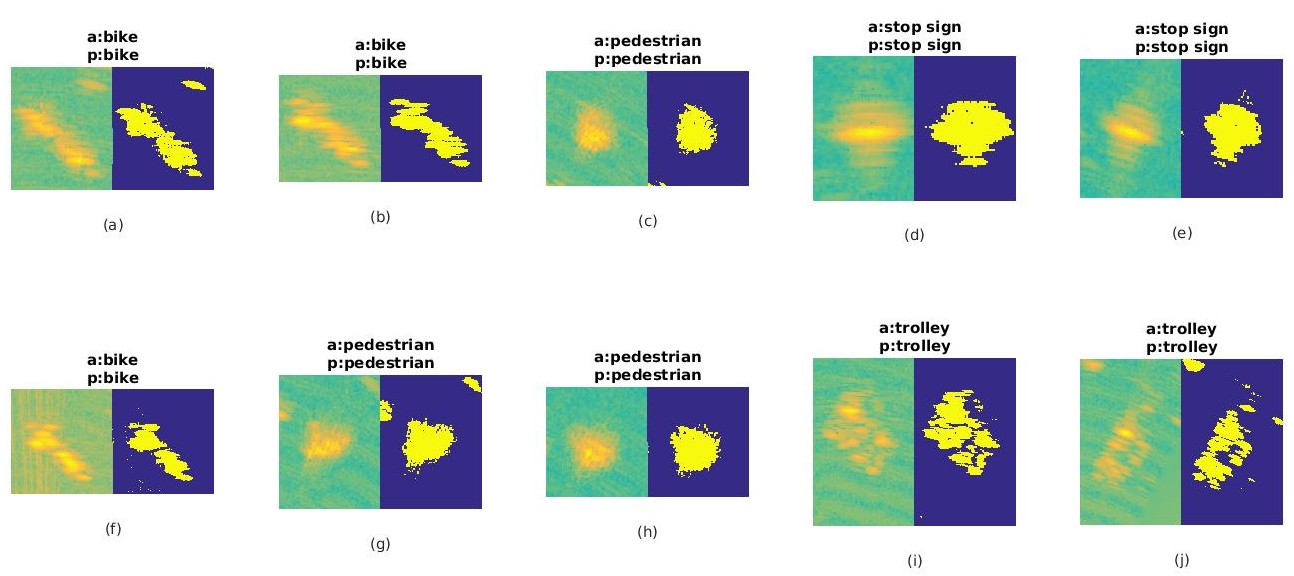}
\caption{Correct recognitions with correspondent detections from the dataset 2. \textit{a:} means actual labels and \textit{p:} means predicted labels.}
\label{fig:correct_test}
\end{figure}

The accuracy for the dataset 2 is 77.3 \%. The accuracy results were  14.5 \% lower than the dataset 1. The dataset 2 is a bit more challenging than the dataset 1 set. Occlusion between objects occurs in this scenario and some objects are close to each other. Since the scenarios are not the same, when the features are extracted, the data points do not correspond to the same distribution, giving different accuracy results. Some examples of correct classification can be seen in Figure \ref{fig:correct_test}. The correct examples were similar to the training examples, classifying the objects correctly.

\begin{figure}[ht]
\centering
\includegraphics[width=\textwidth]{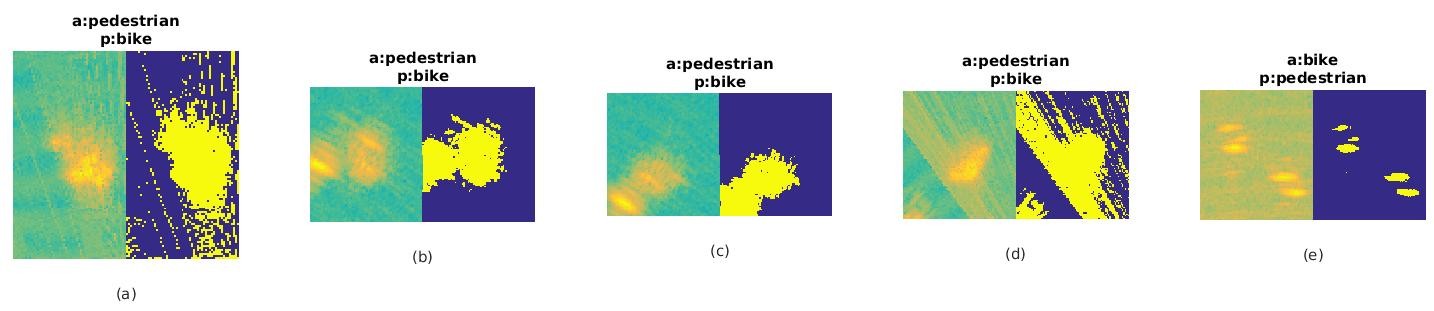}
\caption{Misrecognitions with correspondent detections from the test set. \textit{a:} means actual labels and \textit{p:} means predicted labels.}
\label{fig:miss_test}
\end{figure}

Figure \ref{fig:miss_test} shows some examples of misclassification. Figure \ref{fig:miss_test}a and \ref{fig:miss_test}d are caused by corrupted image acquisition. It is not known why it happened, but it only happened in very few examples. Figures \ref{fig:miss_test}b and \ref{fig:miss_test}c show the result when two objects are close to generating misclassification. Figure \ref{fig:miss_test}e is a misclassification caused by occlusion. The bike in the image is occluded by other objects, causing its return reflection to be low. The threshold method to retrieve a good detection was not enough for that case.

\section{Conclusions}

A methodology of object recognition on a 150 GHz radar is developed. Since we had a very small dataset, hand-crafted features based and on shape and intensity were used to discriminate the objects, opposed to using deep learning techniques.

The shape features were enough for most of the cases. For objects with similar features like a stop sign and a pedestrian, the classification can distinguish between them by their amplitude features, since the stop sign has a high reflectivity due to its metal material. Objects like a trolley and a bike have similar amplitude values due to reflective material, but they have different shapes.

We can list the main problems of the technique based on shape and amplitude:

\begin{itemize}
    \item Different signatures depending on the object pose and range.
    \item Difficulty to resolve when two objects are close.
    \item Corrupted data that is sometimes given by the radar.
    \item Occluded objects will have low return, making it more difficult to have accurate detection in that case.
\end{itemize}

It was decided to add this chapter to the appendix since it was a preliminary work developed in the beginning of my PhD. This work was used as proof of a concept that it is possible to recognise objects using high-resolution radar imaging. In Chapters \ref{ch:300dl} and \ref{ch:radio}, we introduced a prototypical higher resolution radar sensor (300 GHz) and developed an object recognition methodology based on deep neural networks.